\documentclass[11pt, a4paper, logo, onecolumn, nonumbering]{googledeepmind}

\pdftrailerid{redacted}

\usepackage{kantlipsum, lipsum}
\usepackage{dsfont}
\usepackage{multirow}
\usepackage{array}
\usepackage{amssymb} %
\usepackage{pifont} %
\usepackage{makecell}
\usepackage{algorithm}
\usepackage{algpseudocode}
\usepackage{amsmath}
\usepackage{nicefrac}
\usepackage{multicol}
\usepackage{float}
\usepackage{soul}
\usepackage{siunitx}
\usepackage{tabularx}
\usepackage{booktabs}
\usepackage{tabularray}
\UseTblrLibrary{booktabs} %
\usepackage{enumitem}
\usepackage{wrapfig}
\usepackage[T1]{fontenc}

\usepackage{xcolor}
\definecolor{wrongred}{HTML}{D32F2F}
\definecolor{correctgreen}{HTML}{28A745}
\definecolor{figtextcolor}{HTML}{333333}
\definecolor{baselineorange}{HTML}{FDAE61}
\definecolor{vipeblue}{HTML}{2c7bb6}
\definecolor{vipelightblue}{HTML}{77B2E5}
\definecolor{ablationgrey}{HTML}{767676}

\usepackage{subcaption}
\usepackage{opensans}
\DeclareCaptionFont{subfigcapsize}{\fontsize{8pt}{8pt}\selectfont}  %
\usepackage[numbers, sort&compress]{natbib}
\usepackage{xurl}
\usepackage[colorlinks=true, allcolors=vipeblue]{hyperref}
\usepackage[capitalise]{cleveref}
\crefname{figure}{Fig.}{Figs.}  %
\crefname{table}{Tab.}{Tabs.}  %
\crefname{section}{Sec.}{Secs.}  %
\crefname{subsection}{Sec.}{Secs.}  %
\crefname{subsubsection}{Sec.}{Secs.}  %
\crefname{appendix}{App.}{Apps.}  %

\usepackage{graphicx}
\graphicspath{{figures/}}

\usepackage[breakable, skins, most]{tcolorbox}
\newtcolorbox{calloutbox}{
    colframe=vipeblue,
    colback=vipeblue!10!white,
    coltext=vipeblue!30!black,
    arc=5pt,
    width=\columnwidth,
    left=2mm,
    right=2mm,
    top=1mm,
    bottom=1mm,
    boxrule=.5mm,
    fontupper=\normalfont
}
\newcounter{takeaway}
\newcommand{\takeaway}[1]{%
    \stepcounter{takeaway}
    \begin{calloutbox}
        \textbf{Takeaway\;\thetakeaway}\hspace{1em}#1
    \end{calloutbox}
}

\newtcolorbox{promptbox}[1]{
    enhanced jigsaw,            %
    colback=black!5!white,       %
    arc=5pt,                     %
    boxrule=0pt,                 %
    coltitle=black,              %
    fonttitle=\small,            %
    colbacktitle=black!10!white, %
    fontupper=\footnotesize,     %
    title=#1,                    %
    before upper=\setlength{\parskip}{0.5em},
    breakable,
}
\usepackage{listings}

\newtcolorbox{borderedimage}[1][blue]{
    colframe=#1,    %
    boxrule=0.8pt, %
    sharp corners,  %
    boxsep=0pt,     %
    left=0pt, right=0pt, top=0pt, bottom=0pt,
    enhanced,
    nobeforeafter,
    hbox
}

\newcommand{\taskname}[1]{{\itshape{#1}}}

\usepackage{tikz}
\usetikzlibrary{chains, positioning, fit, calc, backgrounds, arrows.meta}
\newlength{\tikzwidth}
\newlength{\boxpad}
\newlength{\boxgap}
\newlength{\boxtextheight}
\usepackage{placeins}

\usepackage{CJKutf8} %
\usepackage[export]{adjustbox}
\title{\centering Visual prompt engineering for video models}

\correspondingauthor{geirhos@google.com}

\usepackage{tabularx}
\usepackage{booktabs}

\reportnumber{} %

\author[1]{Robert~Geirhos\textsuperscript{*}}
\author[1]{Yuxuan~Li\textsuperscript{*}}
\author[1]{Thaddäus~Wiedemer\textsuperscript{*}}
\author[1]{Neha Kalibhat}
\author[1]{Zi Wang}
\author[1]{Mani Malek}
\author[1]{Oyvind~Tafjord}
\author[1]{Kevin~Swersky}
\author[1]{Been~Kim}
\author[1]{Priyank~Jaini\textsuperscript{*}}

\affil[]{Google DeepMind}
\newcommand{\projectpageurl}{https://visual-prompt-engineering.github.io/}

\begin{document}

\begin{abstract}
In the age of foundation models, a model is only as good as its prompt. For this reason, prompt engineering has become an essential technique for improving language model performance. Since video models are currently becoming foundation models for visual tasks (e.g., visual reasoning), we here ask whether they similarly benefit from \emph{visual} prompt engineering: automatically modifying the task image to improve model performance. For example, for a visual physics reasoning task (``Where does the ball land, after passing a set of obstacles?''), an abstract sketch-like scene can be turned into a photorealistic version with a simple call to an image editing model. We find that visual prompt engineering, or VIPE for short, improves video reasoning performance across tasks. In fact, for video models, visual prompt engineering can be even more effective than  classic text-based prompt engineering or test-time scaling. Ultimately, just as text-based prompt engineering systematically improves language model performance, visual prompt engineering can serve as a simple, compute-efficient approach to elicit better visual reasoning performance from video models.
Example videos on our \href{\projectpageurl}{project page}.
\end{abstract}

\newpage
\maketitle

\section{Introduction}

Prompt engineering has become a crucial component for improving language model performance~\citep{khattab2022demonstrate,white2023prompt,giray2023prompt,reynolds2021prompt,khattab2024dspy,yuksekgonul2024textgrad}. By carefully optimizing the text input, one can systematically elicit better outputs without modifying the model itself. Indeed, prompt design significantly affects model accuracy across a range of tasks, from question answering to code generation~\citep{marvin2023prompt,mesko2023prompt}. This is based on the observation that whenever a model's input space is rich enough, there is an opportunity to optimize that input for better downstream performance. Is this principle limited to language models, or does it extend to other modalities, such as the visual domain?

Generative video models have been proposed as foundation models for visual reasoning~\citep{wiedemer2025video,yang2024video,acuaviva2025rethinking,wang2026very,zeller2026mentisoculi,newman2026video,guo2026video,wang2026demystifying,li2026thinking,cheng2026vlms,tong2026thinking}. Recent works have demonstrated that they can tackle surprisingly challenging tasks, from maze solving to logic puzzles and physics simulations~\citep{wiedemer2025video}. While far from perfect, their abilities are rapidly improving. Unlike language models, which receive only text input, video models accept a \emph{text} prompt~(describing what should happen) and a \emph{visual} prompt~(typically an image that becomes the first frame of the generated video). The model then generates a video that depicts a solution attempt for the task.

Yet, while text prompts have been extensively studied and optimized for LLMs and VLMs~\citep{white2023prompt,reynolds2021prompt,wang2023review,gu2023systematic}, the visual input to video models has been treated as fixed. This asymmetry is surprising. For visual reasoning tasks, the image is arguably the dominant conditioning signal: it accounts for many tokens and defines the visual context, spatial layout, objects, and constraints that the model must reason over. Prior work has explored modifying visual inputs for image classifiers, for instance by overlaying shapes onto input images~\citep{shtedritski2023does,yao2024cpt}, learning pixel-level perturbations for CLIP~\citep{bahng2022exploring,zellerhighlight}, or inserting learnable tokens into vision transformers~\citep{jia2022visual}. Prompting via inpainting~\citep{bar2022visual,wang2023images}, visual cloze tasks~\citep{li2025visualcloze}, or visual instructions~\citep{lin2025draw} has also been explored. However, engineering the visual input to improve video model performance remains unexplored.

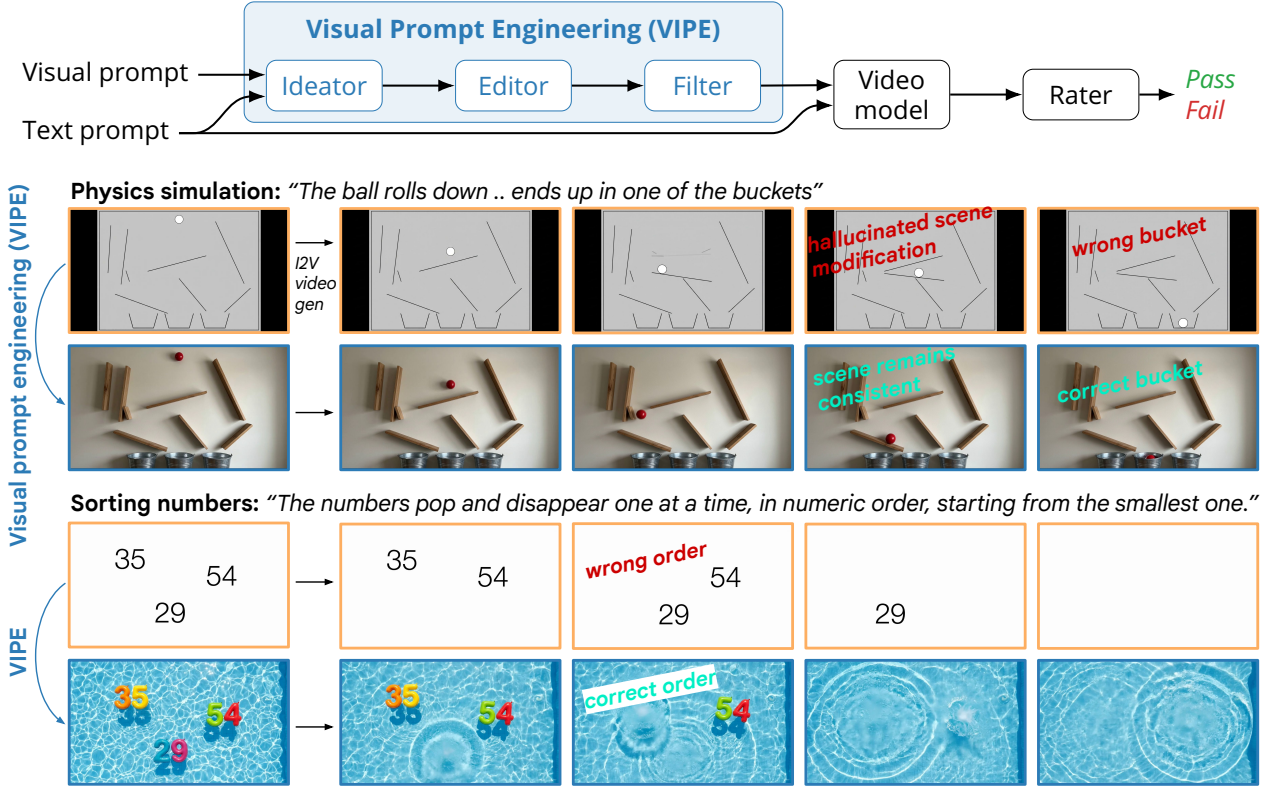
\begin{figure}[ht]
    \centering
    \resizebox{\textwidth}{!}{
        \begin{tikzpicture}[
    every node/.style={font=\small\opensans},
    box/.style={draw, rectangle, rounded corners, minimum width=1.6cm, minimum height=0.7cm, align=center, fill=white},
    solidbox/.style={box, solid},
    arrow/.style={-Latex, thick},
]
    \begin{scope}[local bounding box=pipeline]
        \begin{scope}[start chain=main going {right=of \tikzchainprevious.north east, anchor=north west}, node distance=1cm]
            \node[solidbox, draw=vipeblue, text=vipeblue, on chain]  (ideator) {Ideator};
            \node[solidbox, draw=vipeblue, text=vipeblue, on chain]  (editor)  {Editor};
            \node[solidbox, draw=vipeblue, text=vipeblue, on chain] (filter)  {Filter};
            \node[solidbox, on chain]  (vid)     {Video\\model};
        \end{scope}
        \node[solidbox, right=1cm of vid]  (rater)   {Rater};    
        \draw[arrow] (ideator.east) -- (editor.west);
        \draw[arrow] (editor.east) -- (filter.west);
        \draw[arrow] (filter.east) -- ([yshift=-0.35cm]vid.north west);
        \draw[arrow] (vid.east) -- (rater.west);

        \begin{scope}[on background layer]
            \coordinate (padding) at ([yshift=0.65cm]editor.north);
            \node[fit=(ideator) (filter) (padding), draw, rectangle, rounded corners, inner xsep=0.3cm, inner ysep=0.15cm, fill=vipeblue!10, draw=vipeblue] (vp) {};
            \node[anchor=north, font=\small\bfseries\opensans, text=vipeblue, yshift=-0.1cm] at (vp.north) {Visual Prompt Engineering (VIPE)};
        \end{scope}

        \coordinate[above left=0.15cm and 3.5cm of ideator.west]                  (img_spine);
        \coordinate[below left=0.3cm and 3.5cm of ideator] (txt_spine);
        \node[anchor=west] (img) at (img_spine) {Visual prompt};
        \node[anchor=west] (txt) at (txt_spine) {Text prompt};
        \draw[arrow] (img.east) -- ([yshift=0.15cm]ideator.west);
        \draw[arrow] (txt.east) -- ($(txt.east -| ideator.west) + (-1cm, 0)$) to[out=0, in=180] ([xshift=-0.2cm, yshift=-0.15cm]ideator.west) -- ([yshift=-0.15cm]ideator.west);
        \draw[arrow] (txt.east) -- ($(txt.east -| vid.west) + (-0.85
        cm, 0)$) to[out=0, in=180] ([xshift=-0.2cm, yshift=-0.15cm]vid.west) -- ([yshift=-0.15cm]vid.west);;

        \node[right=0.5cm of rater, align=left, font=\small\itshape\opensans] (out) {\textcolor{correctgreen}{Pass} \\ \textcolor{wrongred}{Fail}};
        \draw[arrow] (rater.east) -- ++(0.2,0) |- (out.west);
    \end{scope}
\end{tikzpicture}
    }
    \\\vspace{0.75em}
    \includegraphics[width=\linewidth]{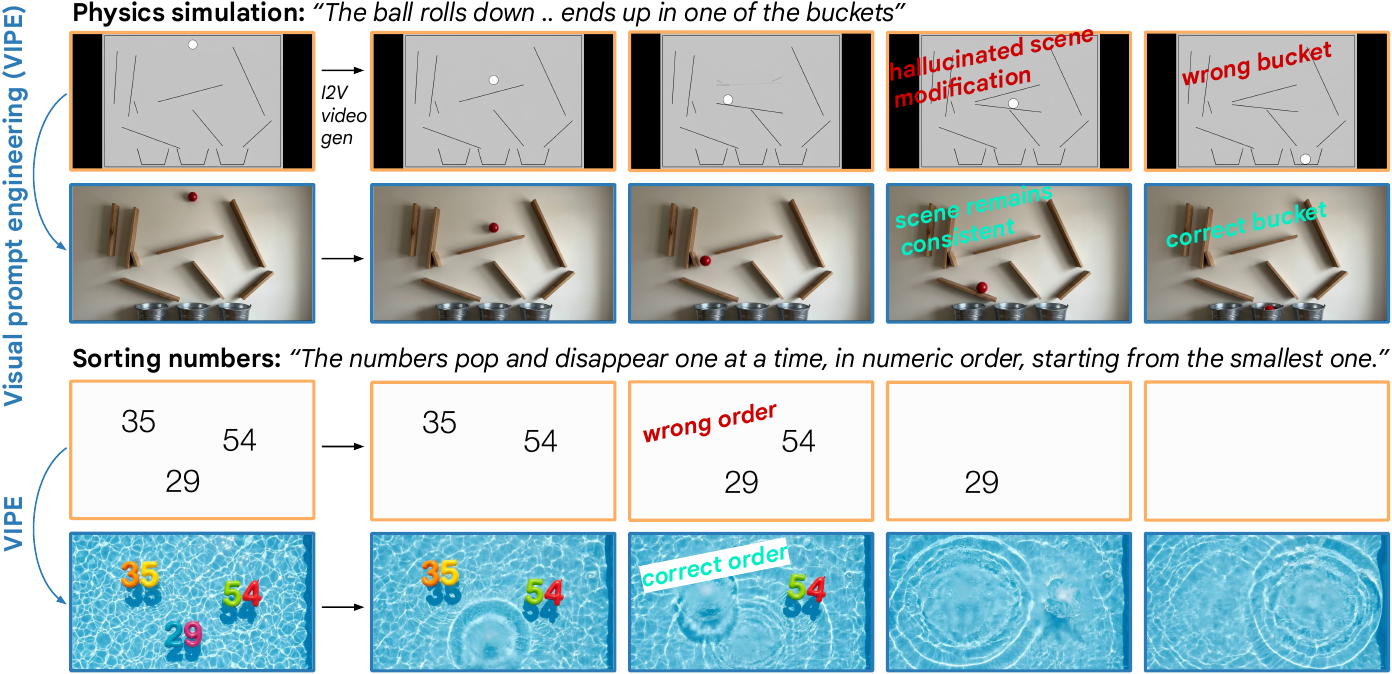}\hfill
    \caption{\textbf{Visual prompt engineering (VIPE)} starts with a visual prompt~(e.g., a ball-and-ramp sketch image) and a text prompt~(e.g., ``The ball rolls down~[...] ends up in one of the buckets''). Instead of directly generating a video, an \textit{ideator} proposes variants of the task~(e.g., replacing the sketch ball with a realistic billiard ball, and the sketch obstacles with wooden shelves). The task variants are then generated with an image \textit{editor}~(e.g., Nano~Banana), followed by an optional \textit{filter} for quality control (e.g., a VLM). After completing visual prompt engineering, the video model generates a video for the modified scenario, which is then scored as a pass/fail. See videos on our \href{\projectpageurl}{project page}.}
    \label{fig:teaser}
\end{figure}

We address this gap by asking: \emph{Can we improve visual reasoning capabilities of video models by transforming the input task image?} We call this \emph{visual prompt engineering}, or VIPE. Concretely, VIPE transforms the visual prompt (e.g., replacing an abstract ball-and-ramp sketch with a photorealistic scene of billiard balls rolling down wooden shelves) using an image editing model, and then passes the transformed image to the video model for reasoning~(e.g., generating a visual prediction for where the ball would roll). The underlying task remains identical; only its visual presentation changes, with the goal of providing a visual context that the model can reason over more effectively.

\begin{calloutbox}
Our \textbf{main contributions} are as follows:
\begin{itemize}[leftmargin=*]
    \item We introduce visual prompt engineering (VIPE), a simple approach that consistently improves visual reasoning capabilities of video models.
    \item We show that VIPE is easily automated and an effective test-time scaling strategy.
    \item We demonstrate that refining the visual context is often more effective for video models than optimizing text instructions.
    \item We reveal that video models systematically prefer photorealistic context over abstract inputs: sketch-based benchmarks can significantly underestimate model capabilities.
\end{itemize}
\end{calloutbox}

\section{Visual prompt engineering (VIPE)}
\paragraph{Definition of visual prompt engineering for video models}
We define a task sample as a pair $(x_i, t_i)$, where $x_i \in \mathcal X$ is a visual prompt (here: always an image) and $t_i \in \mathcal T$ is a text prompt. A video model attempts to solve each task sample by generating a video. Visual prompt engineering (VIPE) aims to improve downstream reasoning performance by replacing an original image $x_i$ with a variant $v_i^*$ without altering the task's underlying logic.

\paragraph{Step 1: Ideator}
Conditioned on an ideation prompt $t_\text{ideate} \in \mathcal{T}$ and $k$ task samples, the ideator (say, a human or a language model) $I: \left(\mathcal{X} \times \mathcal{T}\right)^k \times \mathcal{T} \to \mathcal T$ describes a visual edit in natural language as
\begin{equation}
    t_\text{edit} \sim I\left(X, t_\text{ideate}\right), \quad \text{where} \quad X = \left\{(x_i, t_i)\right\}_{i=1}^k.
\end{equation}
The edit instructions $t_\text{edit}$ are only constrained by the ideation prompt $t_\text{ideate}$, which ensures that the core task remains unchanged, e.g., by specifying what scene elements may not change.

\paragraph{Step 2: Editor}
Let $E: \mathcal{X} \times \mathcal{T} \to \mathcal{X}$ be a general image editor (in our case an image editing model; though in principle this could also be a human), conditioned on edit instructions $t_\text{edit}$. For an original prompt image $x_i$, we sample $m$ candidate variants
\begin{equation}
    V_i = \left\{v_{i,j}\right\}_{j=1}^m, \quad \text{where} \quad v_{i,j} \sim E\left(x_i, t_\text{edit}\right).
\end{equation}

\paragraph{Step 3: Filter (optional)}
To select the highest quality variant, let $S: \mathcal{X} \times \mathcal{X} \to \mathbb{R}$ be a scoring function (e.g., a vision-language model) that evaluates a candidate's quality and its faithfulness to the original prompt image $(x_i)$. The optimal candidate $v_i^*$ is selected by maximizing this scoring function:
\begin{equation}
    v_i^* = \arg\max_{v \in V_i} S\left(v, x_i\right)
\end{equation}

\noindent
Due to its stochastic nature, the entire pipeline can be repeated to obtain $n$ variants $v_{i,j}^*$ for the same original visual prompt $x_i$, e.g., for ensembling.

\section{Text-based prompt engineering helps language models for language tasks. Does visual prompt engineering help video models for visual tasks?}
\label{sec:vpct}
\begin{figure*}[t]
    \centering
    \includegraphics[width=\linewidth]{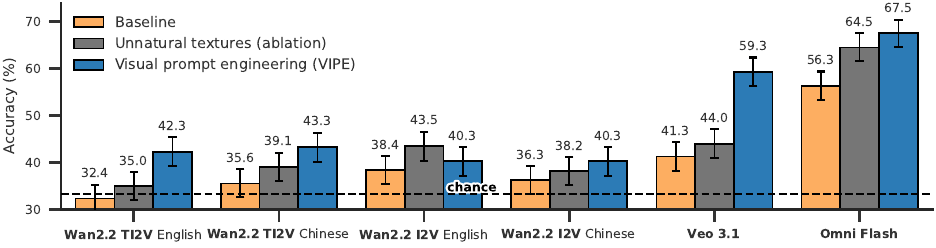}\vspace{1mm}

    \begin{borderedimage}[baselineorange]
        \includegraphics[width=0.19\linewidth]{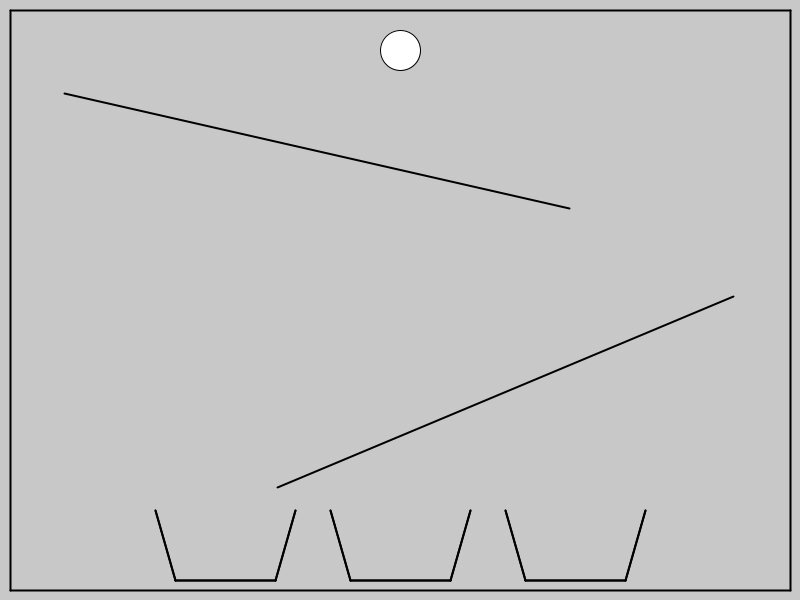}
    \end{borderedimage}%
    \hfill
    \begin{borderedimage}[baselineorange]
        \includegraphics[width=0.19\linewidth]{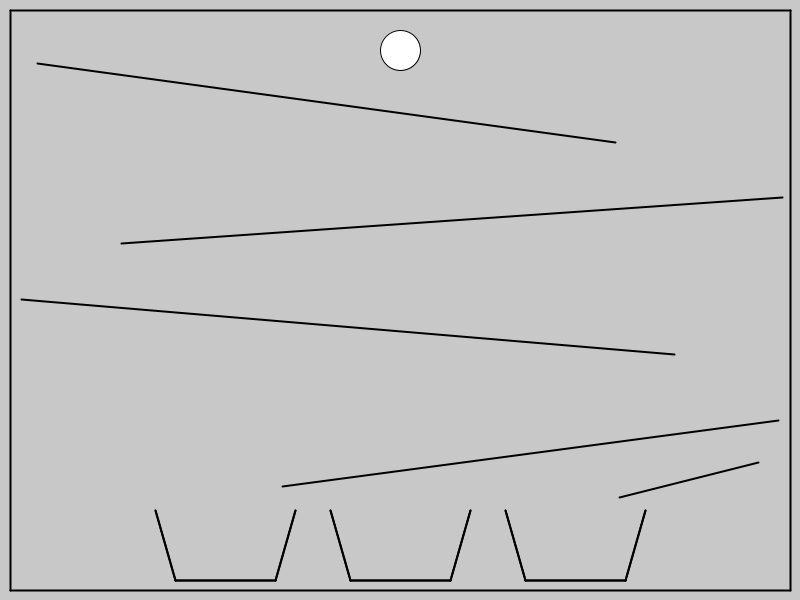}
    \end{borderedimage}%
    \hfill
    \begin{borderedimage}[baselineorange]
        \includegraphics[width=0.19\linewidth]{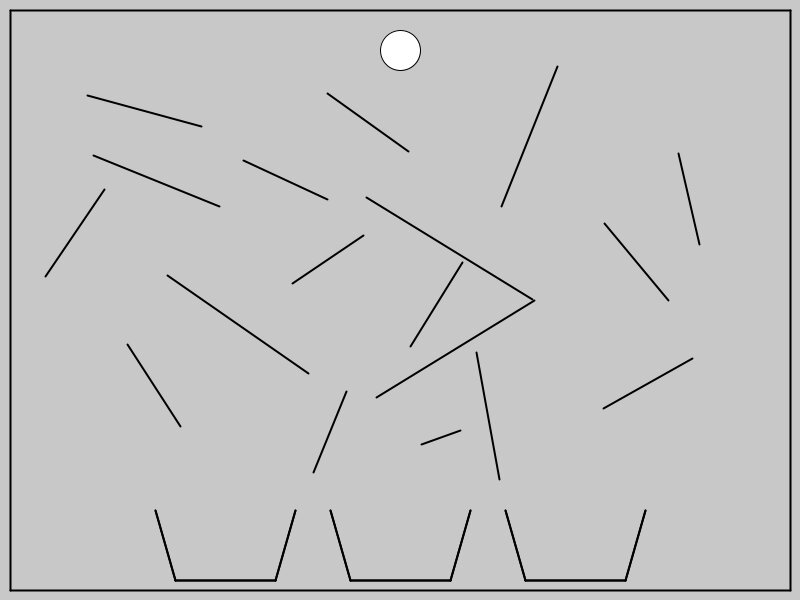}
    \end{borderedimage}%
    \hfill
    \begin{borderedimage}[baselineorange]
        \includegraphics[width=0.19\linewidth]{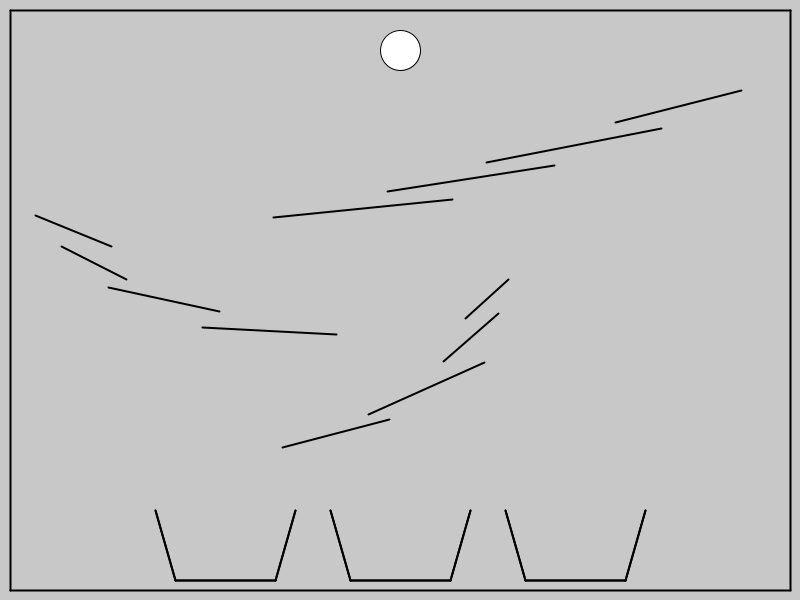}
    \end{borderedimage}%
    \hfill
    \begin{borderedimage}[baselineorange]
        \includegraphics[width=0.19\linewidth]{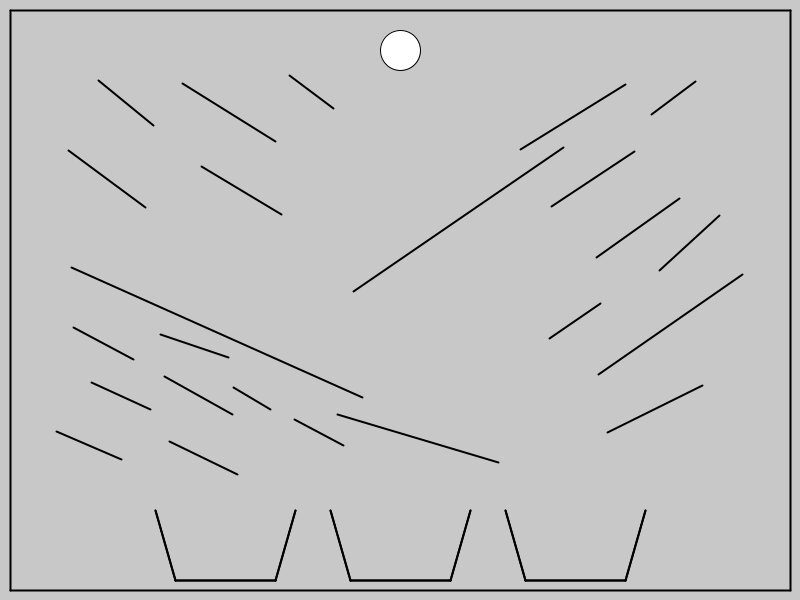}
    \end{borderedimage}%
    \vspace{1mm}

    \begin{borderedimage}[vipeblue]
        \includegraphics[width=0.19\linewidth]{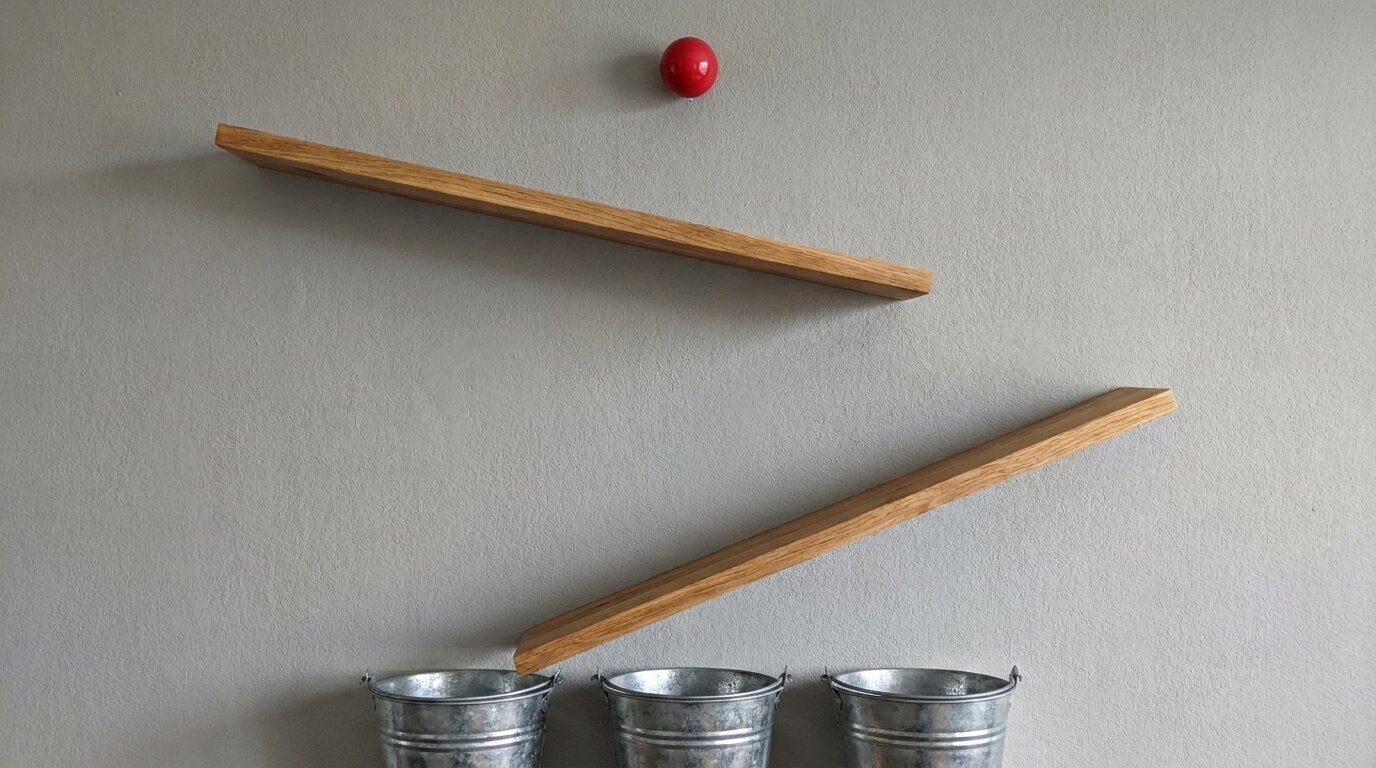}
    \end{borderedimage}%
    \hfill
    \begin{borderedimage}[vipeblue]
        \includegraphics[width=0.19\linewidth]{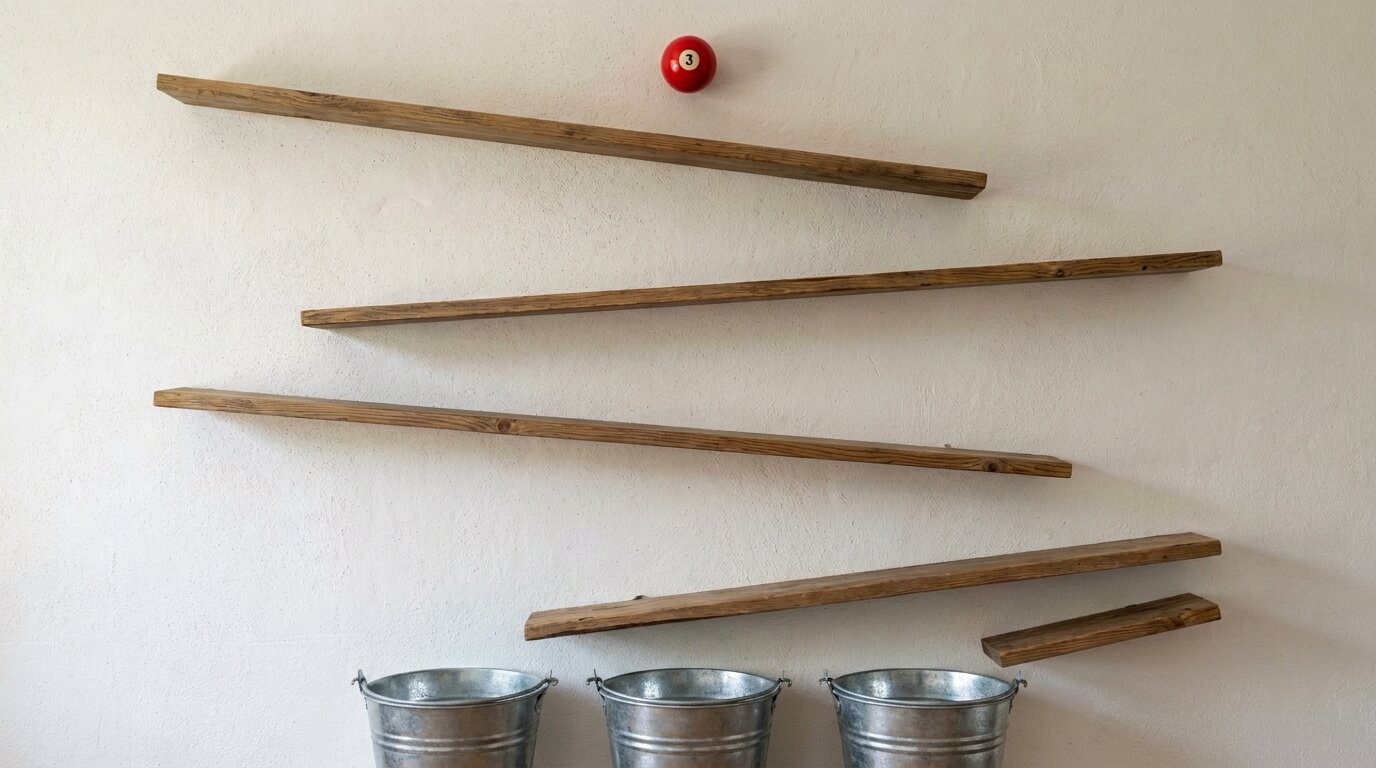}
    \end{borderedimage}%
    \hfill
    \begin{borderedimage}[vipeblue]
        \includegraphics[width=0.19\linewidth]{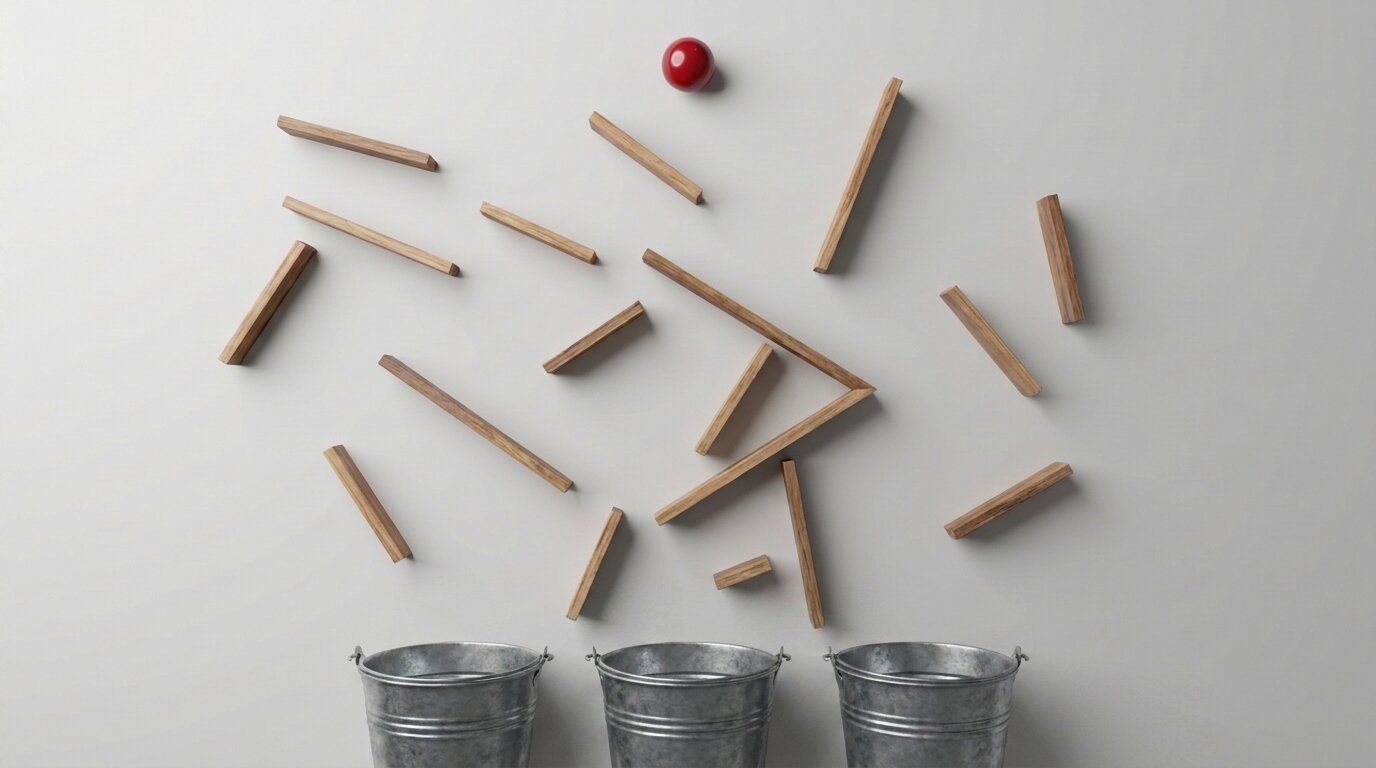}
    \end{borderedimage}%
    \hfill
    \begin{borderedimage}[vipeblue]
        \includegraphics[width=0.19\linewidth]{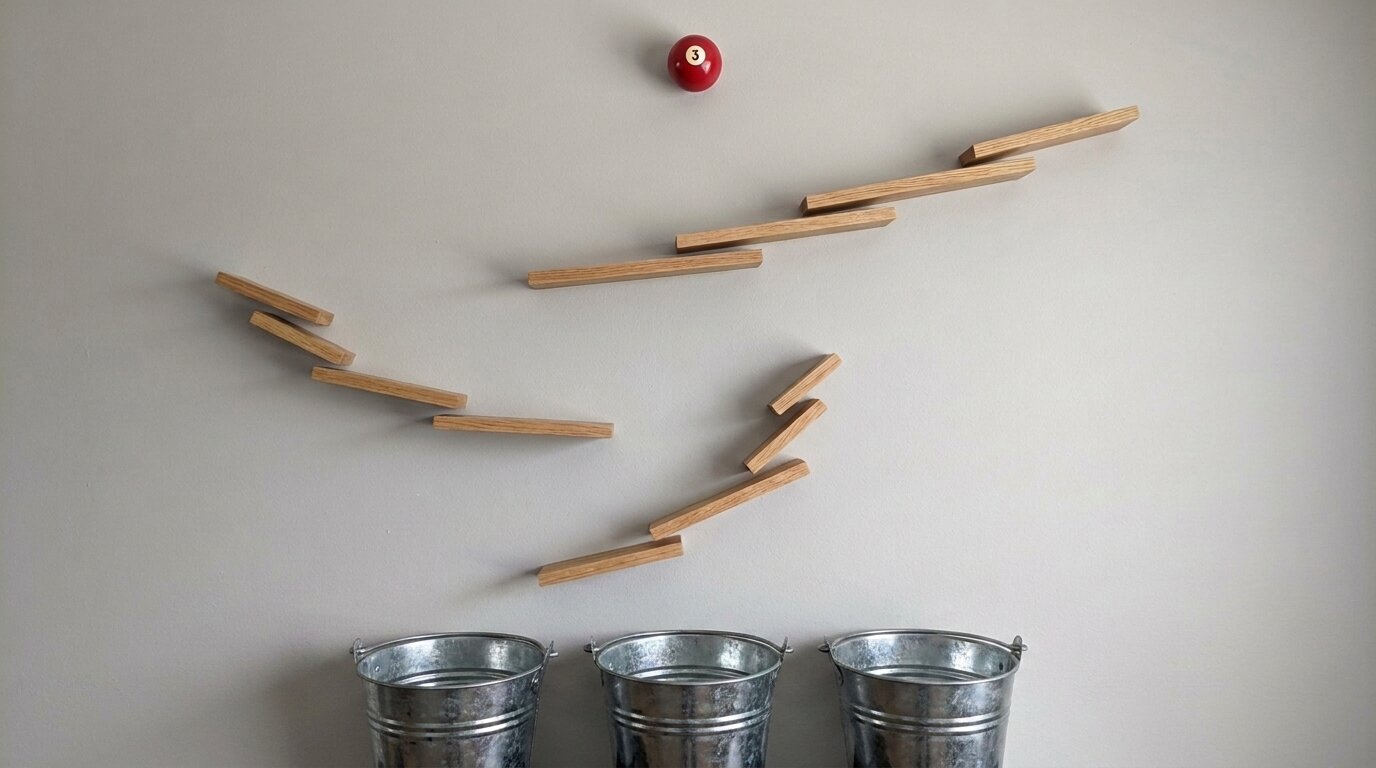}
    \end{borderedimage}%
    \hfill
    \begin{borderedimage}[vipeblue]
        \includegraphics[width=0.19\linewidth]{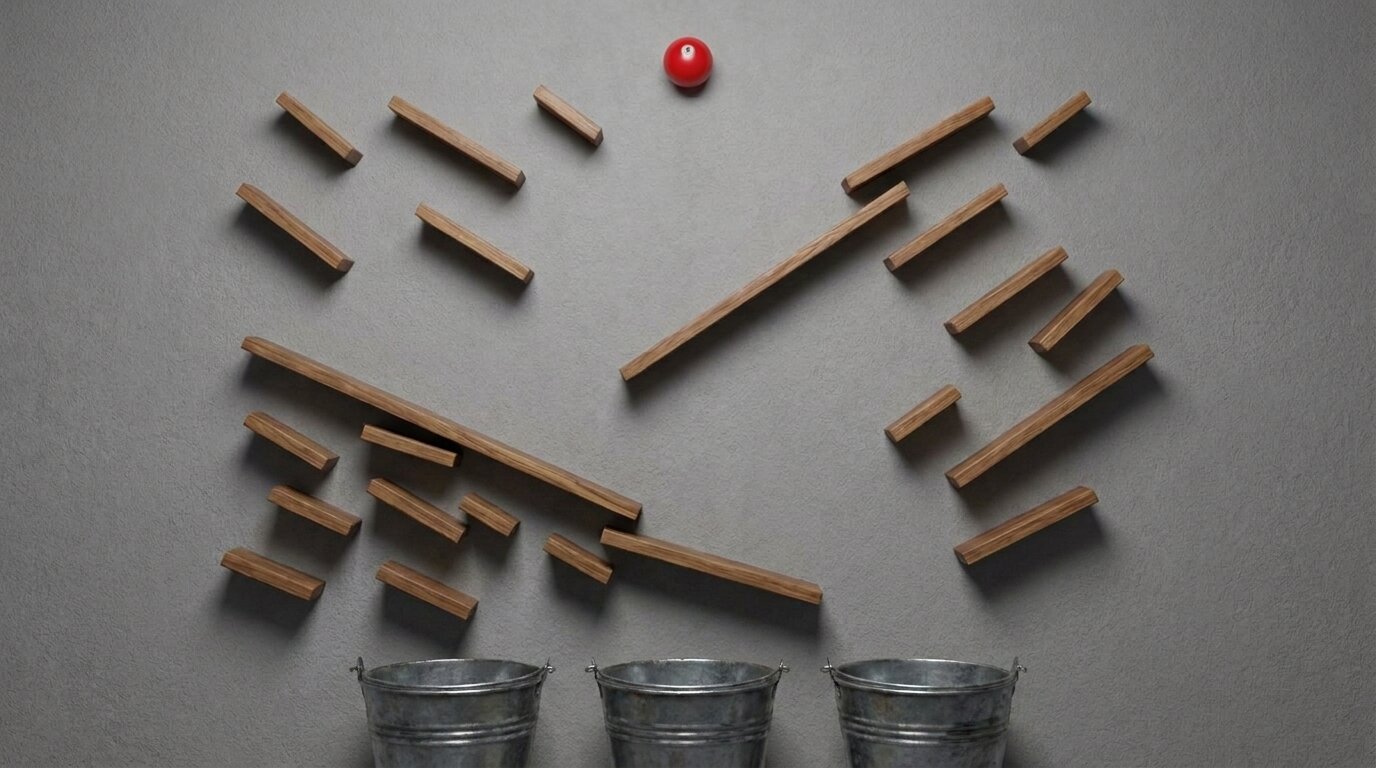}
    \end{borderedimage}%
    \vspace{1mm}

    \begin{borderedimage}[ablationgrey]
        \includegraphics[width=0.19\linewidth]{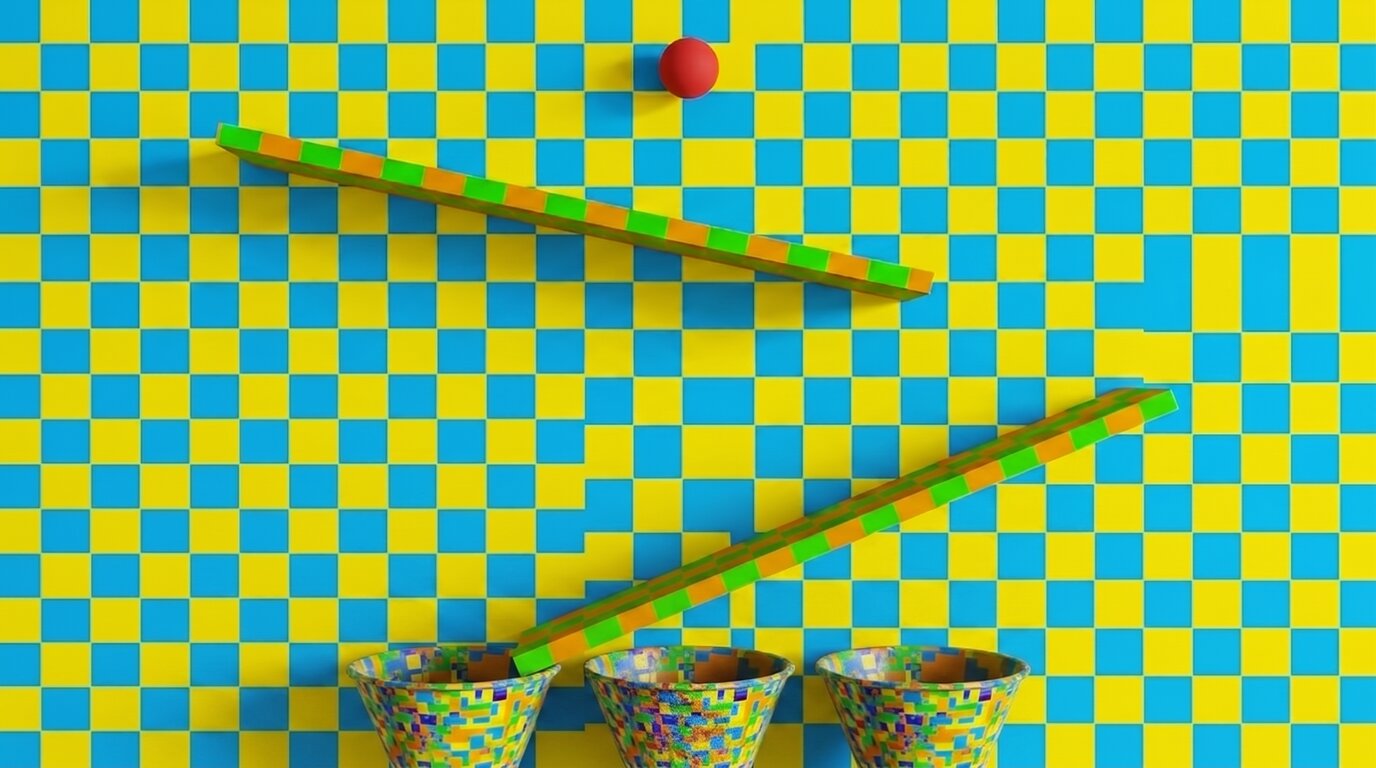}
    \end{borderedimage}%
    \hfill
    \begin{borderedimage}[ablationgrey]
        \includegraphics[width=0.19\linewidth]{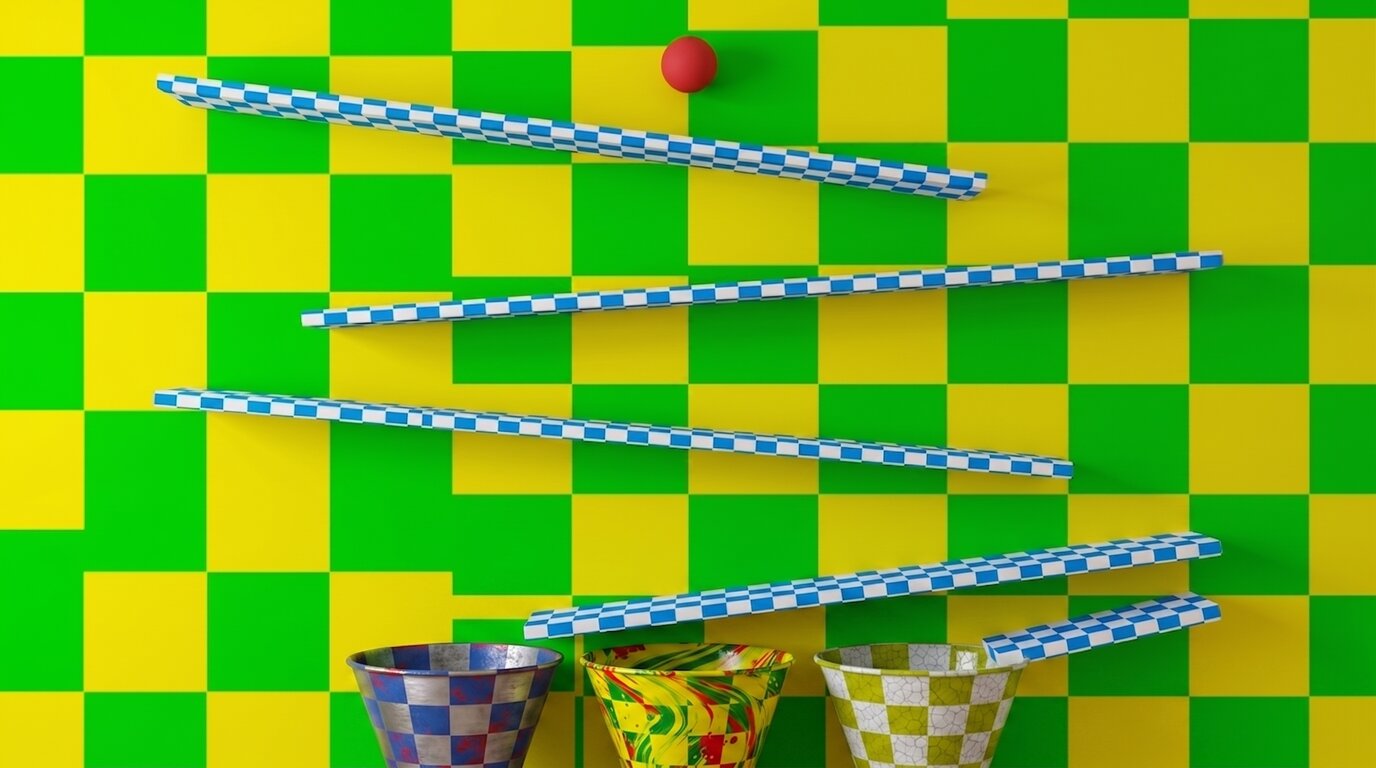}
    \end{borderedimage}%
    \hfill
    \begin{borderedimage}[ablationgrey]
        \includegraphics[width=0.19\linewidth]{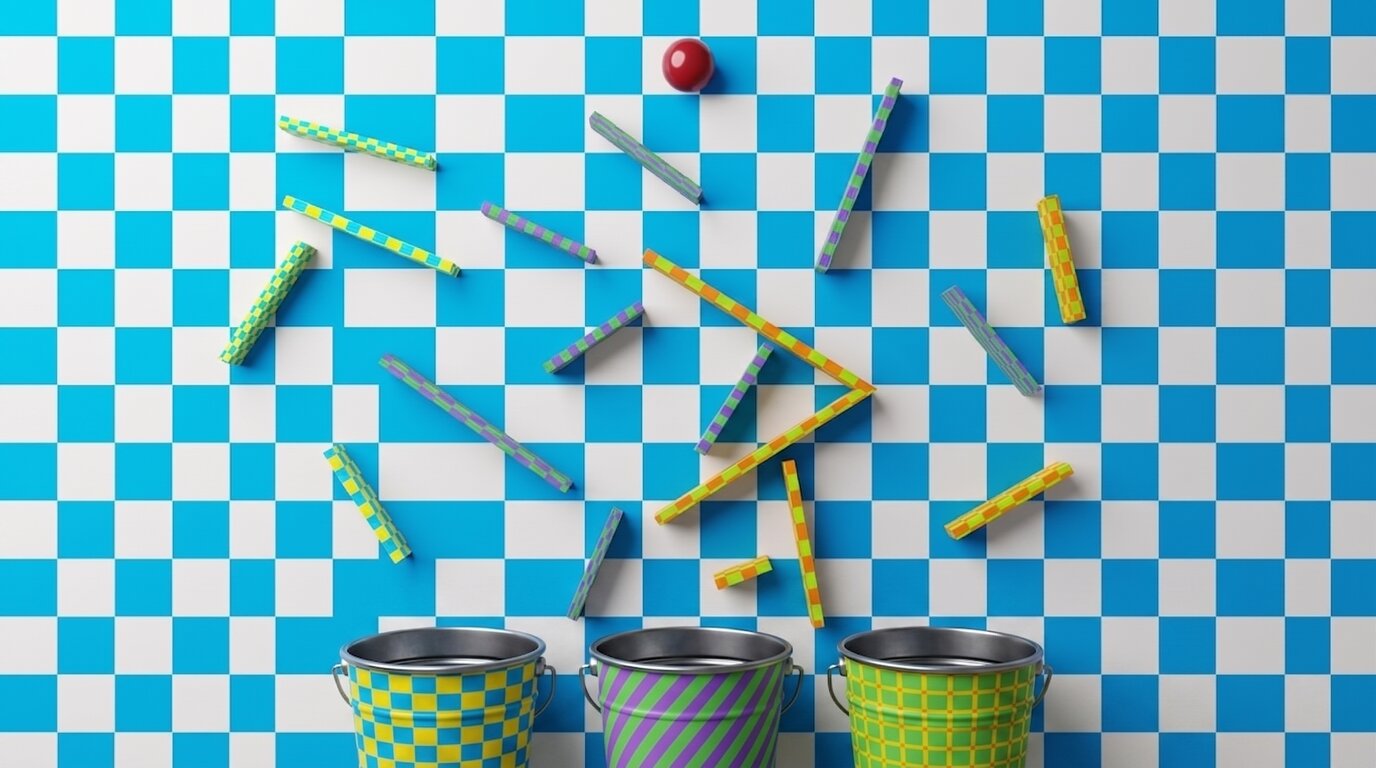}
    \end{borderedimage}%
    \hfill
    \begin{borderedimage}[ablationgrey]
        \includegraphics[width=0.19\linewidth]{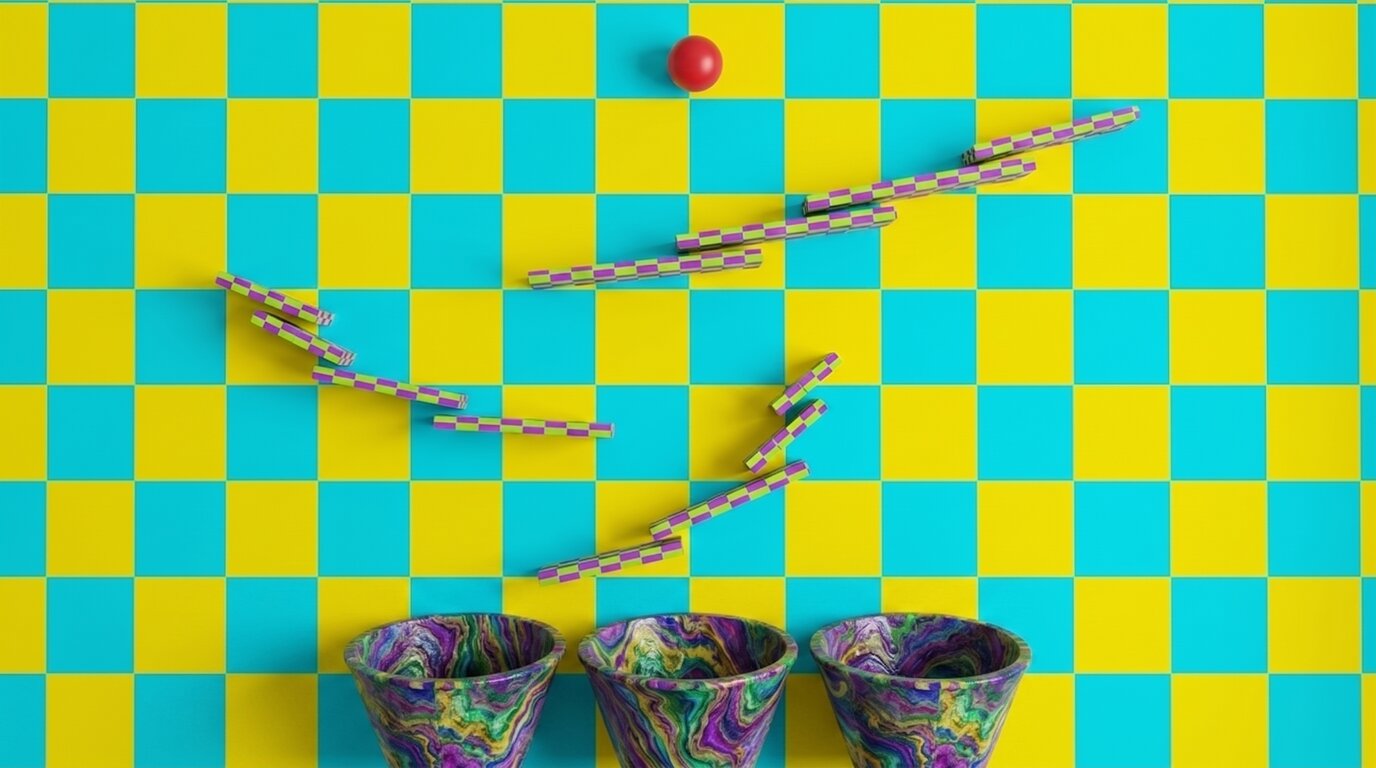}
    \end{borderedimage}%
    \hfill
    \begin{borderedimage}[ablationgrey]
        \includegraphics[width=0.19\linewidth]{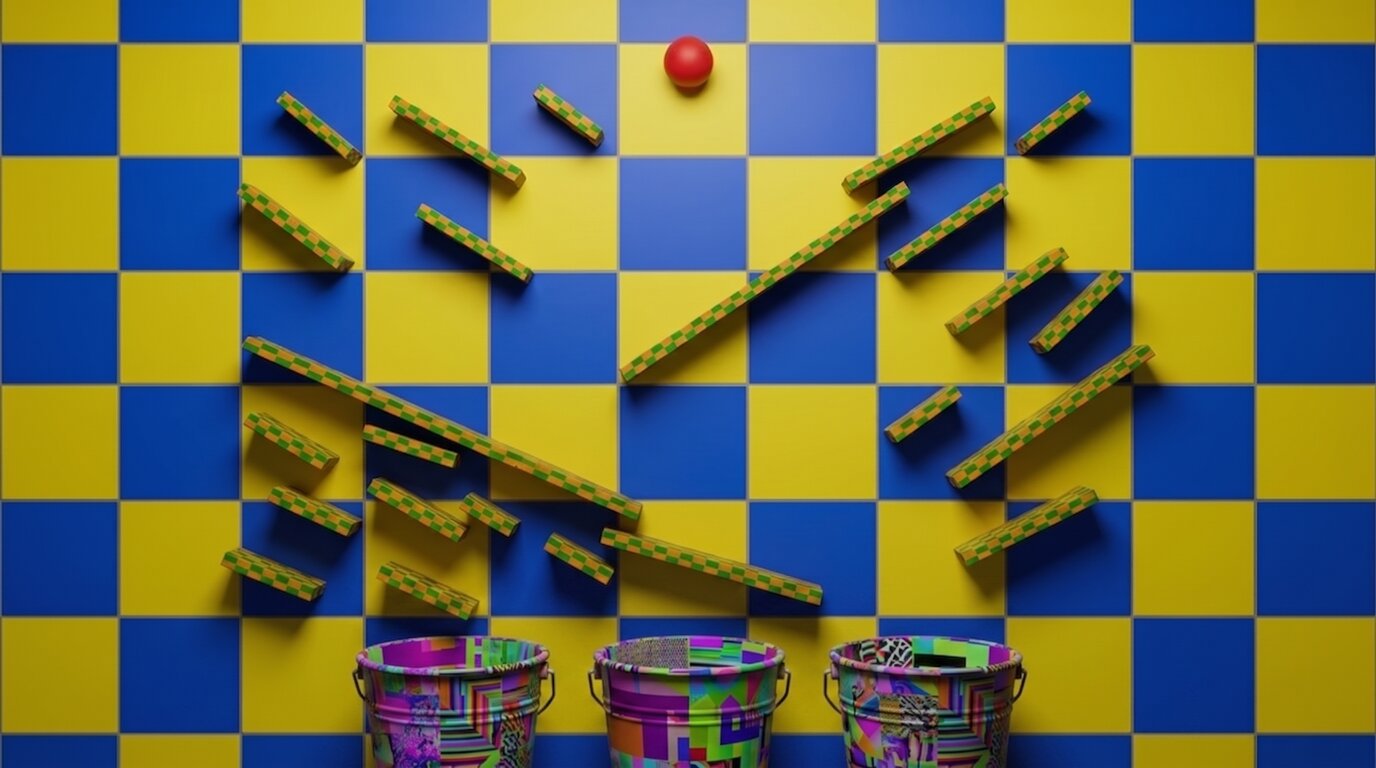}
    \end{borderedimage}%
    \caption{\textbf{Visual prompt engineering improves physics reasoning on \taskname{VPCT}.} The barplots visualize performance of video models on a ball-and-ramp dataset (goal: predict the bucket that the ball will fall into) across three conditions. Samples from each condition are visualized below. \textbf{Top row}: original sketch-like dataset samples (realism \ding{55} 3D \ding{55}). \textbf{Middle row}: visual prompt engineering samples (realism \ding{51} 3D \ding{51}). \textbf{Bottom row}: unnatural textures ablation (realism \ding{55} 3D \ding{51}). Each barplot represents the average across 10 runs of the entire dataset (a total of 1,000 samples); whiskers indicate 95\% confidence intervals.}
    \label{fig:vpct}
\end{figure*}

We start our investigation with visual reasoning on the \taskname{VPCT}~\citep{camelcase2025vpct} dataset. It requires predicting whether a white ball will drop into buckets 1, 2, or 3 after rolling down a series of ramps (represented as black lines). Samples are shown in~\cref{fig:vpct}. The dataset is interesting because it requires models to reason through a challenging visual setup while taking into account which physical forces apply. At the same time, the problem can be represented in its original sketch-style setup, or alternatively in a photorealistic way. We here ask whether these choices influence how well video models perform on this task. To this end, we take the original sketch dataset and perform visual prompt engineering (VIPE) to make it look photorealistic. Until recently, changing the style of an entire dataset would have been prohibitively expensive, requiring days of work by a professional. Fortunately, we can now simply preprocess the dataset with an image editing model.
For \taskname{VPCT}, we instantiate the editor $E$ with Nano~Banana~Pro~\citep{NanoBananaPro2026} and propose an edit instruction $t_\text{edit}$ ourselves: \textit{``Transform this into a realistic photograph, without changing the camera perspective that is facing the wall dead on. Replace the white circle with a small red billiard ball. Replace each black line by a wooden shelf board of exactly the same length, orientation, and placement. Do not add anything else.''} Filtering is performed with Gemini~3.1~Pro~\citep{Gemini3p1Pro2026}, selecting the best out of $m=5$ proposal variants. The resulting dataset was verified by the authors to ensure that visual prompt engineering did not accidentally make the problem easier than the original dataset. We expect that as image editing models become ever more faithful, filtering / verification steps will no longer be required.

After completing the visual prompt engineering step, we can now compare video model performance before and after VIPE. Detailed prompts and evaluation details are in~\cref{app:vpct}; essentially the video model is asked to simulate the ball's trajectory, and an autorater detects the bucket in which the ball lands. The results are shown in~\cref{fig:vpct}: Overall, both open- and closed-source models benefit from increased realism via visual prompt engineering. For example, performance of Wan2.2 is close to chance (=random guessing) in the baseline sketch condition for both the TI2V and I2V checkpoints and across English and Chinese prompts. Since Wan performance may differ between English and Chinese text prompts, we evaluate across both languages. TI2V is a 5B checkpoint that can perform both T2V (not used here) and I2V; while the I2V checkpoint is a MoE model with 14B active parameters~\citep{wan2025}, see also~\cref{app:wan}. With visual prompt engineering, Wan performes above chance across all conditions. For Veo~3.1~\citep{Veo2025}, which already starts with higher baseline performance, the effect of visual prompt engineering is even more pronounced, taking it from 41.3\% to 59.3\%. Finally, Omni Flash~\citep{Omni2026} improves from 56.3\% to 67.5\% via VIPE. In short, visual prompt engineering improves video reasoning on this task---similar to how text-based prompt engineering improves language model performance on language tasks. Visual prompt engineering is, in many ways, the natural and obvious extension of text-based prompt engineering. Given its effectiveness, we find it all the more surprising that it isn't common practice for video models yet.

\takeaway{Similar to how \emph{text-based} prompt engineering can improve language model performance on language tasks, \emph{visual} prompt engineering can systematically improve video model performance on visual reasoning tasks.}

\paragraph{Ablation: unnatural textures} In order to find out whether it is indeed the photorealism that improves video model reasoning, or alternatively the difference between 2D sketch and 3D-looking input, we performed an ablation where samples have completely unrealistic textures (=no realism), while still having depth and therefore ``looking 3D''. In~\cref{fig:vpct}, most models do not show a significant difference between the baseline and unnatural texture condition, suggesting that 3D alone is insufficient and it is indeed visual realism that drives the performance improvement.

\FloatBarrier
\section{How can visual prompt engineering be automated?}
\label{sec:automated-vipe}
The \taskname{VPCT} results from~\cref{sec:vpct} demonstrate that a good visual prompt can systematically improve video reasoning performance. However, we may not always have a good prior intuition on what constitutes a good prompt, and the visual prompt search space is vast. Here, we therefore test whether VIPE can be automated by comparing two approaches, freeform ideation and step-by-step edits (ACE). To be clear: our main contribution is the insight that video model reasoning improves with visual prompt engineering (VIPE). \emph{How} VIPE is instantiated is up to the community, will likely change as models get better, and most importantly, builds on standard, established components. We regard the fact that existing models and methods can be plugged in as an advantage, and of course don't claim any novelty regarding these existing techniques. Since video generation is currently expensive, we consistently perform all of the following experiments with Veo~3.1.

\begin{figure}[t]
    \centering
    \includegraphics[width=\linewidth]{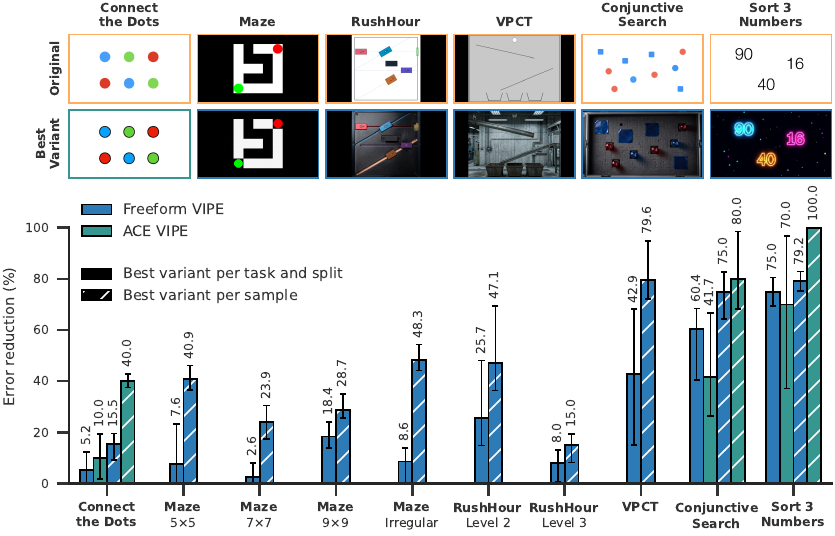}
    \caption{\textbf{Automated visual prompt engineering improves performance on a wide range of tasks}. Performance is plotted as error reduction (in percent) relative to the baseline of no visual prompt engineering. For example, if the baseline achieves 20\% accuracy and VIPE achieves 60\% accuracy, the error rate is halved (80\% errors $\rightarrow$ 40\% errors), resulting in a 50\% reduction. Solid bars show performance for the overall best variant per task and split (e.g., \taskname{Maze 3$\times$3}), out of $n=20$ total variants. Successful variants range from complex 3D renders (e.g., \taskname{RushHour} is re-contextualized as an anodized metal puzzle box) to small appearance changes (e.g., the \taskname{Maze} is interpreted as an arcade game screen recording with subtle pixel texture). Hatched bars show performance when letting an oracle pick the best variant per task sample. See~\cref{app:automated-vipe} for details and more results. Whiskers indicate 95\% confidence intervals. Example videos for each task and variant are on our \href{\projectpageurl}{project page}.}
    \label{fig:automated-vipe}
\end{figure}

\paragraph{Tasks}
We investigate this question on six visual reasoning tasks; an example of each is included in~\cref{fig:automated-vipe}. \taskname{VPCT} is introduced in~\cref{sec:vpct}. \taskname{Maze} includes four different kinds of mazes from~\citet{wiedemer2025video} with different complexity. We also use three other tasks from the same article: asking the model to \taskname{Connect the Dots} with equal color, \taskname{Sort 3 Numbers} by manipulating them in sequence, and performing \taskname{Conjunctive Search} by identifying target objects matching specific criteria in a cluttered scene. Finally, we test \taskname{RushHour}~\citep{zeller2026mentisoculi}, which asks the model to rearrange cars to free a path out of a congested parking lot. In total, we evaluate 18,160 videos for this experiment; details in~\cref{app:automated-vipe}. In order to score whether a given video successfully solves the task, we employ either VLM-based or deterministic autoraters. Details can be found in~\cref{sec:autorater-details,app:automated-vipe}.

\subsection{Freeform ideation by a VLM}
\label{subsec:freeform-ideation}
We start with a naive approach: generating prompt engineering ideas $t_\text{edit}$ with a VLM ideator. The VLM's general ideation prompt $t_\text{ideate}$ seeks to transfer the task to a new visual domain while preserving spatial layout and object correspondence. In practice, we found the resulting image edits to be faithful to this instruction, only changing the visual appearance of the images without altering the task. Here, we instantiate the VIPE ideator with Gemini~3.1~Pro~\citep{Gemini3p1Pro2026}, the image editor as Nano~Banana~2 (Gemini~3.1~Flash~Image), and the filtering model again as Gemini~3.1~Pro (selecting the best out of 5 attempts). The ideator does not receive feedback from downstream evaluation, i.e., operates open-loop. Full prompts and details are provided in~\cref{sec:freeform-ideation-prompts}. The results are shown in~\cref{fig:automated-vipe}. Across a wide range of tasks, we show that this simple freeform, open-loop visual prompt engineering approach successfully finds many alternative visual prompts that elicit better performance in Veo~3.1. Some task variants allow Veo~3.1 to achieve surprisingly large performance gains (reducing the error rate by over $75\%$ compared to the baseline of no visual prompt engineering), simply by altering the visual context in which the model's reasoning occurs.

\subsection{Step-by-step edits: Atomic Concept Editing (ACE)}
\label{subsec:ace}
An alternative, more structured approach for visual prompt engineering is editing the original image step-by-step, one concept at a time (e.g., original image $\rightarrow$ change background to blue $\rightarrow$ increase object size $\rightarrow$ ...). To this end, we adapt the ACE framework~\citep{kalibhat2026ace} to the visual prompt engineering setting. An atomic concept edit (ACE) is a simple edit that either adds, removes or replaces a single concept at a time. Prompt exploration is performed in a tree-like fashion: Starting from the root node (original image), ACEs are generated by Gemini~3.1~Pro to verbally describe the possible edits to the current image. The edits are then applied to the image, using the image editor Nano~Banana~2, to obtain a number of child-node images. For each of these child nodes, exploration again branches out in different directions. The ACE exploration stops when a pre-determined number of successful nodes are generated. In contrast to freeform ideation, ACE operates in a closed loop that iteratively adapts the edits according to autorater feedback. More details can be found in~\cref{app:ACE}.

Since ACE is expensive, we applied it only to three best/worst performing tasks: \taskname{Connect the Dots}, \taskname{Conjunctive Search} and \taskname{Sort 3 Numbers}. \cref{fig:automated-vipe} shows the results for ACE in shades of green. Overall, we observe systematic improvements over the ``no visual prompt engineering'' baseline, for a similar final performance to freeform VIPE. ACE's more structured edits yield more optimized per-sample variants, however, which lead to a 100\% reduction rate on \taskname{Sort 3 Numbers}.

Our analysis of the prompt mutations generated by ACE revealed several task-agnostic edit patterns that consistently improved model performance. One major pattern involved increasing realism of material properties to improve temporal object permanence, e.g., transforming abstract elements into tangible 3D objects using descriptors like ``matte clay'' or ``textured chalkboard''. Another major strategy relied on high-contrast target isolation to guide the model's attention, e.g., utilizing solid dark backgrounds or applying thick borders. A final distinct trend showed mutations moving toward semantic simplification, e.g., replacing formal vocabulary and abstract modifiers with direct action verbs such as replacing ``same-colored'' with ``matching'' or ``the two'' with ``both''. Furthermore, we observed that these strategies are rarely applied in isolation; rather, successful mutations frequently combine multiple patterns. An in-depth analysis is presented in ~\cref{app:ACE}.

\takeaway{Visual prompt engineering does not need to rely on human intuition: it can be successfully automated in different ways, from freeform ideation with a VLM to structured, step-by-step concept edits.}

\section{Can visual prompt engineering be used for test-time scaling?}
\label{sec:test-time-scaling}
Language models benefit from scaling during pre-training, post-training, and test-time. While video models are being scaled along the first two axes, test-time scaling has received limited attention. Notable exceptions are \citet{li2026thinking}, characterizing the number of generated frames as a form of test-time compute; and \citet{newman2026video} who use a heuristic on multiple early-decoded videos to select the most promising attempt.

\begin{wrapfigure}{r}{0.5\textwidth}
    \centering
    \includegraphics[width=\linewidth]{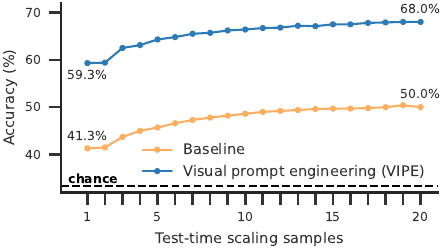}
    \caption{
        \textbf{Visual prompt engineering can be combined with test-time scaling to further improve performance.} For Veo~3.1 on \taskname{VPCT}, majority voting across 20 independent runs improves performance.
    }
    \label{fig:vpct_tt_scaling}
\end{wrapfigure}

How does VIPE compare to test-time scaling? \cref{fig:vpct_tt_scaling} shows Veo~3.1's base performance on \taskname{VPCT} improving with self-consistency (majority voting)~\citep{wang2022self} across up to 20 test-time samples: from 41.3\% to 50.0\% (+8.7 percentage points). In comparison, a single test-time sample on the engineered visual prompts from~\cref{sec:vpct} already achieves 59.3\% (+18 percentage points). While this indicates that VIPE is more effective than test-time scaling via self-consistency, we need not pick one or the other: self-consistency scales just as well on top of the engineered visual prompts as on the original prompts, yielding a compounded 68.0\% (+27.7 percentage points) with 20 test-time samples.

\begin{figure}[h]
    \centering
    \includegraphics[width=\linewidth]{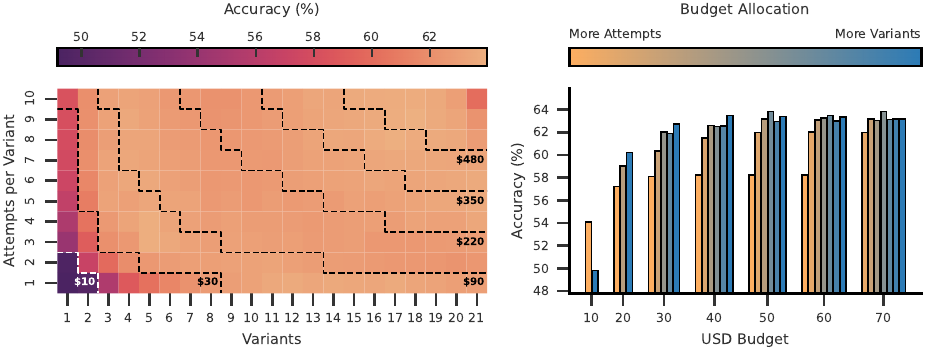}
    \caption{\textbf{VIPE variants are more cost-effective than additional attempts on the same prompt}. \textbf{Left}. Accuracy on \taskname{VPCT} with test-time scaling via self-consistency (majority voting) as a function of the number of VIPE variants (x axis) and the number of attempts (generated videos) per variant (= traditional test-time scaling, y axis). Budget isolines indicate total cost assuming USD~3.20 per video and USD~0.40 per VIPE with $m=5$ edits~\citep{google_gemini_api_pricing}; see~\cref{app:cost} for details. \textbf{Right}. For a given budget, it is generally better to allocate that budget to maximize the number of VIPE variants (right-most blue bars, corresponding to cells at $y=1$ in the heatmap) as opposed to generating more attempts with the same visual prompt. Similar results for \taskname{Maze} and \taskname{RushHour} in~\cref{app:test-time-scaling}.}
    \label{fig:budget-tradeoffs}
\end{figure}

However, the above experiment assumes that we already know an effective visual prompt variant. If that's not the case, can (automated) visual prompt engineering itself be an effective test-time scaling approach? \cref{fig:budget-tradeoffs}~Left shows performance when majority-voting over multiple generated videos \textit{and} VIPE variants for a fixed, given budget derived from API inference cost. Due to VIPE's effectiveness and low cost compared to video generation (even when accounting for $m$ initial proposals and filtering, a VIPE variant currently costs only $\nicefrac{1}{8}$ of a video, see~\cref{app:cost}), performance is generally optimized by exploring as many VIPE variants as possible. This is also shown in \cref{fig:budget-tradeoffs}~Right: it is more cost-effective to explore more VIPE variants than to generate more videos per variant.

\takeaway{Visual prompt engineering is a more cost-effective test-time scaling strategy than simply generating more videos; both approaches can be combined for best results.}

\FloatBarrier
\section{Is language or visual prompt engineering more effective for video models?}
\label{sec:images-vs-words}
We showed that visual prompt engineering can effectively improve video model reasoning. However, video models receive both a visual prompt and a text prompt as input, and both are natural levers for influencing model behavior. In the following, we directly compare how text prompt engineering vs.~visual prompt engineering influence a video model's ability to solve the same task.

\paragraph{Text prompt engineering for video models}
We adapt the prompt engineering ideator from~\cref{subsec:freeform-ideation} to propose alternative text prompts instead of image edits, keeping all source images from a baseline task fixed. Similar to the image edit ideator, we instruct the text ideator to propose text prompts that preserve the task structure. For example, if the original prompt is ``The two blue balls begin to glow'', the ideator may propose ``The two blue balls shatter into tiny fragments'' instead. We then compare how well these proposed alternative text prompts and alternative image prompts elicit better reasoning performance. The comparison is performed on four challenging datasets: \taskname{VPCT}, \taskname{Connect the Dots}, and \taskname{Conjunctive Search}, and \taskname{Sort 3 Numbers}. Besides \taskname{VPCT}, Veo~3.1 achieves a baseline accuracy of around 3--4\% on the other tasks, making them suitable for measuring improvement headroom. See~\cref{app:freeform_ideation} for full details.

\begin{figure*}[!ht]
    \centering
    \scriptsize
    \setlength{\tabcolsep}{3pt}
    \begin{tblr}{
        colspec = {Q[c,m] Q[j,m,0.2\linewidth] Q[j,m,0.2\linewidth] Q[j,m,0.2\linewidth] Q[c,m,0.3\linewidth]},
        colsep = 4pt,
        row{1} = {halign = c},
    }
    & \textbf{Baseline} & \textbf{Best Text Variant} & \textbf{Best Image Variant} & \textbf{All Variants} \\
    \midrule

    \rotatebox[origin=c]{90}{\textbf{\taskname{VPCT} (subset)}}
        & \begin{adjustbox}{valign=m}
            \begin{minipage}{\linewidth}
                \begin{borderedimage}[baselineorange]
                    \includegraphics[width=\linewidth]{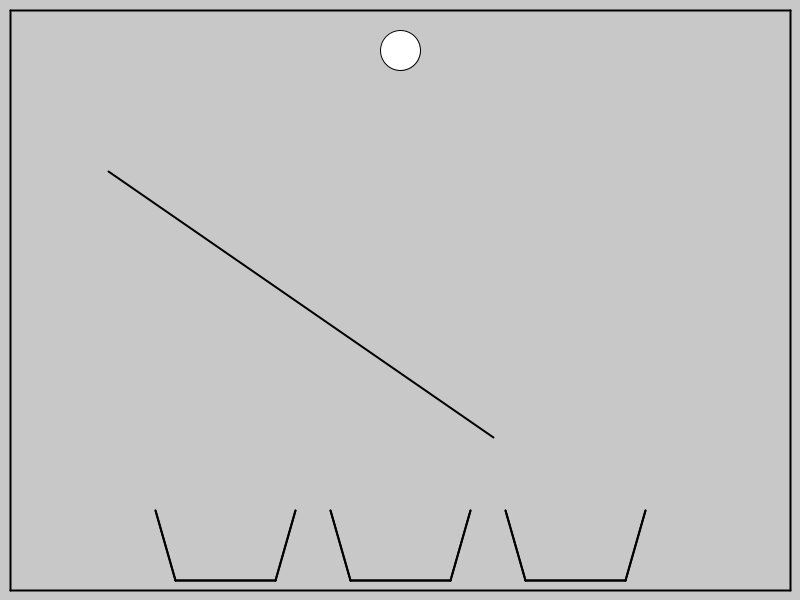}
                \end{borderedimage}\par
                The ball moves down in a physically plausible way, sliding down the obstacles, and ends up in one of the three containers at the bottom.\strut\hfill\textbf{57\%}
            \end{minipage}
        \end{adjustbox}
        
        & \begin{adjustbox}{valign=m}
            \begin{minipage}{\linewidth}
                \begin{borderedimage}[baselineorange]
                    \includegraphics[width=\linewidth]{figures/image_vs_words/ideator_3p1pro/qualitative/vpct_baseline.jpg}
                \end{borderedimage}\par
                \textcolor{vipeblue}{Animate the white circle as a hailstone falling through slanted gutters. It must roll down the black lines under gravity and drop into one of the bottom basins.}\strut\hfill\textbf{62\%}
            \end{minipage}
        \end{adjustbox}
        
        & \begin{adjustbox}{valign=m}
            \begin{minipage}{\linewidth}
                \begin{borderedimage}[vipeblue]
                    \includegraphics[width=\linewidth]{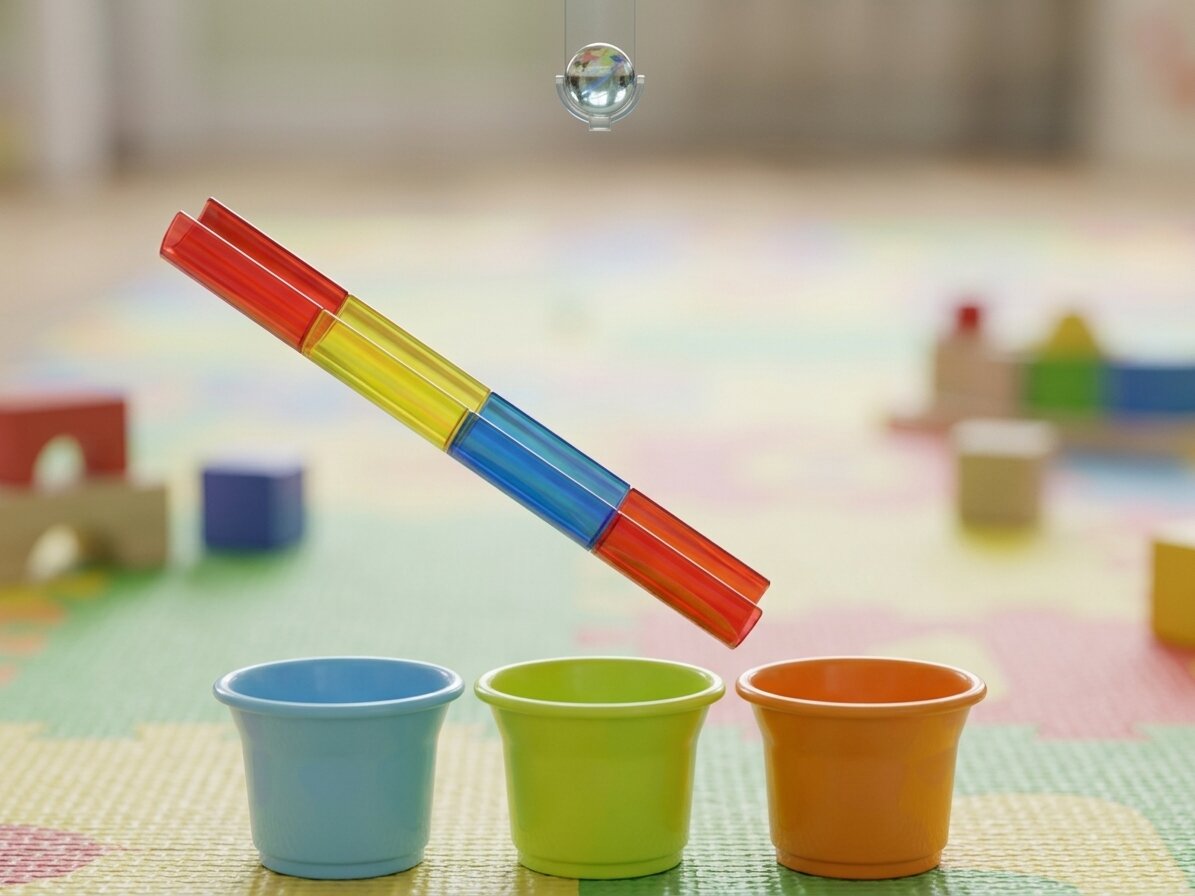}
                \end{borderedimage}\par
                The ball moves down in a physically plausible way, sliding down the obstacles, and ends up in one of the three containers at the bottom.\strut\hfill\textbf{73\%}
            \end{minipage}
        \end{adjustbox}
        
        & \includegraphics[width=\linewidth, valign=m]{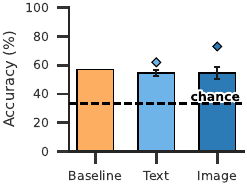} \\
    \midrule

    \rotatebox[origin=c]{90}{\textbf{\taskname{Conjunctive Search}}}
        & \begin{adjustbox}{valign=m}
            \begin{minipage}{\linewidth}
                \begin{borderedimage}[baselineorange]
                    \includegraphics[width=\linewidth]{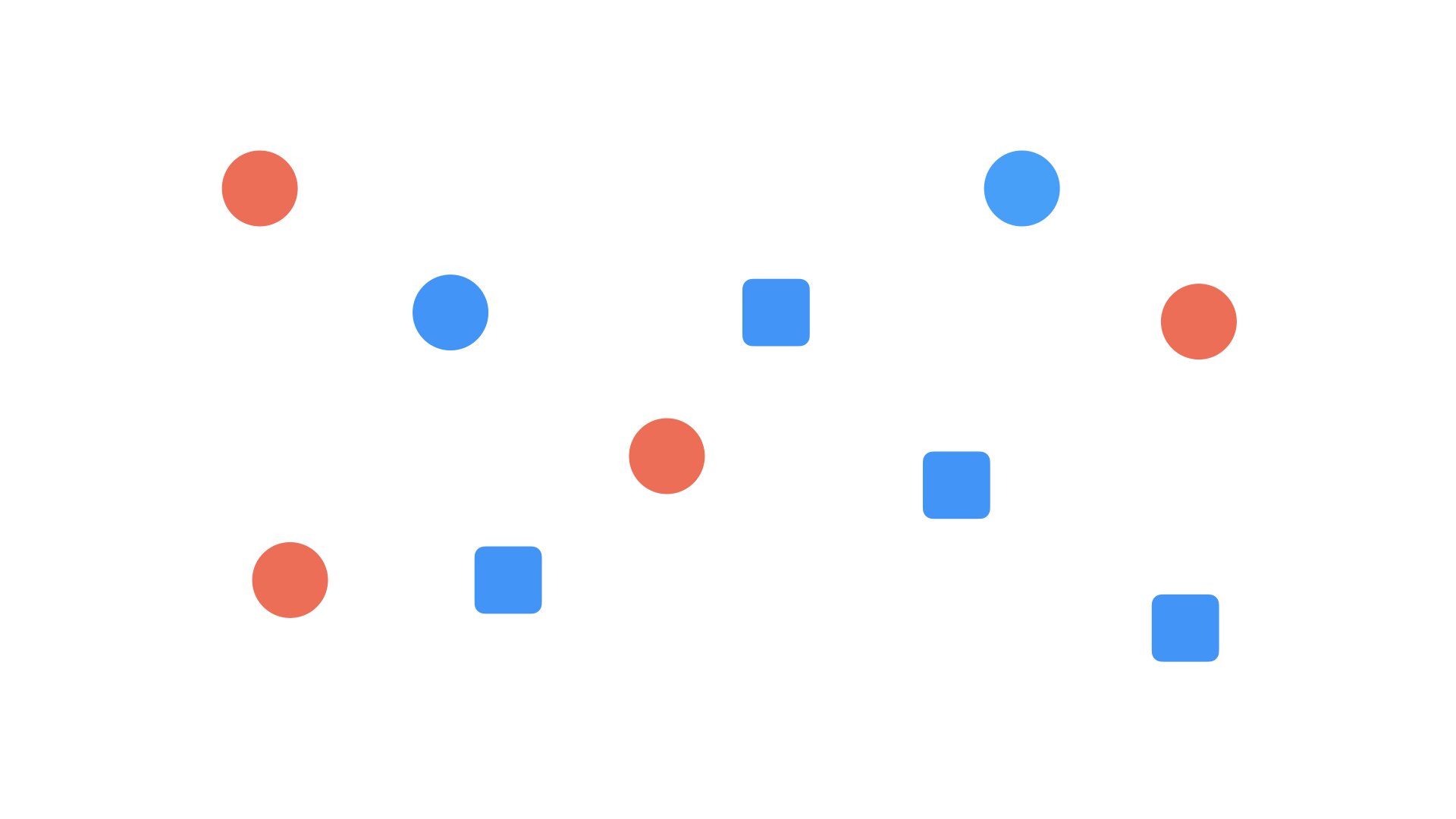}
                \end{borderedimage}\par
                The two blue balls begin to glow.\strut\hfill\textbf{4\%}
            \end{minipage}
        \end{adjustbox}
        
        & \begin{adjustbox}{valign=m}
            \begin{minipage}{\linewidth}
                \begin{borderedimage}[baselineorange]
                    \includegraphics[width=\linewidth]{figures/image_vs_words/ideator_3p1pro/qualitative/conjunctive_search_baseline.jpg}
                \end{borderedimage}\par
                The two blue balls \textcolor{vipeblue}{wobble and jiggle [...]}\strut\hfill\textbf{12\%}
            \end{minipage}
        \end{adjustbox}

        & \begin{adjustbox}{valign=m}
            \begin{minipage}{\linewidth}
                \begin{borderedimage}[vipeblue]
                    \includegraphics[width=\linewidth]{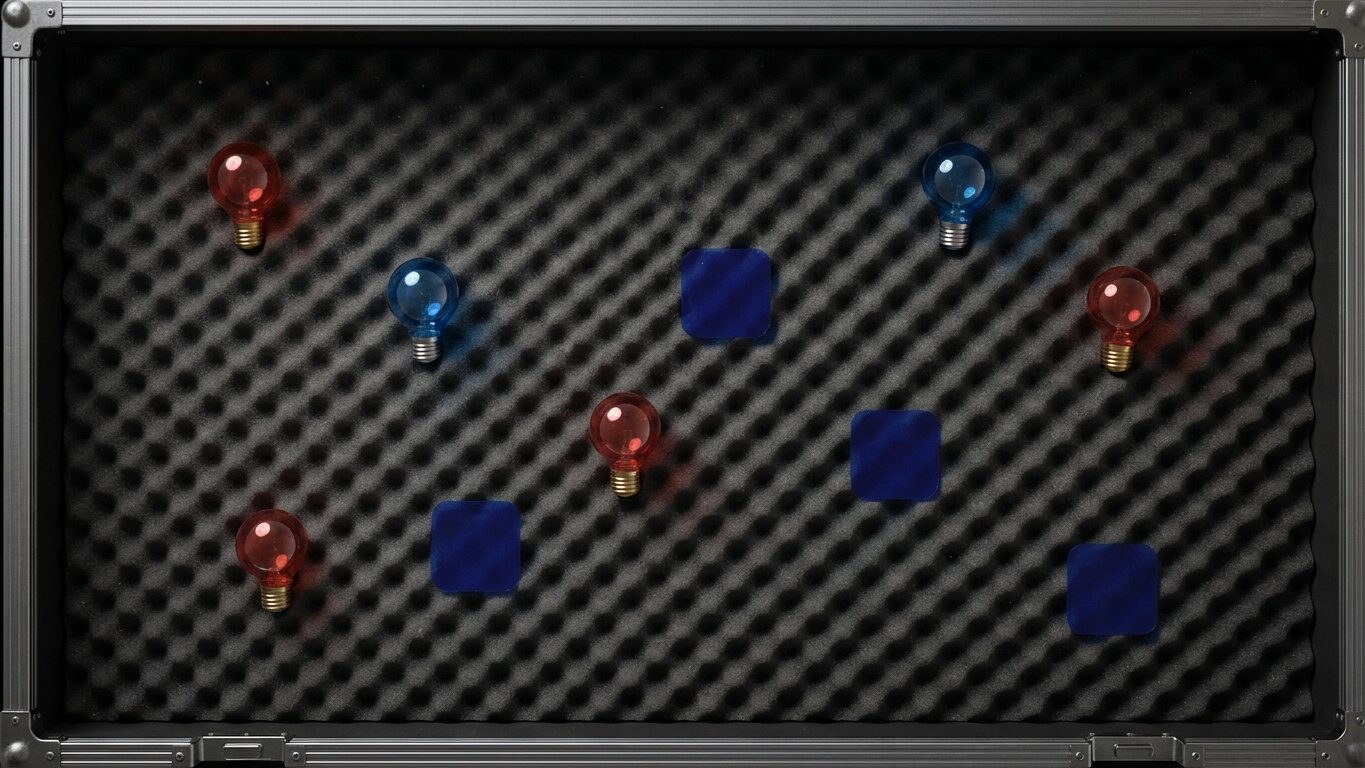}
                \end{borderedimage}\par
                The two blue balls begin to glow.\strut\hfill\textbf{62\%}
            \end{minipage}
        \end{adjustbox}
        
        & \includegraphics[width=\linewidth, valign=m]{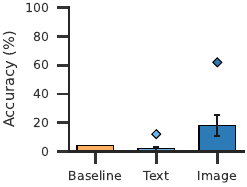} \\
    \midrule

    \rotatebox[origin=c]{90}{\textbf{\taskname{Sort 3 Numbers}}}
        & \begin{adjustbox}{valign=m}
            \begin{minipage}{\linewidth}
                \begin{borderedimage}[baselineorange]
                    \includegraphics[width=\linewidth]{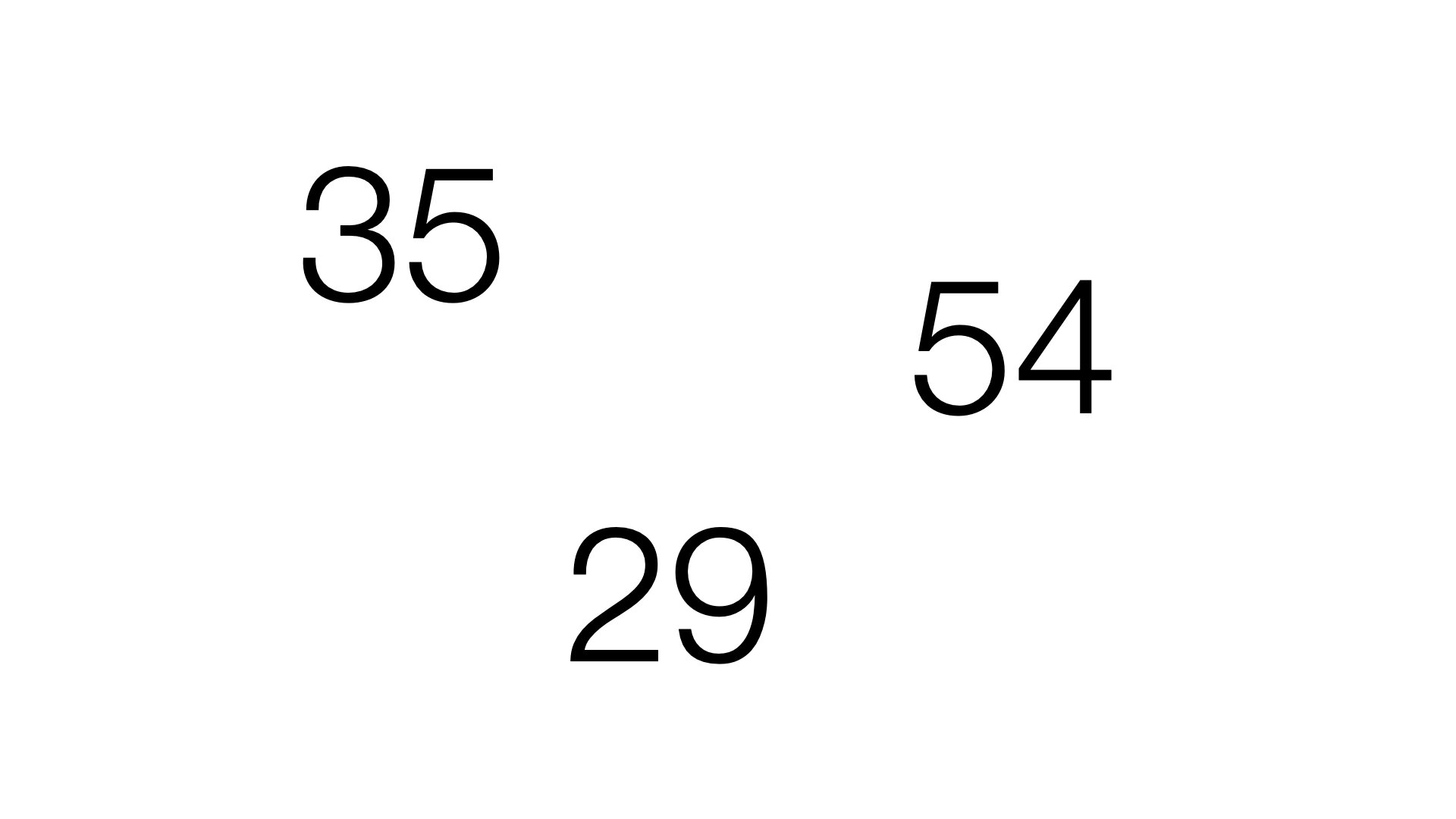}
                \end{borderedimage}\par
                The numbers pop and disappear one at a time, in numeric order, starting from the smallest one.\strut\hfill\textbf{4\%}
            \end{minipage}
        \end{adjustbox}
        
        & \begin{adjustbox}{valign=m}
            \begin{minipage}{\linewidth}
                \begin{borderedimage}[baselineorange]
                    \includegraphics[width=\linewidth]{figures/image_vs_words/ideator_3p1pro/qualitative/sort_3num_baseline.jpg}
                \end{borderedimage}\par
                \textcolor{vipeblue}{One by one, in numeric order [...] each number morphs into a solid five-pointed star.}\strut\hfill\textbf{86\%}
            \end{minipage}
        \end{adjustbox}
        
        & \begin{adjustbox}{valign=m}
            \begin{minipage}{\linewidth}
                \begin{borderedimage}[vipeblue]
                    \includegraphics[width=\linewidth]{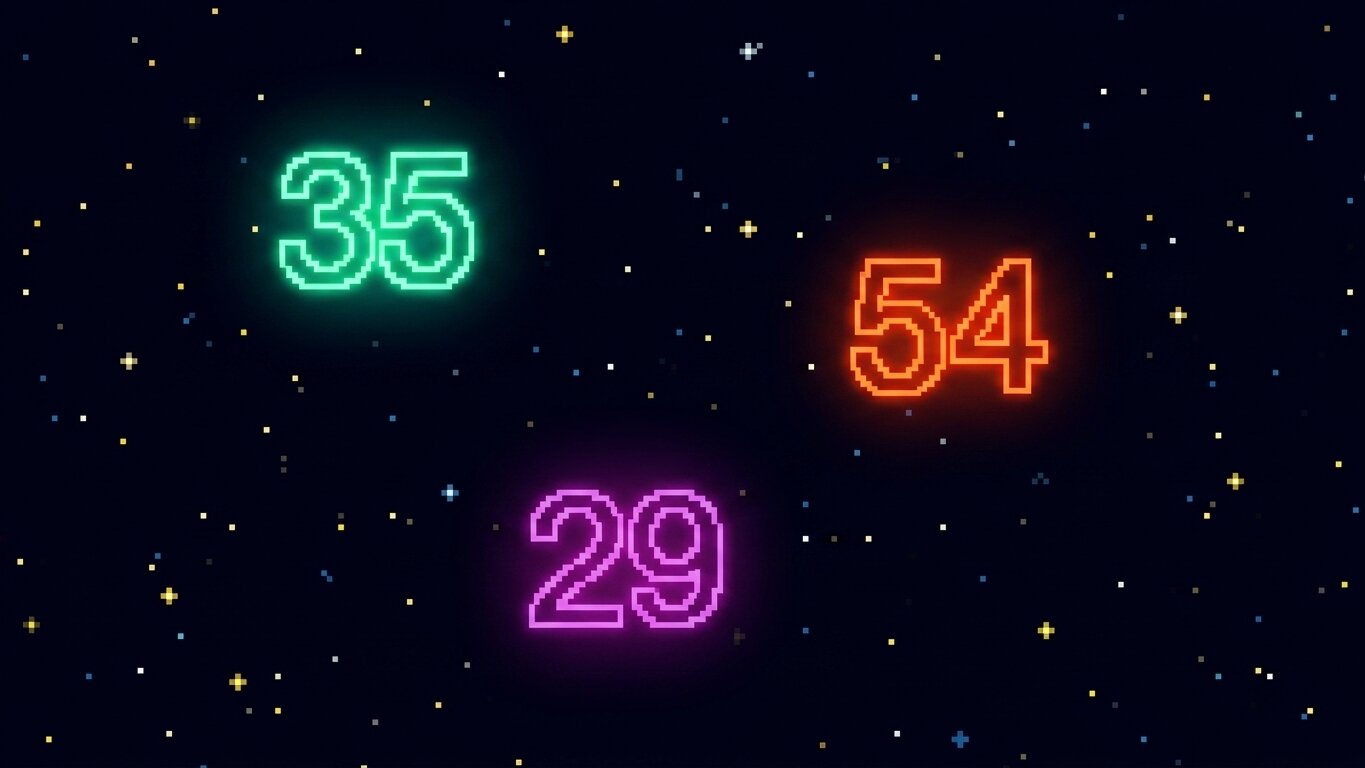}
                \end{borderedimage}\par
                The numbers pop and disappear one at a time, in numeric order, starting from the smallest one.\strut\hfill\textbf{76\%}
            \end{minipage}
        \end{adjustbox}
        
        & \includegraphics[width=\linewidth, valign=m]{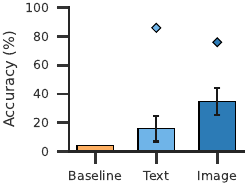} \\
    \midrule

    \rotatebox[origin=c]{90}{\textbf{\taskname{Connect the Dots}}}
        & \begin{adjustbox}{valign=m}
            \begin{minipage}{\linewidth}
                \begin{borderedimage}[baselineorange]
                    \includegraphics[width=\linewidth]{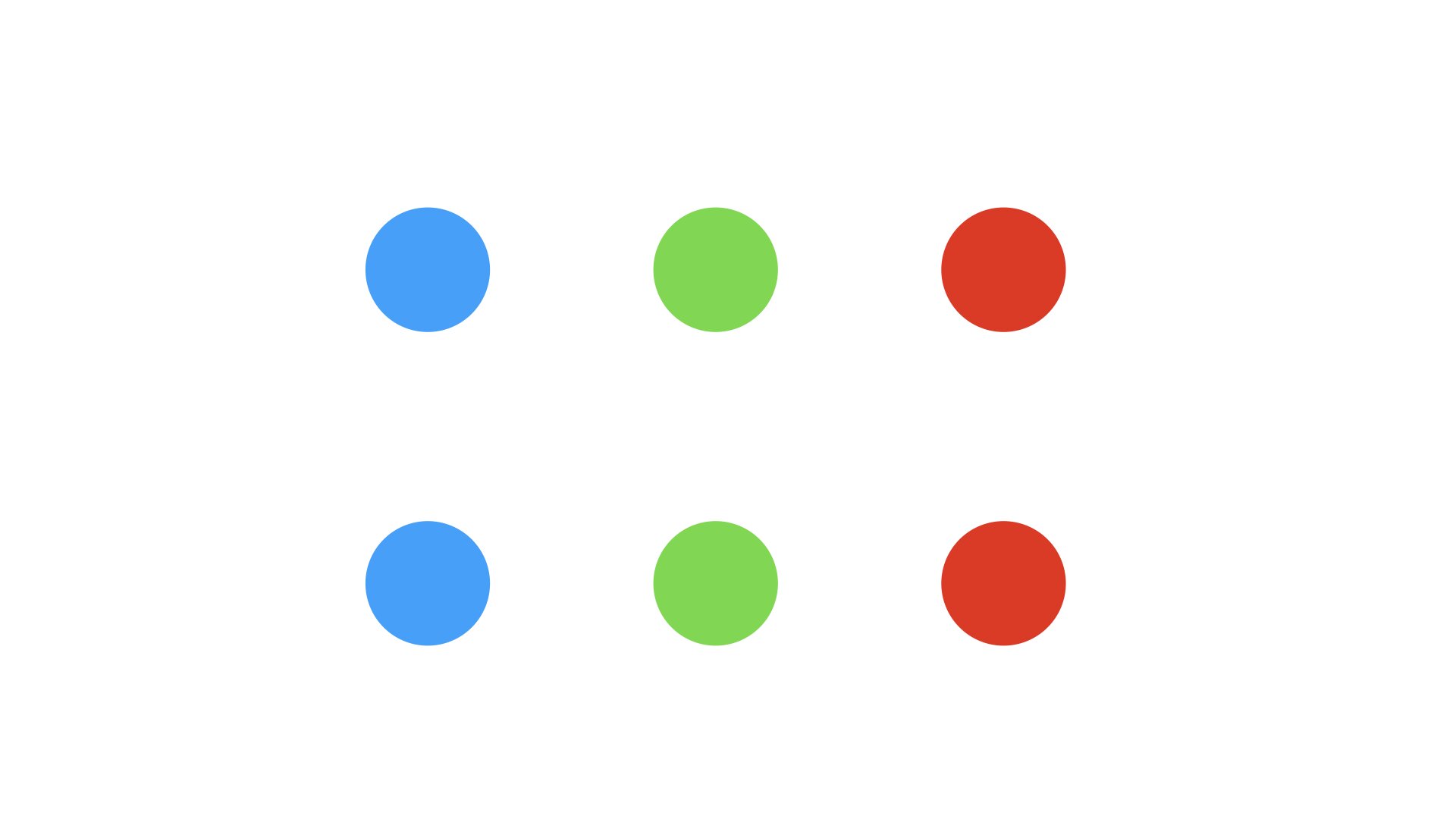}
                \end{borderedimage}\par
                Connect each pair of same-colored circles with a line.\strut\par\vspace{\baselineskip}\hfill\textbf{3\%}
            \end{minipage}
        \end{adjustbox}
        
        & \begin{adjustbox}{valign=m}
            \begin{minipage}{\linewidth}
                \begin{borderedimage}[baselineorange]
                    \includegraphics[width=\linewidth]{figures/image_vs_words/ideator_3p1pro/qualitative/connect_dots_baseline.jpg}
                \end{borderedimage}\par
                \textcolor{vipeblue}{[...] spark at one circle [...] traveling to the other circle of the exact same color [...]}\strut\par\hfill\textbf{35\%}
            \end{minipage}
        \end{adjustbox}
        
        & \begin{adjustbox}{valign=m}
            \begin{minipage}{\linewidth}
                \begin{borderedimage}[vipeblue]
                    \includegraphics[width=\linewidth]{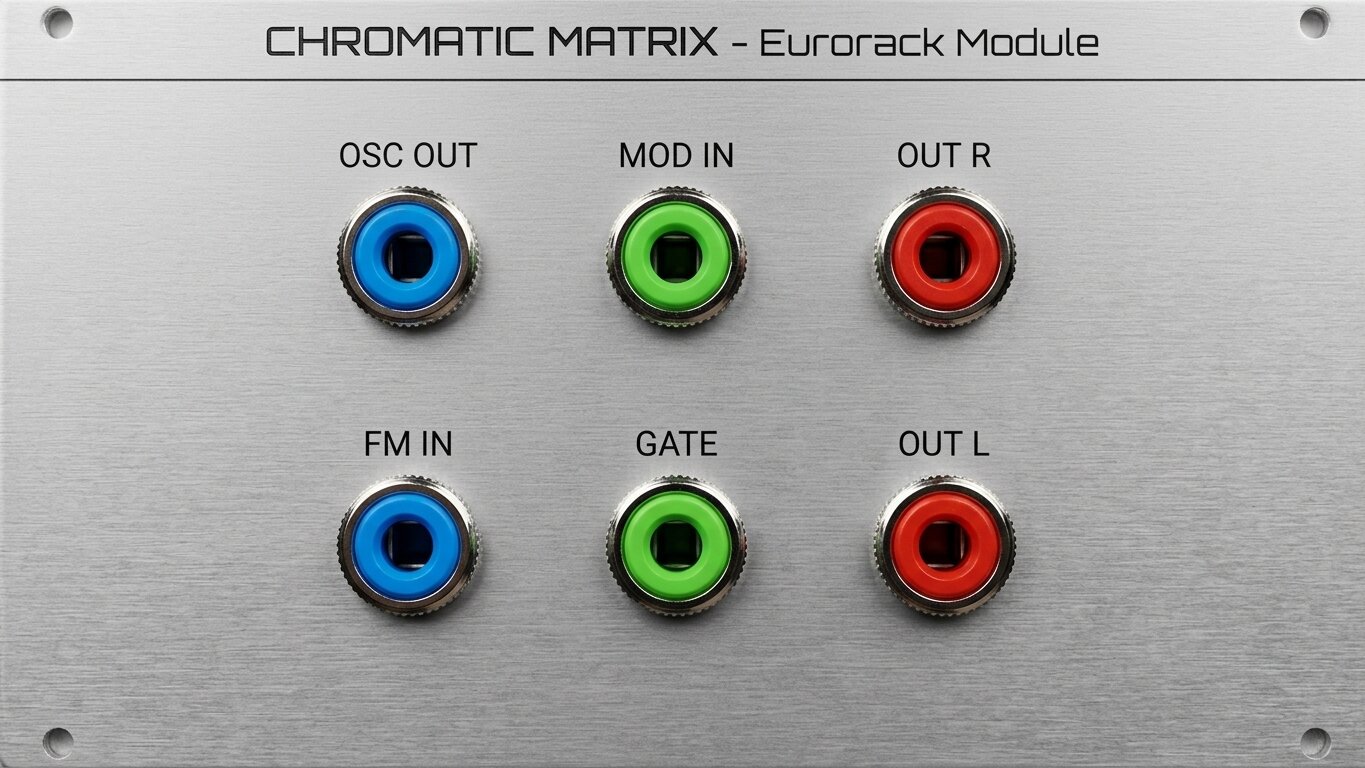}
                \end{borderedimage}\par
                Connect each pair of same-colored circles with a line.\strut\par\vspace{\baselineskip}\hfill\textbf{8\%}
            \end{minipage}
        \end{adjustbox}
        
        & \includegraphics[width=\linewidth, valign=m]{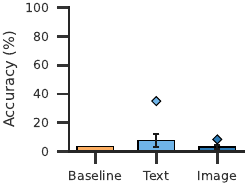} \\
    \end{tblr}
    \caption{\textbf{Visual prompt engineering can be more effective than text prompt engineering for video models.} Each row represents a distinct task, showing the visual input and text prompt for the baseline, the best text variant, and the best image variant, with their respective accuracies in parentheses. The rightmost column summarizes the accuracy of all text and image proposals. The diamond markers show the best variant among $n=20$. Error bars indicate 95\% CI. See~\cref{sec:freeform-more-variants} for more example text and image variants.} 
    \label{fig:imagewords}
\end{figure*}

\paragraph{Visual vs. text prompt engineering} Results are shown in~\cref{fig:imagewords}. Overall, we find that both text and visual prompt engineering substantially improve performance over the baseline across tasks, confirming that each modality is a viable lever for eliciting better video reasoning. For example, on \taskname{Sort 3 Numbers}, the best text variant and the best image variant each achieved roughly a 20-fold improvement (86\% and 76\%, vs. 4\% without prompt engineering). On \taskname{Conjunctive Search} and \taskname{Sort 3 Numbers}, the image variants on average led to substantially larger improvements than text variants, while the single best text variant sometimes exceeds the single best image variant. On \taskname{VPCT}, we did not observe substantial improvement from the text and image variants proposed by the freeform ideators, with a particular image edit among $n=20$ proposals as notable exception, raising Veo~3.1 accuracy from 57\% to 73\%. \taskname{Connect the Dots} remains a challenging task for Veo even with prompt engineering, with generally low performance under diverse image and text variant proposals. These results suggest that for most tasks, a well-chosen text or image prompt improve visual reasoning performance, though finding such a prompt requires non-trivial exploration. We also explored joint text and image prompt engineering (see~\cref{sec:freeform-joint-vipe} for more details). Our simple joint text and image prompt engineering ideator generated coherent and interesting alternative ways to represent a task, but the resulting task variants were less effective in eliciting better video reasoning compared to changing the images alone. Taken together, visual prompt engineering reliably improves video reasoning, and is frequently more effective than traditional language-based prompt engineering.

\takeaway{Depending on the task, visual prompt engineering can be more powerful than text prompt engineering.}

\section{Does visual prompt engineering help native image generation models, too?}
\label{sec:image-gen-models}
Throughout this paper, VIPE has been applied by using an image generation model (Nano~Banana~Pro) as the \emph{editor}, and a video model (Veo~3.1) as the \emph{reasoner}. A natural question is: what happens when the image generation model itself becomes the reasoner? Since Nano~Banana can both understand \emph{and} generate images, it can be asked to directly produce an image depicting the predicted outcome e.g., an image of the ball landing in the predicted bucket for \taskname{VPCT}. If the realism bias hypothesis is correct, and VIPE helps because it translates inputs into a model's native representation, then models that already operate in the photorealistic space should benefit less from VIPE. We test this on \taskname{VPCT} by evaluating three configurations:
\begin{enumerate}[nosep]
    \item \textbf{Nano~Banana~Pro (Gemini~3~Pro~Image), image output};
    \item \textbf{Nano~Banana~2 (Gemini~2.5~Flash~Image), image output};
    \item \textbf{Nano~Banana~2 (Gemini~2.5~Flash~Image), text output} (``Left'', ``Center'', or ``Right'').
\end{enumerate}

\noindent
Each configuration is evaluated on both the original sketch dataset and the
VIPE (photorealistic) dataset, averaged across 10 runs. Results report the
best prompt per configuration and are summarized in~\cref{tab:nb_vpct_results}. For a detailed discussion we refer to~\cref{app:nano_banana}, but overall, it is clear that VIPE does not systematically improve image model reasoning performance. Why is that the case? We hypothesize that VIPE is effective when it bridges a \emph{representation gap} between the input format and the model's internal representation. One possible explanation is that video models like Veo~3.1 may be trained predominantly on realistic videos, thus presenting them with abstract sketches forces reasoning in unfamiliar territory. VIPE helps to bridge this gap. Native image generation models, on the other hand, may be trained on more diverse styles spanning both abstract sketches and realistic images.

\takeaway{We hypothesize that VIPE is most effective when it bridges a representation gap between the input and the model's internal representation space. The gap between abstract and realistic input may be much larger for today's video models than for image generation models.}

\begin{table}[h]
\centering
\small
\begin{tabular}{lllcccc}
\toprule
\textbf{Configuration} & \textbf{Output} & \textbf{Dataset} & \textbf{Pass@1} & \textbf{Pass@5} & \textbf{Pass@10} & \textbf{Maj.\ Vote} \\
\midrule
Nano~Banana~Pro & Image & VIPE   & 50.0\% & 94.0\% & 99.0\% & 54.0\% \\
Nano~Banana~Pro & Image & Sketch & 45.0\% & 86.0\% & 99.0\% & 48.0\% \\
\addlinespace
Nano~Banana~2   & Image & VIPE   & 21.0\% & 69.0\% & 91.0\% & 32.0\% \\
Nano~Banana~2   & Image & Sketch & 37.0\% & 87.0\% & 98.0\% & 48.0\% \\
\addlinespace
Nano~Banana~2   & Text  & VIPE   & 50.0\% & 93.0\% & 97.0\% & 58.0\% \\
\textbf{Nano~Banana~2} & \textbf{Text} & \textbf{Sketch} & \textbf{58.0\%} & \textbf{94.0\%}
    & \textbf{99.0\%} & \textbf{80.0\%} \\
\addlinespace
\textit{Chance} & & & \textit{33.3\%} & \textit{86.8\%} & \textit{98.3\%} & \textit{33.3\%} \\
\bottomrule
\end{tabular}
\caption{\textbf{\taskname{VPCT} results for native image generation models}, best prompt
per configuration. Bold indicates best overall. For comparison, Veo~3.1 achieves 41.3\% (sketch) / 59.3\% (VIPE) accuracy~(\cref{fig:vpct}), and Gemini~3.1~Pro (text-only VLM) achieves 96\% on the sketch dataset~(\cref{tab:vpct_control_conditions}).}
\label{tab:nb_vpct_results}
\end{table}

\section{Why does visual prompt engineering help?}
Throughout this article, we have seen evidence that the effectiveness of visual prompt engineering primarily stems from bridging the domain gap between abstract and realistic input. For example, the \taskname{VPCT} experiments from~\cref{sec:vpct} showed that the best performance is achieved when sketches are transformed into input that has both 3D structure and realistic textures, while the ``unnatural textures ablation'' confirmed that adding 3D depth alone is not sufficient. The automated prompt engineering methods from~\cref{sec:automated-vipe} repeatedly surfaced creative ways to improve realism, from improving material properties to transforming abstract scenes into tangible 3D objects like a chalkboard. Video models clearly have a \emph{realism bias}. When comparing video generations with and without VIPE, we noticed a recurring pattern: scene consistency appears to be drastically improved in realistic scenes, while objects tend to randomly appear, disappear, or change shape in abstract settings. This naturally impedes any downstream reasoning since it changes the task setup, just like changing a number at the beginning of a calculation would lead to the wrong result. 

\begin{figure}[htbp]
    \centering
    \captionsetup[subfigure]{labelformat=empty}

    \setlength{\fboxsep}{0pt}
    
    \begin{subfigure}[b]{0.24\textwidth}
        \centering
        0\%\par\vspace{2pt} 
        \begin{borderedimage}[baselineorange]
            \includegraphics[width=\textwidth]{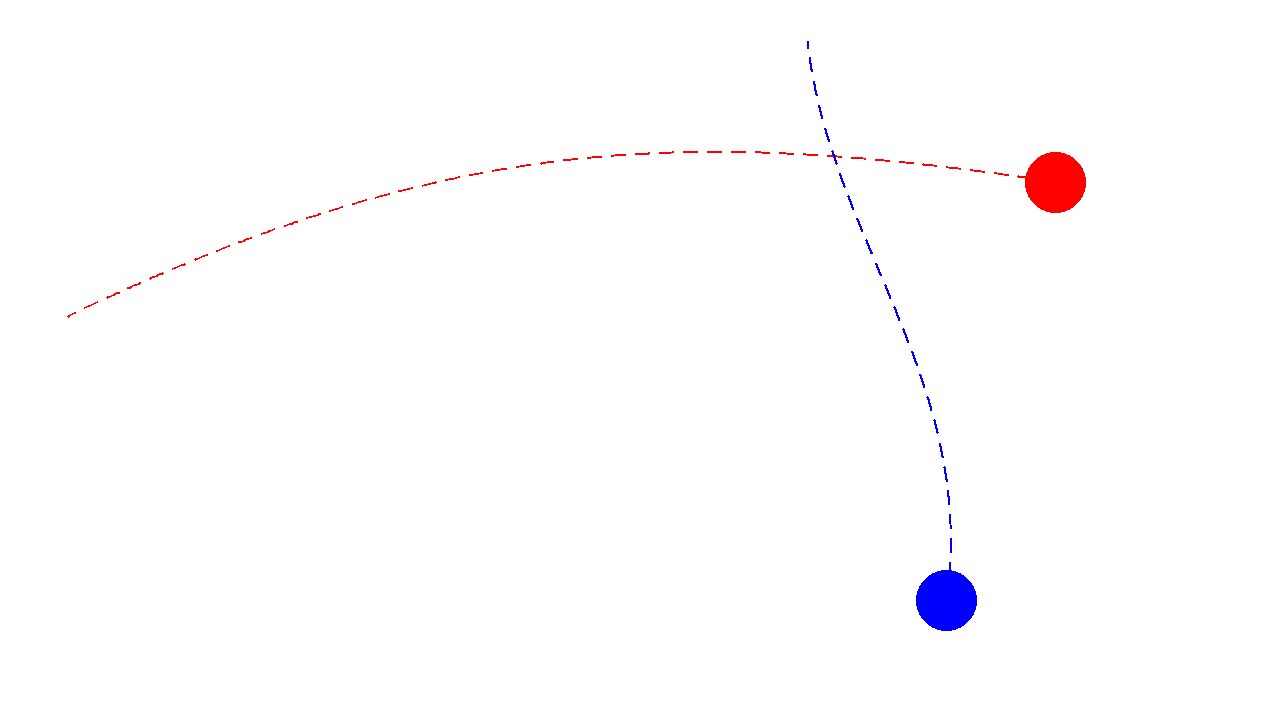}
        \end{borderedimage}
        \caption{Baseline: synthetic}
    \end{subfigure}
    \hfill
    \begin{subfigure}[b]{0.24\textwidth}
        \centering
        11\%\par\vspace{2pt}
        \begin{borderedimage}[vipeblue]
            \includegraphics[width=\textwidth]{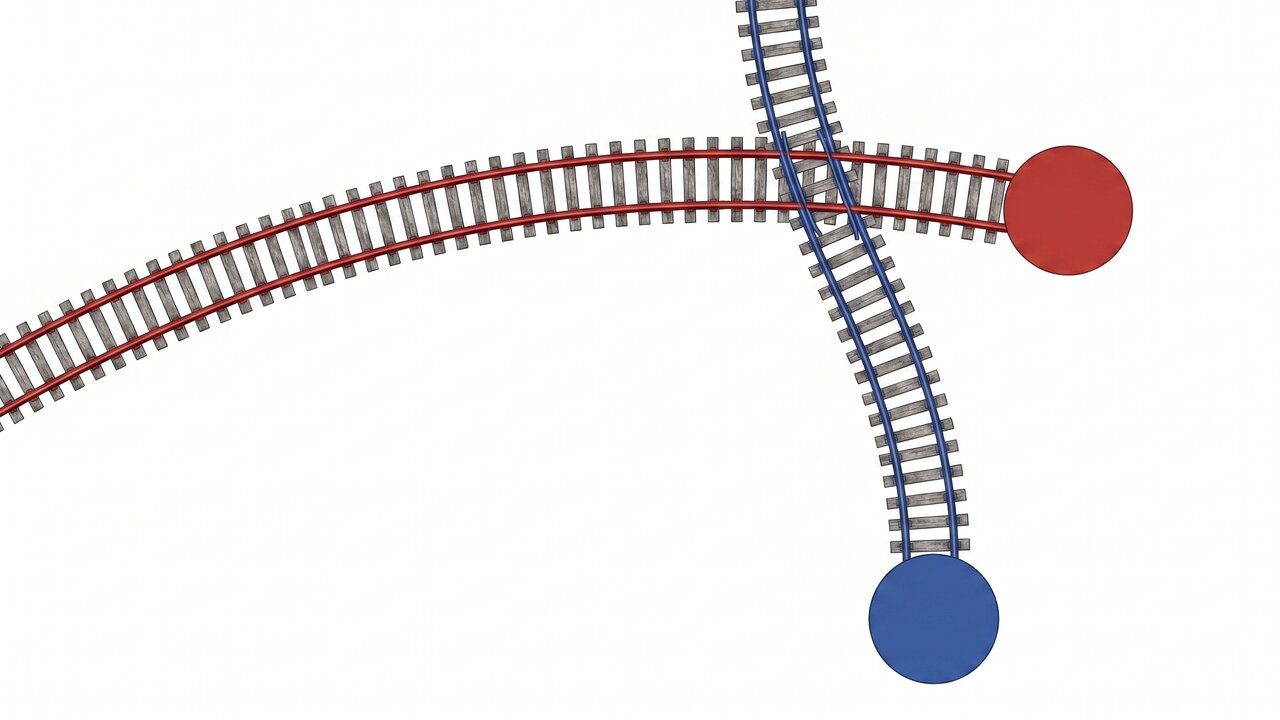}
        \end{borderedimage}
        \caption{+ railway tracks}
    \end{subfigure}
    \hfill
    \begin{subfigure}[b]{0.24\textwidth}
        \centering
        24\%\par\vspace{2pt}
        \begin{borderedimage}[vipeblue]
            \includegraphics[width=\textwidth]{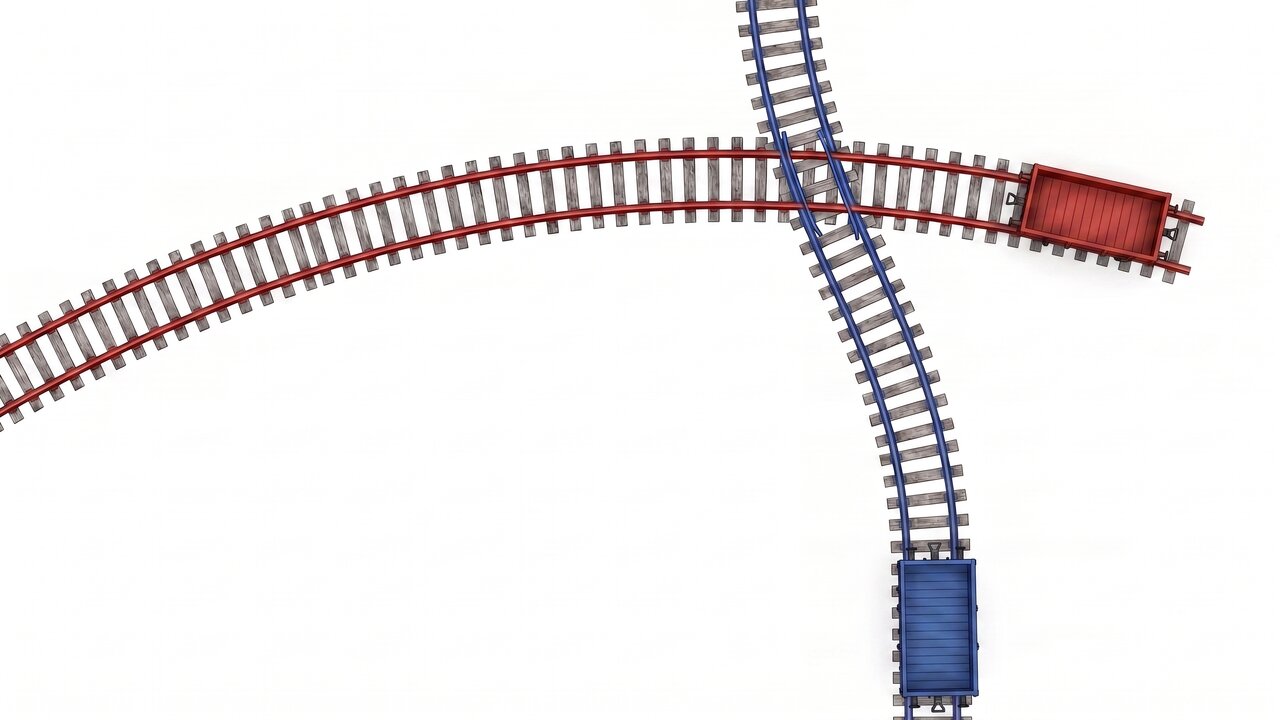}
        \end{borderedimage}
        \caption{+ rail wagons}
    \end{subfigure}
    \hfill
    \begin{subfigure}[b]{0.24\textwidth}
        \centering
        59\%\par\vspace{2pt}
        \begin{borderedimage}[vipeblue]
            \includegraphics[width=\textwidth]{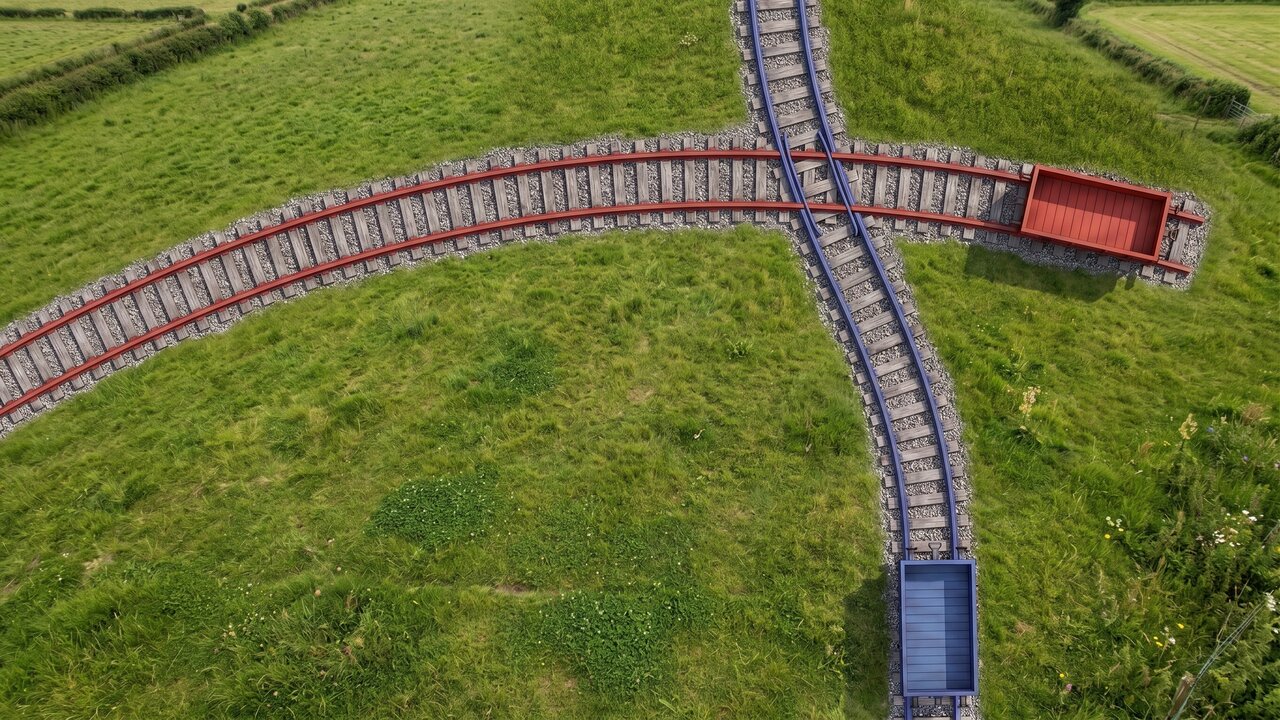}
        \end{borderedimage}
        \caption{+ background}
    \end{subfigure}
    \caption{\textbf{Each step towards realism improves generation consistency.} A simple dataset of red/blue objects and paths is scored on scene consistency (100 videos per condition). Prompt: \emph{``The red and blue $\{$circles, rail wagons$\}$ move along the $\{$dashed lines, railway tracks$\}$ until they reach the intersection point. Static shot, no zoom no pan no dolly. NO CAMERA MOVEMENT AT ALL.''} Starting from the baseline (synthetic) condition, three VIPE variants systematically introduce more realism. When videos generated by Veo~3.1 are rated as a pass/fail on scene consistency by humans (success if objects/lines/tracks don't randomly appear or disappear, change shape, morph etc.), it becomes clear that increased realism causes video model generations to become more consistent.}
    \label{fig:collision}
\end{figure}

To understand better how scene consistency is influenced by VIPE, we design a simple experiment that visually prompt-engineers a synthetic dataset step-by-step in the direction of a fully realistic scene, shown in~\cref{fig:collision}. This is based on the fact that realism is not necessarily a binary choice but rather a spectrum. Human ratings of scene consistency improved from 0\% (!) in the synthetic setting to 59\% in the fully realistic setting. Interestingly, every single step towards increased realism directly causes video model generations to become more consistent, thereby offering an explanation for why visual prompt engineering helps.

\section{Discussion}
\paragraph{Summary} Inspired by the fact that text-based prompt engineering has become a foundational technique for improving language model performance, we set out to understand whether video model reasoning can similarly be improved by visual prompt engineering, i.e., transforming the task image via image editing. We find that good ``visual prompts'' indeed systematically improve video reasoning across tasks. To the author's surprise, VIPE works even better than traditional test-time scaling methods like self-consistency (for a given inference budget), and can also be more effective than optimizing the text prompt alone. We identified patterns that reliably improve performance, such as the \emph{realism bias} of video models: video models prefer to reason in photorealistic scenes, as opposed to impoverished, abstract or synthetic ones. Taken together, VIPE is a cheap and effective lever for improving visual reasoning with video models.

\paragraph{Realism bias: stop evaluating video models on abstract, impoverished tasks}
Video model accuracy improves when transforming abstract settings into realistic-looking scenes. On one hand, this is great: it means that visual prompt engineering in the direction of increased photorealism can substantially improve video model reasoning across many tasks. As a consequence, reasoning performance on datasets that look abstract / synthetic (which is the case for most of these datasets) is often only a lower bound on the model's true, underlying reasoning ability. This type of distinction has been described in the literature as the ``performance vs.\ competence'' dilemma~\citep{firestone2020performance}: e.g., in order to measure the true maximum speed of a fish, it's worth testing the fish in the water as opposed to on land. Similarly, our results indicate that video models may sometimes have the \emph{competence} (or ability) to solve a task better than what is indicated by their poor \emph{performance} on an abstract, impoverished version of the task. Consequently, we recommend to stop evaluating video models on abstract tasks when there's the option to visually prompt-engineer a more realistic version of the same task.

\takeaway{Whenever feasible, don't evaluate video models on abstract/impoverished tasks. Video models have a realism bias, thus unrealistic-looking tasks underestimate their true performance. Instead, use visual prompt engineering to transform abstract tasks into realistic setups.}

\noindent
On the other hand, the fact that video models have a realism bias is clearly suboptimal. It would be preferable for a video model to solve any task specified in any way, shape or form. To a certain degree, this feels reminiscent of the brittleness of early language models, where a lot of attention to detail went into formulating tasks in a way that elicited the best possible performance \citep[e.g.][]{jiang2020can,bouraoui2020inducing,reynolds2021prompt}. For example, to borrow an example from~\citet{jiang2020can}, the text prompts ``Obama worked as a \_\_\_'' vs. ``Obama is a \_\_\_ by profession'' can systematically influence language model performance, despite being variants of the exact same underlying task. In this sense, realism bias might be characterized as an unwanted shortcut~\citep{geirhos2020shortcut,lapuschkin2019unmasking} or dataset bias~\citep{torralba2011unbiased} that models pick up from their training data. It would be interesting to understand whether scene consistency is highest for realistic scenes because unrealistic/abstract scenes take the model out-of-distribution, or whether it simply takes them to an in-distribution setting with different priors (e.g., abstract scenes may resemble cartoon movies where it is perfectly normal for objects to appear or disappear in implausible ways).

\paragraph{When does VIPE help?}
Our evaluation of native image generation models on \taskname{VPCT} (\cref{sec:image-gen-models})
reveals a complementary perspective on the realism bias. VIPE helps video models
because it translates inputs from an unfamiliar representation (e.g., abstract sketches)
into a familiar one (photorealistic scenes). But for image generation models that are already covering both representation spaces better, VIPE provides little benefit. This suggests that the realism bias may not be a universal property of all visual models, but rather a consequence of a representation mismatch between the input and the model's training distribution.

\paragraph{Limitations}
The biggest limitation, at the moment, is that the quality of a prompt engineering variant is highly dependent on the quality of the image editing model. If the image editor changes the nature of the task or introduces artifacts, then it can become impossible for the downstream video model to solve the original task. We mitigated this risk through a filtering / quality control step where e.g.\ a VLM reviews (and possibly rejects) image edits. In the future, with further improvements in image editing models, this step may not be necessary anymore. Another limitation is the potential cost overhead of visual prompt engineering. However, since the cost of image editing pales in comparison to the cost of video generation, in practice this often turns out to be an advantage of VIPE, since a (comparatively cheap) image editing call can significantly reduce the (comparatively expensive) chance of a video generation that fails to solve the task.

\paragraph{Outlook} The success of visual prompt engineering points toward a shifting paradigm where visual instructions are treated as an optimizable space rather than fixed inputs, just like text prompts are carefully crafted and optimized to elicit better language model performance. As frontier video models evolve towards accepting richer multi-modal context \citep[e.g.][]{agarwal2026cosmos,GeminiOmniFlash2026}, visual prompt engineering might extend beyond editing a single image---perhaps involving multiple reference images or videos. As models continue to improve, an interesting question arises: Will we need less and less visual prompt engineering, since video models might just understand even abstract tasks more easily? Or, alternatively, will visual prompt engineering become more powerful since image editing is becoming better and better? Only time will tell, but if there's anything to be learned from the history of language model development, it's the lesson that a model can only ever be as good as its prompt.

\section*{Acknowledgements}
We would like to thank Zhengxuan Wu, Shane Gu, David Fleet, and Mike Mozer for helpful discussions/feedback, and the Veo, Wan, Gemini, Omni, and Nano~Banana teams for developing these models.

\bibliographystyle{unsrtnat}
\bibliography{refs}

\newpage
\appendix
\crefalias{section}{appendix}
\section*{Appendix}

\section{Datasets}
\label{app:datasets}
\paragraph{VPCT} MIT license, dataset on HuggingFace by~\citet{camelcase2025vpct}, 100 samples. For the results in~\cref{sec:automated-vipe}, we use the first ten samples. For~\cref{sec:vpct}, the entire dataset is used.

\paragraph{Conjunctive Search, Sort 3 Numbers, Connect the Dots} Task ideas were drawn with permission from the authors of~\citet{wiedemer2025video}. Datasets were generated manually in Keynote by the authors. \taskname{Conjunctive Search} contains 5 image samples, each showing 2 blue circles (targets), 4 red circles (distractors) and 4 blue squares (distractors) at different positions. \taskname{Sort 3 Numbers} contains 5 image samples of different sets of double-digit numbers. \taskname{Connect the Dots} contains 6 image samples of 3 same-colored circles with varying colors. 

\paragraph{Maze} With permission from the authors, we use the four maze solving splits from~\citet{wiedemer2025video}. The first three splits with 50 samples each are mazes on square grids with size $5\times5$, $7\times7$, and $9\times9$. The last splits contains 40 non-axis-aligned mazes with curved paths. We report results on the first ten samples of each split in~\cref{sec:automated-vipe}.

\paragraph{RushHour} One task from the MentisOculi benchmark~\citep{zeller2026mentisoculi}, published under an Apache~2.0 license. We select this task as it is the only one used to test video models in the original paper. The task is organized in five splits (difficulty levels) with 50 samples each, where the level corresponds to the number of cars that have to be moved for a minimum-length solution. Level~1 is easily solved by Veo~3.1, while Levels~4~and~5 prove too difficult, largely because the model fails to maintain consistency in complex scenes~\citep{zeller2026mentisoculi}. We consequently focus our analysis on Levels~2~and~3, reporting results on the first ten samples of either in~\cref{sec:automated-vipe}.

\section{Inference details}
\label{app:cost}
We run inference with Gemini, Veo and Omni models via the Vertex AI API. Images and videos are generated at 720p. Videos are generated with a duration of 8s. At the time of writing, generation costs USD~0.40 per second for videos with Veo~3.1 (with or without audio), for a total of USD~3.20 per video. Image editing costs around USD~0.005 for a 720p input image and USD~0.067 for a 720p output image for Gemini~3.1~Flash~Image (Nano~Banana~2). A complete VIPE loop with five ideated edits and filtering costs around USD~0.40, dominated by the image editing cost~\citep{google_gemini_api_pricing}.

\section{VPCT: evaluation details}
\label{app:vpct}

\paragraph{MSE-based evaluator (for original sketches)} To determine in which of three containers (left, center, right) the ball lands, we partition the bottom portion of each frame into three regions of interest and measure the mean squared error (MSE) between the first frame and each subsequent frame independently within each region. A sliding window of the last five MSE values is maintained per region. After an initial warmup period of 12 frames to initialize the sliding window average, the evaluator identifies the first frame on which exactly one region's MSE exceeds its running mean by at least a factor of two, and returns that region as the predicted landing container.

\paragraph{Color-based evaluator (for visual prompt engineering variants)} Since visual prompt engineering introduces more variation and would thereby reduce the robustness of an MSE-based evaluation, we here instead track the location of the ball throughout the video. Each frame is converted to HSV space and a threshold isolates pixels matching the ball's color (red). Contours are extracted from the resulting binary mask and filtered by area and circularity to reject background noise. The centroid of the largest valid contour is computed and mapped to one of the three container regions based on its horizontal position. The evaluator returns the first frame on which the ball is detected inside a container region. If the ball is detected on the final frame but falls outside the container regions, the nearest container is assigned as the model's guess. In order to make sure that a VLM rewriter doesn't simply spell out the solution bucket (e.g., `... and finally, the ball drops into the middle bucket') and instead, the video model is required to simulate the ball's trajectory, we preprocessed VIPE VPCT images such that the buckets are removed / replaced with the background color. That way, the model still needs to simulate the trajectory, but the ball doesn't end up in a bucket that can be pre-specified. Since this setup makes the task a bit harder to understand for humans, we retained buckets for visualization purposes in figures; keeping in mind that the core task is still exactly the same and the ball's location is scored against the target regions where the buckets would have been.

\paragraph{Text prompts} Here are prompt details for \taskname{VPCT}.

\begin{promptbox}{Original prompt (English)}
\texttt{[image]}

The ball moves down in a physically plausible way, sliding down the obstacles, and ends up in one of the three containers at the bottom. Static shot, no zoom no pan no dolly.
\end{promptbox}

\begin{promptbox}{Original prompt (Chinese used for the corresponding Wan2.2 experiments)}
\texttt{[image]}

\begin{CJK*}{UTF8}{gbsn}
球以符合物理规律的方式向下移动，滑下障碍物，并最终落在底部的三个容器之一中。静态镜头，无缩放无平移无推拉。
\end{CJK*}
\end{promptbox}

\begin{promptbox}{VIPE and unnatural textures prompt (English)}
\texttt{[image]}

The single red ball moves down in a physically plausible way, sliding down the obstacles (always down, never up), and after passing the obstacles, it finally drops to the ground. Apart from the red ball, the scene and wooden boards remain static and unchanged. Static shot, no zoom no pan no dolly. NO CAMERA MOVEMENT AT ALL.
\end{promptbox}

\begin{promptbox}{VIPE and unnatural textures prompt (Chinese used for the corresponding Wan2.2 experiments)}
\texttt{[image]}

\begin{CJK*}{UTF8}{gbsn}
单个红球以符合物理规律的方式向下移动，滑下障碍物（总是向下，从不向上），在穿过障碍物后，它最终掉到地面上。除了红球之外，场景和木板保持静态不变。静态镜头，无缩放无平移无推拉。完全没有相机移动。
\end{CJK*}
\end{promptbox}

\paragraph{Test-time scaling details}
To assess how model accuracy on the physics \taskname{VPCT} benchmark scales by test-time scaling the number of independent runs, we computed majority-vote accuracy for $k = 1, 2, \ldots, 20$ runs using a leave-out subsampling procedure. The benchmark consists of 100 samples, each assigned to one of three classes (chance level: 33.3\%). For each value of $k$, we enumerated all $\binom{20}{k}$ possible subsets of $k$ runs drawn without replacement from the 20 available; where this number exceeded 5,000, we instead drew 5,000 random subsets. For each subset, we then computed a majority vote across the $k$ run predictions per item, resolving ties by sampling uniformly at random among the tied classes. The reported accuracy for each $k$ is the mean over all evaluated subsets.

\paragraph{Have VLMs solved physics reasoning on VPCT?}
Over the course of one year, vision-language models (VLMs) have shown rapid progress on the \taskname{VPCT} dataset, from below 50\% accuracy in April 2025 to now 91\% accuracy by the leading model on the \href{https://epoch.ai/benchmarks/vpct/}{public leaderboard}, Gemini~3~Pro~Preview. An updated version, Gemini~3.1~Pro, even scores 96\% on this dataset in our own experiments. Have VLMs, therefore, essentially solved physics reasoning on this type of data?

\begin{figure*}[t]
    \centering
    \begin{subfigure}[t]{0.31\textwidth}
        \centering
        \includegraphics[width=\linewidth]{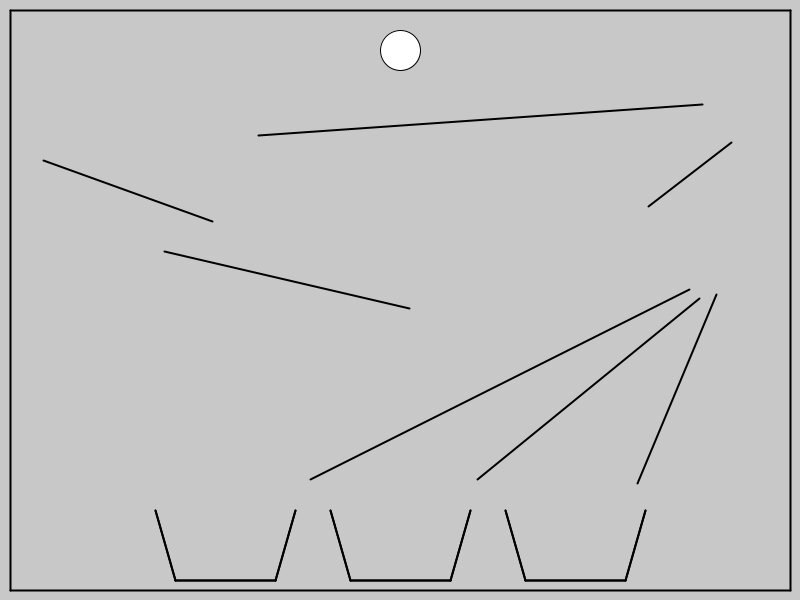}
        \caption{\textbf{Original:}\\``...~this vertical drop leads directly \textcolor{correctgreen}{into the first bucket on the left}''}
    \end{subfigure}
    \hfill
    \begin{subfigure}[t]{0.31\textwidth}
        \centering
        \includegraphics[width=\linewidth]{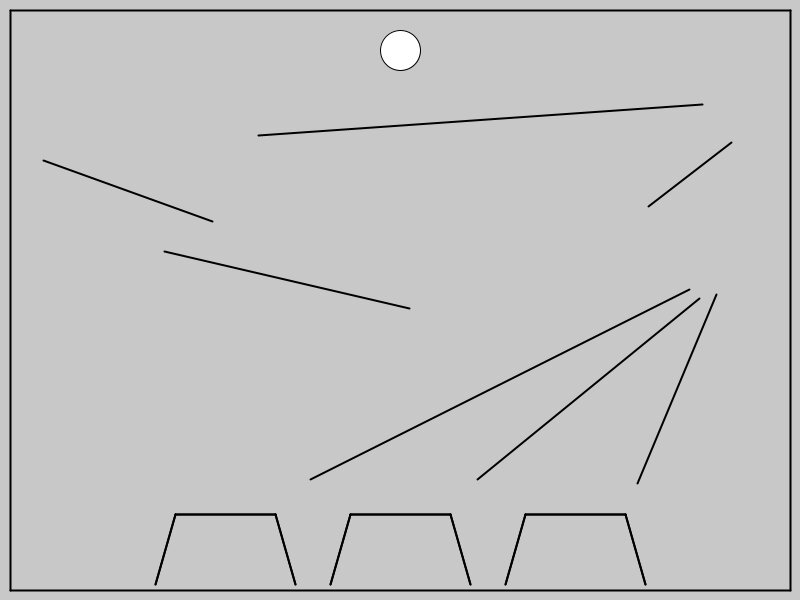}
        \caption{\textbf{Upside-down buckets:}\\ ``...~this final slope guides the ball \textcolor{wrongred}{directly into the left bucket}''}
    \end{subfigure}
    \hfill
    \begin{subfigure}[t]{0.31\textwidth}
        \centering        \includegraphics[width=\linewidth]{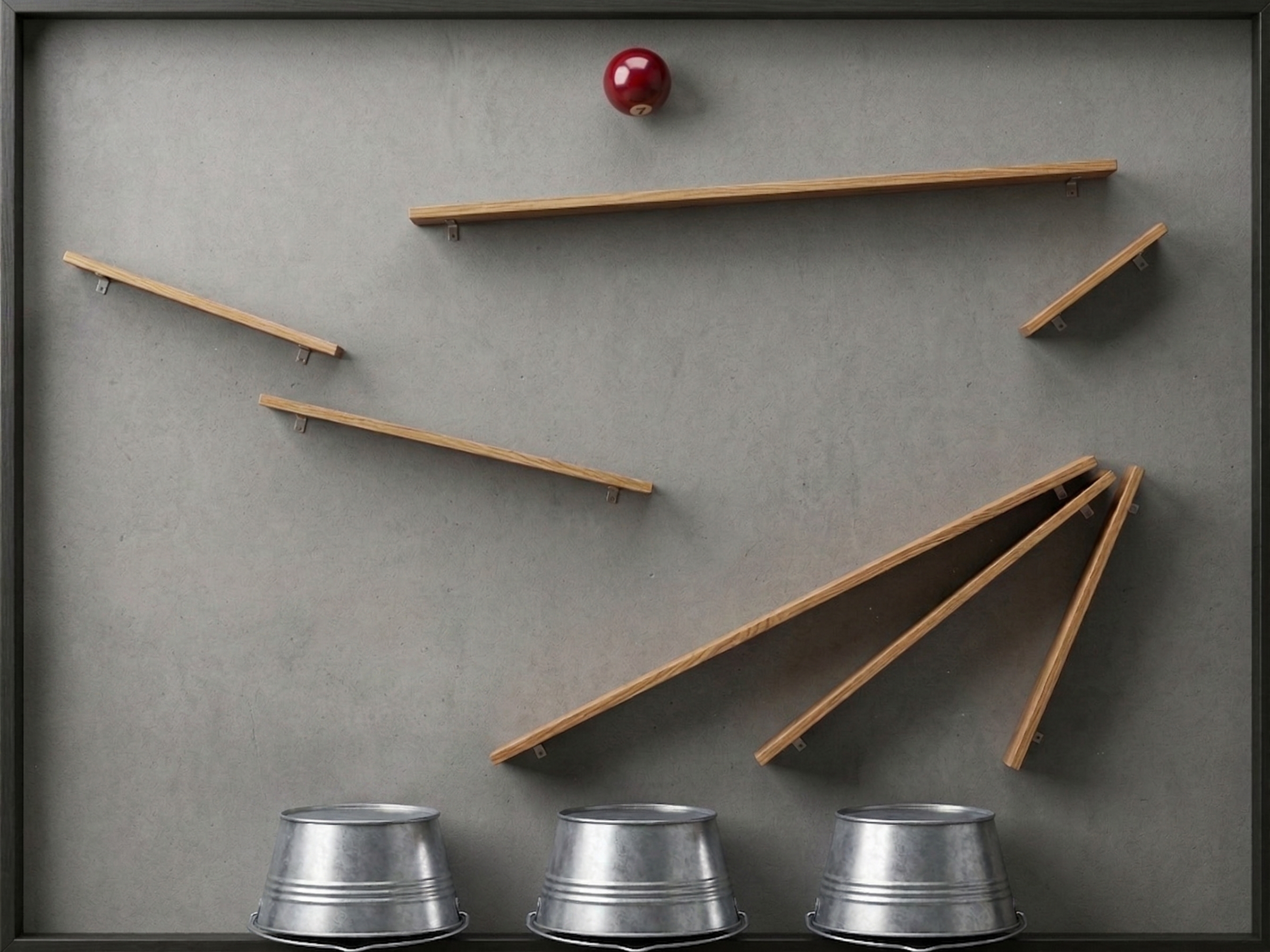}
        \caption{\textbf{Realistic upside-down buckets:}\\``...~the ball will land \textcolor{wrongred}{in the leftmost bucket}''}
    \end{subfigure}
    \caption{\textbf{Control conditions reveal shallow physics understanding of vision-language models.} Prompt: \textit{``Predict which bucket (numbered 1, 2, or 3 from left to right) the ball will eventually fall into. If the ball will not fall into any bucket, output `answer(0)'; if the ball falls into a bucket, output `answer(X)' where X is the bucket number (1, 2, or 3).'} \textbf{Original (a)}: VLMs appear to understand the setup and correctly predict the bucket. \textbf{Control condition upside-down buckets (b):} The model still believes that the ball will drop \textbf{into} a bucket, revealing a shallow understanding. \textbf{Control condition with realistic upside-down buckets (c)}: In order to rule out that the VLM has, in fact, a good understanding of physics and just misunderstands the abstract sketch setup of condition (b), a realistic control condition confirms that the model's understanding is indeed shallow: it still does not get the answer right.}
    \label{fig:vpct_shallow_physics}
\end{figure*}

We decided to put their physics reasoning abilities to a test by creating a modified version \taskname{VPCT}, visualized in~\cref{fig:vpct_shallow_physics}. The result is striking: when buckets are turned upside-down, model accuracy drops down to a mere 3--4\% (see~\cref{tab:vpct_control_conditions} for accuracies). The model completely ignores that a ball cannot drop \emph{into} a bucket that is upside-down. This provides strong evidence that VLMs only have a superficial, not a robust understanding of physics. Vision-language models, on this task, appear to use a shortcut~\citep{lapuschkin2019unmasking,geirhos2020shortcut} and lack an understanding of the underlying physical principles.

In order to rule out that the model simply isn't able to perceive whether the bucket is upside-down or not (or somehow misunderstands the sketch), we separately asked the model to identify whether the buckets are upside down in any given sample with the following prompt: \emph{``Are the buckets upside down - True or False? Output `answer(X)' where X is either `True' (= buckets are upside down) or `False' (= buckets are not upside down).''} Gemini~3.1~Pro accuracy on this task is 100\%. The model correctly identifies all upside-down buckets as being upside-down, and also identifies all original buckets as not being upside-down. The model, therefore, has full access to the visual information required to solve the task---but it fails to put the pieces together. In summary, despite reaching deceptively high accuracies on the original \taskname{VPCT}, VLMs are far from solving physics reasoning.

\begin{table}[h]
\caption{Accuracy of vision-language models on \taskname{VPCT} control conditions. Values in brackets indicate whether the model-selected bucket (even though it's upside-down) would have been the correct bucket in the original task. The binary classification task asks the model to identify whether the buckets are upside-down or not in a balanced dataset (chance: 50\%).}
\label{tab:vpct_control_conditions}
\centering
\begin{tabular}{lrrr}
\toprule
& \multicolumn{3}{c}{\textbf{Condition}}\\
\cmidrule(lr){2-4}
\textbf{Model} & \makecell[bc]{\textbf{Original}\\\textbf{Sketch}} & \makecell[bc]{\textbf{Upside-Down}\\\textbf{Sketch}} & \makecell[bc]{\textbf{Upside-down}\\\textbf{Classification}}\\
\midrule
Gemini~3.1~Pro          & 96\%     & 4\% (87\%)  & 100\%          \\
Gemini~2.5~Pro          & 48\%     & 3\% (42\%)  &  82\%          \\
\bottomrule
\end{tabular}
\end{table}

\section{Wan}
\label{app:wan}
\subsection{Wan rewriter} The Wan model has the option enable/disable the prompt rewriter~(Qwen2.5-Plus). To understand whether this helps or hurts performance, we performed an ablation of the Wan rewriter regarding \taskname{VPCT} performance. With rewriter enabled, \taskname{VPCT} with visual prompt engineering performance was 37.7\% for the TI2V model, which is lower than the comparison conditions 42.3\% (TI2V English) or 43.3 (TI2V Chinese) from~\cref{fig:vpct}. Thus, for this task, the model performs better without rewritten prompts. For this reason, numbers in the main paper are reported without rewriter enabled.

\section{Automated VIPE details}
\label{app:automated-vipe}

\cref{tab:automatde-vipe-datapoints} summarizes how many variants and final videos were generated for each task in~\cref{sec:automated-vipe,fig:automated-vipe}. Each bar in~\cref{fig:automated-vipe} summarizes only the best variant: the overall best for an entire task and split (e.g., \taskname{Connect the Dots}, \taskname{Maze 3$\times$3}, \taskname{RushHour} Level 2) as solid bars, the best per sample-variant as hatched bars.

\begin{table}[]
    \caption{Breakdown of the \textbf{\num{18160} generated videos} for~\cref{sec:automated-vipe,fig:automated-vipe}.}
    \centering
    \begin{tabular}{llrrrr}
    \toprule
    \textbf{VIPE} & \textbf{Task (Split)} & \makecell[bc]{\textbf{Task}\\ \textbf{Samples}} & \textbf{Variants} & \makecell[bc]{\textbf{Videos}\\\textbf{per Sample}} & \makecell[bc]{\textbf{Videos}\\\textbf{Total}} \\
    \midrule
    Freeform & \taskname{RushHour} (Level 2) & 10 & 20 & 10 & 2000 \\
    & \taskname{RushHour} (Level 3) & 10 & 20 & 10 & 2000 \\
    & \taskname{Maze} ($5 \times 5$) & 10 & 20 & 10 & 2000 \\
    & \taskname{Maze} ($7 \times 7$) & 10 & 20 & 10 & 2000 \\
    & \taskname{Maze} ($9 \times 9$) & 10 & 20 & 10 & 2000 \\
    & \taskname{Maze} (irregular) & 10 & 20 & 10 & 2000 \\
    & \taskname{VPCT} & 10 & 20 & 10 & 2000 \\
    & \taskname{Conjunctive Search} & 5 & 20 & 10 & 1000 \\
    & \taskname{Connect the Dots} & 6 & 20 & 10 & 1200 \\
    & \taskname{Sort 3 Numbers} & 5 & 20 & 10 & 1000 \\
    \midrule
    ACE & \taskname{Conjunctive Search} & 5 & 20 & 3 & 300 \\
    & \taskname{Connect the Dots} & 6 & 20 & 3 & 360 \\
    & \taskname{Sort 3 Numbers} & 5 & 20 & 3 & 300 \\
    \bottomrule
    \end{tabular}
    \label{tab:automatde-vipe-datapoints}
\end{table}

\subsection{Freeform ideation}
\label{app:freeform_ideation}
We iterate the entire automatic visual prompter $n$ times in sequence, with each iteration's final edits instructions $t_\text{edit}$ being appended to the next iteration's ideation prompt $t_\text{ideate}$ to encourage diversity.

\begin{figure*}[h]
  \centering
  \includegraphics[width=0.19\linewidth]{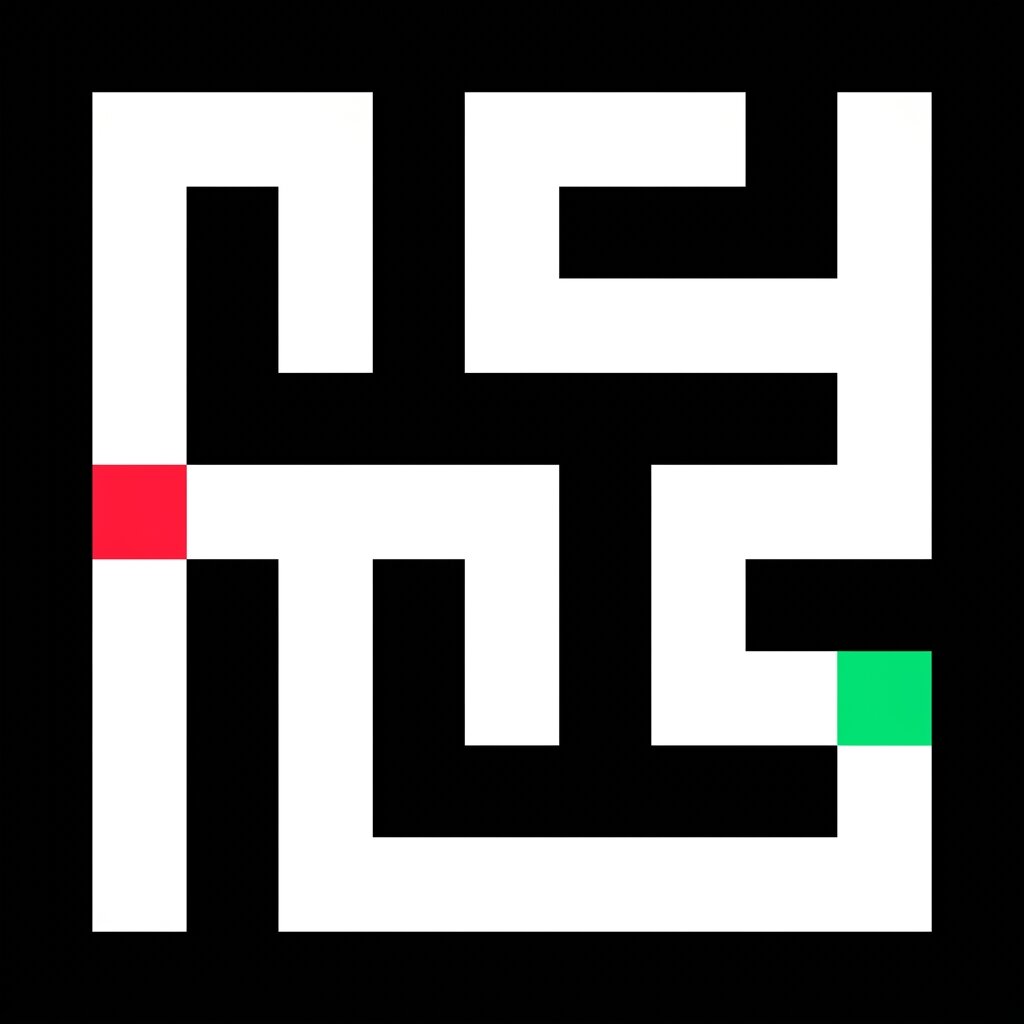}\hfill
  \includegraphics[width=0.19\linewidth]{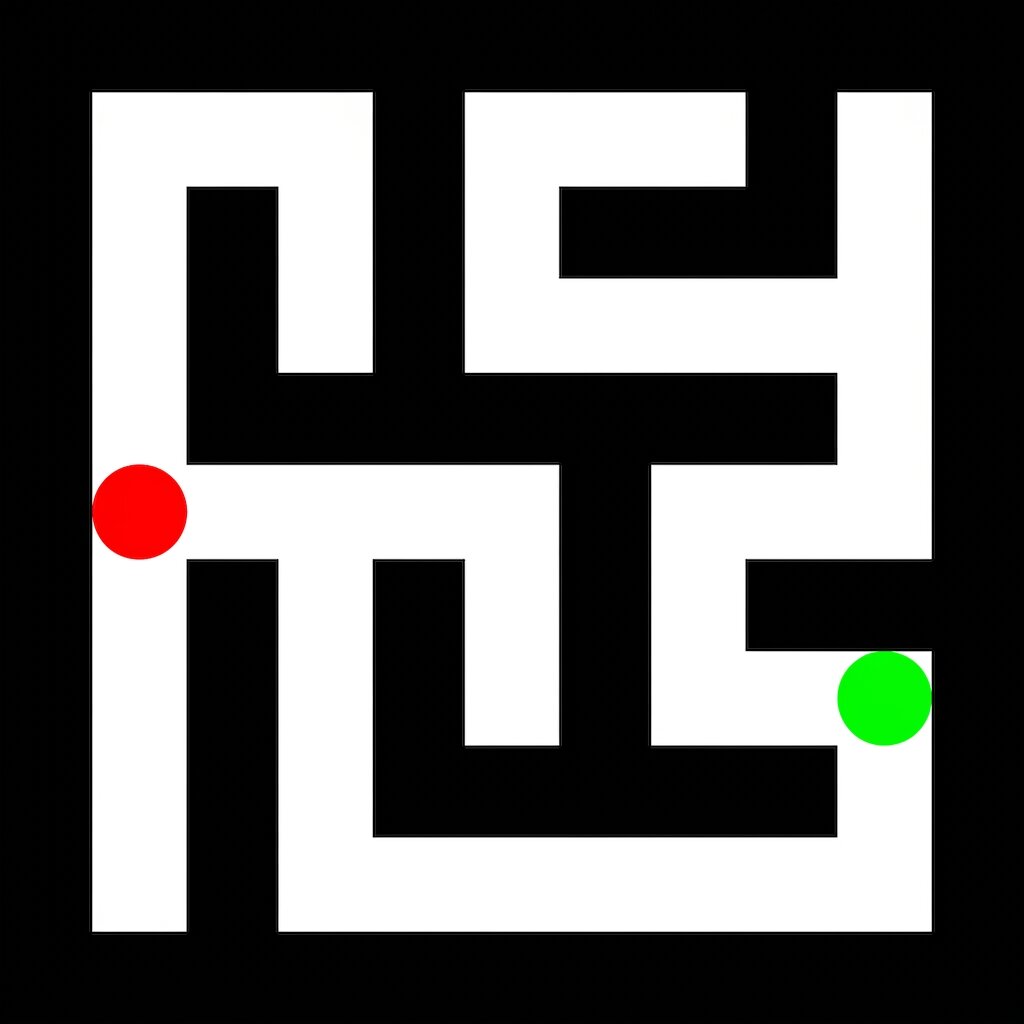}\hfill
  \includegraphics[width=0.19\linewidth]{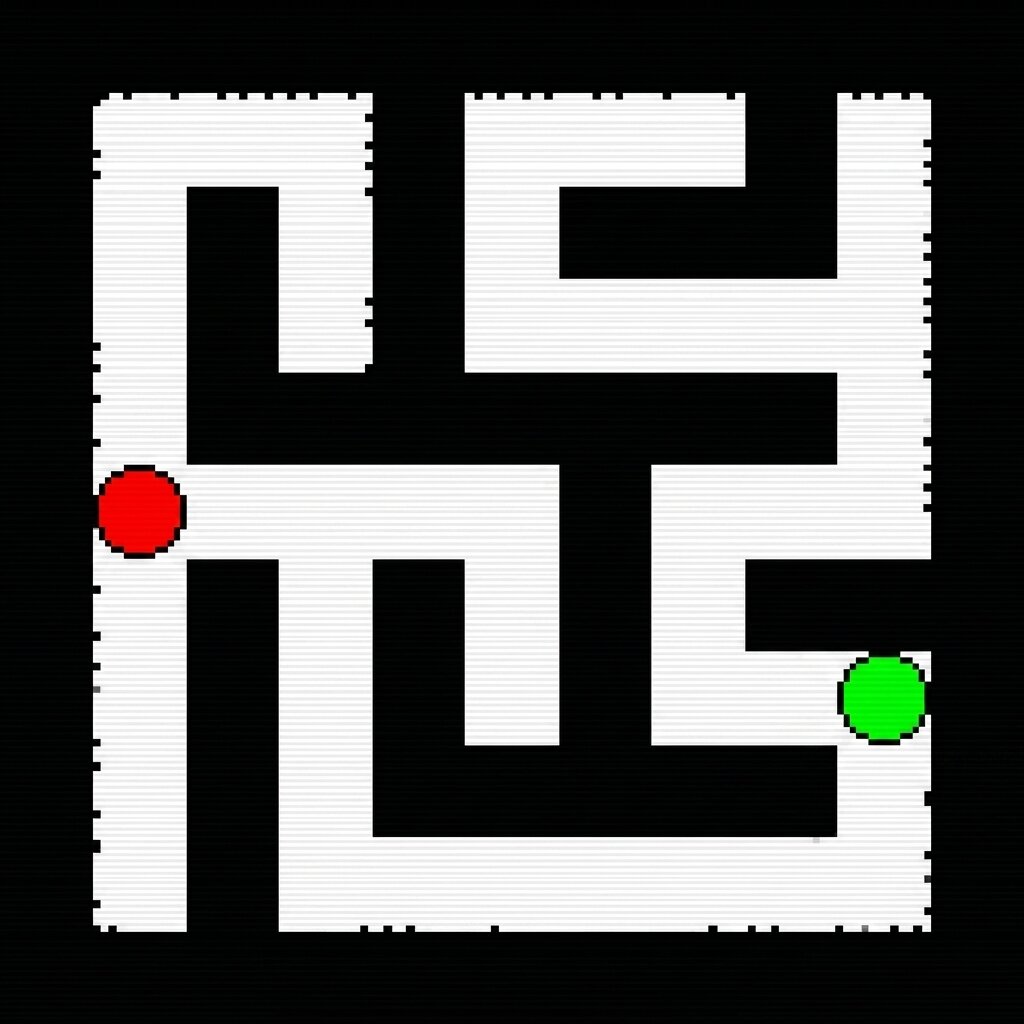}\hfill
  \includegraphics[width=0.19\linewidth]{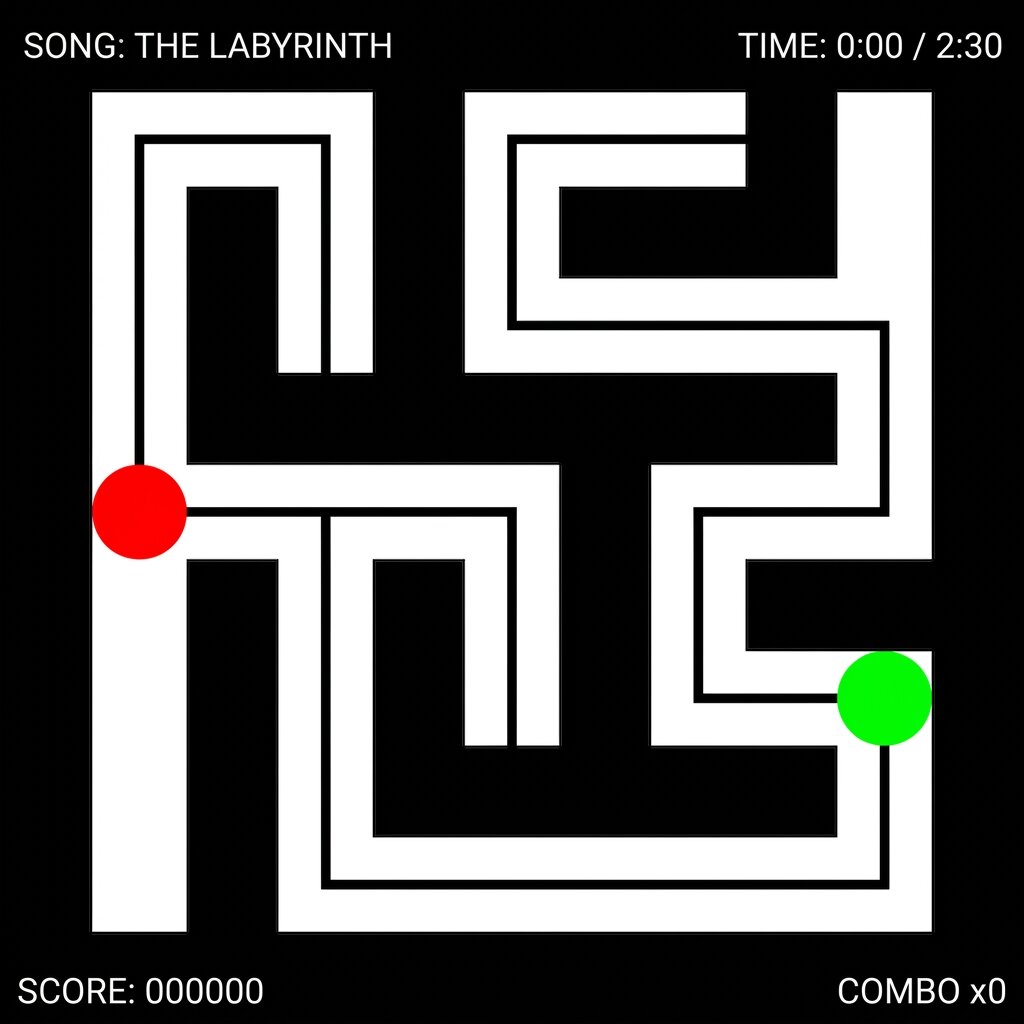}\hfill
  \includegraphics[width=0.19\linewidth]{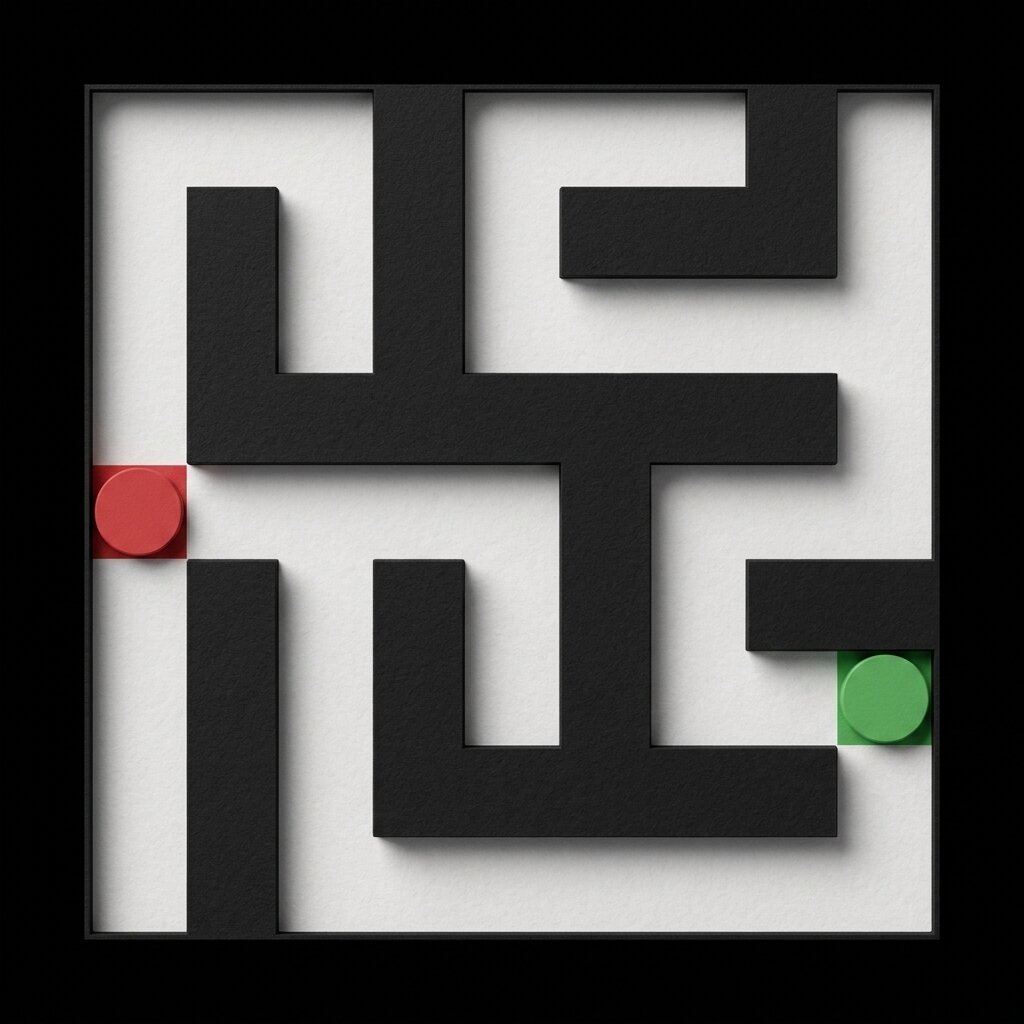}
  
  \vspace{10pt} %

  \includegraphics[width=0.19\linewidth]{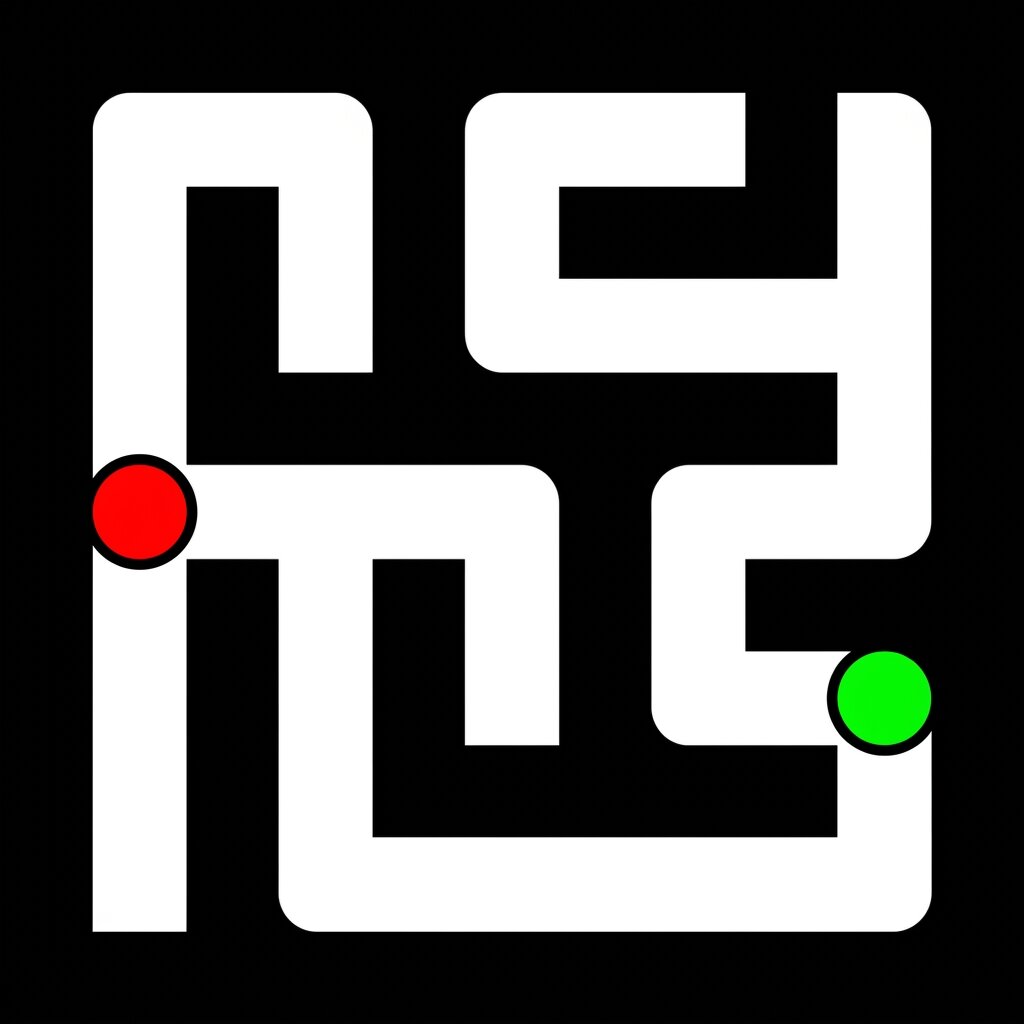}\hfill
  \includegraphics[width=0.19\linewidth]{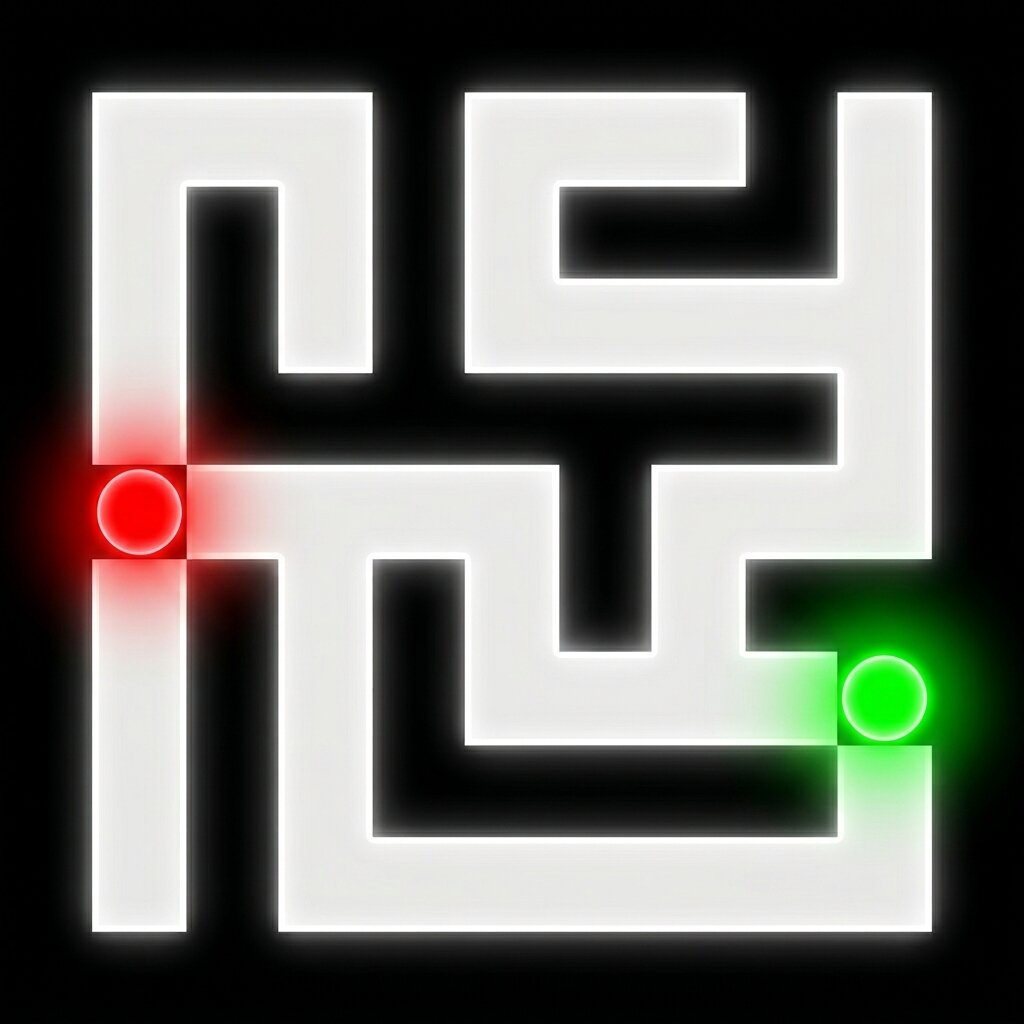}\hfill
  \includegraphics[width=0.19\linewidth]{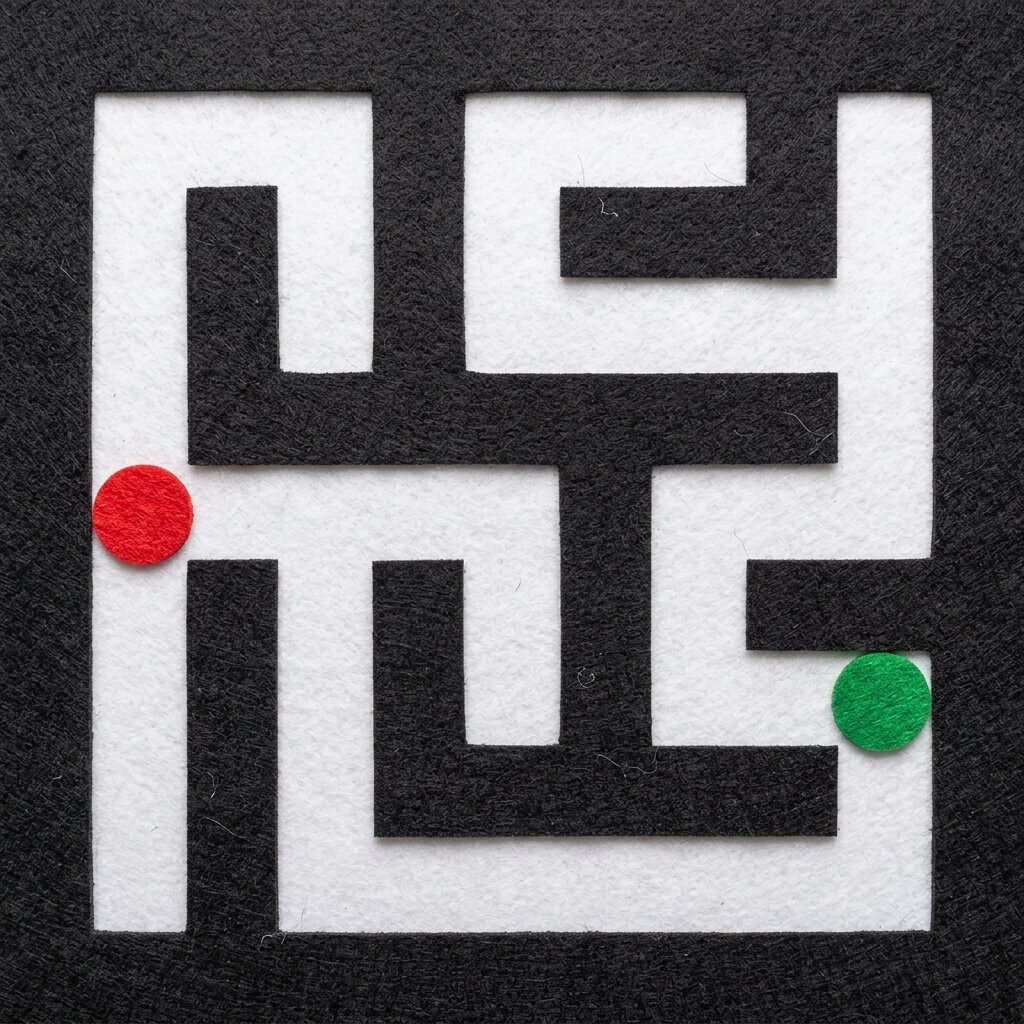}\hfill
  \includegraphics[width=0.19\linewidth]{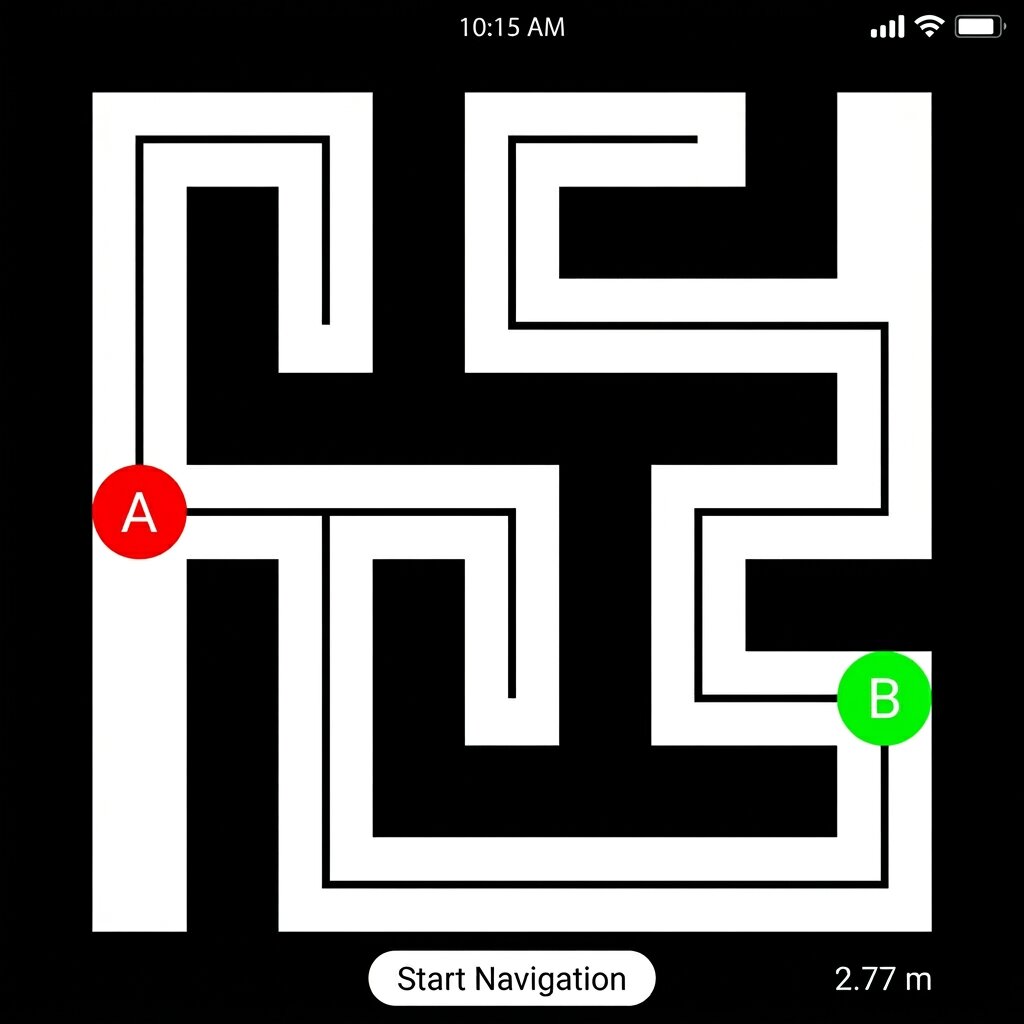}\hfill
  \includegraphics[width=0.19\linewidth]{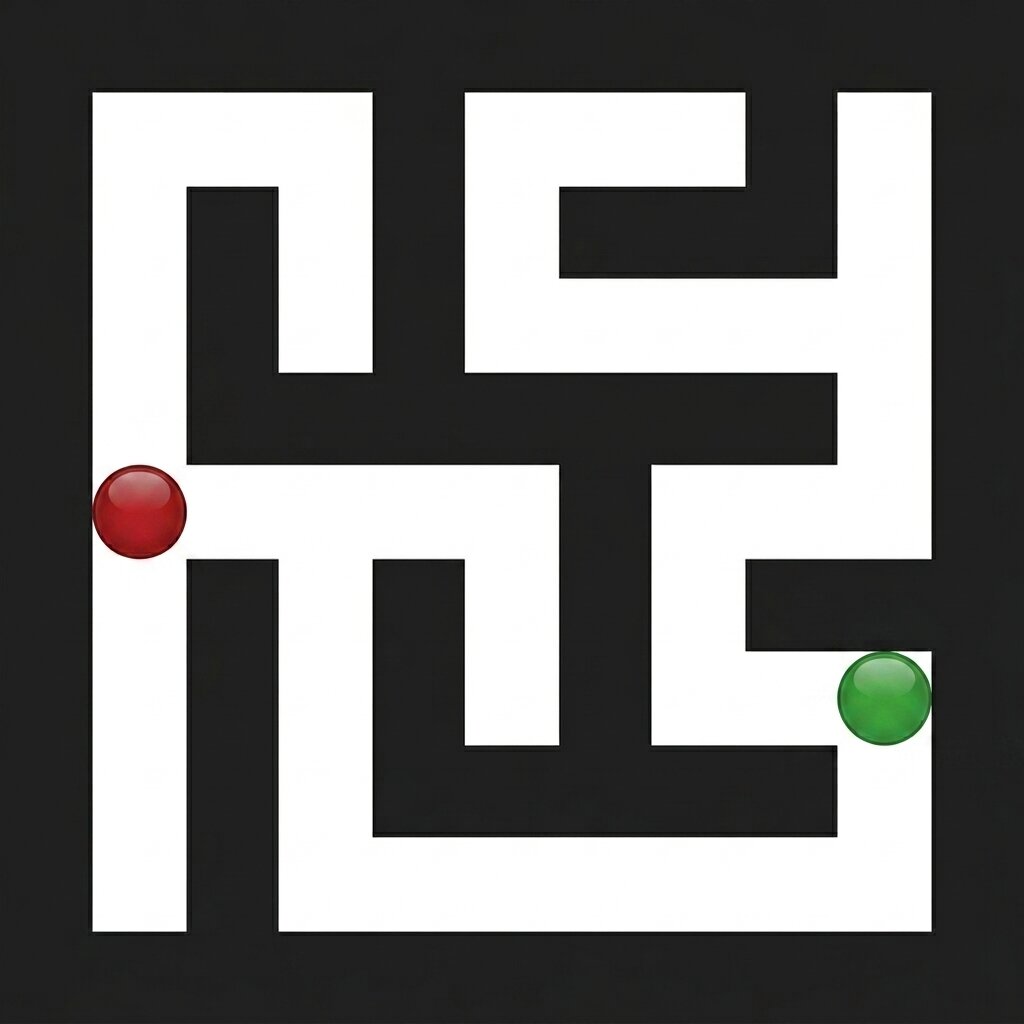}

  \caption{Image variants for \taskname{Maze}.}
  \label{fig:images_maze}
\end{figure*}

We find this iterated approach to be generally effective in avoiding similar visual settings. However, we notice on the \taskname{Maze} task (\cref{fig:images_maze}) that a very detailed base task description that specifies the appearance of many parts of the scene (see below) can constrain the editor's ability to realize proposed visual edits, resulting in only subtle visual changes. Nevertheless, even subtle changes can boost performance, as demonstrated by the ``pixel arcade game'' maze variant shown in~\cref{fig:automated-vipe} which merely modifies the scene to have a low-contrast screen pixel texture.

We also observe that freeform ideation struggles to find a single variant that is effective across task splits (e.g., \taskname{Maze} sizes or \taskname{RushHour} levels).

\begin{wrapfigure}{r}{0.5\textwidth}
    \centering
    \includegraphics[width=\linewidth]{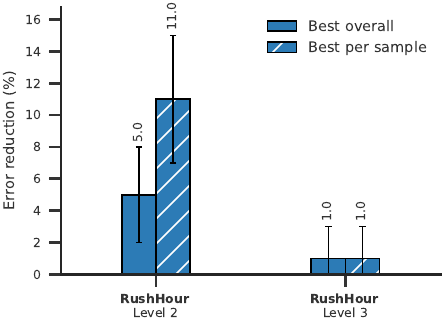}
    \caption{Freeform VIPE results on \taskname{RushHour} for Wan2.2~TI2V.}
    \label{fig:automated-vipe-wan}
\end{wrapfigure}

Finally, we show results for freeform ideation with Wan2.2~TI2V~(see~\cref{app:wan}) in~\cref{fig:automated-vipe-wan}, demonstrating its effectiveness beyond Veo~3.1.

\paragraph{Language vs. visual prompt engineering details}
We base this comparison on the \taskname{Conjunctive Search}, \taskname{Sort 3 Numbers} (\taskname{Sort 3 Numbers} in numerical order), and \taskname{Connect the Dots} tasks, as well as the first 10 images from \taskname{VPCT}. For each task, we collect $n{=}20$ sequential proposals from the text prompt engineering ideator and $n{=}20$ sequential proposals from the image prompt engineering ideator. We used Gemini~3.1~Pro for both ideators. All $n$~proposals are evaluated on the full image set with 10~video attempts per image. Each video is scored against binary pass (0 or 1) by Gemini~3.1~Pro as a VLM autorater.

\subsubsection{Prompts}
\label{sec:freeform-ideation-prompts}

We provide the full prompts used by the freeform prompt engineering ideator described in~\cref{sec:automated-vipe,sec:images-vs-words} below.

\begin{promptbox}{Image ideator prompt}
\textsc{system instruction}

You are an expert AI researcher designing image variants for a visual reasoning benchmark. The video event is fixed---your goal is to explore diverse alternative visual scenes where a video generation model might perform better, while preserving the core reasoning challenge.

Each proposal should be distinct from previous ones. Aim for diversity in visual domains and artistic styles.
\tcbline
\textsc{user content}

Based on the base task and the sample images provided, propose a new visual domain for the task images. The text instruction will NOT be changed.

The variant should:
\begin{itemize}[nosep,leftmargin=*]
\item Preserve spatial structure, element count, and element correspondence
\item Present the task in a visual style where a video generation model might have better priors
\item Be compatible with the fixed text instruction
\end{itemize}

Output your proposal as a JSON block:
\begin{verbatim}
{"variant_id": "short_snake_case_id",
 "variant_description": "Brief description of the visual
  domain change",
 "image_edit_prompt": "Detailed prompt for an image editing
  model to transform the original task images into this
  variant's visual style. Be specific about what each visual
  element should become.",
 "autorater_rubric": "Criteria for evaluating the generated
  video. Describe desired behavior (what correct/partial/
  incorrect outcomes look like) AND undesired behavior to
  penalize (e.g., visual artifacts, glitches, unrealistic
  motion)."}
\end{verbatim}

\begin{itemize}[nosep,leftmargin=0pt]
\item[] Fixed video instruction (do NOT change): \texttt{[base task]}
\item[] Sample images: \texttt{[sample images]}
\item[] Prior proposals: \texttt{[descriptions of all prior proposals]}
\end{itemize}

\end{promptbox}

\begin{promptbox}{Image filter prompt}
\textsc{system instruction}

I will provide an original abstract image, an edit instruction, and multiple attempts at editing the image.
Please evaluate which attempt best follows the edit instruction while preserving the ESSENTIAL STRUCTURE of the original image. Essential structure includes:
\begin{itemize}[nosep]
\item Spatial layout: positions and arrangements of elements
\item Element count: the same number of distinct elements
\item Element correspondence: each element in the original maps to exactly one element in the edit
\end{itemize}
Edit instruction: \texttt{\{edit\_prompt\}}
After reviewing all images, explain your reasoning, then answer with exactly this statement: `For this reason, the \{first / second / third / fourth / ...\} attempt is the most faithful edit.'
\tcbline
\textsc{user content}
\begin{itemize}[nosep, leftmargin=0pt]
\item[] Original image: \texttt{[original image]}
\item[] Attempt 1: \texttt{[edited image 1]}
\item[] Attempt 2: \texttt{[edited image 2]}
\item[] \ldots
\end{itemize}
\end{promptbox}

\begin{promptbox}{Text ideator prompt}
\textsc{system instruction}

You are an expert AI researcher optimizing text prompts for a visual reasoning benchmark. The visual scene is fixed---your goal is to explore diverse alternative video events (what happens to express the answer) where a video generation model might perform better, while preserving the core reasoning challenge.

Each proposal should be distinct from previous ones. Aim for diversity in how the video event is described and framed.
\tcbline
\textsc{user content}

Based on the base task and the sample images provided, propose a new text instruction variant for video generation. The images will NOT be changed.

The variant should:
\begin{itemize}[nosep,leftmargin=*]
\item Describe what should happen in the video without leaking the solution
\item Preserve the core reasoning challenge
\item Consider alternative video events the model may handle better
\end{itemize}

Output your proposal as a JSON block:
\begin{verbatim}
{"variant_id": "short_snake_case_id",
 "variant_description": "Brief description of this instruction
  variant and how it differs from the base",
 "instruction_variant": "Text prompt for the video generation
  model. Describe what event should unfold over time without
  leaking the solution to the task.",
 "autorater_rubric": "Criteria for evaluating the generated
  video. Describe desired behavior (what correct/partial/
  incorrect outcomes look like) AND undesired behavior to
  penalize (e.g., visual artifacts, glitches, unrealistic
  motion)."}
\end{verbatim}

\begin{itemize}[nosep,leftmargin=0pt]
\item[] Base task description: \texttt{[base task]}
\item[] Sample images: \texttt{[sample images]}
\item[] Prior proposals: \texttt{[descriptions of all prior proposals]}
\end{itemize}

\end{promptbox}

\subsubsection{Base task descriptions}
The base task descriptions (corresponding to the original task text prompts) for \taskname{VPCT}, \taskname{Conjunctive Search}, \taskname{Sort 3 Numbers}, and \taskname{Connect the Dots} are detailed in~\cref{tab:prompts_vpct,tab:prompts_conjunctive,tab:prompts_sort3num,tab:prompts_connect_dots}. Task descriptions for \taskname{Maze} and \taskname{RushHour} are detailed below.

\begin{promptbox}{Maze task description}
    Create a 2D animation based on the provided image of a maze. The red circle slides smoothly along the white path, stopping perfectly on the green circle. The red circle never slides or crosses into the black areas of the maze. The camera is a static, top-down view showing the entire maze.
    
    \taskname{Maze}:
    \begin{itemize}[nosep,leftmargin=*]
        \item The maze paths are white, the walls are black.
        \item The red circle moves to the goal position, represented by a green circle.
        \item The red circle slides smoothly along the white path.
        \item The red circle never slides or crosses into the black areas of the maze.
        \item The red circle stops perfectly on the green circle.
    \end{itemize}

    Scene:
    \begin{itemize}[nosep,leftmargin=*]
        \item No change in scene composition.
        \item No change in the layout of the maze.
        \item The red circle travels along the white path without speeding up or slowing down.
    \end{itemize}
     
     Camera:
    \begin{itemize}[nosep,leftmargin=*]
        \item Static camera.
        \item No zoom.
        \item No pan.
        \item No glitches, noise, or artifacts.
    \end{itemize}
\end{promptbox}

\begin{promptbox}{RushHour task description}
    Look at this image:
    
    Below is the first image, showing the initial configuration of a congested parking lot. Each colored rectangle represents a vehicle, and the red car is the one that must reach the exit. The exit is marked with a green area at the border:
    
    Rules:
    \begin{enumerate}[nosep,leftmargin=*]
        \item Each vehicle can only move forward or backward with straight sliding motion along its own axis.
        \item No rotation is allowed at any time.
        
        \item A vehicle continues to move in the chosen direction until it touches another vehicle or a boundary.
        
        \item Only one vehicle moves per action.
        
        \item The goal is for the red car to reach the exit located on the edge of the grid.
        
        \item Vehicle shapes, colors, exit, and outlines must not change throughout the solution.
        
        \item No camera motion: no zoom, no pan, no rotate, no tilt, no dolly.
        
        \item Do not add or remove anything: no new objects, labels, lights, shadows, reflections, textures, markings, or UI elements.
        
        \item The background, grid, exit, and all pieces remain perfectly static, except for the piece currently sliding.
    \end{enumerate}
    
     Task: Plan the minimal sequence of moves needed to free the red car and allow it to exit the parking lot.
     
     Output: A video demonstrating the full solution to the puzzle, one move at a time.
\end{promptbox}

\subsubsection{Example proposals from freeform prompt engineering}
\label{sec:freeform-more-variants}

In \taskname{VPCT}, \taskname{Conjunctive Search}, \taskname{Sort 3 Numbers}, and \taskname{Connect the Dots}, the freeform ideation pipeline proposes text variants based on the original prompts listed below, and does not specifically ideate on camera instructions. Instead, we append "Static shot, no zoom or pan." to the original and generated prompts before sending to the video generation model.

\FloatBarrier

\begin{table}[h]
\centering
\small
\caption{Text prompt variants for \taskname{VPCT}.}
\label{tab:prompts_vpct}
\begin{tabular}{p{\linewidth}}
\toprule
\textbf{Prompt Text} \\
\midrule
Original: The ball moves down in a physically plausible way, sliding down the obstacles, and ends up in one of the three containers at the bottom. \\
\midrule
Simulate 2D downward gravity. The white circle drops, collides with the angled black lines, rolls along them, and settles into one of the bottom receptacles. \\
Simulate a gravity drop test on this schematic. The white circle must fall, interact rigidly with the drawn lines, and settle into one of the bottom containers. \\
Animate this cross-section: the white boulder drops, rolls down the slanted black cliff ledges under gravity, and settles into one of the three valley basins below. \\
Animate this side-view transparent coin box. The white round coin falls, rolling and sliding naturally down the slanted lines. Apply accurate rigid-body gravity and collision physics until it drops into one of the three bottom bins. \\
Animate this gravity pinball setup. The white sphere drops, deflecting off the angled barriers under realistic gravity, until it settles into the correct bottom receptacle. \\
Animate this internal gumball machine mechanism. The white ball drops, rolls down the angled black ramps, and settles into one of the three bottom collection trays. \\
Animate the white circle as a hailstone falling through slanted gutters. It must roll down the black lines under gravity and drop into one of the bottom basins. \\
A spherical payload drops into the industrial sorting chute. Gravity pulls it down, causing it to roll along and deflect off the angled guide rails until it settles into one of the collection bins below. \\
Animate this kinetic desk toy. The solid white sphere drops, rolling smoothly down the angled black barriers under natural gravity, and settles into one of the three bottom catch-basins. \\
Animate this 2D schematic. Apply downward gravity to the white circle. It must fall, bounce, and roll along the static black lines until coming to rest inside one of the three bottom receptacles. \\
\bottomrule
\end{tabular}
\end{table}

\begin{figure*}[h]
  \centering
  \includegraphics[width=0.19\linewidth]{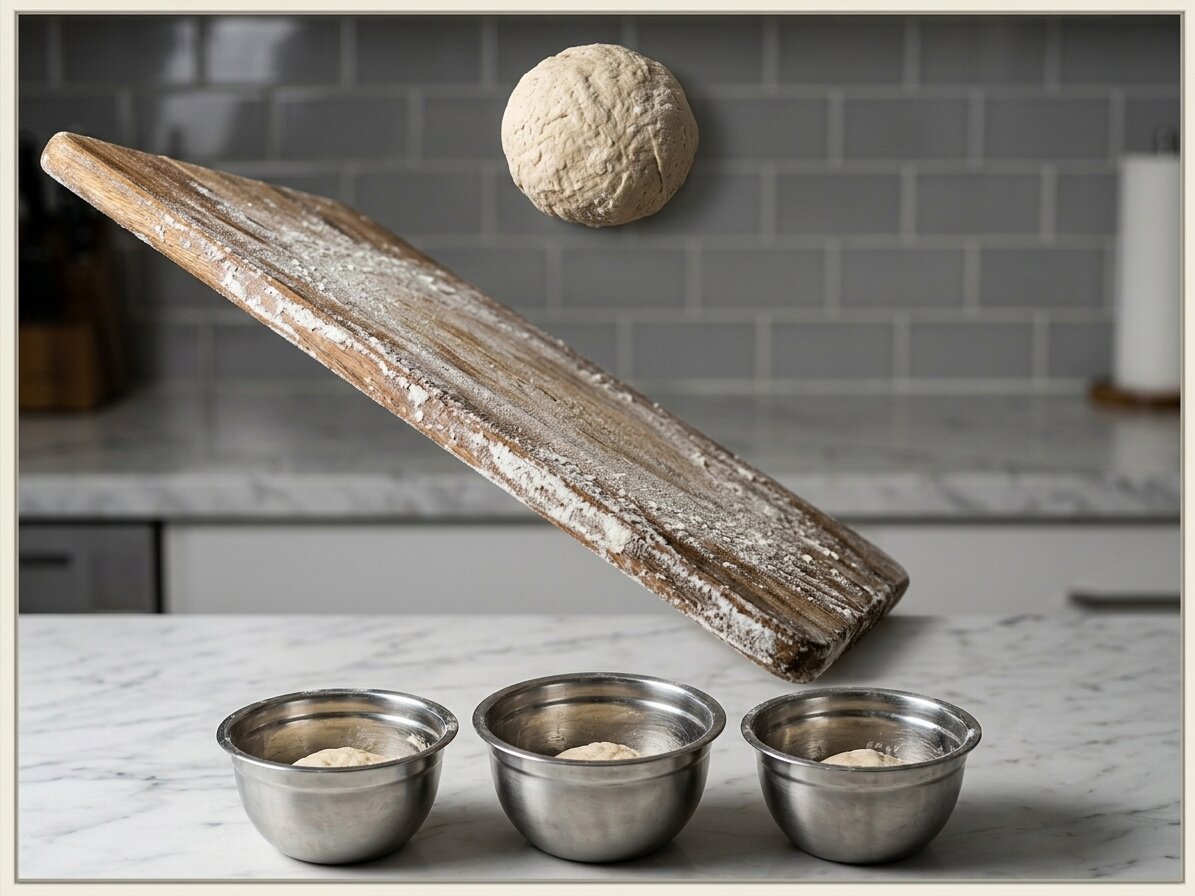}\hfill
  \includegraphics[width=0.19\linewidth]{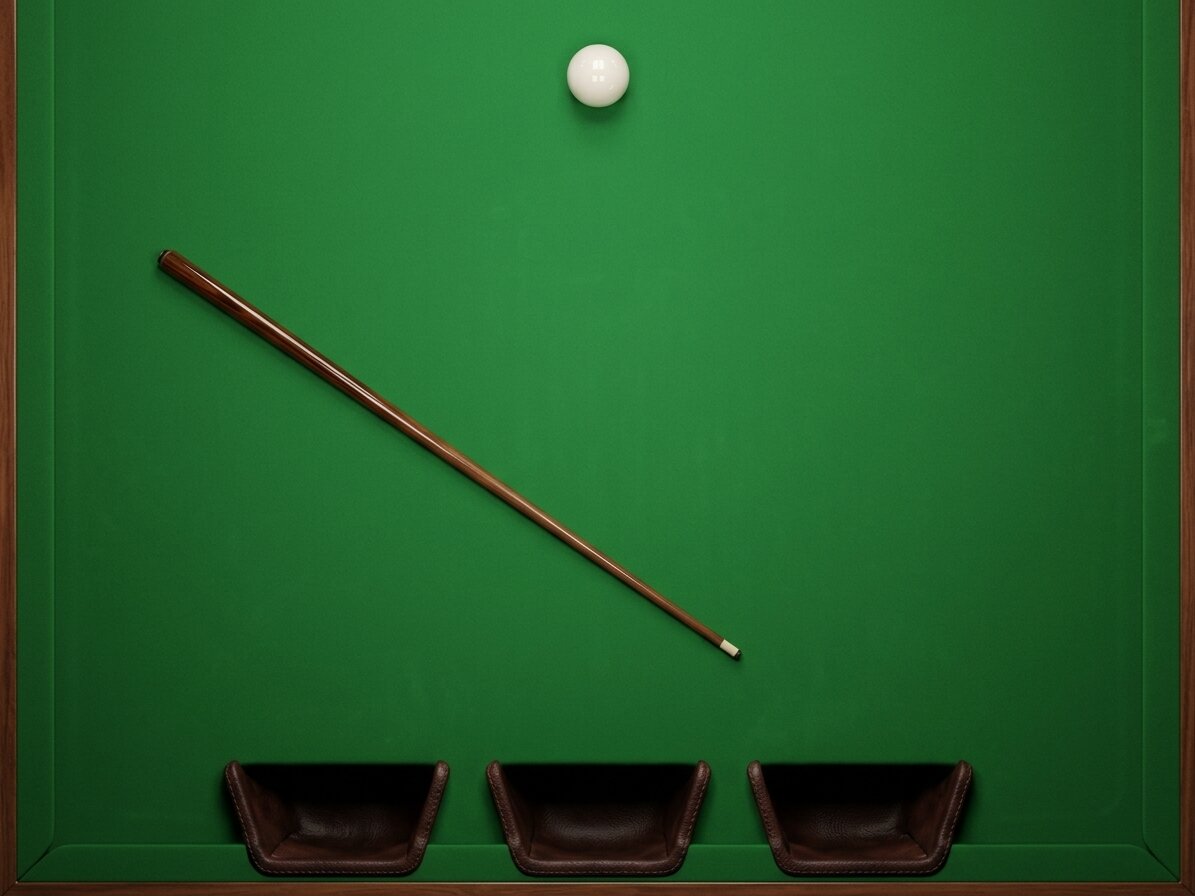}\hfill
  \includegraphics[width=0.19\linewidth]{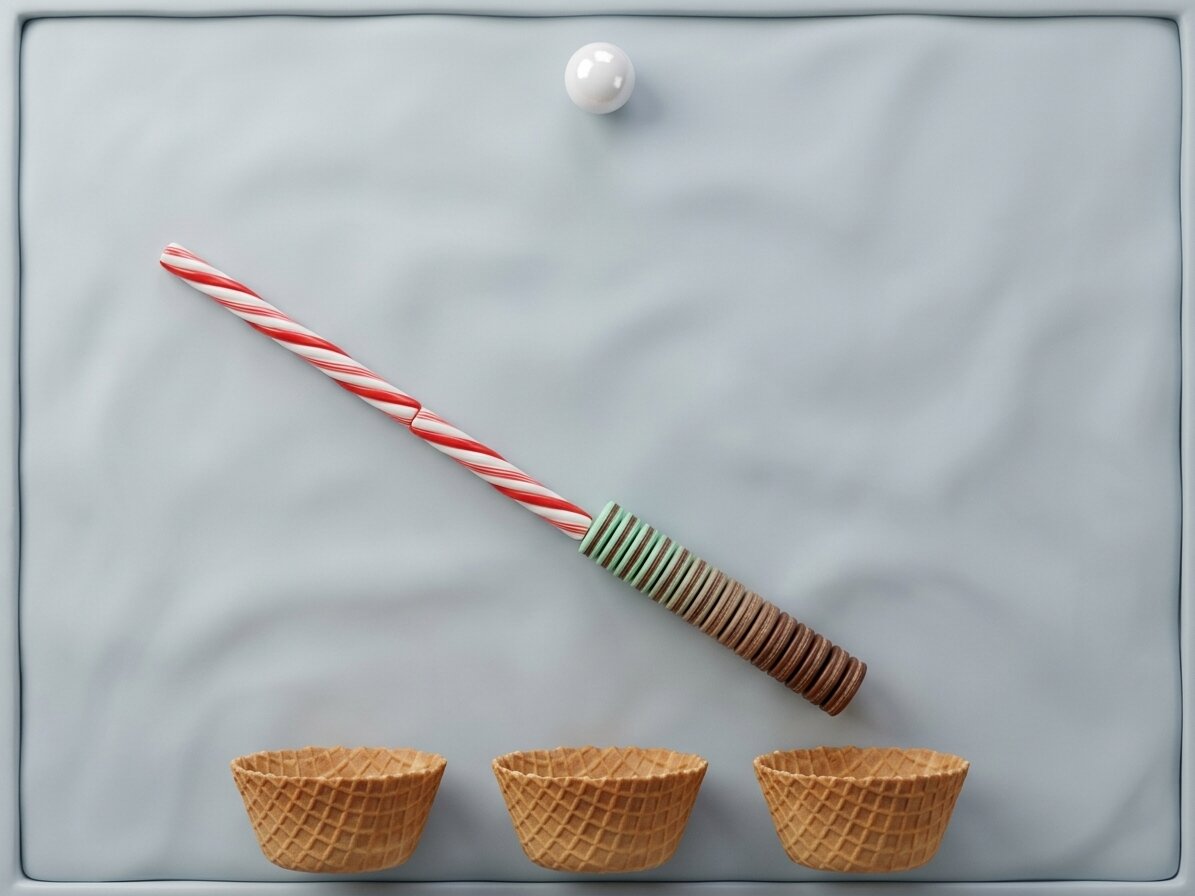}\hfill
  \includegraphics[width=0.19\linewidth]{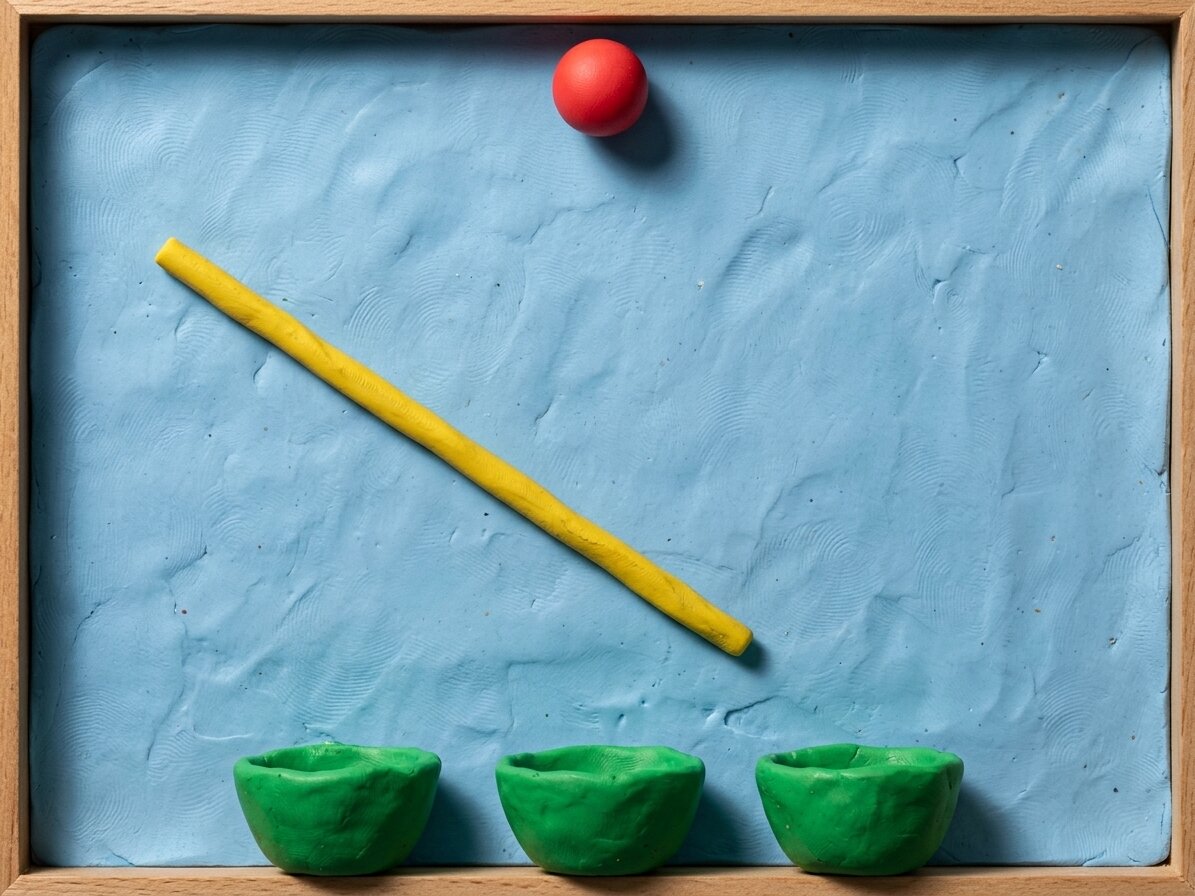}\hfill
  \includegraphics[width=0.19\linewidth]{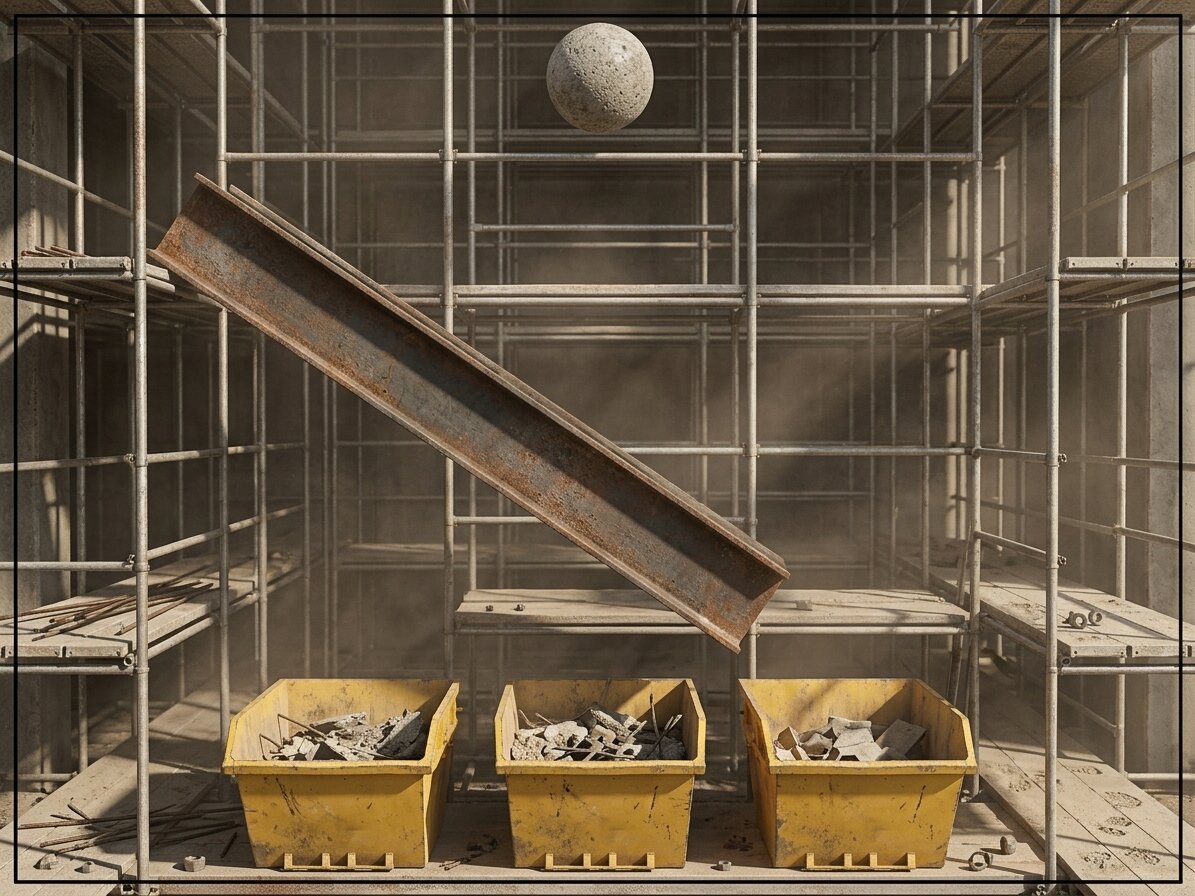}
  
  \vspace{10pt} %

  \includegraphics[width=0.19\linewidth]{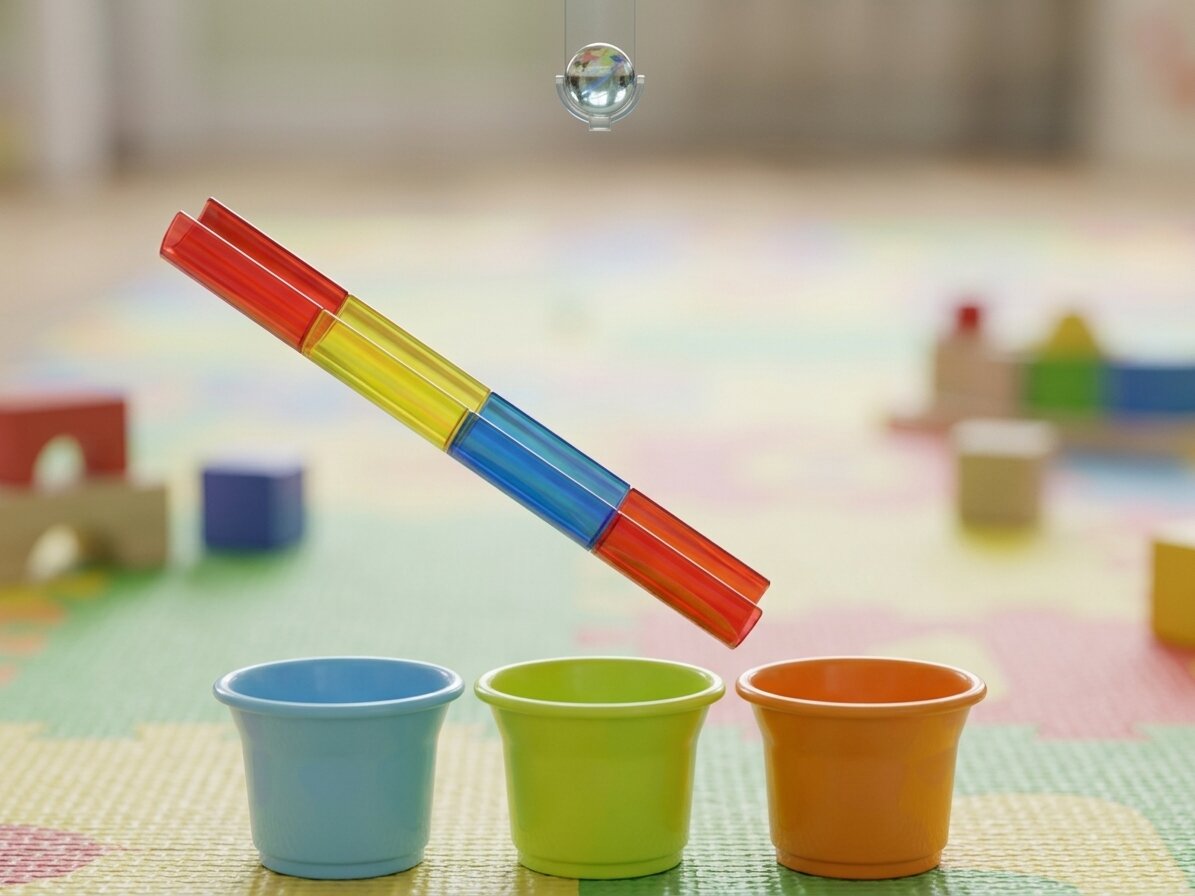}\hfill
  \includegraphics[width=0.19\linewidth]{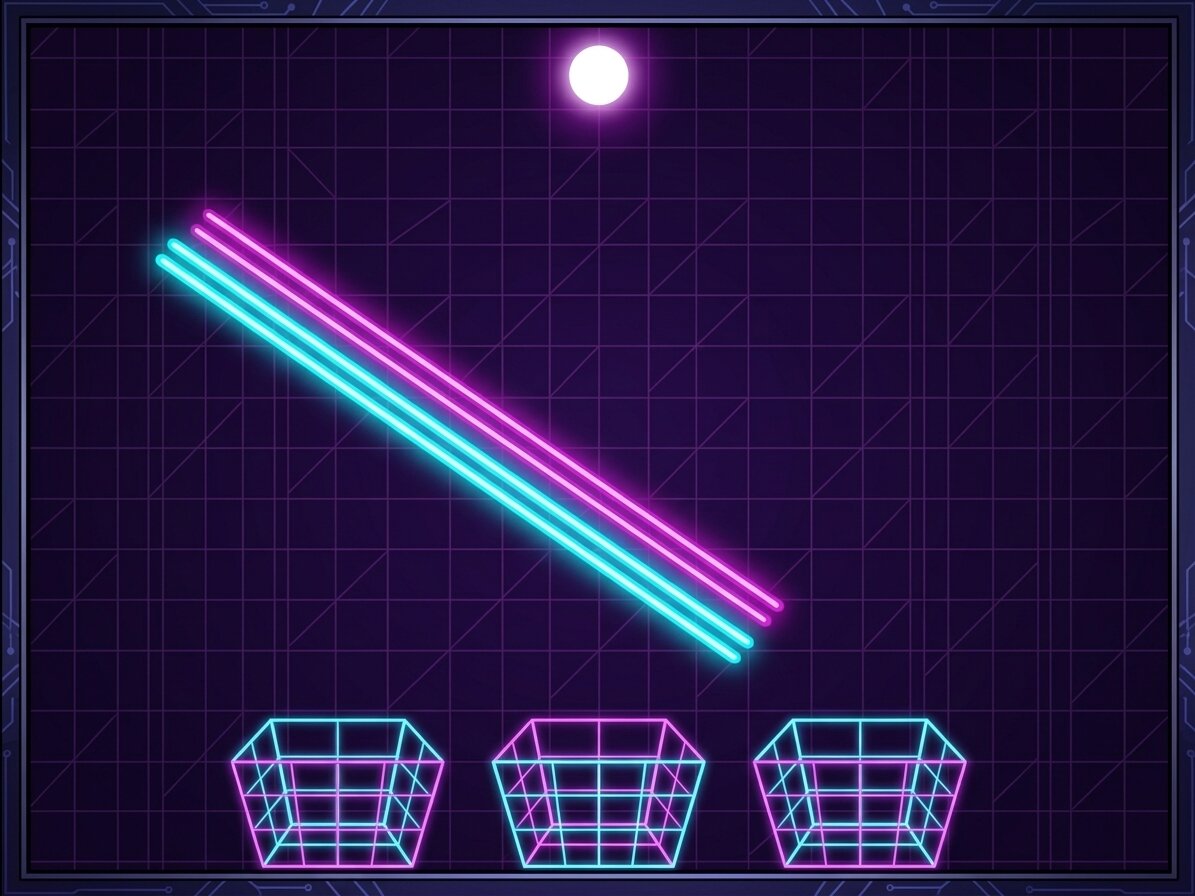}\hfill
  \includegraphics[width=0.19\linewidth]{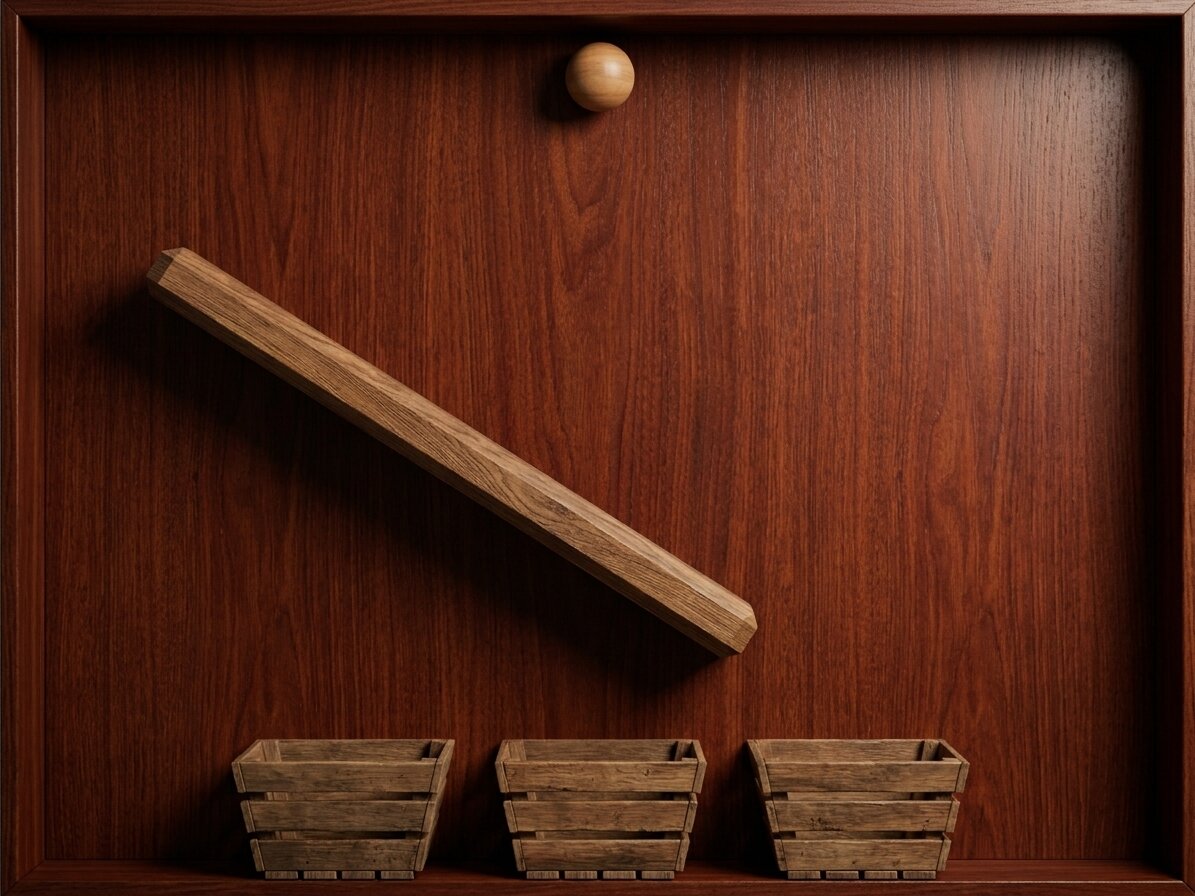}\hfill
  \includegraphics[width=0.19\linewidth]{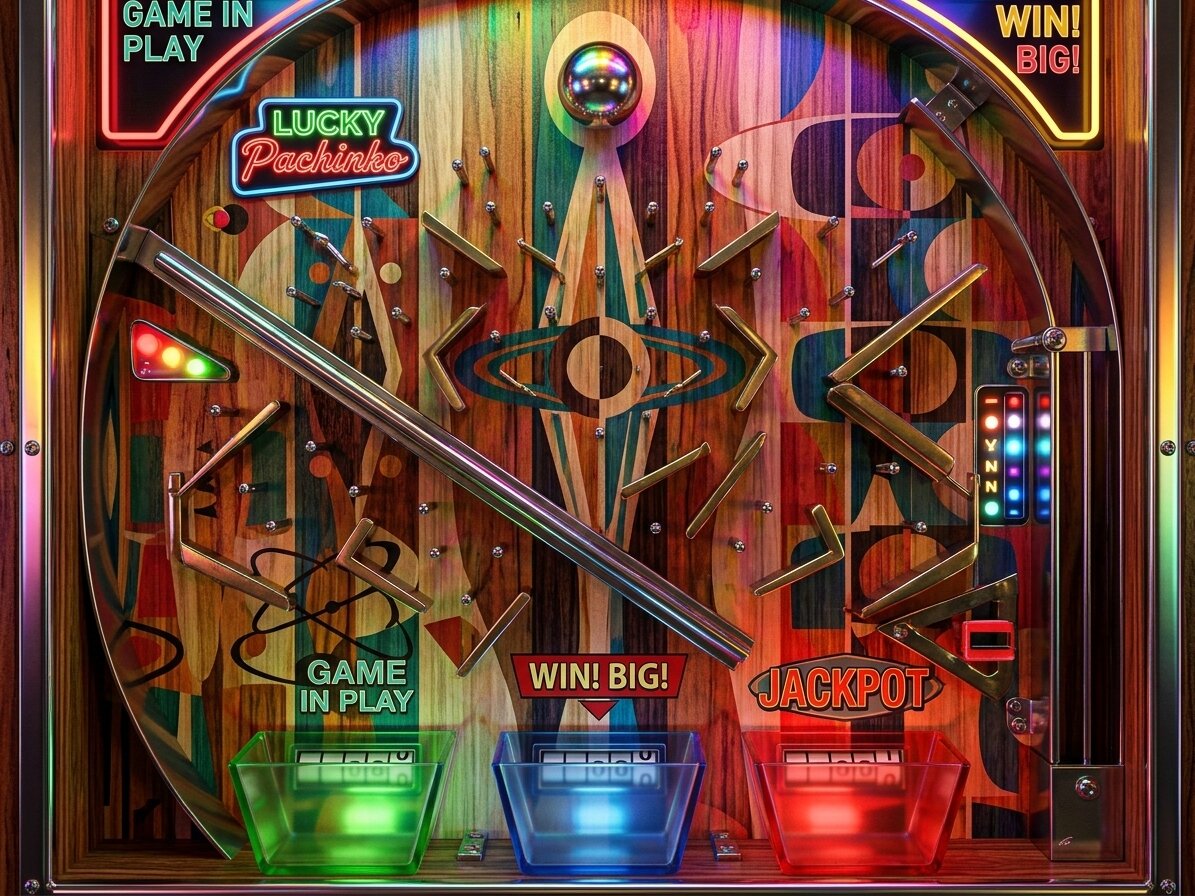}\hfill
  \includegraphics[width=0.19\linewidth]{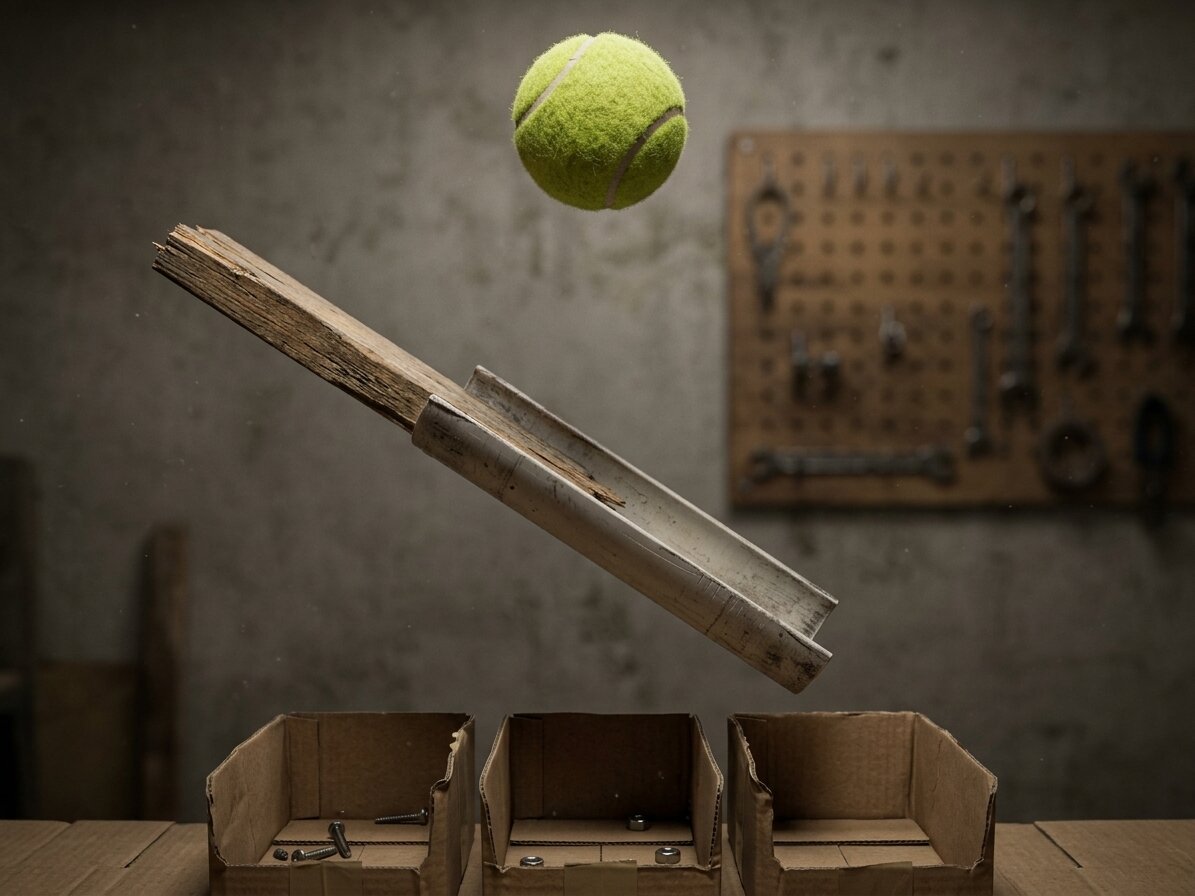}

  \caption{Image variants for \taskname{VPCT}.}
  \label{fig:images_vpct}
\end{figure*}

\begin{table}[h]
\centering
\small
\caption{Text prompt variants for \taskname{Conjunctive Search}.}
\label{tab:prompts_conjunctive}
\begin{tabular}{p{\linewidth}}
\toprule
\textbf{Prompt Text} \\
\midrule
Original: The two blue balls begin to glow. \\
\midrule
The two blue balls become increasingly blurry until they are completely out of focus. \\
The two blue balls bounce continuously up and down. \\
The two blue balls turn bright green. \\
A black square outline is drawn around each of the two blue balls. \\
A solid black line grows to connect the two blue balls. \\
Smoke begins to billow from the two blue balls. \\
The two blue circles slowly expand to twice their original size. \\
The two blue balls slowly fade away until they completely disappear. \\
The two blue balls flip over repeatedly like spinning coins. \\
A glowing glowing golden ring appears and continuously orbits around each of the two blue balls. \\
\bottomrule
\end{tabular}
\end{table}

\begin{figure*}[h]
  \centering
  \includegraphics[width=0.19\linewidth]{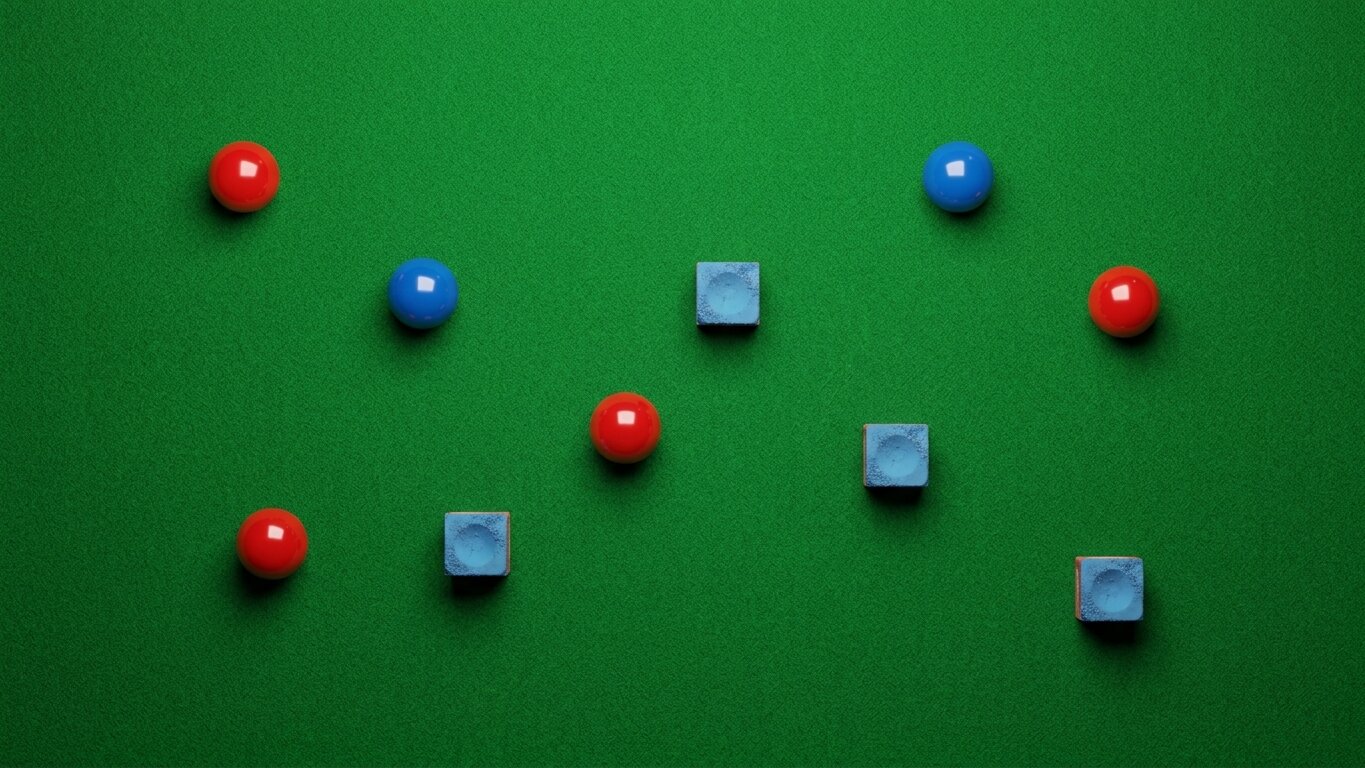}\hfill
  \includegraphics[width=0.19\linewidth]{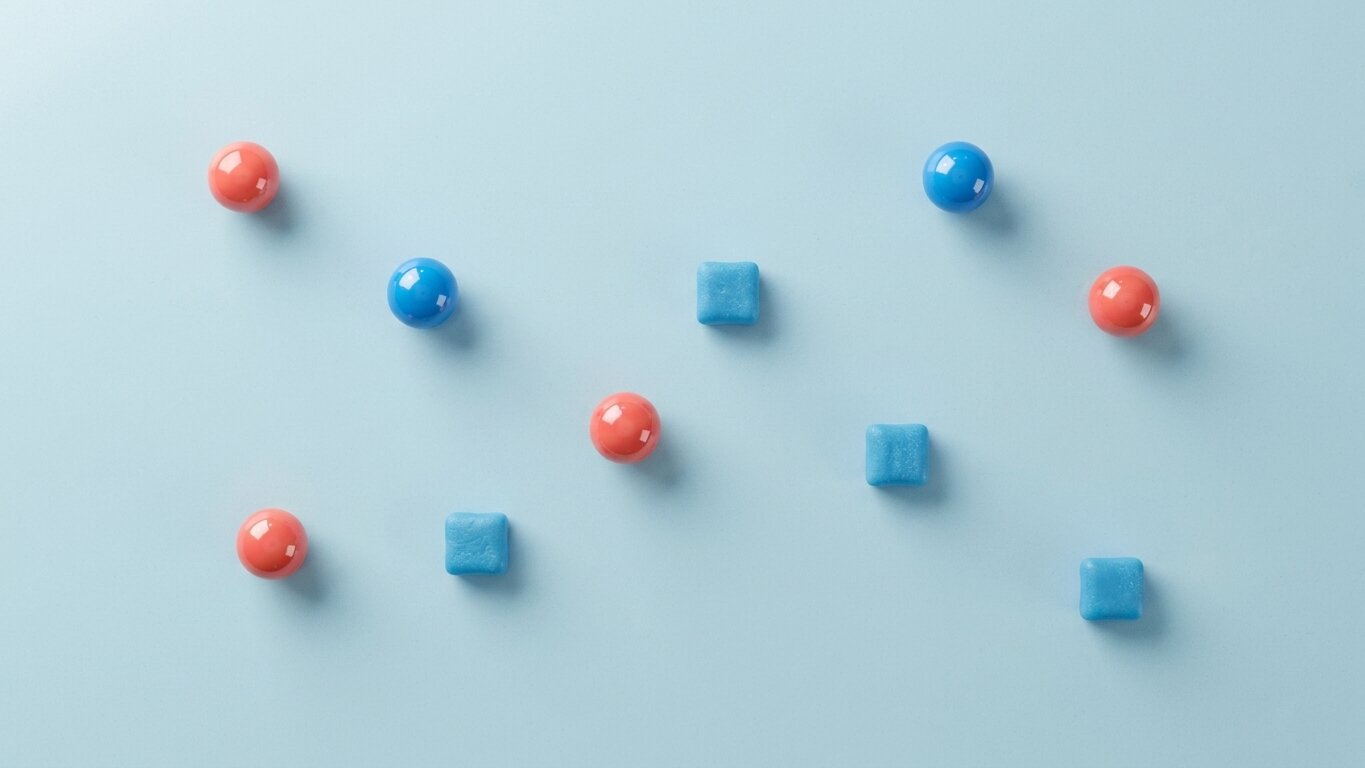}\hfill
  \includegraphics[width=0.19\linewidth]{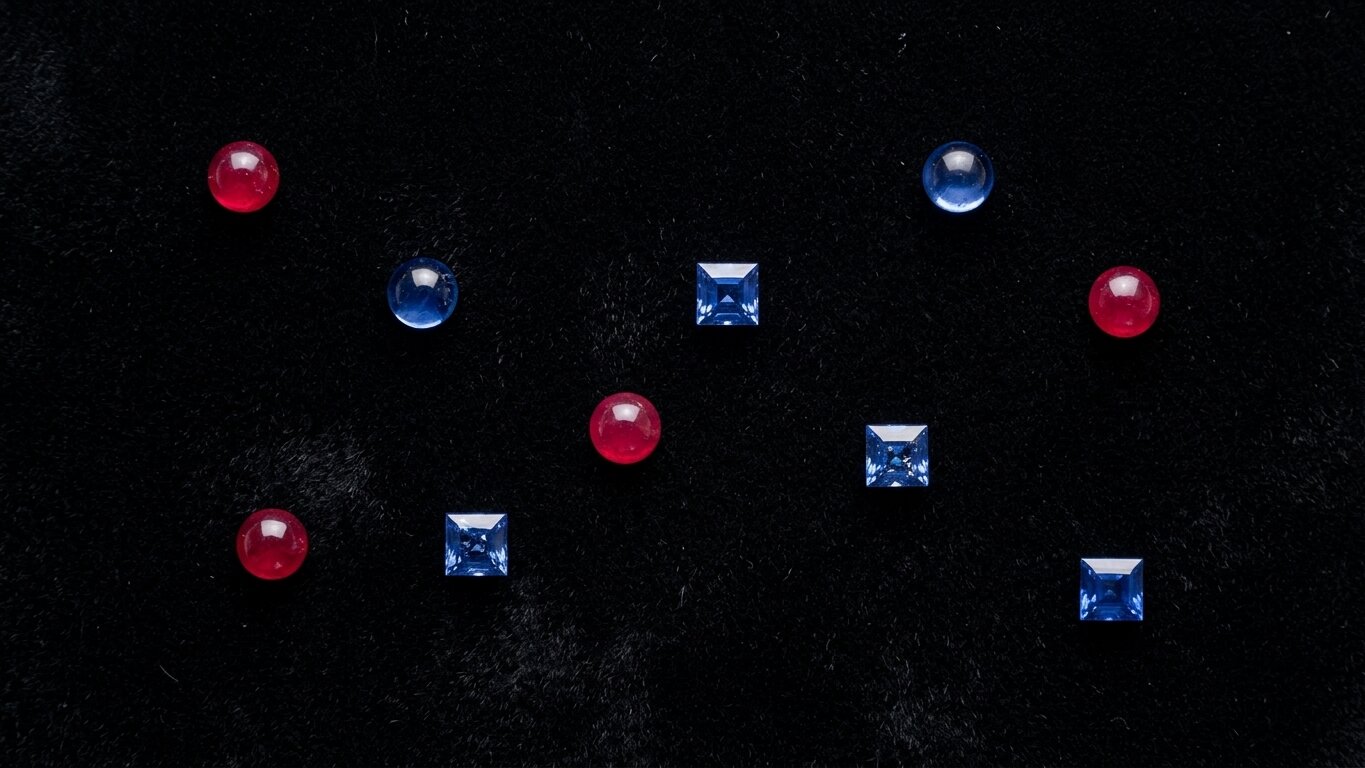}\hfill
  \includegraphics[width=0.19\linewidth]{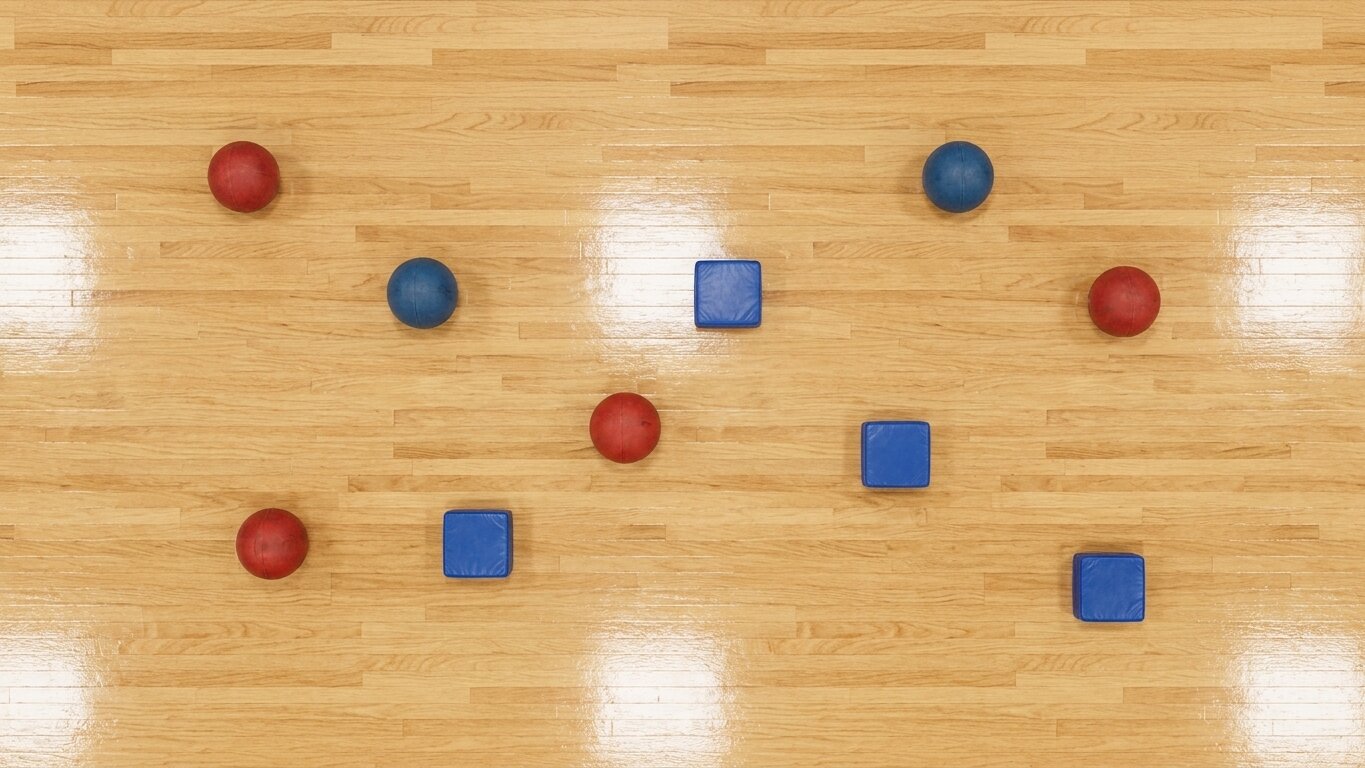}\hfill
  \includegraphics[width=0.19\linewidth]{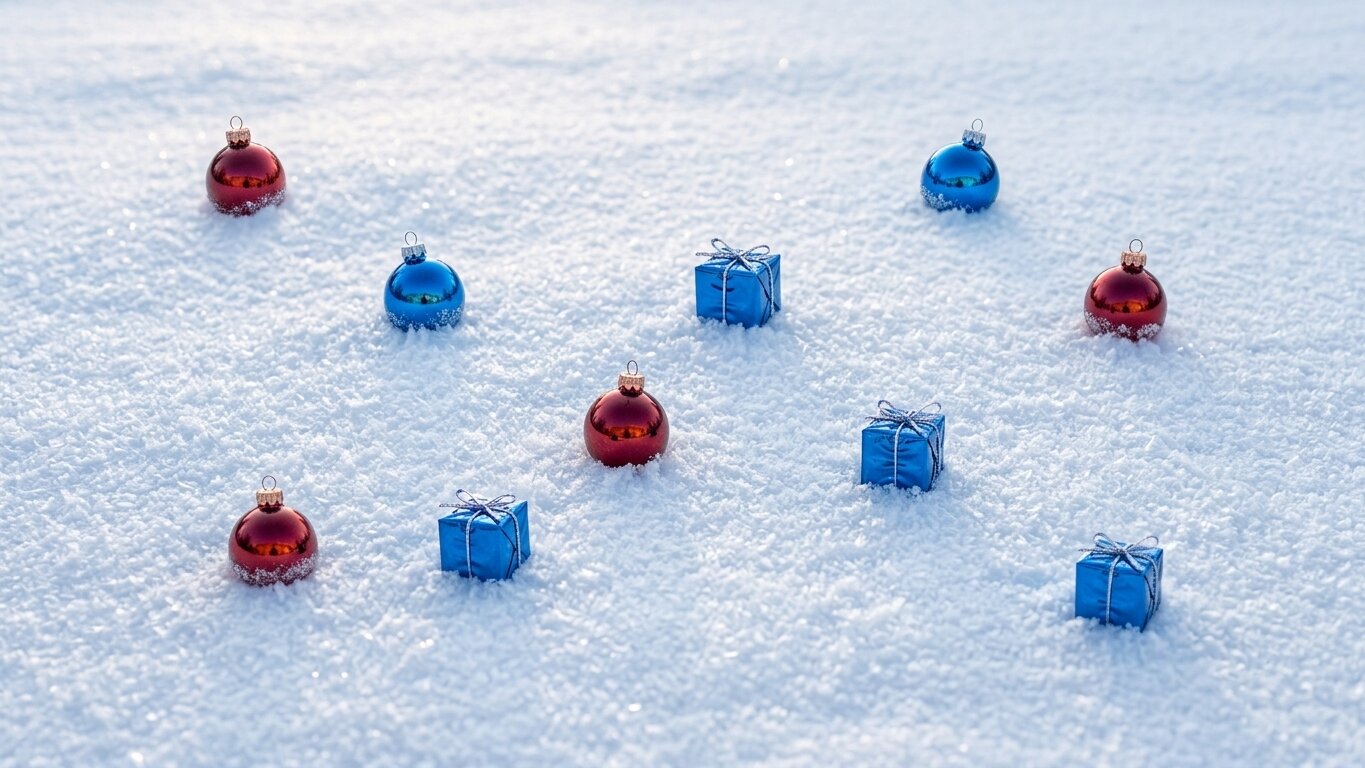}
  
  \vspace{10pt} %

  \includegraphics[width=0.19\linewidth]{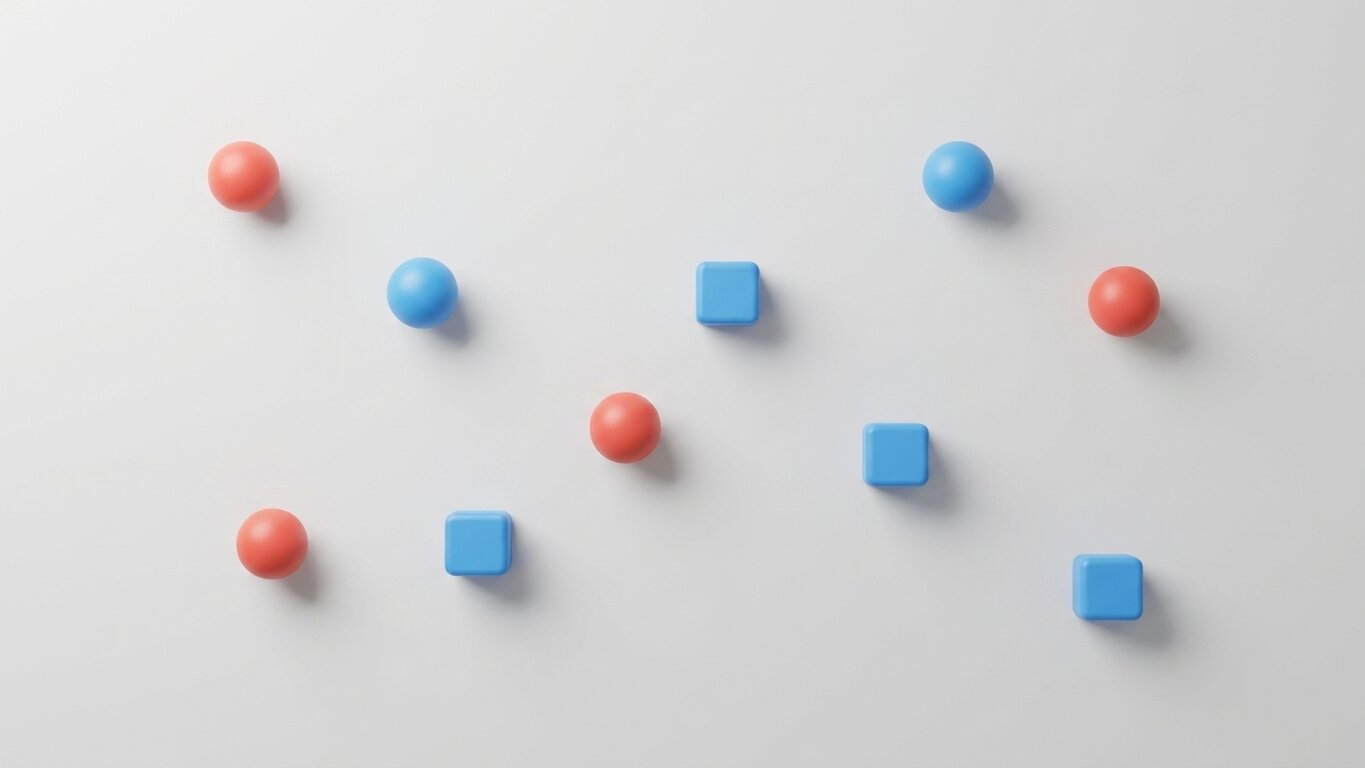}\hfill
  \includegraphics[width=0.19\linewidth]{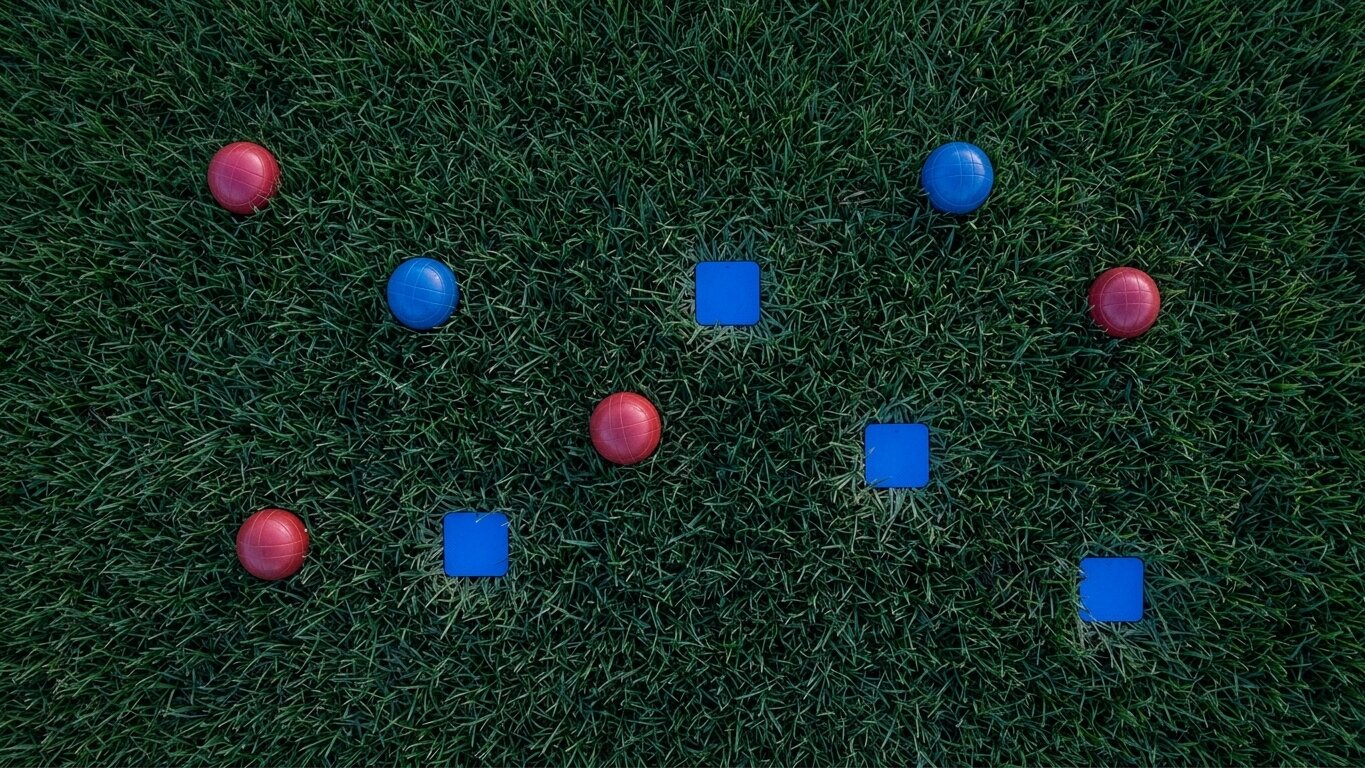}\hfill
  \includegraphics[width=0.19\linewidth]{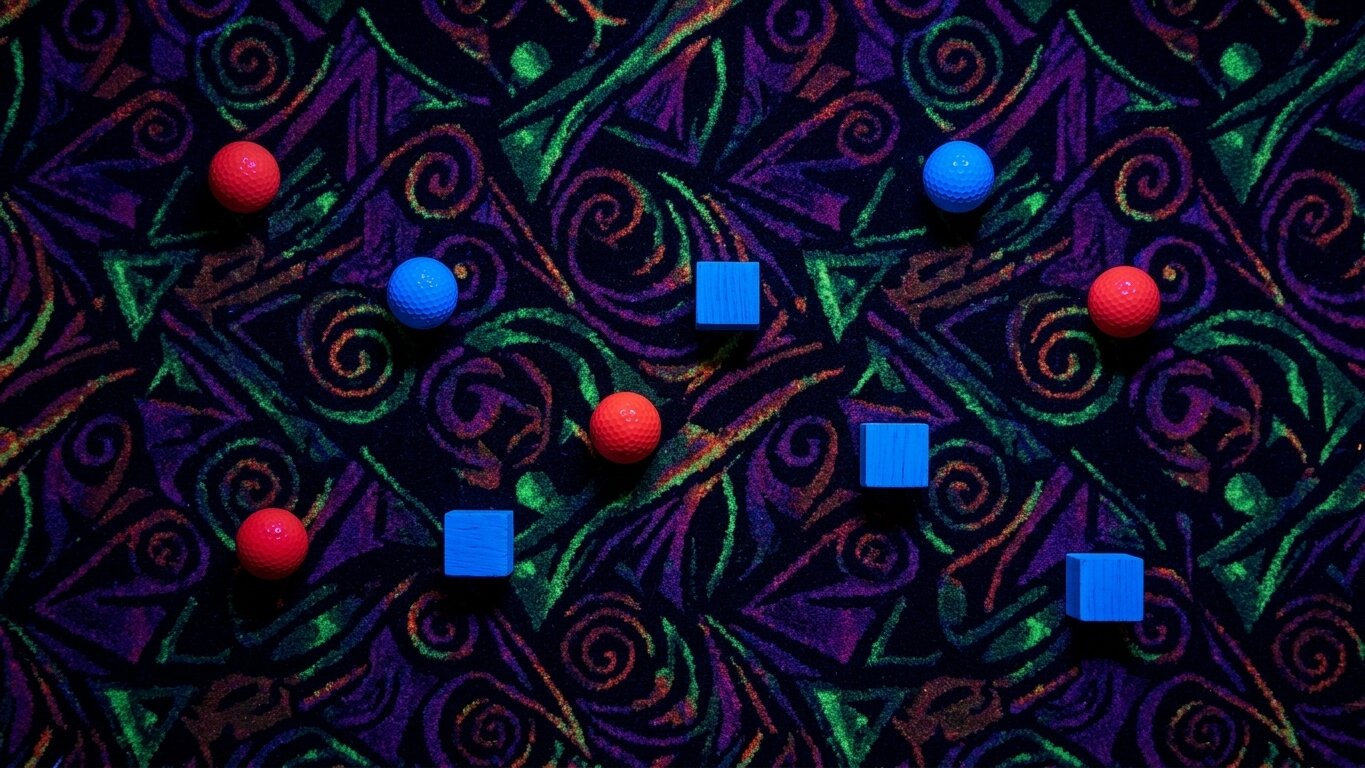}\hfill
  \includegraphics[width=0.19\linewidth]{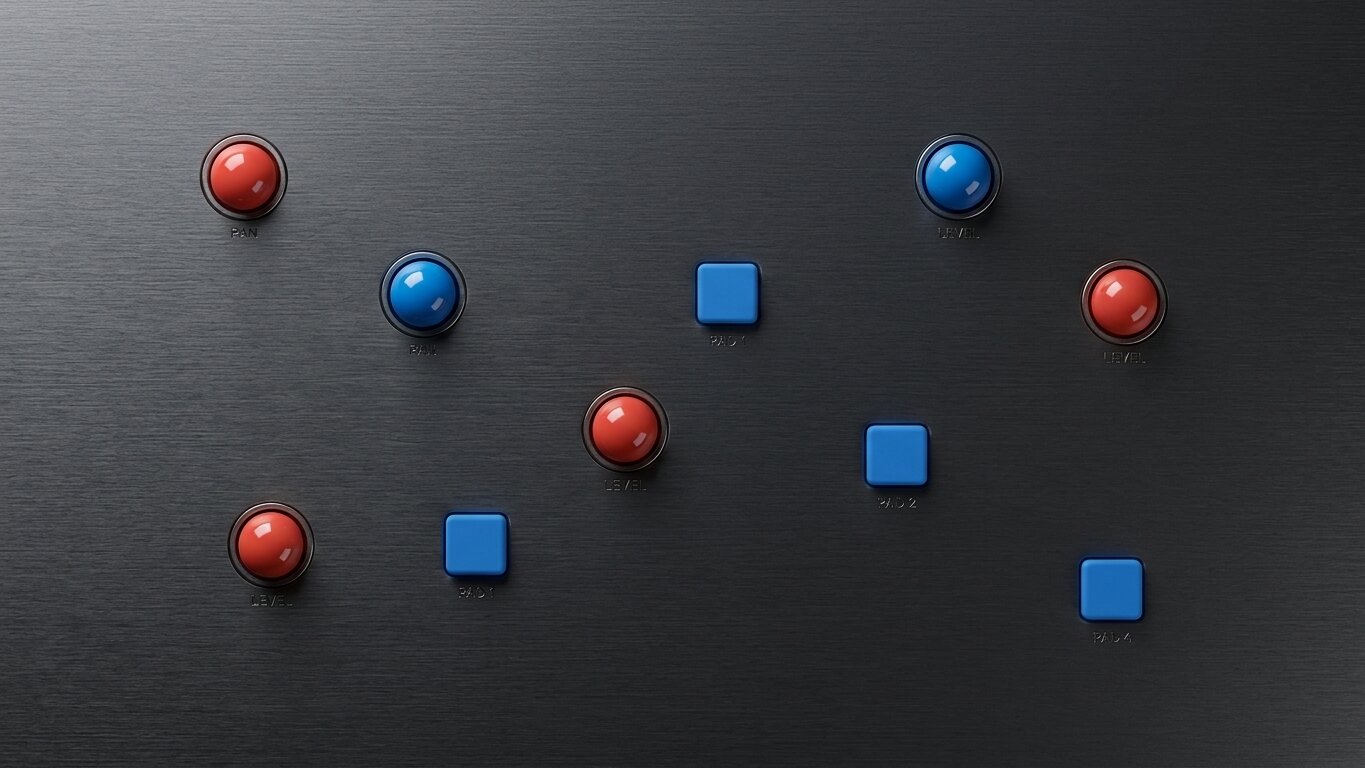}\hfill
  \includegraphics[width=0.19\linewidth]{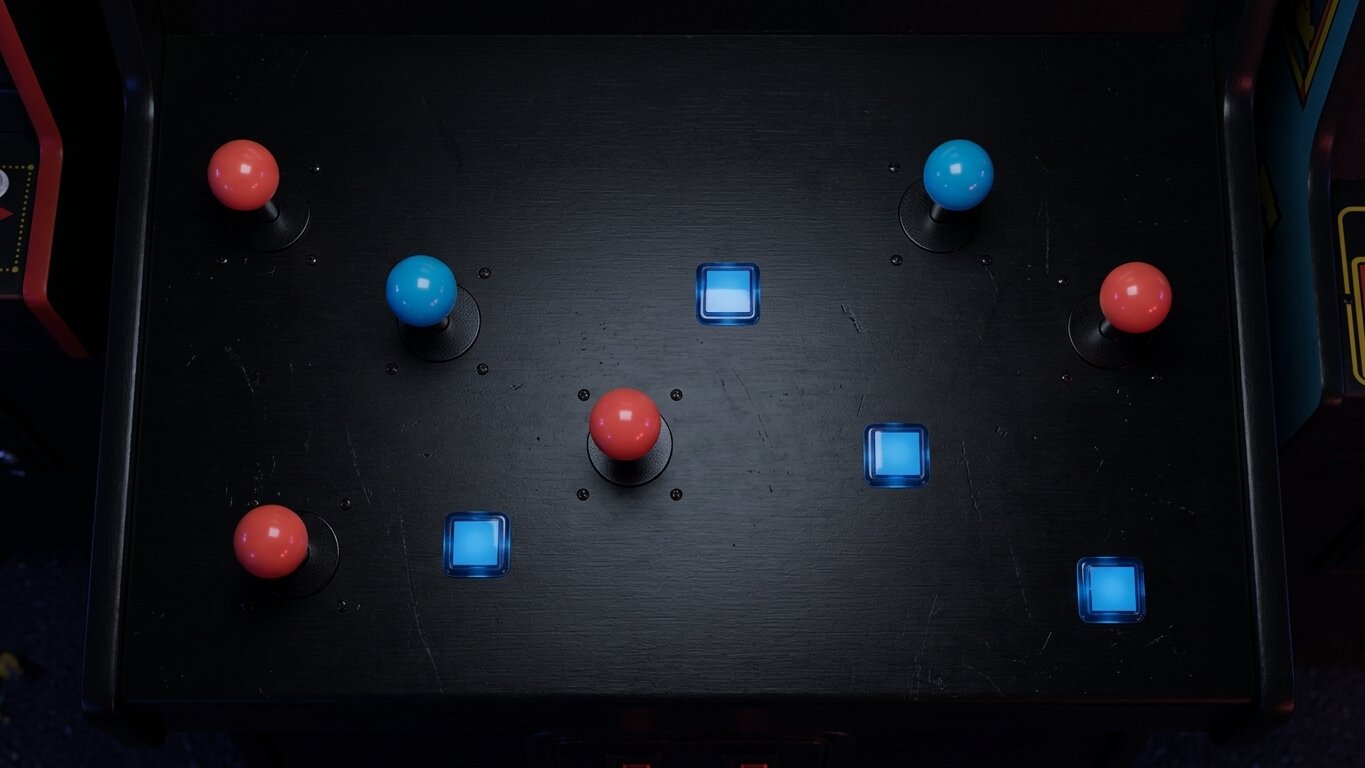}

  \caption{Image variants for \taskname{Conjunctive Search}.}
  \label{fig:images_conjunctive}
\end{figure*}

\begin{table}[h]
\centering
\small
\caption{Text prompt variants for \taskname{Sort 3 Numbers}.}
\label{tab:prompts_sort3num}
\begin{tabular}{p{\linewidth}}
\toprule
\textbf{Prompt Text} \\
\midrule
Original: The numbers pop and disappear one at a time, in numeric order, starting from the smallest one. \\
\midrule
The numbers become heavily blurred and unreadable one at a time, in numeric order, starting from the smallest one. \\
The numbers perform a single vertical jump and land back in place, one at a time, in numeric order, starting from the smallest. \\
The numbers sequentially catch fire and burn away, starting from the lowest value and increasing. \\
A red circle is drawn around the numbers one at a time, in numeric order, starting from the smallest one. \\
Make the numbers turn bright green one at a time, following strictly ascending numerical order. \\
The numbers fall down off the bottom edge of the screen one at a time, in numeric order, starting from the smallest. \\
The numbers flip horizontally 180 degrees in 3D space one at a time, in numeric order, starting from the smallest one. \\
The numbers experience a brief digital glitch and RGB color split one at a time, in numeric order, starting from the smallest one. \\
The numbers emit a bright glow one at a time, in numeric order, starting from the smallest one. \\
Draw a continuous solid line connecting the numbers in ascending numeric order, starting at the smallest and ending at the largest. The numbers themselves must not move. \\
\bottomrule
\end{tabular}
\end{table}

\begin{figure*}[h]
  \centering
  \includegraphics[width=0.19\linewidth]{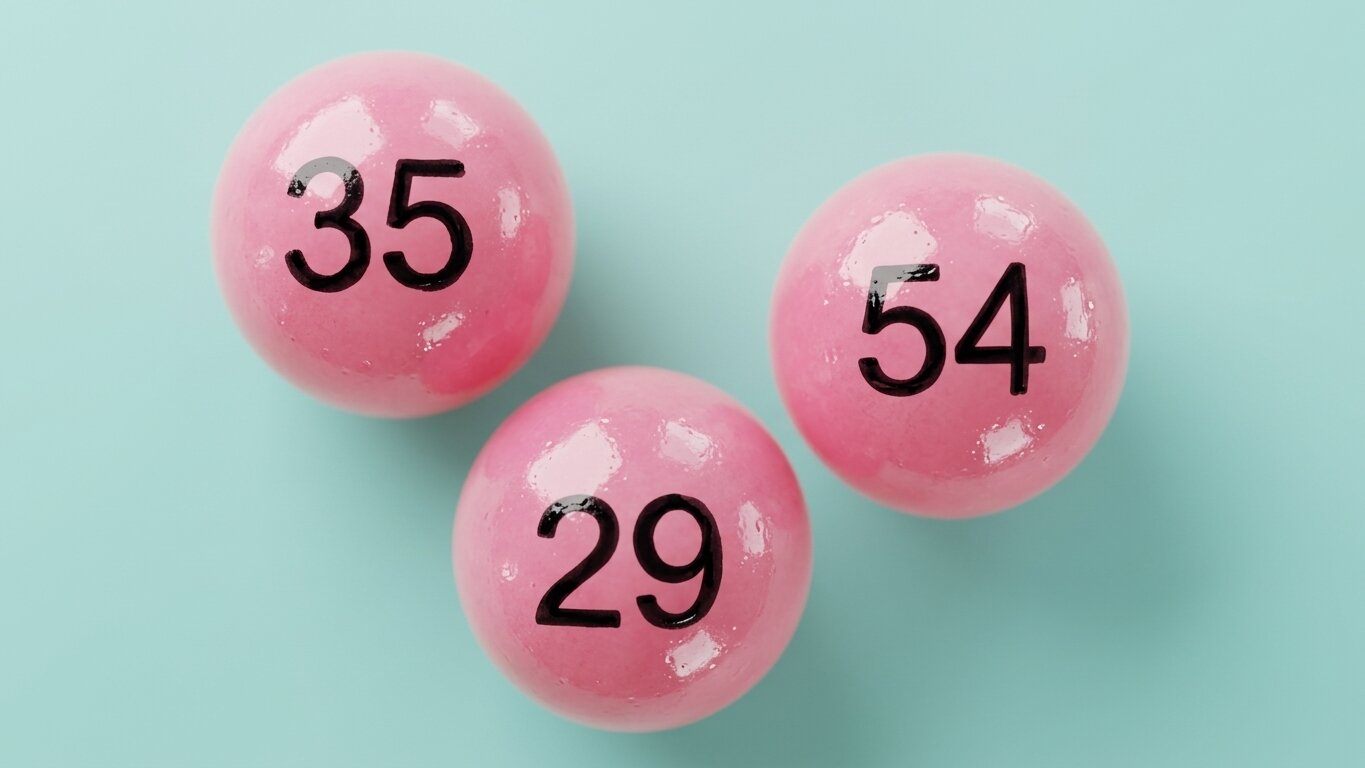}\hfill
  \includegraphics[width=0.19\linewidth]{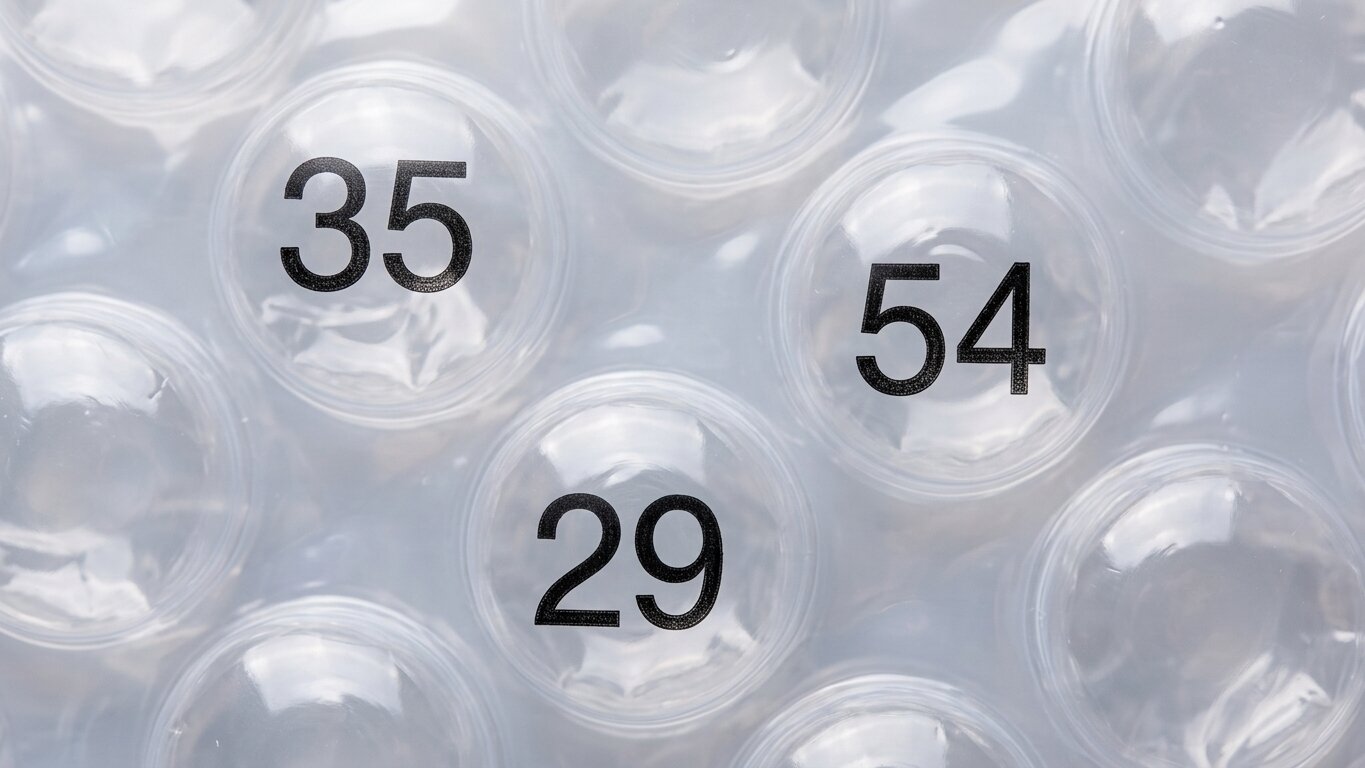}\hfill
  \includegraphics[width=0.19\linewidth]{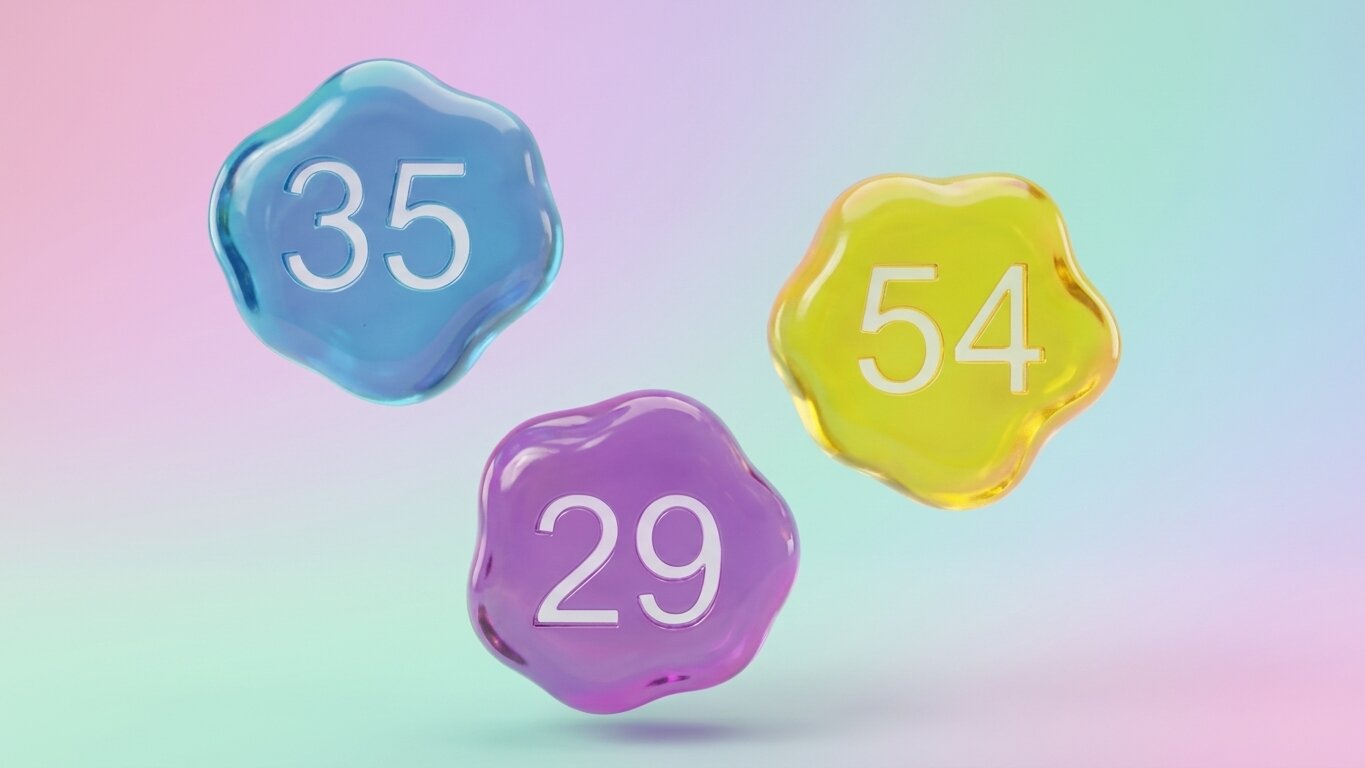}\hfill
  \includegraphics[width=0.19\linewidth]{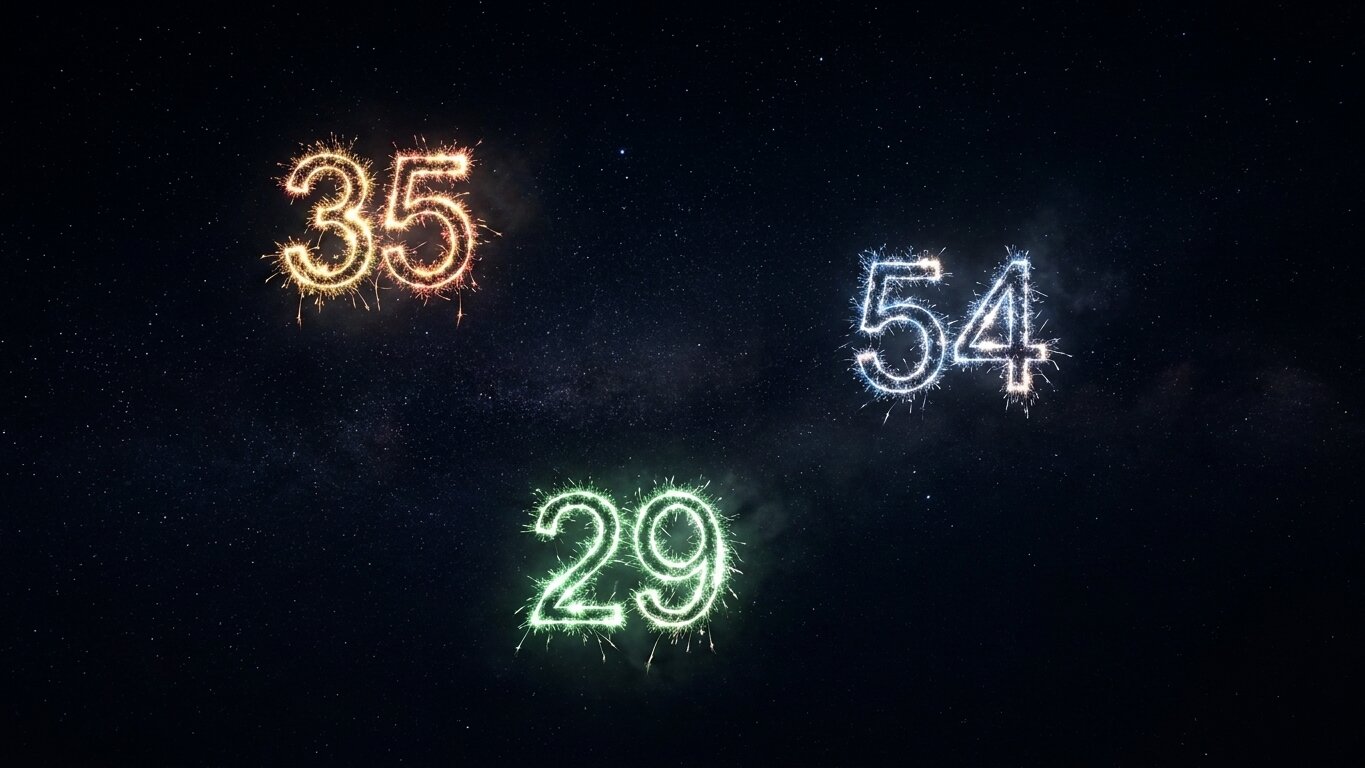}\hfill
  \includegraphics[width=0.19\linewidth]{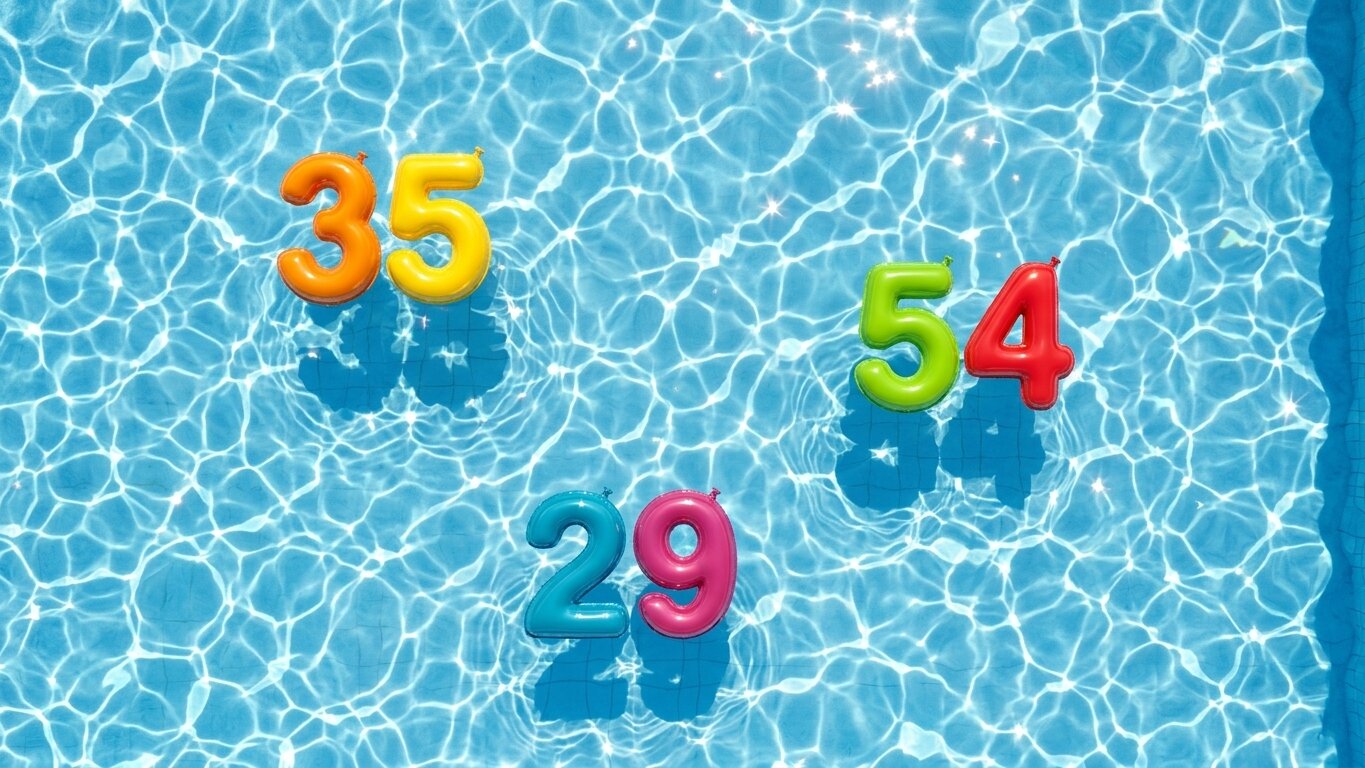}
  
  \vspace{10pt} %

  \includegraphics[width=0.19\linewidth]{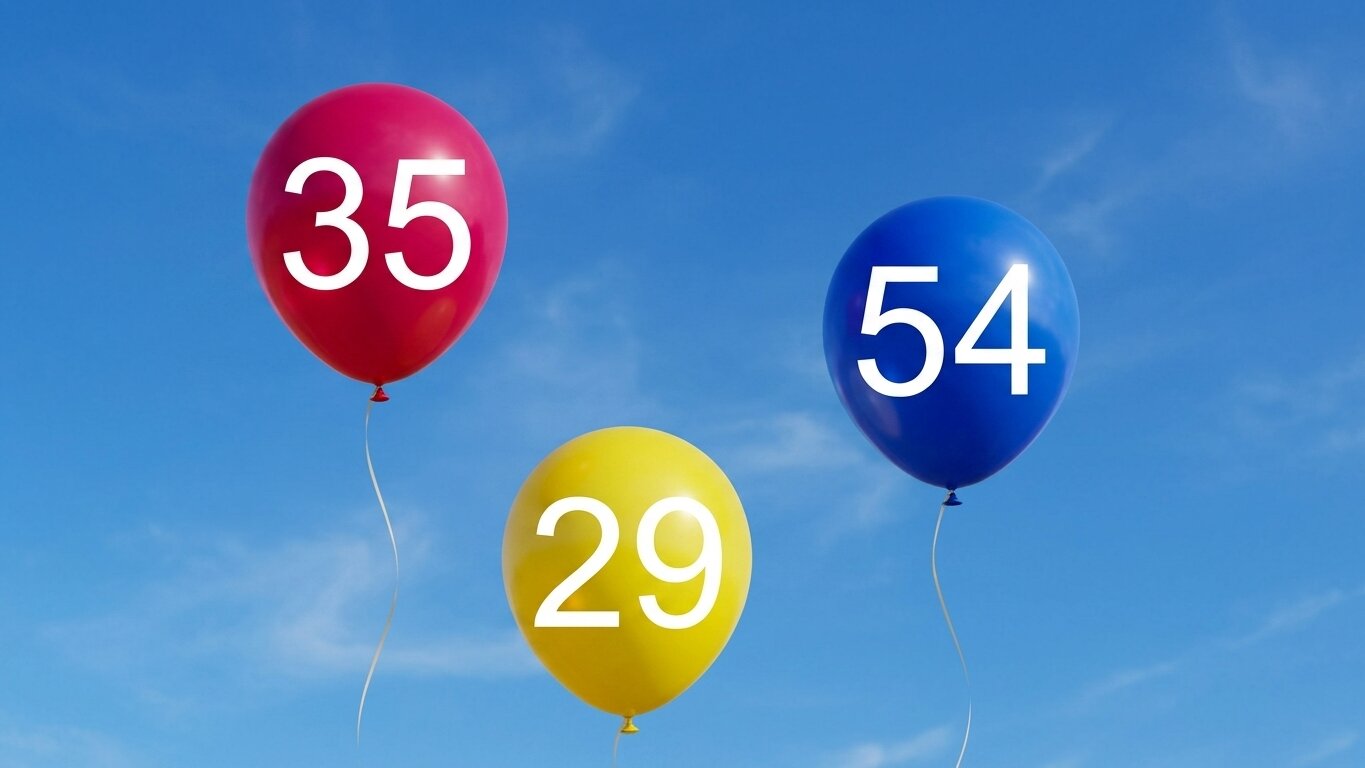}\hfill
  \includegraphics[width=0.19\linewidth]{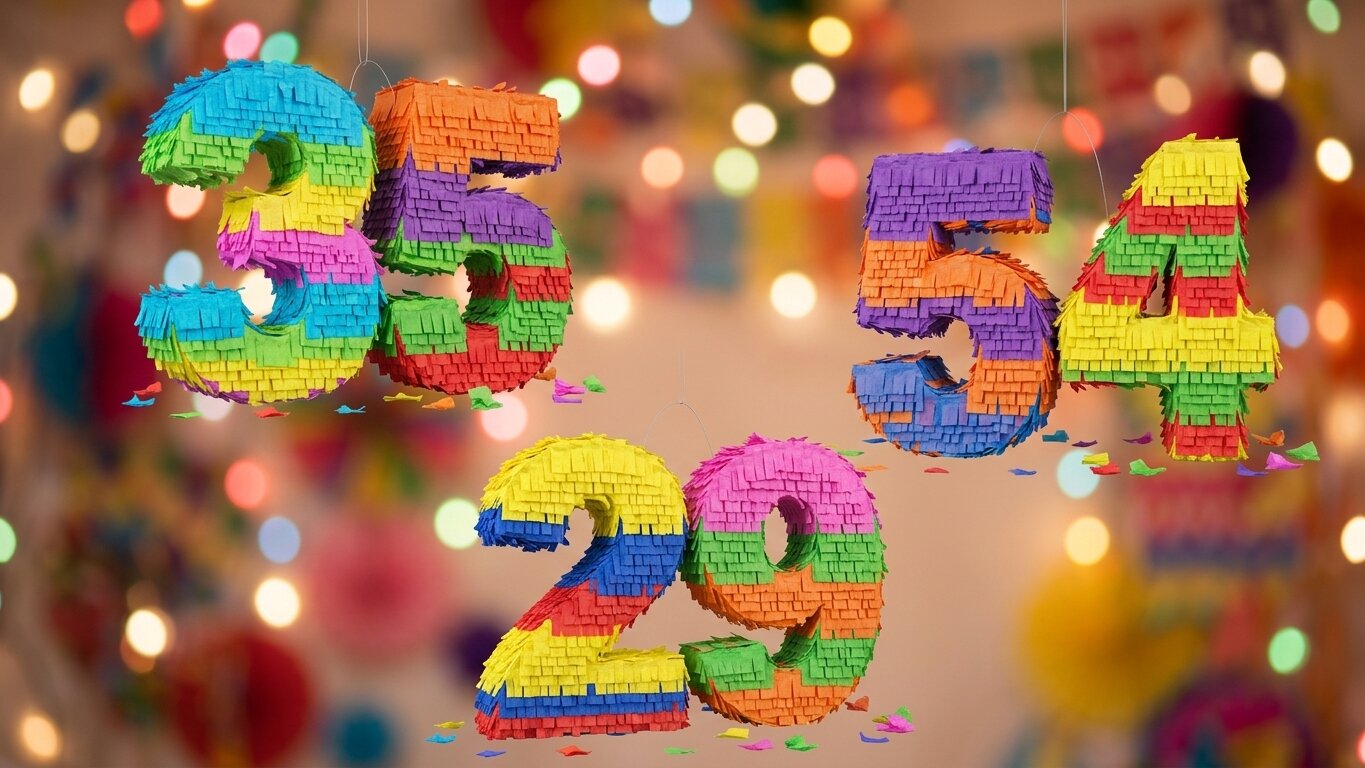}\hfill
  \includegraphics[width=0.19\linewidth]{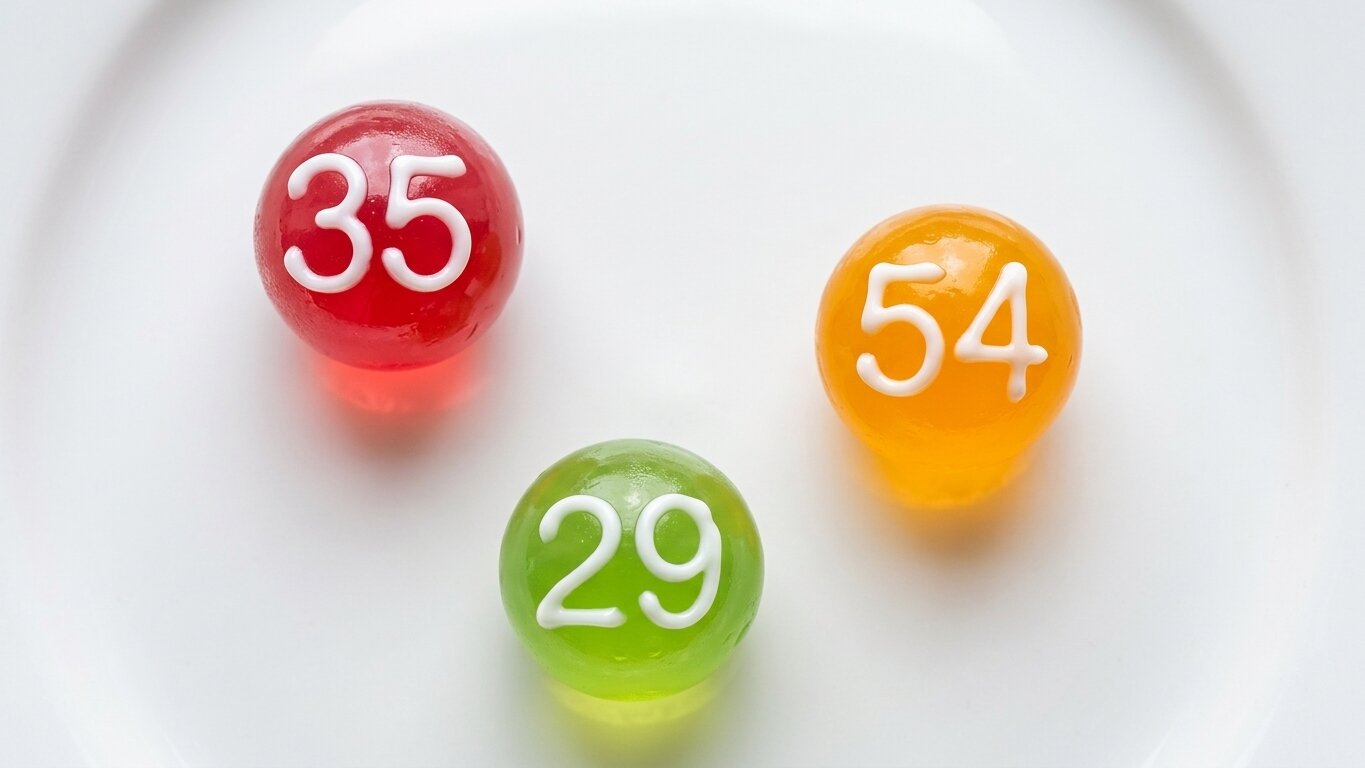}\hfill
  \includegraphics[width=0.19\linewidth]{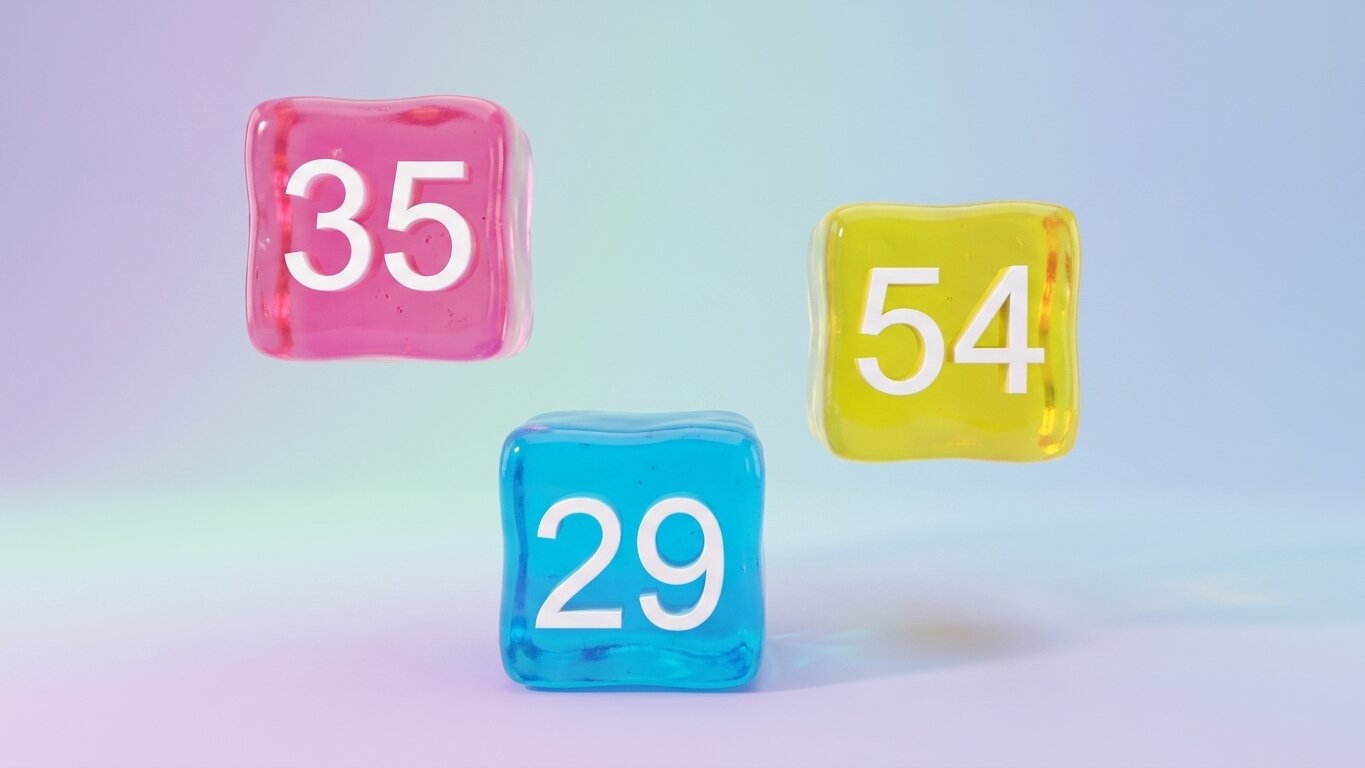}\hfill
  \includegraphics[width=0.19\linewidth]{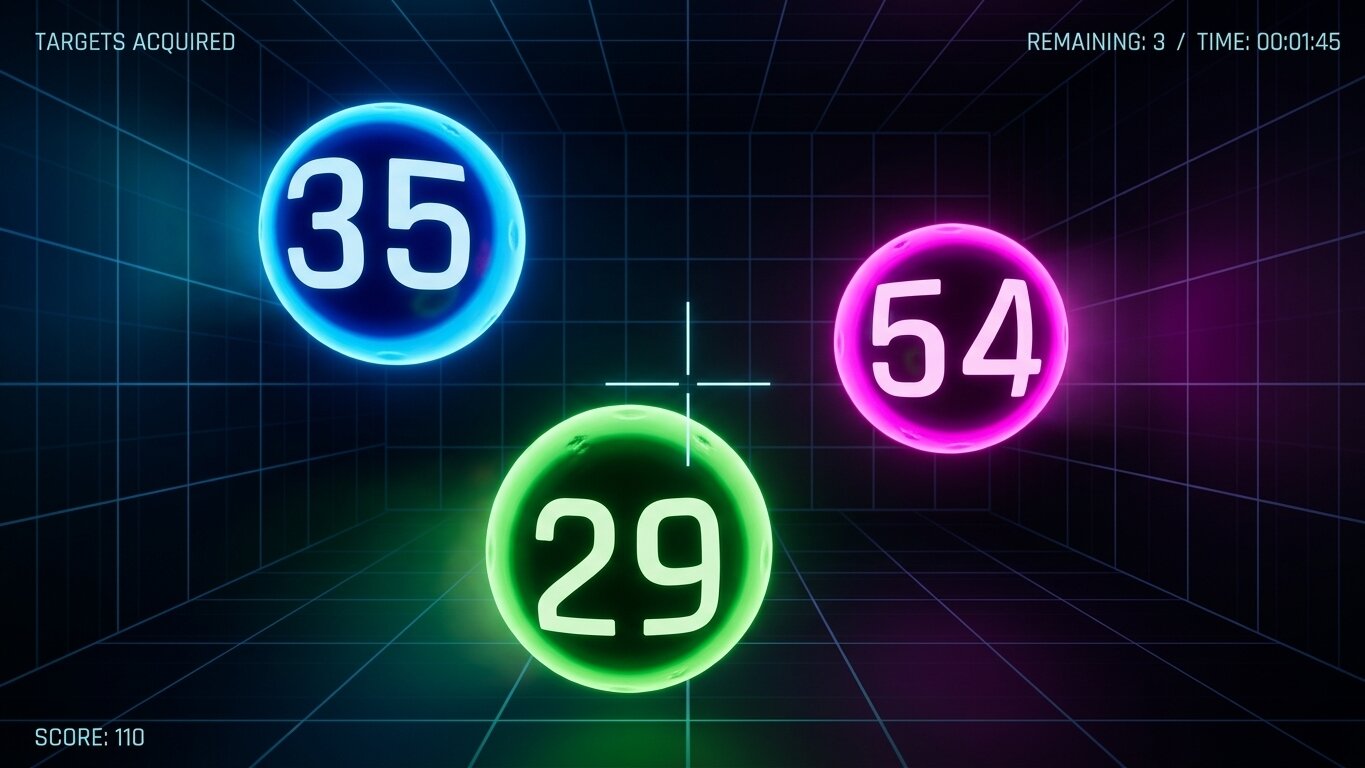}

  \caption{Image variants for \taskname{Sort 3 Numbers}.}
  \label{fig:images_sort}
\end{figure*}

\begin{table}[h]
\centering
\small
\caption{Text prompt variants for \taskname{Connect the Dots}.}
\label{tab:prompts_connect_dots}
\begin{tabular}{p{\linewidth}}
\toprule
\textbf{Prompt Text} \\
\midrule
Original: Connect each pair of same-colored circles with a line. \\
\midrule
Slide the circles in the bottom row horizontally until each circle rests directly underneath the top-row circle of the exact same color. \\
Animate the image so that for each color, the two matching circles bounce up and down simultaneously. Do this sequentially, one pair at a time. \\
Make each circle fly toward the other circle of the exact same color. They must collide and merge into a single circle. \\
Animate the circles so that each pair of identical colors flips over in 3D space simultaneously, like a coin. Animate one matching pair at a time. \\
Animate the image so pairs of the exact same color simultaneously invert to their negative/complementary colors. Sequence this so only one matched pair inverts at a time, holding briefly before reverting. \\
Animate a translucent capsule drawing itself around each pair of matching circles. Each enclosure must contain exactly two identical items. \\
Sequentially for each color pair, expand the two same-colored circles in size until their outer edges perfectly touch, leaving their center coordinates fixed. \\
Animate the same-colored circles as liquid drops. For each matching pair, have them organically expand outward until they touch and fuse together into a single fluid shape. \\
Sequentially spawn a glowing spark at one circle and animate it traveling to the other circle of the exact same color. The spark must disappear upon arrival, leaving no trail. \\
Sequentially select each pair of same-colored circles. Simultaneously drain the color from both circles in the pair until they turn solid gray, then proceed to the next pair. \\
\bottomrule
\end{tabular}
\end{table}

\begin{figure*}[h]
  \centering
  \includegraphics[width=0.19\linewidth]{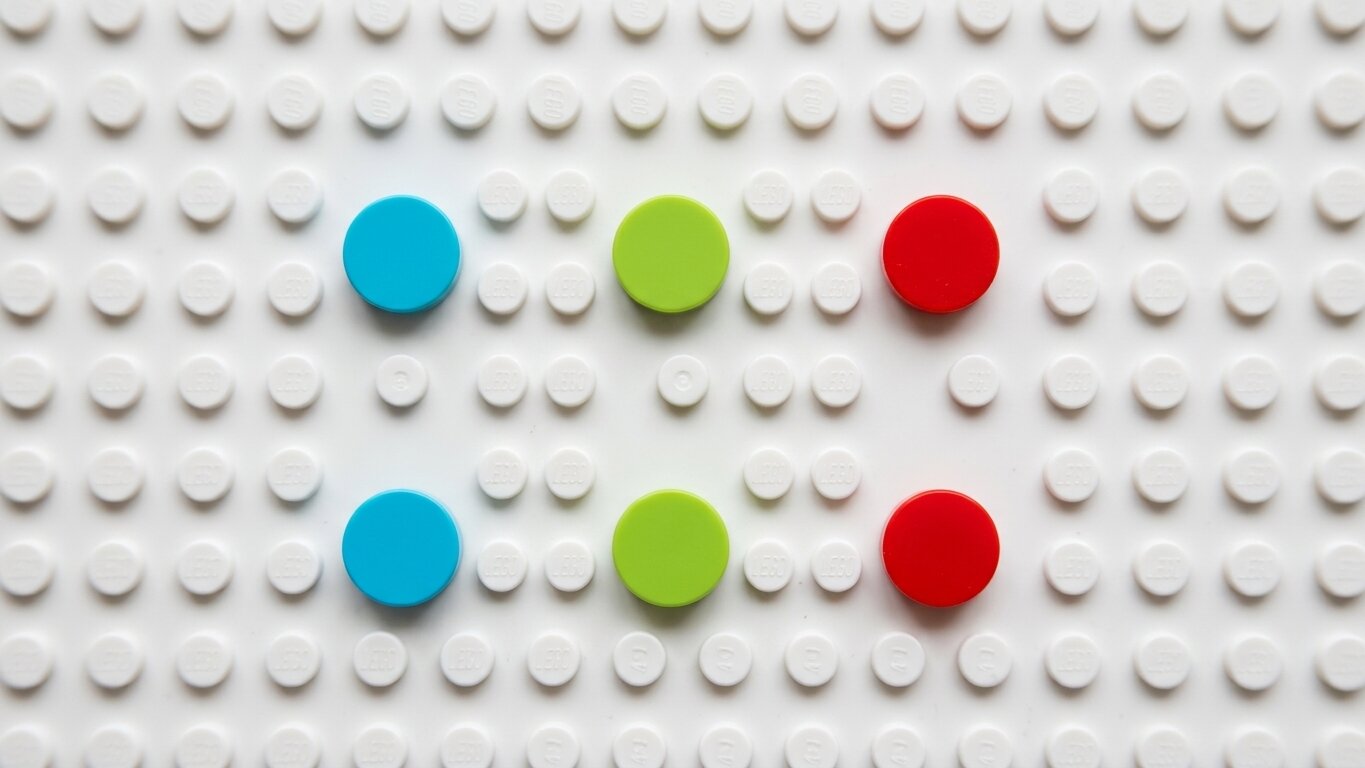}\hfill
  \includegraphics[width=0.19\linewidth]{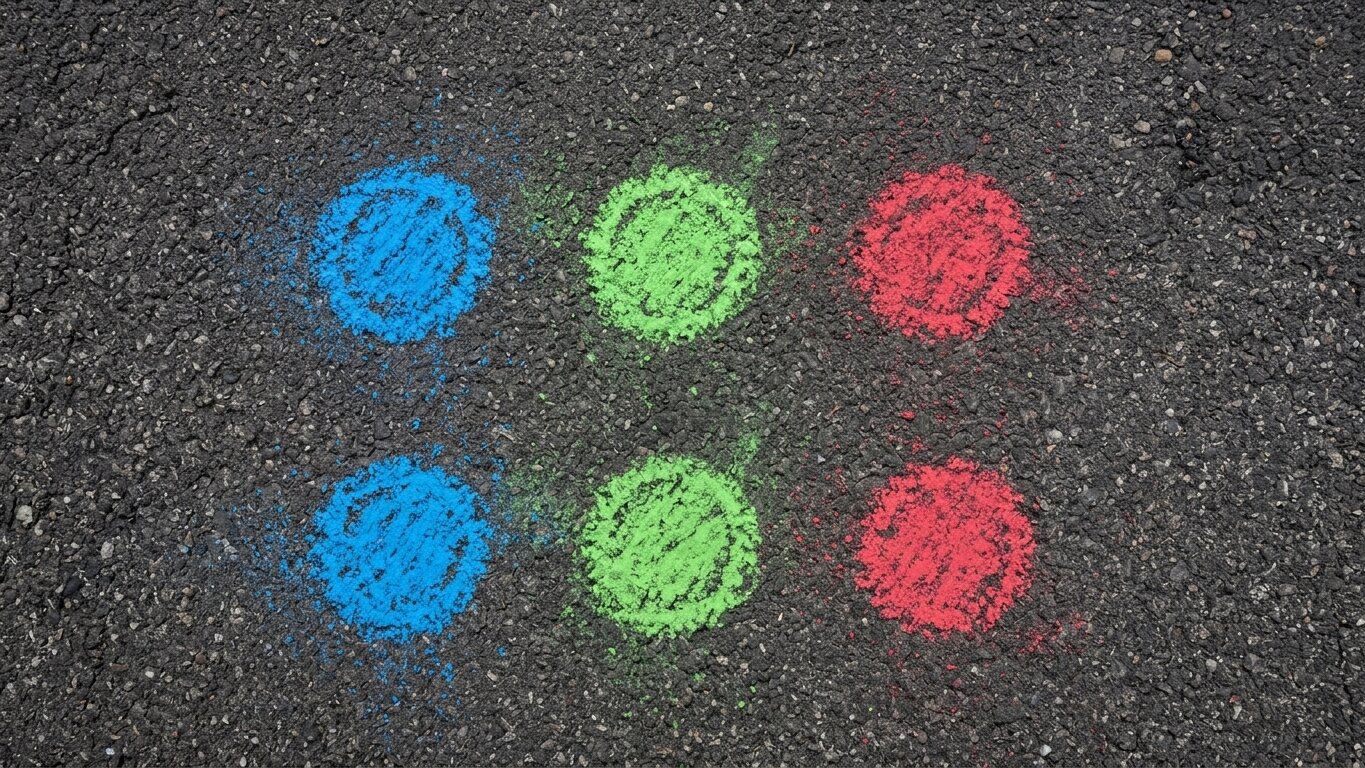}\hfill
  \includegraphics[width=0.19\linewidth]{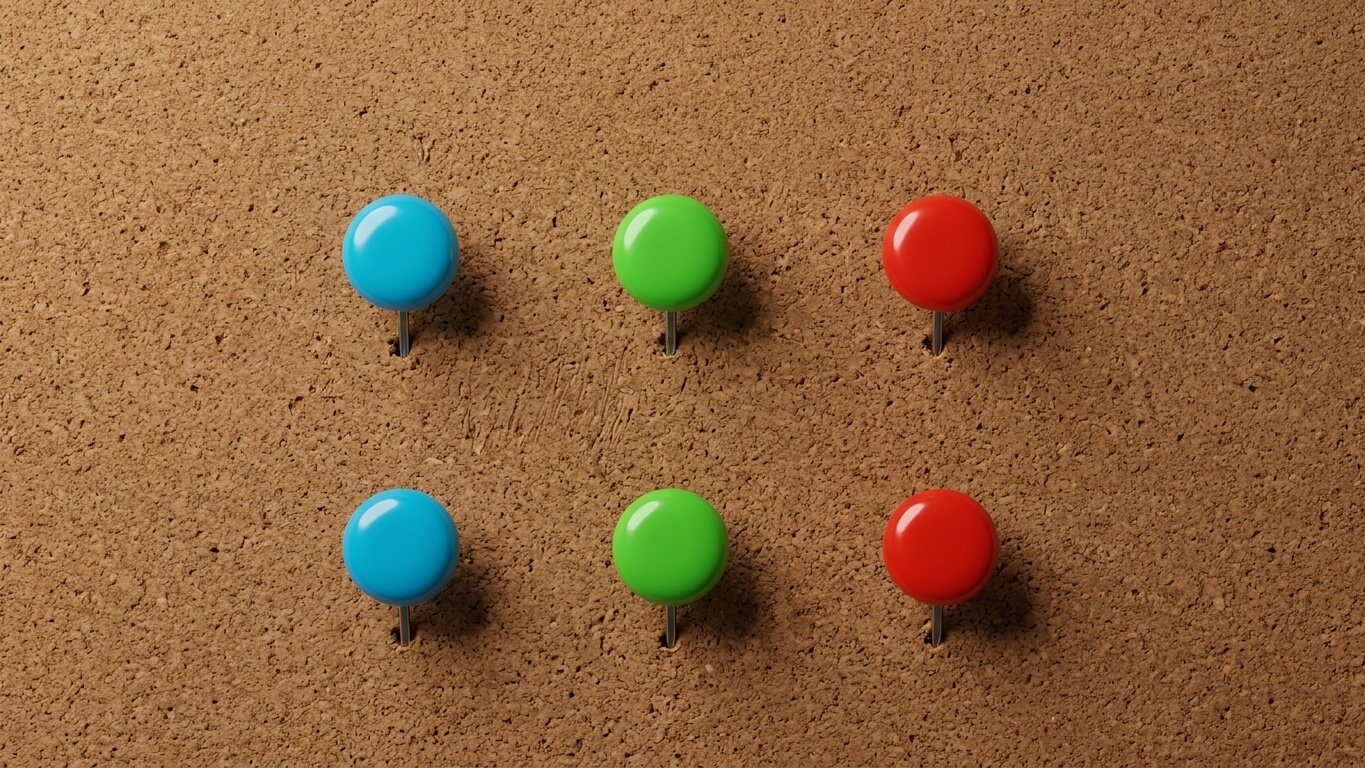}\hfill
  \includegraphics[width=0.19\linewidth]{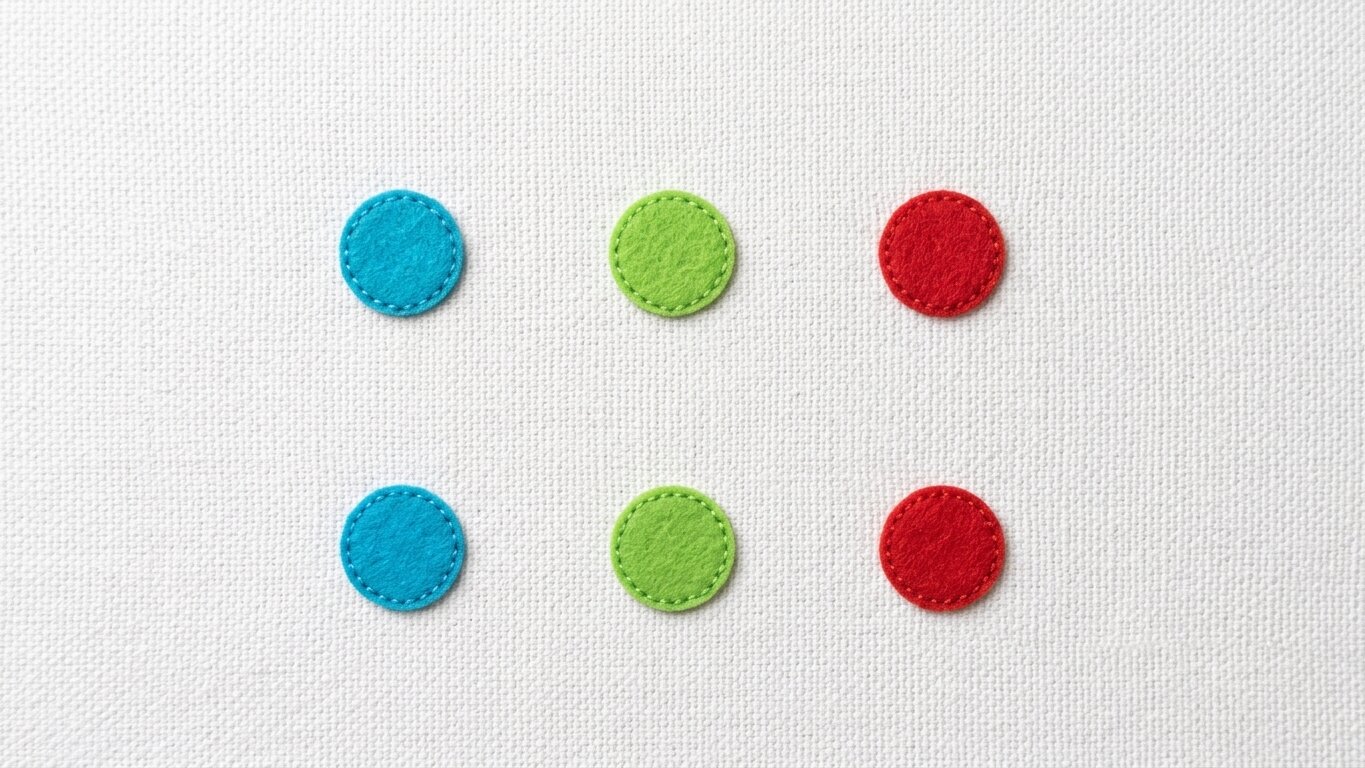}\hfill
  \includegraphics[width=0.19\linewidth]{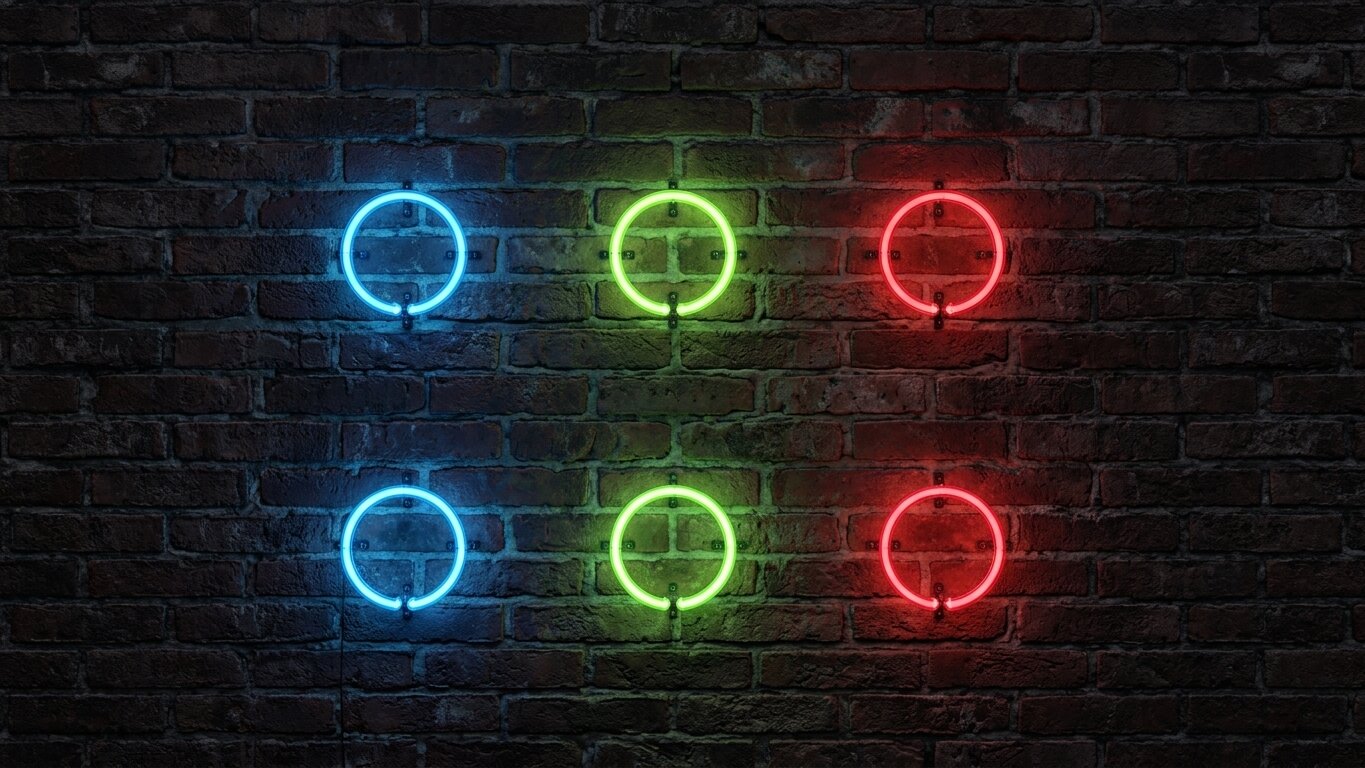}
  
  \vspace{10pt} %

  \includegraphics[width=0.19\linewidth]{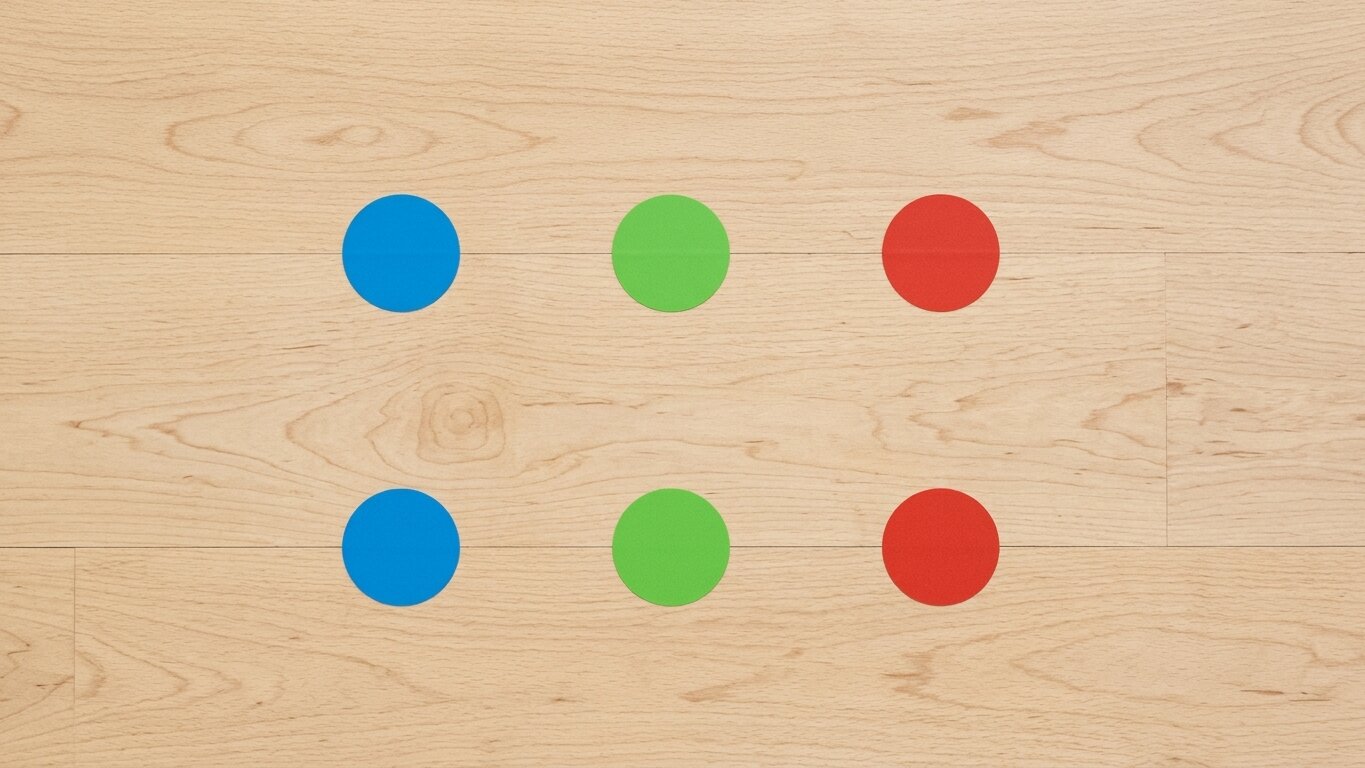}\hfill
  \includegraphics[width=0.19\linewidth]{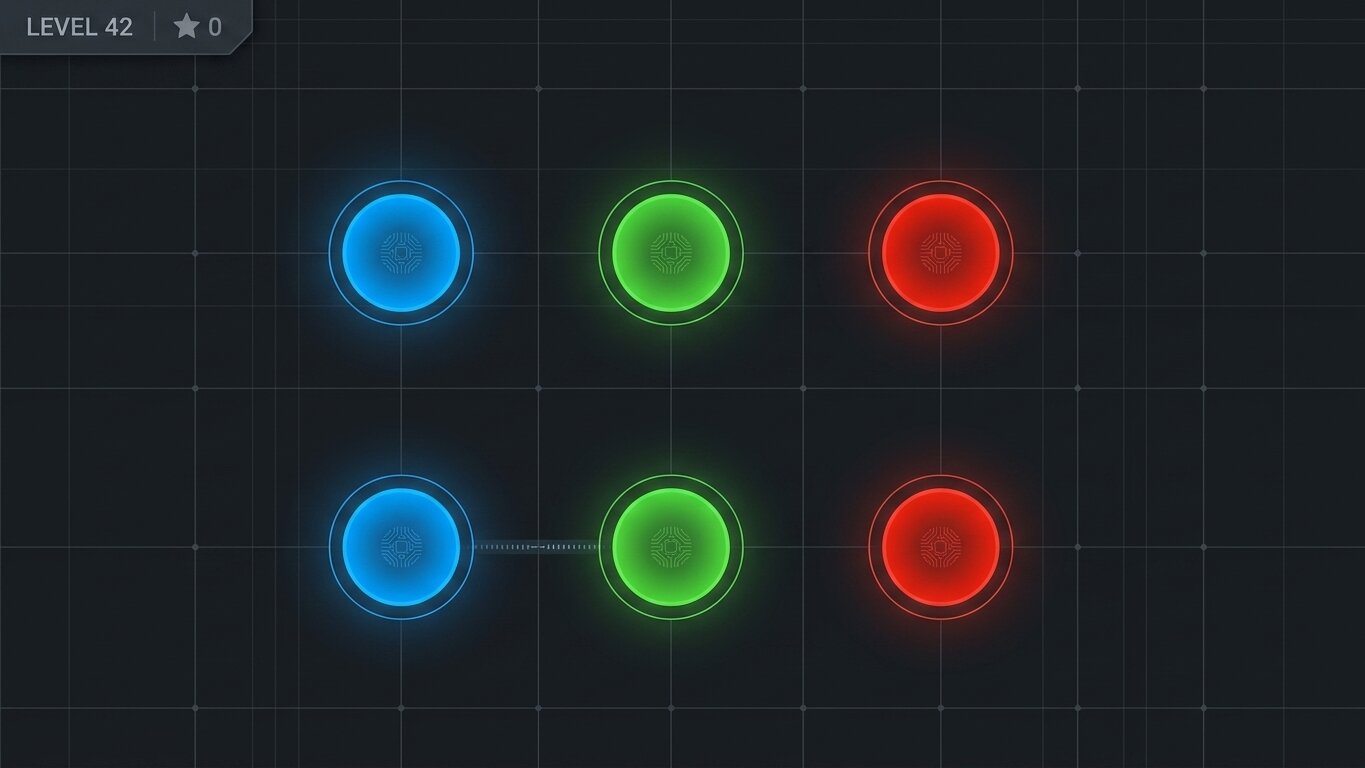}\hfill
  \includegraphics[width=0.19\linewidth]{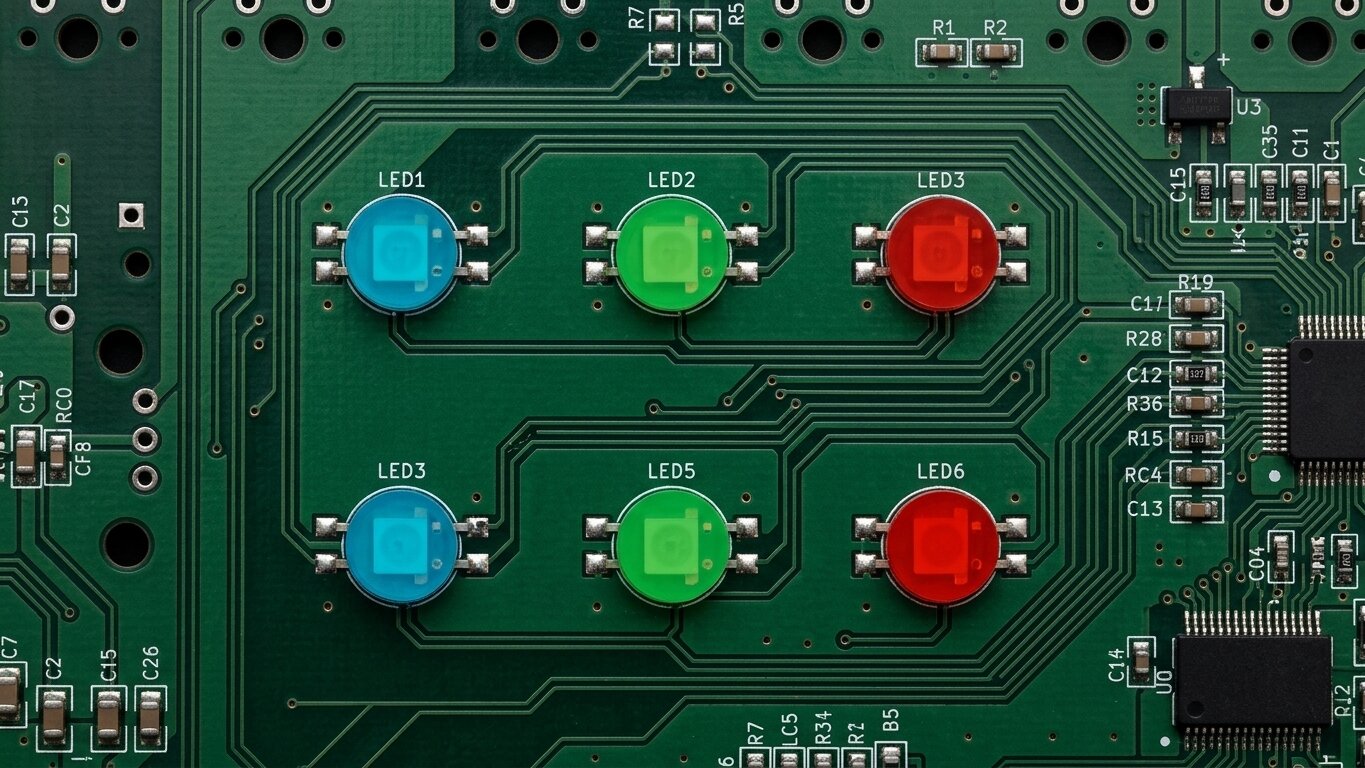}\hfill
  \includegraphics[width=0.19\linewidth]{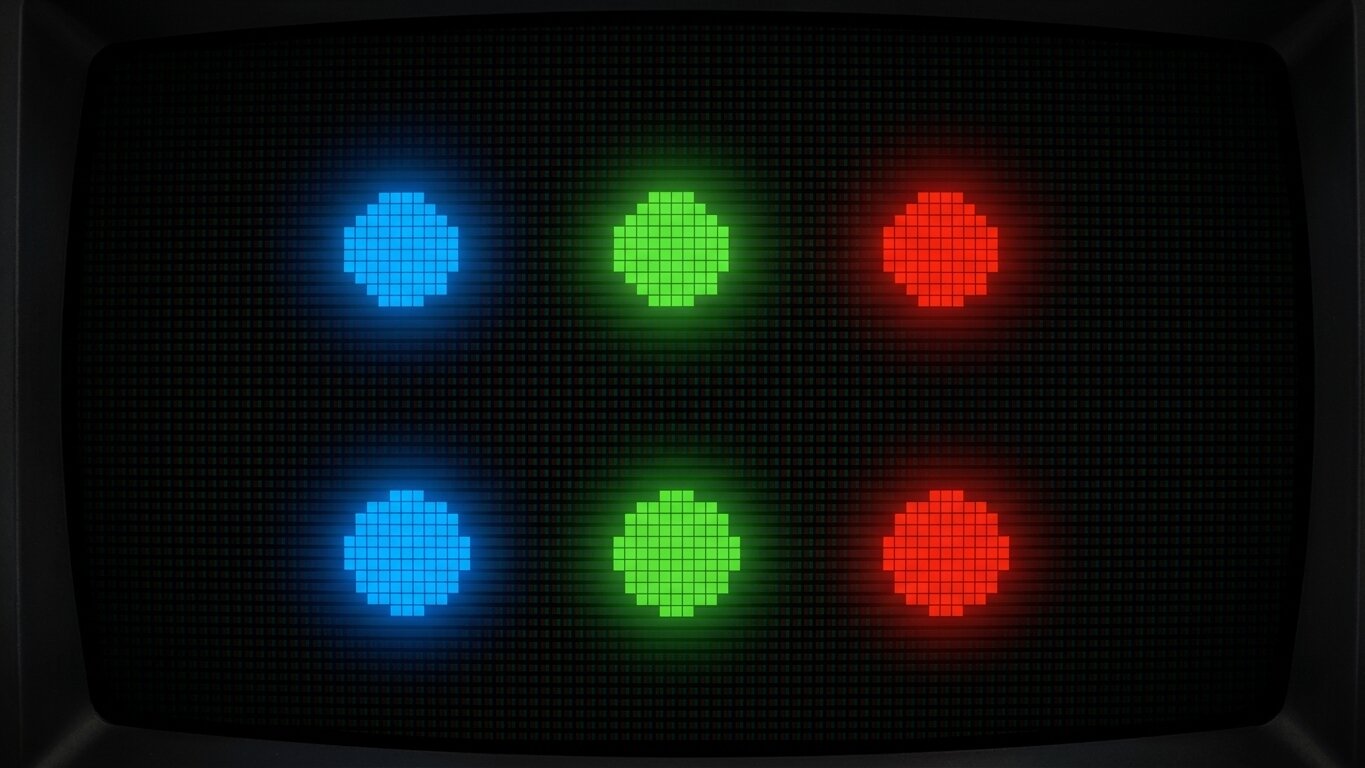}\hfill
  \includegraphics[width=0.19\linewidth]{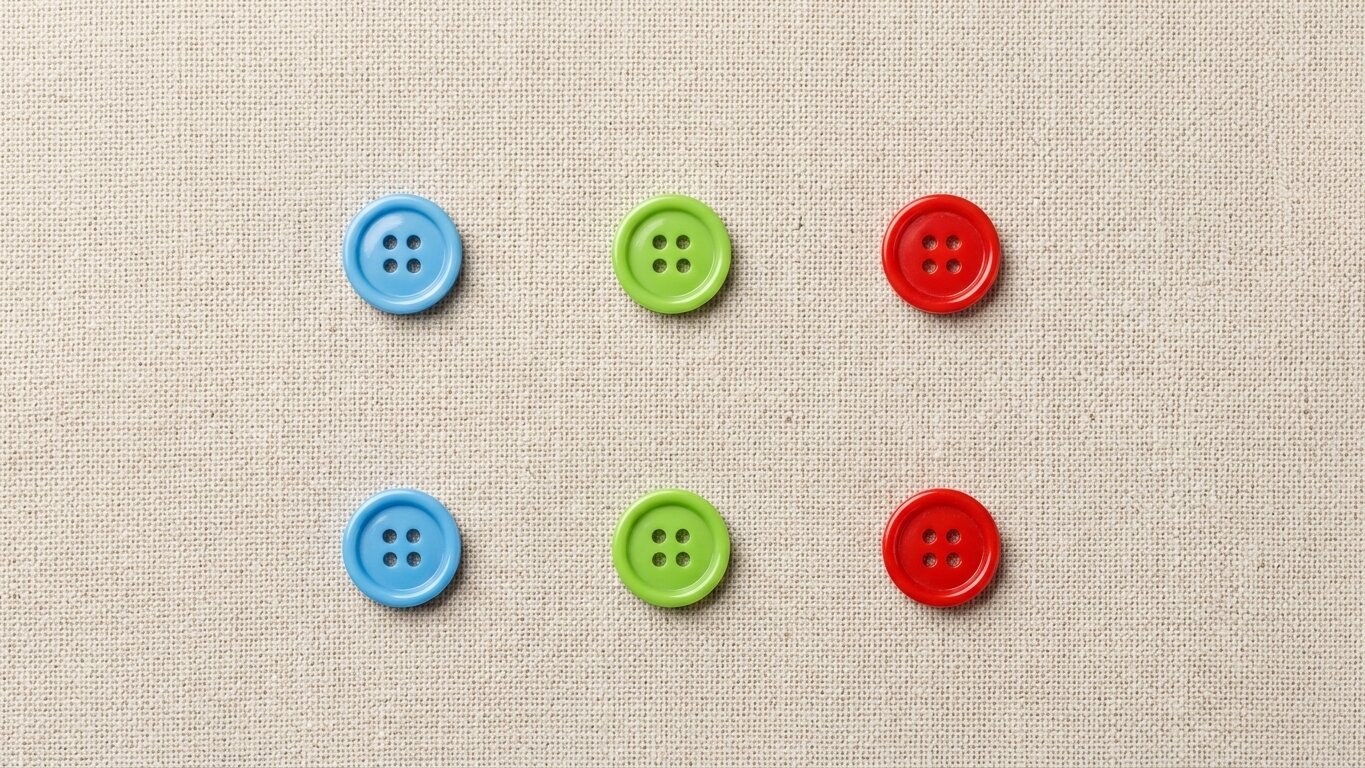}

  \caption{Image variants for \taskname{Connect the Dots}.}
  \label{fig:images_connect}
\end{figure*}

\begin{figure*}[h]
  \centering
  \includegraphics[width=0.19\linewidth]{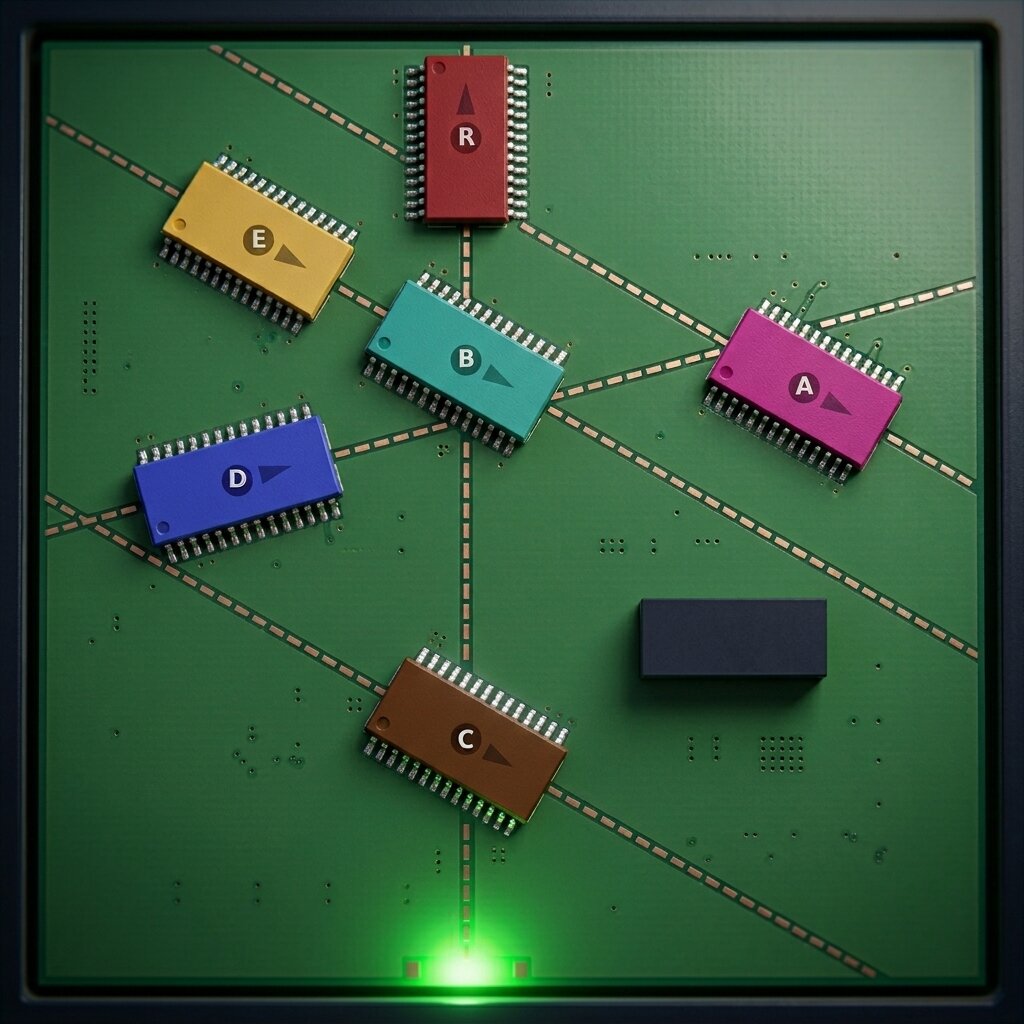}\hfill
  \includegraphics[width=0.19\linewidth]{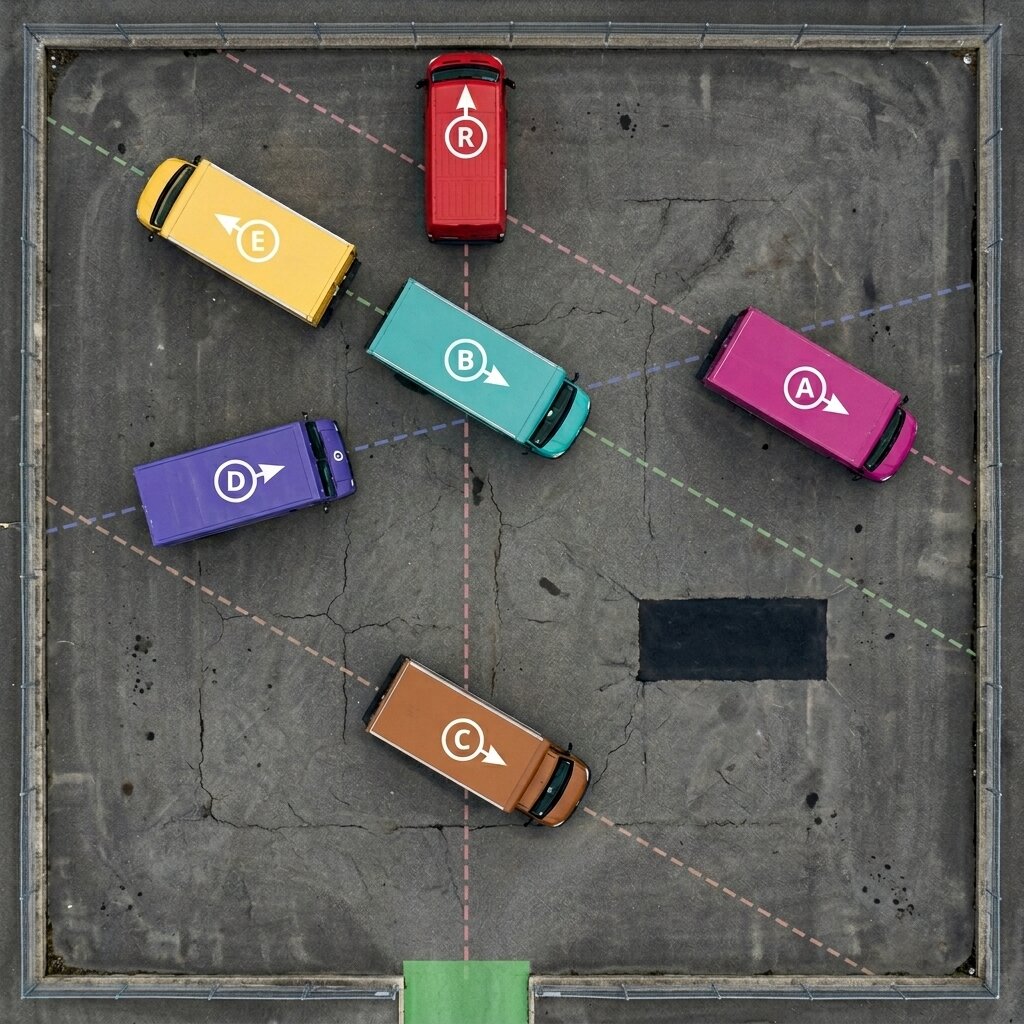}\hfill
  \includegraphics[width=0.19\linewidth]{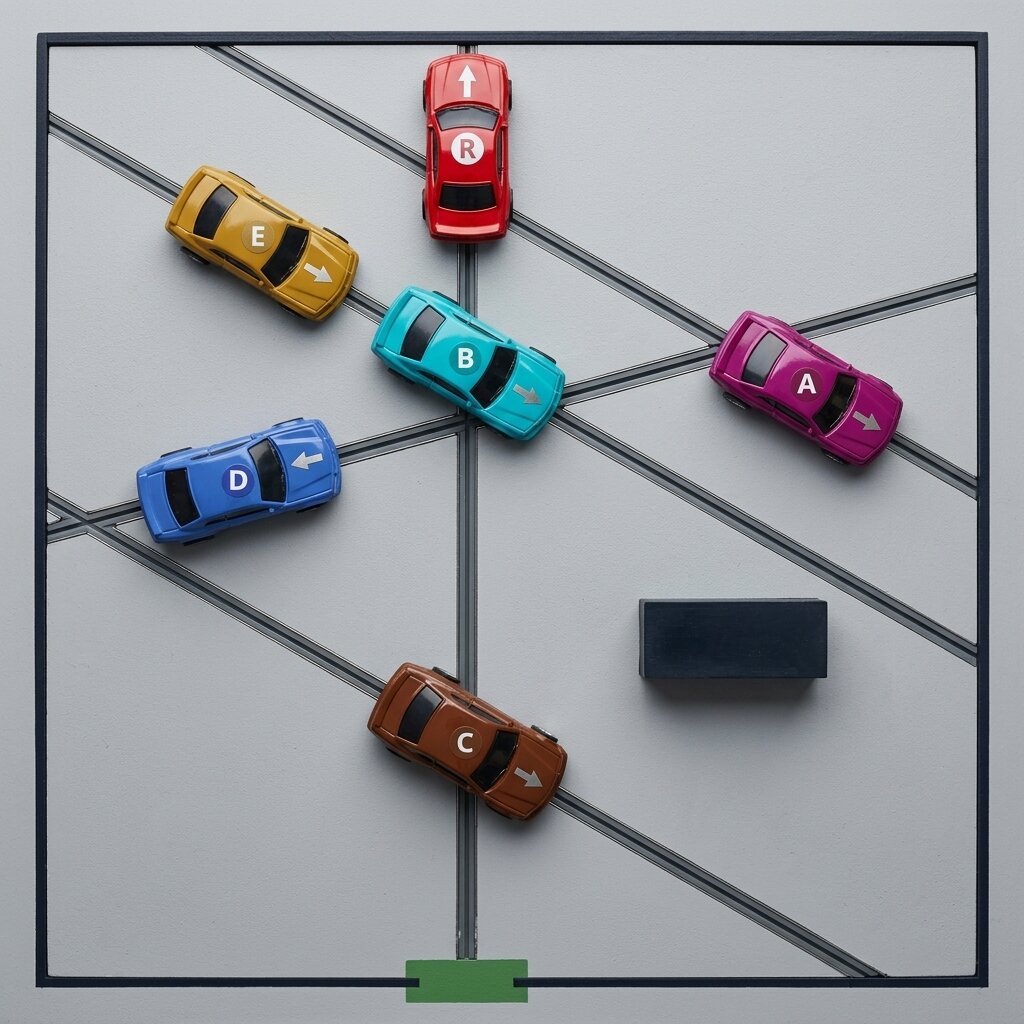}\hfill
  \includegraphics[width=0.19\linewidth]{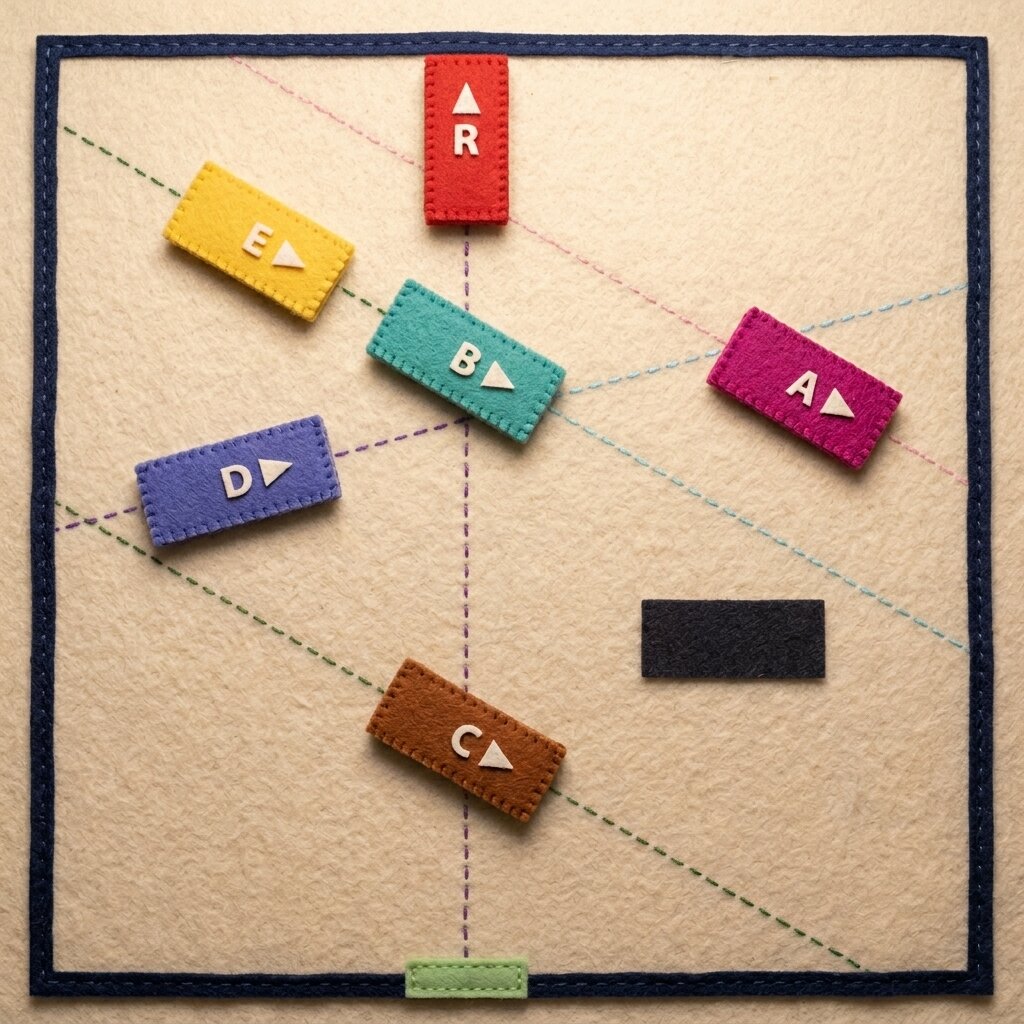}\hfill
  \includegraphics[width=0.19\linewidth]{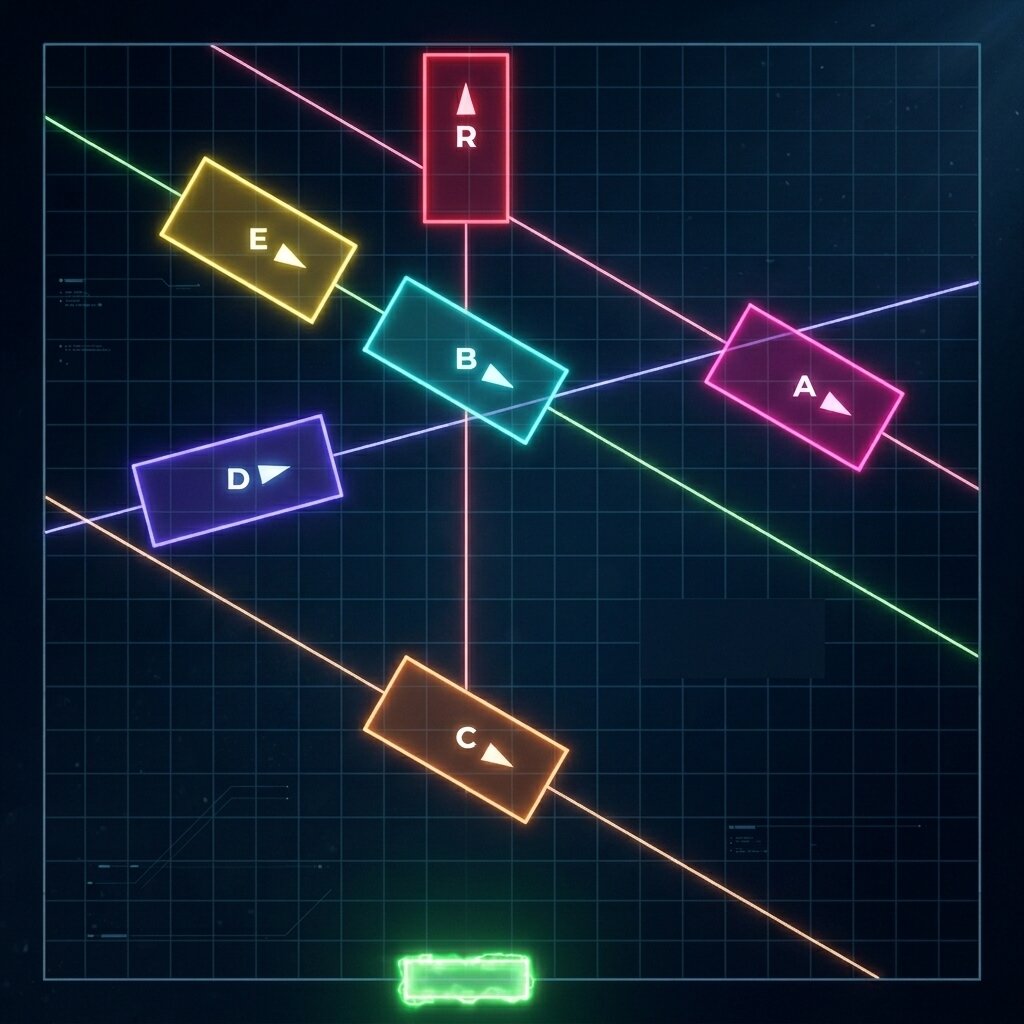}
  
  \vspace{10pt} %

  \includegraphics[width=0.19\linewidth]{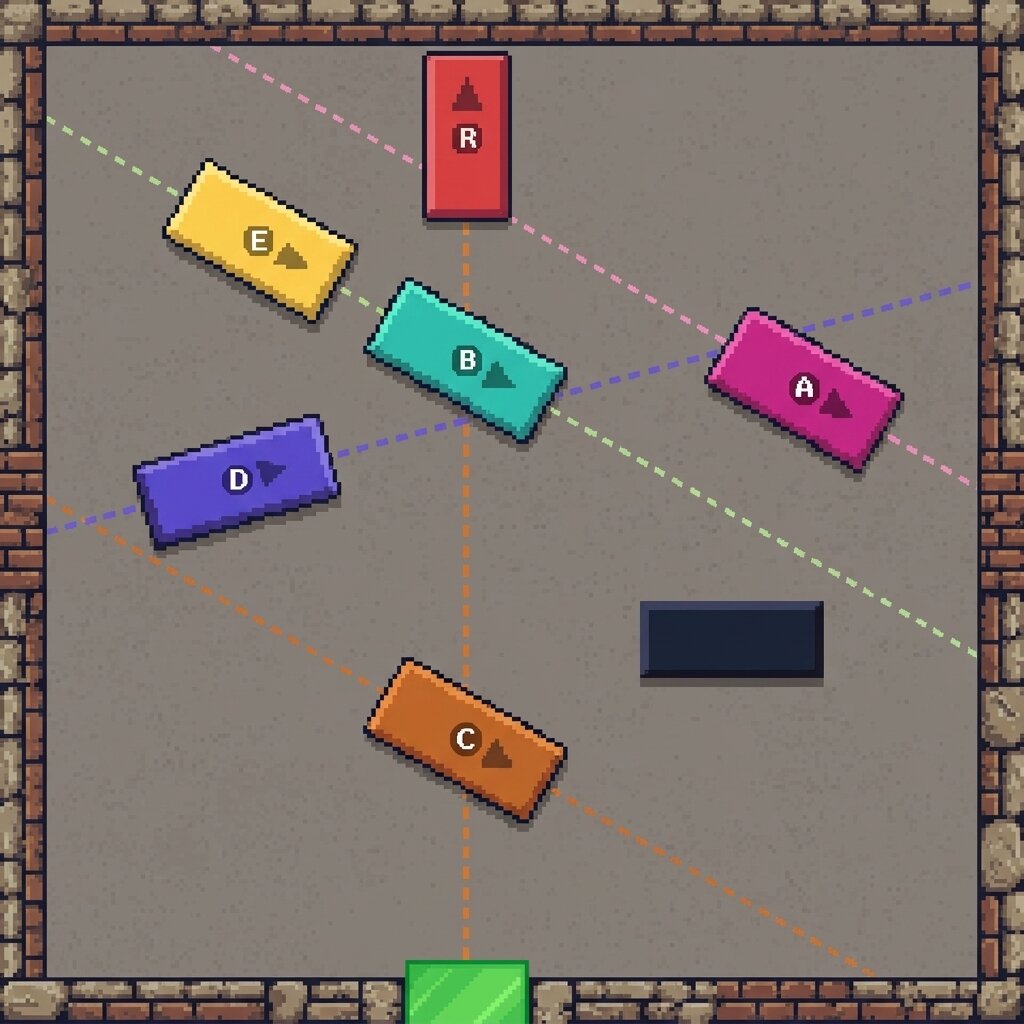}\hfill
  \includegraphics[width=0.19\linewidth]{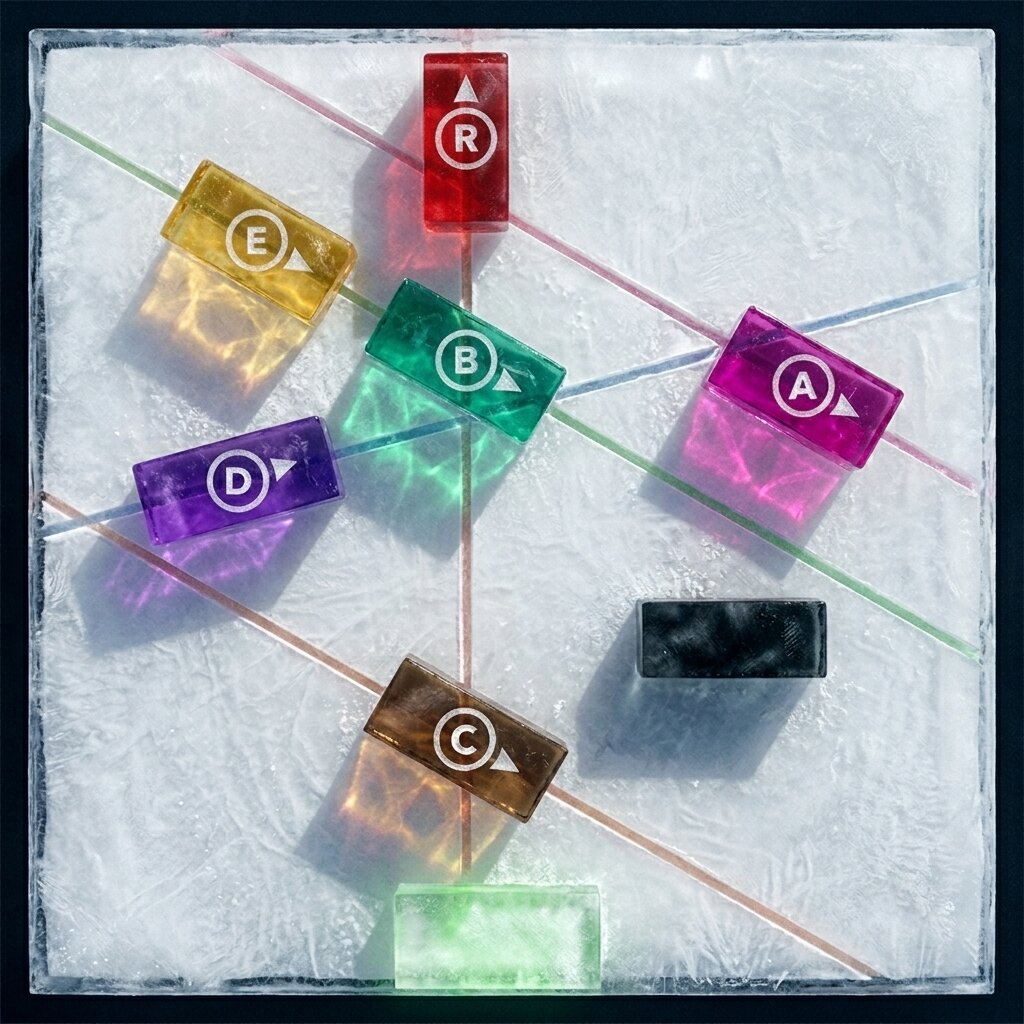}\hfill
  \includegraphics[width=0.19\linewidth]{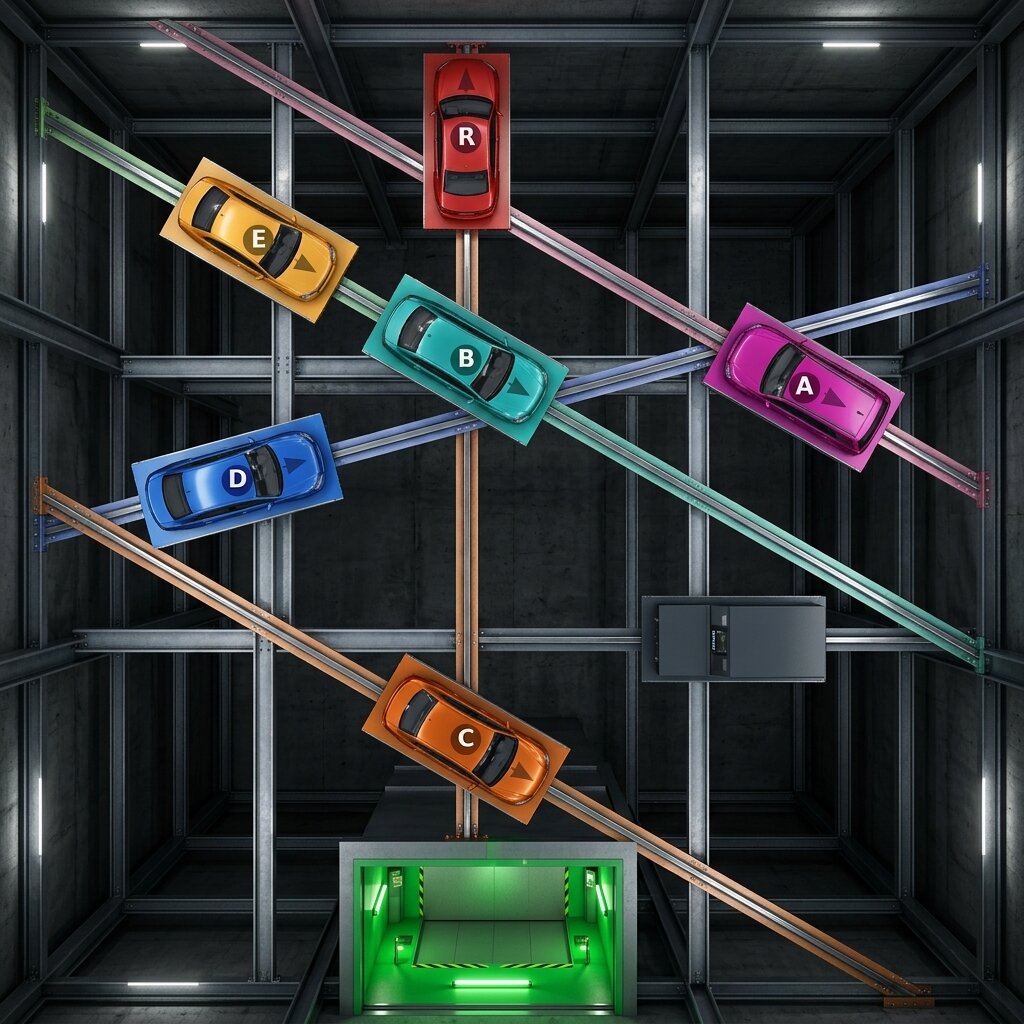}\hfill
  \includegraphics[width=0.19\linewidth]{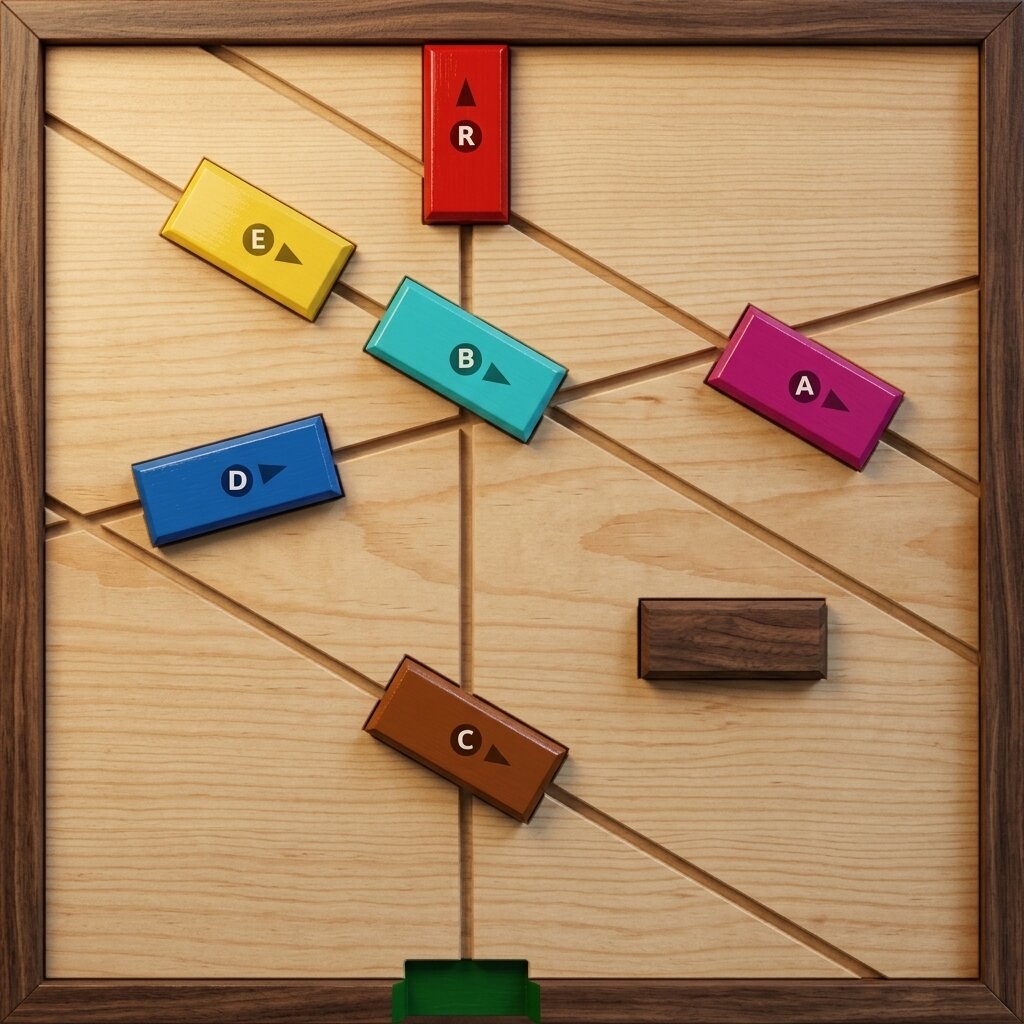}\hfill
  \includegraphics[width=0.19\linewidth]{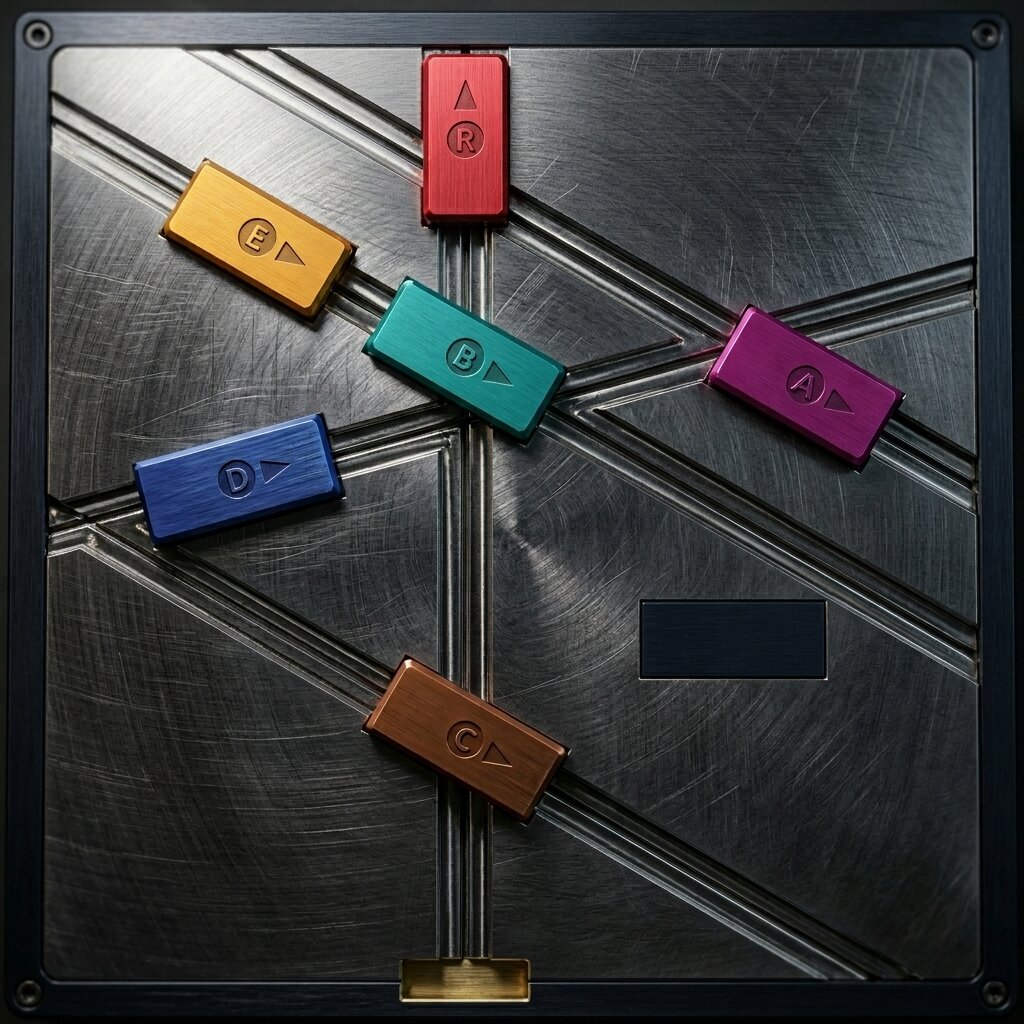}

  \caption{Image variants for \taskname{RushHour}.}
  \label{fig:images_rushhour}
\end{figure*}

\FloatBarrier

\subsection{Step-by-step edits: ACE}
\label{app:ACE}

\begin{figure}
    \centering
    \includegraphics[width=0.7\linewidth]{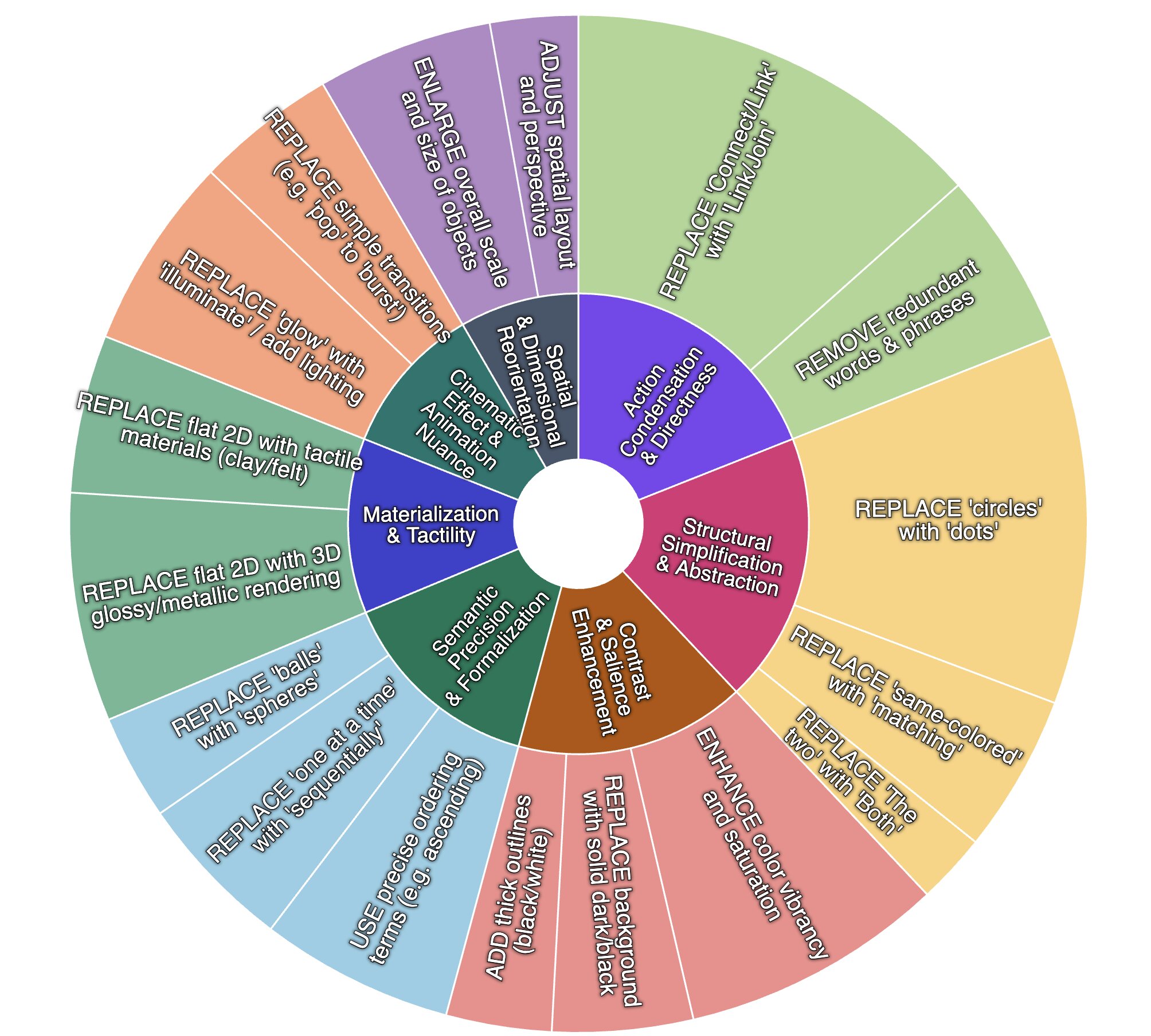}
    \caption{Hierarchical taxonomy of the most effective ACE prompt mutations. Data is aggregated across \taskname{Conjunctive Search}, number sorting, and connecting dots tasks.}
    \label{fig:ace_actions}
\end{figure}

In this section, we provide more details for~\cref{subsec:ace}. We apply the atomic concept editing (ACE) framework~\citep{kalibhat2026ace} to both \emph{text variant} and \emph{image variant} settings. They differ in which dimension of the input is mutated during exploration. In both cases, the goal is to discover prompt or image modifications that improve a video generation model's success rate on visual reasoning tasks while preserving the core reasoning challenge. \cref{fig:ace_actions} summarizes the effective edits.

Besides prompt exploration with simple edits, a critical component of the ACE method is constitution learning, where causal insights into which editing strategies result in the greatest (and least) score improvements are summarized to reconstruct the underlying ``constitution'' of the model's behavior (essentially a summary of model-generated insights). The exact learning algorithm can be found in~\citet{kalibhat2026ace}. See the learned constitutions below. Once an constitution is learned, the prompt exploration can be done by conditioning on the constitution, so that more successful edits are more likely to be generated. ACE receives feedback from the autorater, and operates in an exploration, constitution learning, re-exploration loop. In practice, we can also combine the freeform ideation and ACE methods by treating the ideated prompts as root prompts of ACE.

\subsubsection{Learned constitutions for image variant}

For the \emph{\taskname{Sort 3 Numbers}} task, the learned constitution identified:
\begin{itemize}
    \item \textbf{Effective strategies:}
    \begin{itemize}
        \item \emph{High-Contrast Background Isolation} --- Replace the original background with a solid, high-contrast or dark color. This cleanly isolates foreground elements, reducing visual noise and drastically improving the video model's ability to maintain focus and track objects across frames.
        \item \emph{Subtle Volumetric Dimension Addition} --- Introduce controlled 3D properties such as embossing, volumetric geometry, or neon aesthetics to flat elements. This enhances object visibility and physical presence across time without relying on unpredictable highly reflective surfaces.
        \item \emph{Spatial Boundary Enclosure} --- Enclose individual textual or numerical elements within distinct geometric borders, such as circles or boxes. This creates definitive spatial boundaries that reinforce element correspondence and tracking without altering the element count.
        \item \emph{Uniform Scale Amplification} --- Significantly increase the scale of all focal elements uniformly while preserving their center coordinates. This maximizes legibility and visibility for the video model, serving as an effective temporal tracking anchor despite occasional risks of layout crowding.
        \item \emph{Thematic Surface Contextualization} --- Transform the default sterile environment into a cohesive, recognizable physical medium, such as a chalkboard with chalk writing. This grounds the abstract reasoning task in a stable real-world context, naturally enhancing the model's temporal consistency.
        \item \emph{Controlled Vibrant Color Differentiation} --- Assign distinct, vivid, or moderately vibrant colors to individual elements. This provides clear visual separation to enhance temporal tracking, successfully avoiding the visual overload and artifacting associated with uniquely harsh, highly saturated color modifications.
    \end{itemize}
    \item \textbf{Ineffective strategies:}
    \begin{itemize}
        \item \emph{Basic Font Weight Amplification} --- Apply ultra-heavy or overly bold weights to standard fonts. Drastically thickening plain typography frequently warps element shapes during video generation, blurring boundaries and causing tracking failures.
        \item \emph{Highly Reflective Textures} --- Applying heavily reflective, shiny, or metallic material finishes to the primary objects. These complex textures introduce unpredictable light variations and specular highlights across video frames, which degrades temporal consistency and object permanence.
        \item \emph{Individual Saturated Color Coding} --- Assigning distinct, uniquely saturated hues to multiple homogeneous elements. While intended to visually separate items, excessive chromatic variation frequently overloads the video model's visual attention, causing failures in temporal tracking.
    \end{itemize}
\end{itemize}

For the \emph{\taskname{Connect the Dots}} task, the learned constitution identified:
\begin{itemize}
    \item \textbf{Effective strategies:}
    \begin{itemize}
        \item \emph{High-Contrast Dark Background} --- Replace the background with solid black or very dark colors. This was the single most successful edit, achieving 50\% pass rate (3/6 videos) in one run---the only mutation to ever reach the success threshold.
        \item \emph{Bold Outlines and Borders} --- Add thick, crisp black or white outlines around every circle, reaching 25\% pass rate in multiple runs. Heavy outlines sharpen the boundaries of circles and may help the video model perceive them as distinct connectable endpoints.
        \item \emph{Ultra-Saturated Neon Colors} --- Shift all circle colors to maximally saturated neon equivalents while ensuring each color pair remains distinct. Empirically reached 24\% pass rate. Critical: ensure the color correspondence between paired circles is preserved and amplified, not muddied.
        \item \emph{Visual Connection Cues} --- Add subtle directional hints that guide the eye between same-colored circles: faint dotted guidelines, subtle color gradients in the space between matched circles, or gentle arrow indicators.
        \item \emph{Circle Enlargement with Preserved Spacing} --- Increase circle diameter by 40--60\% while keeping center coordinates fixed. Larger circles are easier for the video model to identify and track as connection endpoints.
        \item \emph{Rendering as Physical Objects} --- Render circles as 3D spheres, buttons, or tokens on a flat surface. Physical-looking objects with depth cues (shadows, highlights) may activate the model's understanding of spatial relationships better than flat 2D discs.
        \item \emph{Color-Matched Labeling} --- Add small text labels, numbers, or symbols inside each circle that are the same for matching pairs. This gives the model a redundant pairing signal beyond color alone.
    \end{itemize}
    \item \textbf{Ineffective strategies:}
    \begin{itemize}
        \item \emph{Background Grid or Graph Paper} --- Adding grids, graph paper, or blueprint patterns achieved 0\% pass rate (0/18 videos). Grid lines visually interfere with the lines the model must draw.
        \item \emph{Spatial Rearrangement} --- Moving or rearranging circles breaks the spatial structure and invalidates the task.
        \item \emph{Element Addition or Removal} --- Adding or removing circles changes the number of color pairs and invalidates the task.
        \item \emph{Heavy Artistic Stylization} --- Extreme stylistic filters (impressionism, heavy pixel art) that blur circle boundaries or make colors ambiguous.
        \item \emph{Neon Glow Without Structure} --- Adding emissive glow effects without also increasing structural clarity reached only 11\% pass rate (2/18)---the diffuse light halos around circles make their boundaries less distinct and can cause color bleeding.
    \end{itemize}
\end{itemize}

For the \emph{\taskname{Conjunctive Search}} task, the learned constitution identified:
\begin{itemize}
    \item \textbf{Effective strategies:}
    \begin{itemize}
        \item \emph{Selective 3D Materialization} --- Transform only the target objects to possess volumetric, glossy, or metallic materials while keeping all other scene elements flat. This strong material contrast effectively isolates the targets and helps the model distinguish them from distractors.
        \item \emph{Distractor Opacity Reduction} --- Lower the transparency of all non-target shapes to make them fade into the background. Suppressing the visual prominence of distractors through opacity alone reliably guides the model's attention toward the fully visible target objects.
        \item \emph{Target Object Enlargement} --- Significantly increase the physical size or scale of the target objects while maintaining their original positions. Amplifying the spatial footprint of the targets makes them undeniably prominent without fundamentally altering the core reasoning challenge.
        \item \emph{Distractor Desaturation} --- Remove the color saturation of non-target shapes to turn them into grayscale elements. Stripping away the color features from distractors creates a stark color contrast that reliably isolates the target objects.
        \item \emph{Target Color Intensification} --- Significantly increase the color vibrancy, brightness, or saturation of the target objects to turn them into intensely vivid elements, strongly drawing the model's focus without relying on external cues or altering scene geometry.
    \end{itemize}
    \item \textbf{Ineffective strategies:}
    \begin{itemize}
        \item \emph{Superficial Target Highlighting} --- Adding artificial external cues such as thick glowing outlines, borders, or localized halos to emphasize targets. These superficial additions are highly unreliable and significantly less effective than modifying the intrinsic physical properties of the objects.
        \item \emph{Global Illumination and Spotlighting} --- Introducing targeted spotlighting or altering the scene's global illumination instead of modifying object properties. Such lighting changes can unpredictably obscure necessary spatial details.
        \item \emph{High-Contrast Background Replacement} --- Replacing the background with solid black or dark colors. This drastic background change is highly unreliable, as it often causes scene elements to blend into the darkness and unpredictably destroys essential spatial structural perception.
        \item \emph{Relative Distractor Shrinking} --- Reducing the physical size of all non-target objects while maintaining their positions. Indiscriminately altering the scale of multiple distractor elements disrupts the model's understanding of element correspondence and scene geometry.
    \end{itemize}
\end{itemize}

\subsubsection{Learned constitutions for text variant}

For the \emph{\taskname{Sort 3 Numbers}} task:
\begin{itemize}
    \item \textbf{Effective strategies:}
    \begin{itemize}
        \item \emph{Value-Based Relational Phrasing} --- Describe the sorting logic by referencing the underlying attributes or relative values of the objects rather than just naming a standard order.
        \item \emph{Process-Oriented Transition Verbs} --- Substitute basic motion or state-change verbs with more descriptive actions that capture the visual flow or the deliberate way objects move and disappear.
        \item \emph{Chronological Sequence Adverbs} --- Incorporate adverbs that explicitly highlight the step-by-step or sequential nature of the actions to clarify the timing and progression of the task.
    \end{itemize}
    \item \textbf{Ineffective strategies:}
    \begin{itemize}
        \item \emph{Stative Result Descriptors} --- Replacing verbs of motion or transition with words that describe a final state or a static arrangement, which fails to capture the actual process being observed.
        \item \emph{Tautological Logical Redundancy} --- Adding descriptive phrases or modifiers that repeat a logic already clearly established by other terms in the prompt, leading to unnecessary wordiness without adding clarity.
        \item \emph{Decontextualized Technical Terminology} --- Substituting common visual nouns or simple descriptors with formal mathematical or categorical terms that lack sensory grounding in the scene.
    \end{itemize}
\end{itemize}

For the \emph{\taskname{Connect the Dots}} task:
\begin{itemize}
    \item \textbf{Effective strategies:}
    \begin{itemize}
        \item \emph{High-Frequency Connective Verbs} --- Using common, high-frequency verbs like ``Link'' or ``Join'' to define the primary instruction. These provide the most consistent results when the rest of the sentence syntax remains standard and free of complex modifiers.
        \item \emph{Simplified Visual Nouns} --- Replacing specific geometric terms with universally understood visual descriptors, such as ``dots'' instead of ``circles.'' This reduction in technical specificity helps the model ground the task in basic visual components.
        \item \emph{Explicit Relational Grounding} --- Favoring basic, explicit phrases like ``same color'' or ``matching color'' over complex or vague adjectives. Directness in describing the exact property shared by objects leads to more consistent reasoning than abstract synonyms.
    \end{itemize}
    \item \textbf{Ineffective strategies:}
    \begin{itemize}
        \item \emph{Semantic Overcrowding} --- Proposing modifications to prompts that already contain non-standard modifiers or prior edits. Success rates drop significantly as the number of deviations from the root task description increases, making multi-mutation prompts fragile.
        \item \emph{Action Path Qualifiers} --- Specifying the physical quality of a required path with adjectives like ``clear,'' ``continuous,'' or ``unbroken.'' These terms often obscure the underlying spatial logic and confuse the model's execution of the core challenge.
        \item \emph{Abstract Numerical Grouping} --- Replacing direct count words like ``each'' or ``pair'' with formal collective nouns like ``duo'' or ``set.'' Such terminology often interferes with the model's ability to process the quantitative requirements.
        \item \emph{Technical Property Synonyms} --- Using high-register or technical synonyms for basic visual properties, such as ``identically hued'' instead of ``same-colored.'' These terms lack the specific grounding needed to focus the model on the correct visual feature.
        \item \emph{Redundant Procedural Modifiers} --- Adding lengthy phrases to describe the method of execution, such as ``by drawing a single stroke'' instead of ``with a line.'' This unnecessary linguistic complexity often dilutes the core instruction.
    \end{itemize}
\end{itemize}

For the \emph{\taskname{Conjunctive Search}} task:
\begin{itemize}
    \item \textbf{Effective strategies:}
    \begin{itemize}
        \item \emph{Geometric Noun Synonym Substitution} --- Replacing everyday object nouns with geometric or abstract synonyms maintains the core semantic identity. This strategy is highly effective when the model fluidly maps the new terminology to the visual input.
        \item \emph{Phrasal Verb Conversion} --- Replacing single-word formal action verbs with equivalent common phrasal verbs preserves task integrity. Visual models often ground colloquial, multi-word phrasal actions more reliably than formal synonyms.
        \item \emph{Contextual Grounding Reinforcement} --- Adding precise spatial framing clauses, definite temporal adverbs, or broad observational directives helps safely constrain the model's visual search without altering the core semantics.
        \item \emph{Aspectual Phase Marker Simplification} --- Removing or simplifying aspectual phase markers (such as ``begin to'' or ``start to'') often preserves the core challenge by stripping away rigid temporal onset constraints, allowing the model to focus purely on the core visual action.
        \item \emph{Qualitative Modifier Simplification} --- Removing subjective adjectives or adverbs eliminates unintended threshold constraints. Stripping these qualitative expectations reliably helps the model focus on the core binary state change or action.
    \end{itemize}
    \item \textbf{Ineffective strategies:}
    \begin{itemize}
        \item \emph{Qualitative Modifier Addition} --- Introducing adverbs or adjectives that imply subjective intensity or specific visual manners inadvertently imposes unsupported threshold constraints, creating precise visual expectations that the model fails to consistently detect.
        \item \emph{Specific Feature Constraint Addition} --- Prepending clauses that mandate focus on a narrow visual property (such as shape or specific object forms) often distracts the video model, creating unintended biases in visual search.
        \item \emph{Elevated Action Verb Substitution} --- Replacing simple, common action verbs with formal or elevated synonyms frequently disrupts basic visual grounding by shifting away from expected colloquial actions.
        \item \emph{Collective Quantifier Substitution} --- Replacing explicit numerical quantities with collective terms or groupings (e.g., swapping exact numbers with ``both'' or ``a pair of'') frequently disrupts the model's precise object grounding.
    \end{itemize}
\end{itemize}

\subsubsection{Mutation prompts}
\label{sec:mutation-prompts}

Below we show the full instructions provided to the mutation model for generating ACEs. The text variant and image variant prompts share a common structure with six parts: concept discovery, atomicity rules, a diversity sweep, action templates, the output format, and a worked example. The key differences are that the text variant instructs the model to mutate the task description while the image variant instructs it to propose image edit instructions.

\begin{promptbox}{Instructions for generating text variant ACEs}
You are an expert Concept Mutation Engine for video reasoning task descriptions.
Your task is to extract an ACESet from a given task description using an OBJECTIVE and CONSTITUTION as guidance.

\textbf{IMPORTANT}: The constitution is provided ONLY as guidance --- it highlights directions that have been useful in the past. You MUST still be exploratory and propose ACEs that go beyond the strategies listed in the constitution. Discover novel mutation dimensions that are not covered by any existing strategy. The best ACESets contain a mix of constitution-guided AND independently discovered mutations.

An ACESet represents a set of atomic actions each of which mutate (add, remove or replace) a single concept in the task description.

\textbf{PART 1: CONCEPT DISCOVERY.}
Read the task description carefully and identify every independently mutable concept present. A ``concept'' is any single semantic unit that can be changed in isolation without affecting other parts of the description. This includes --- but is not limited to --- any nouns, verbs, adjectives, adverbs, phrases, spatial or temporal references, quantifiers, style directives, or any other meaningful element you can identify.

Do NOT limit yourself to a predefined set of concept categories. Exhaustively scan the entire task description and discover all concepts that could be rephrased, replaced, removed, or augmented while preserving the core reasoning challenge.

\textbf{PART 2: THE RULE OF STRICT ATOMICITY.}
Each ACE must change exactly ONE concept in the task description. Compound edits are strictly forbidden.
Rule A (Single-Concept Rule): Each ACE must target exactly one concept. If your edit touches two different semantic units, split it into separate ACEs.
Rule B (The ``AND'' Litmus Test): If your verbalization requires ``and'' to explain the change, it is INVALID. Split it into two separate ACEs.

\textbf{PART 3: THE SWEEP --- DIVERSITY IS MANDATORY.}
Systematically walk through every token and phrase in the task description. For each concept you identified in Part~1, propose at least one atomic mutation.
CRITICAL DIVERSITY RULE: Your ACESet MUST contain ACEs that span multiple DISTINCT semantic dimensions. If all your proposed edits fall under the same 1--2 categories (e.g., all verb replacements, or all phrasing changes), your output is INVALID. Each ACE should feel fundamentally different from the others --- targeting a completely unrelated semantic property.

\textbf{PART 4: ACTION TEMPLATES.}
Use ONLY these literal structures for verbalization:
REPLACE $\langle$Existing\_Concept$\rangle$ with $\langle$New\_Value$\rangle$;
SET $\langle$Attribute\_Name$\rangle$ of $\langle$Concept$\rangle$ as $\langle$Value$\rangle$;
REMOVE $\langle$Existing\_Concept$\rangle$.
\end{promptbox}

\begin{promptbox}{Instructions for generating image variant ACEs}
You are an expert Concept Mutation Engine for image edit prompts in a video reasoning benchmark. Your task is to extract an ACESet from the given input using an OBJECTIVE and CONSTITUTION as guidance.

\textbf{IMPORTANT}: The constitution is provided ONLY as guidance --- it highlights directions that have been useful in the past. You MUST still be exploratory and propose ACEs that go beyond the strategies listed in the constitution. Discover novel visual mutation dimensions that are not covered by any existing strategy. The best ACESets contain a mix of constitution-guided AND independently discovered mutations.

An ACESet represents a set of atomic actions each of which proposes a single image edit concept to apply to the reference images.

IMPORTANT: In IMAGE VARIANT MODE. You are designing image edit instructions, NOT text prompts. The original task description in the State tells you what the video task is --- your image edits must keep the scene compatible with that task.

\textbf{PART 1: CONCEPT DISCOVERY.}
Examine the reference images carefully and identify every independently mutable visual concept present. A ``concept'' is any single visual attribute that can be changed in isolation without affecting other visual properties. This includes --- but is not limited to --- any aspect of how the scene looks: rendering, color, texture, material, borders, backgrounds, lighting, perspective, framing, artistic style, domain, or any other visual property you can identify.

Do NOT limit yourself to a predefined set of concept categories. Exhaustively analyze the reference images and discover all visual concepts that could be altered while preserving the spatial structure, element count, and element correspondence required by the original task description.

\textbf{PART 2: THE RULE OF STRICT ATOMICITY.}
Each ACE must change exactly ONE visual concept. Compound edits are strictly forbidden.
Rule A (Single-Concept Rule): Each ACE must target exactly one visual property. If your edit changes two different visual attributes, split it into separate ACEs.
Rule B (The ``AND'' Litmus Test): If your verbalization requires ``and'' to explain the change, it is INVALID. Split it into two separate ACEs.

\textbf{PART 3: THE SWEEP --- DIVERSITY IS MANDATORY.}
Systematically analyze every visual property of the reference images. For each concept you identified in Part~1, propose at least one atomic image edit.
CRITICAL DIVERSITY RULE: Your ACESet MUST contain ACEs that span multiple DISTINCT visual dimensions. If all your proposed edits fall under the same 1--2 categories (e.g., all color changes, or all style changes), your output is INVALID. Each ACE should feel fundamentally different from the others --- targeting a completely unrelated visual property.

\textbf{PART 4: ACTION TEMPLATES.}
Use ONLY these literal structures for verbalization. Each ACE's updated prompt is the IMAGE EDIT INSTRUCTION (not a video task prompt):
REPLACE $\langle$Existing\_Visual\_Concept$\rangle$ with $\langle$New\_Value$\rangle$;
SET $\langle$Visual\_Attribute$\rangle$ of $\langle$element\_or\_scene$\rangle$ as $\langle$Value$\rangle$;
REMOVE $\langle$Existing\_Visual\_Concept$\rangle$.
\end{promptbox}

\noindent
In both variants, the mutation model also receives the exploration objective (a description of the persona, input setup, task preservation rules, generalizability constraints, and success criteria) and two randomly sampled reference images as multimodal input.

\subsubsection{ACE tree search configuration}
\label{sec:tree-search}

The exploration tree has three depth levels with branching factors of $(6, 5, 3)$ at depths 0, 1, and 2, respectively. At each node, one video is generated per reference image (three videos total), and the autorater scores all of them. The objective is considered satisfied for a node if the mean of its autorater scores exceeds the minimum score threshold. Nodes whose objective is already satisfied are not expanded further (i.e., no child mutations are generated from successful nodes).

The exploration follows an interleaved loop: at each depth, all nodes first receive video generations, then autorater scores, and finally---for nodes that did not satisfy the objective---child mutations are generated. The score aggregation function computes the arithmetic mean of all valid binary autorater scores for a node.

\subsubsection{Re-exploration with learned constitutions}
\label{sec:re-exploration}

After learning task-specific constitutions from the initial exploration, we run a second round of ACE exploration for each task-variant combination. The tree search configuration remains identical (branching factors of 6, 5, 3; three videos per node), but the task-specific success thresholds are adjusted based on observed baseline difficulty.

\FloatBarrier

\subsection{Joint text and image prompt engineering}
\label{sec:freeform-joint-vipe}

\begin{figure*}[t]
  \centering
  \scriptsize
  \setlength{\tabcolsep}{3pt}
  \begin{tabular}{c ccccc}
    \midrule
    \rotatebox[origin=c]{90}{\textbf{Conjunctive\ Search}}
    &
    \begin{minipage}[c]{0.17\linewidth}\centering
      \includegraphics[width=\linewidth]{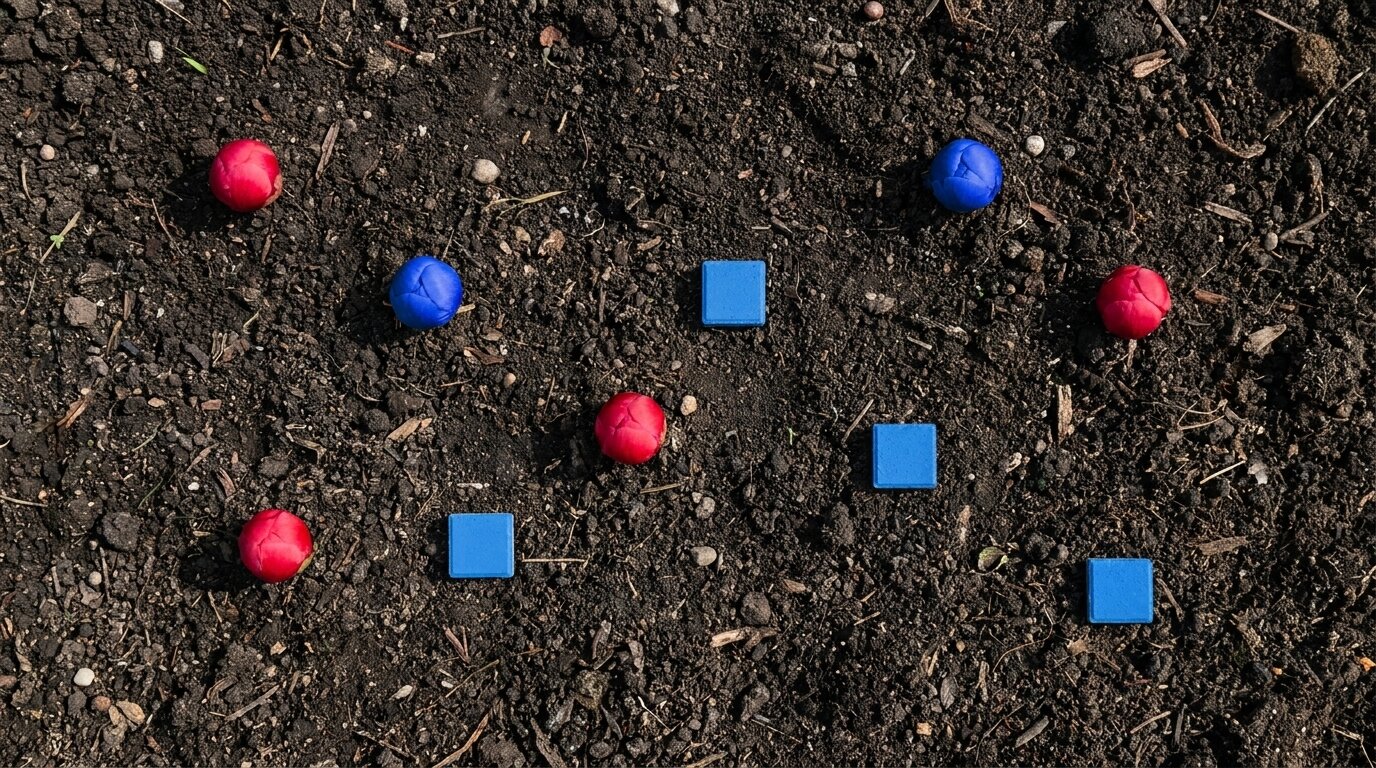}\\[2pt]
      {\tiny In a time-lapse, the two blue flower buds slowly open their petals and bloom into beautiful full flowers.}
    \end{minipage}
    &
    \begin{minipage}[c]{0.17\linewidth}\centering
      \includegraphics[width=\linewidth]{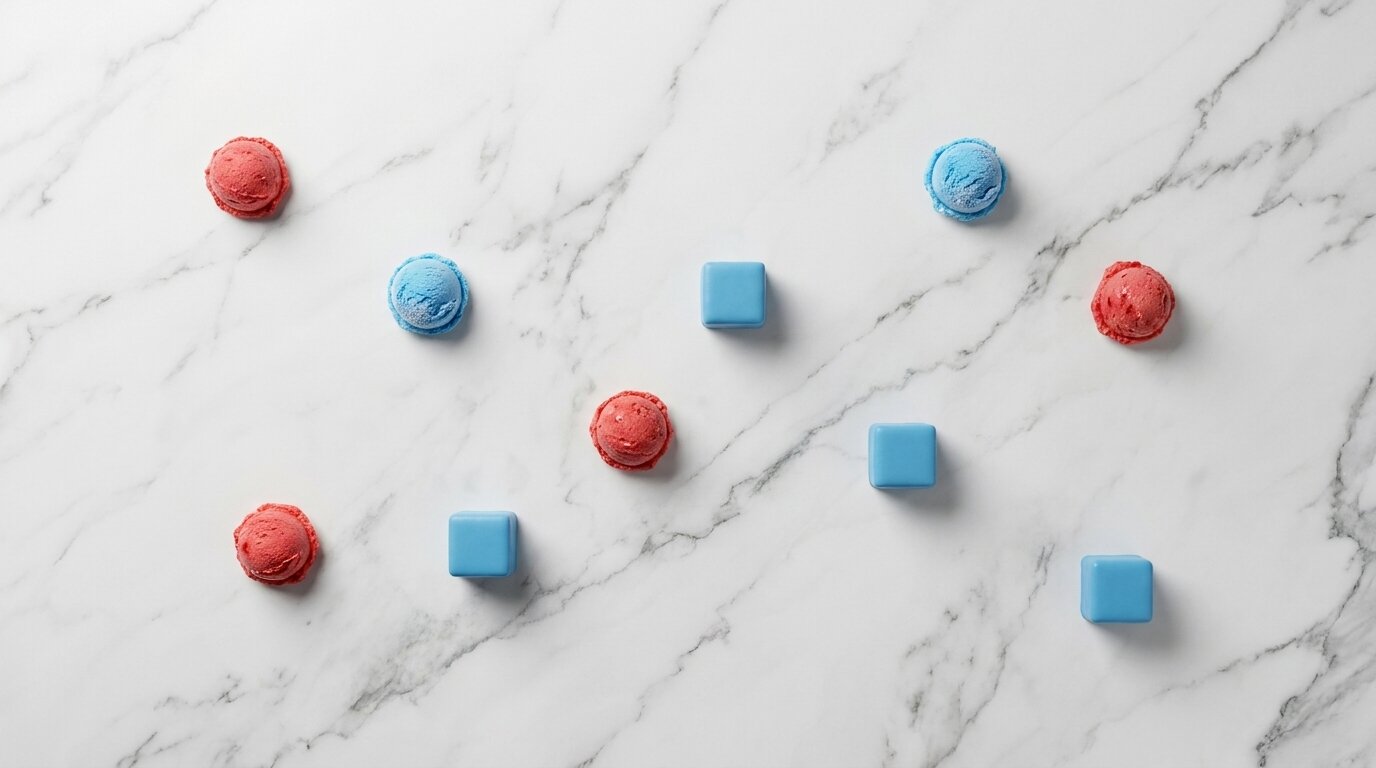}\\[2pt]
      {\tiny The two blue scoops of ice cream begin to melt into liquid puddles.}
    \end{minipage}
    &
    \begin{minipage}[c]{0.17\linewidth}\centering
      \includegraphics[width=\linewidth]{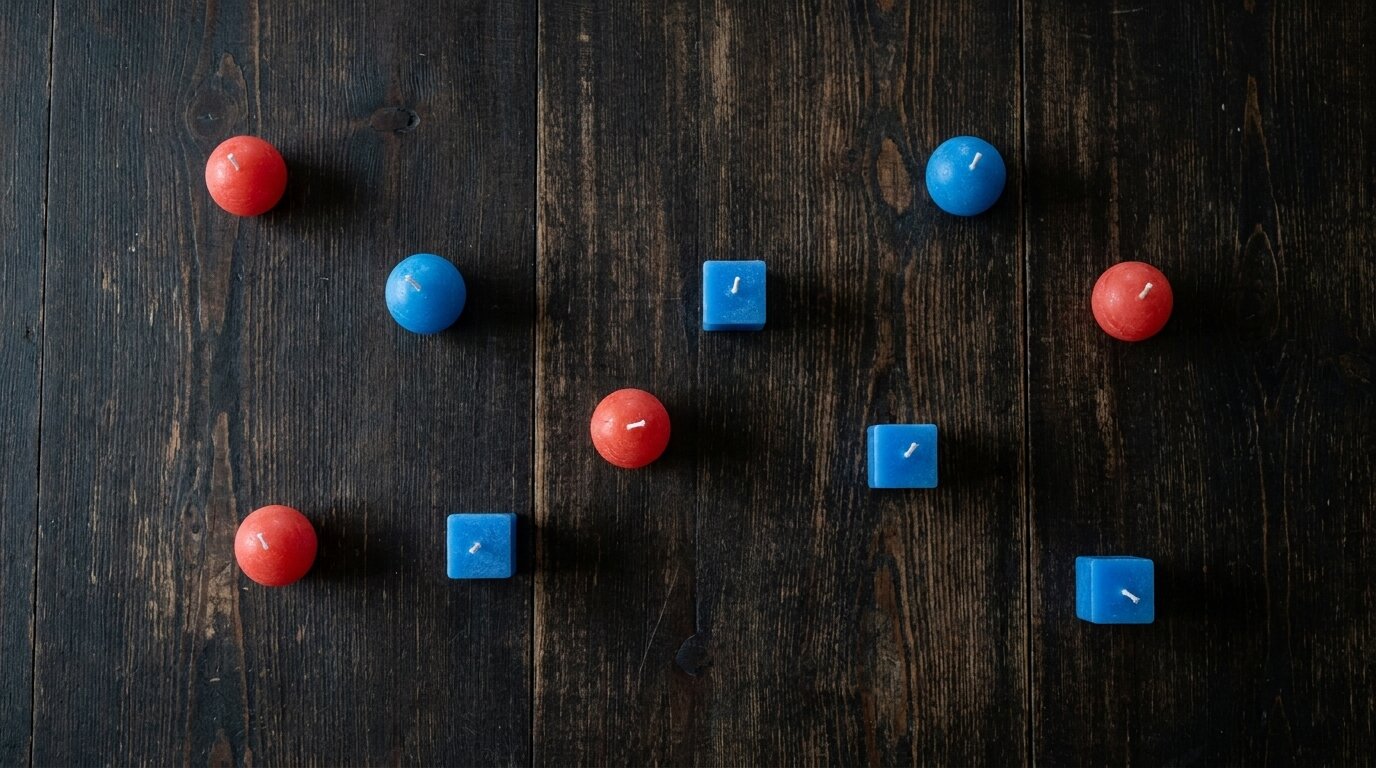}\\[2pt]
      {\tiny The wicks of the two blue spherical candles catch fire and burn with a bright flickering flame.}
    \end{minipage}
    &
    \begin{minipage}[c]{0.17\linewidth}\centering
      \includegraphics[width=\linewidth]{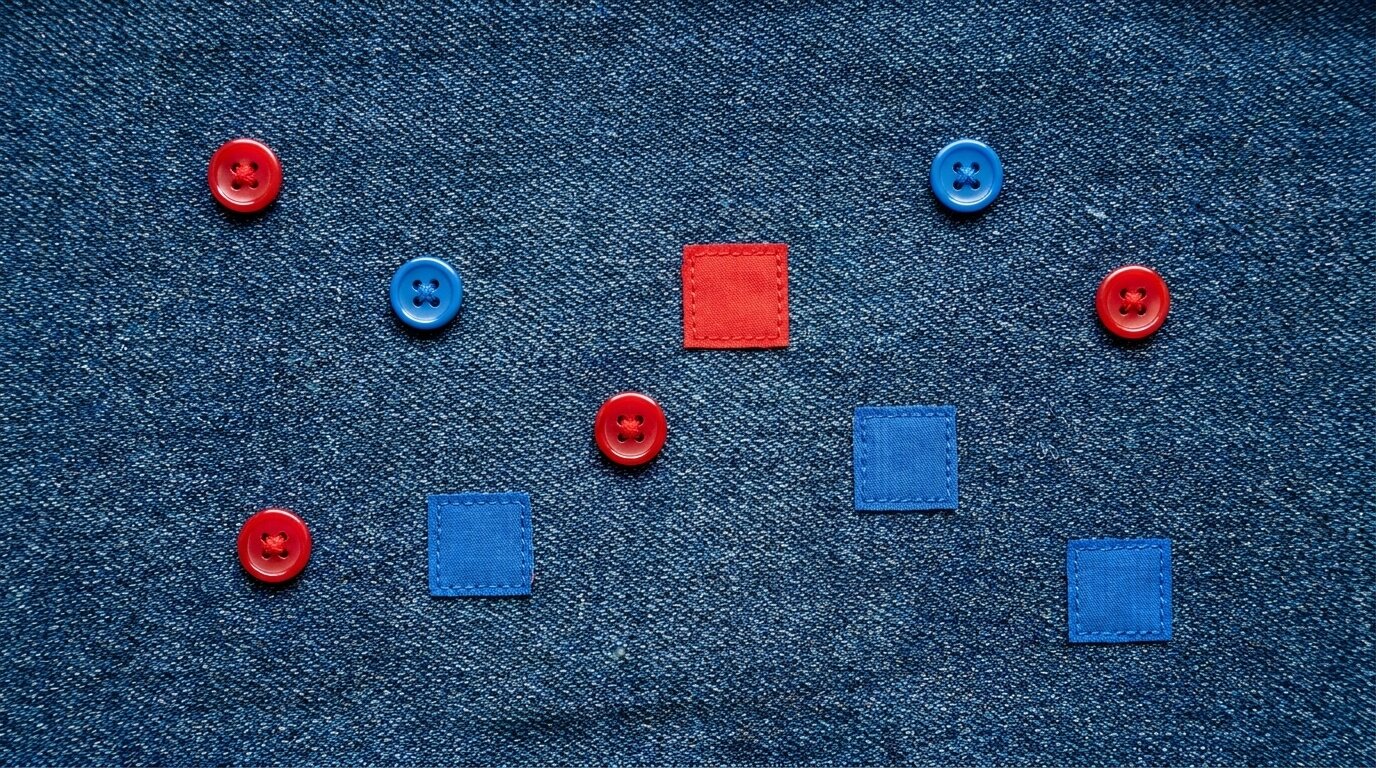}\\[2pt]
      {\tiny The threads holding the two blue buttons suddenly snap, causing them to pop off the fabric and roll away.}
    \end{minipage}
    &
    \begin{minipage}[c]{0.17\linewidth}\centering
      \includegraphics[width=\linewidth]{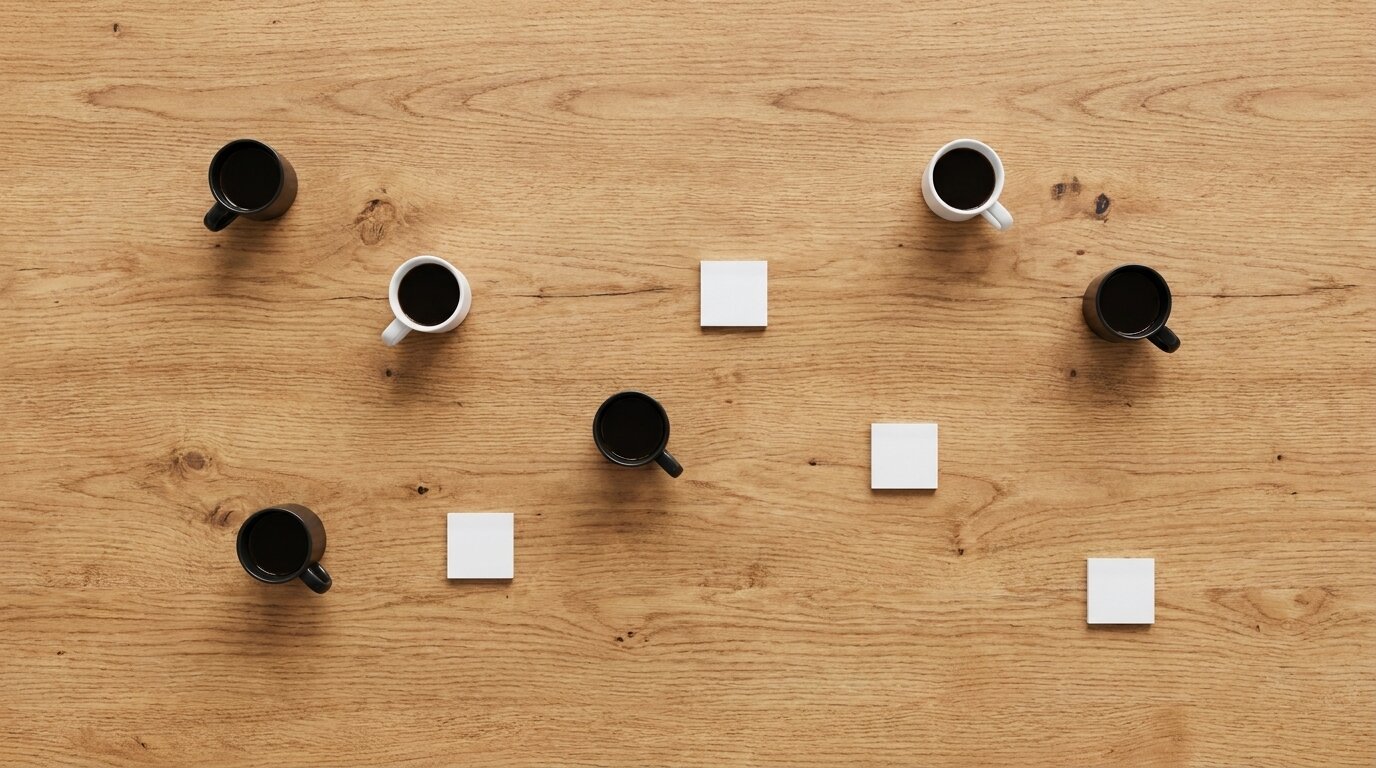}\\[2pt]
      {\tiny The two white coffee mugs begin to emit hot steam.}
    \end{minipage}
    \\
    \midrule
    \rotatebox[origin=c]{90}{\textbf{Sort 3 Num.}}
    &
    \begin{minipage}[c]{0.17\linewidth}\centering
      \includegraphics[width=\linewidth]{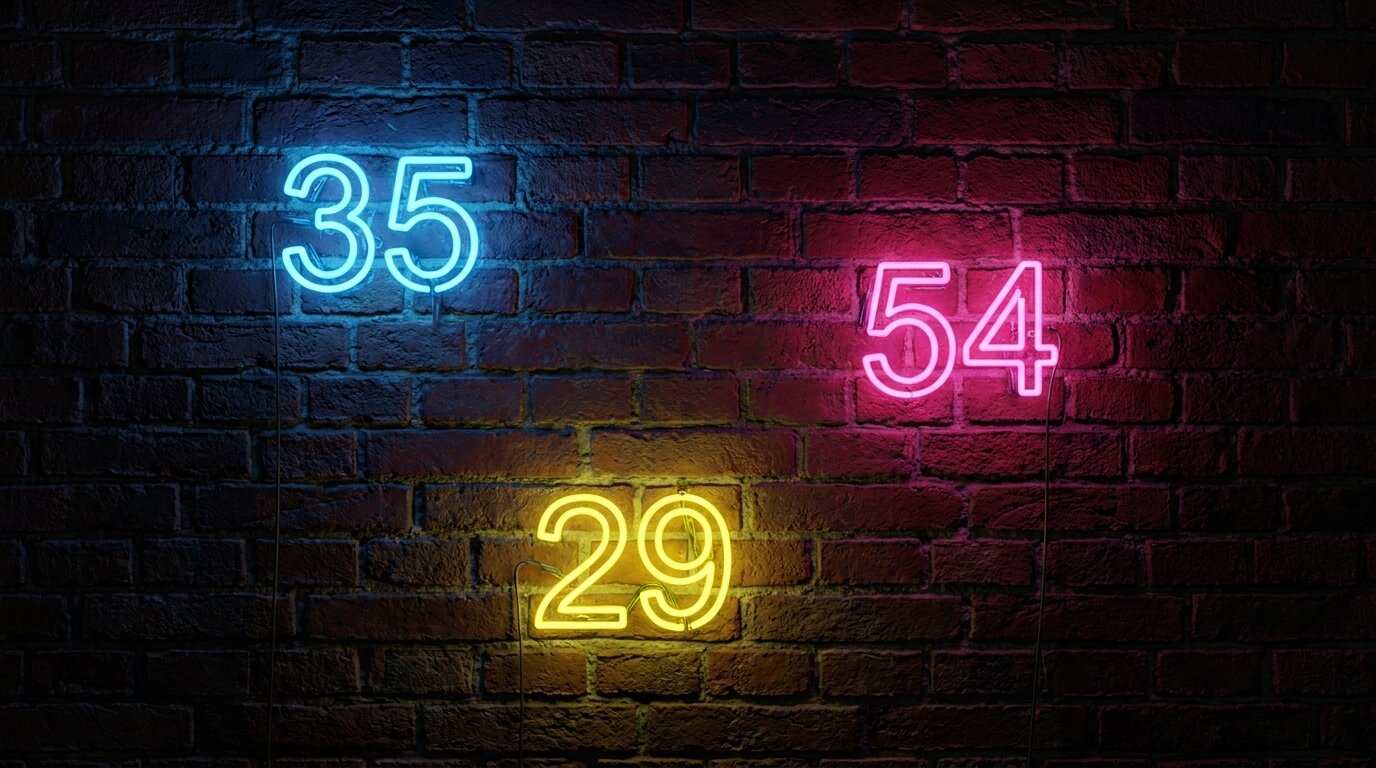}\\[2pt]
      {\tiny The glowing neon numbers flicker and power down, turning completely dark one at a time. The numbers must turn off in numeric order, starting from the smallest number.}
    \end{minipage}
    &
    \begin{minipage}[c]{0.17\linewidth}\centering
      \includegraphics[width=\linewidth]{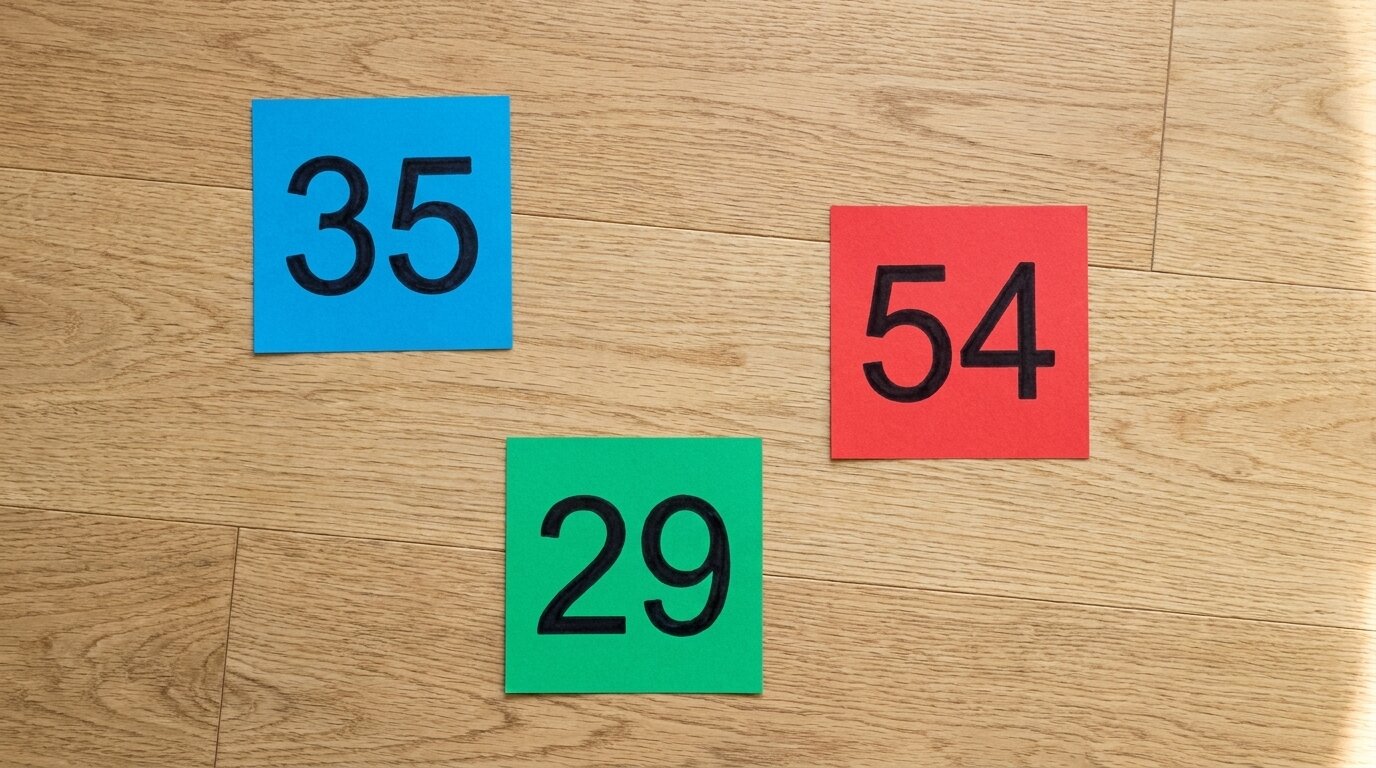}\\[2pt]
      {\tiny A vacuum cleaner hose nozzle enters the frame and sucks up the confetti pieces one at a time. The nozzle must target and remove the confetti in numeric order, starting from the lowest number and ending with the highest.}
    \end{minipage}
    &
    \begin{minipage}[c]{0.17\linewidth}\centering
      \includegraphics[width=\linewidth]{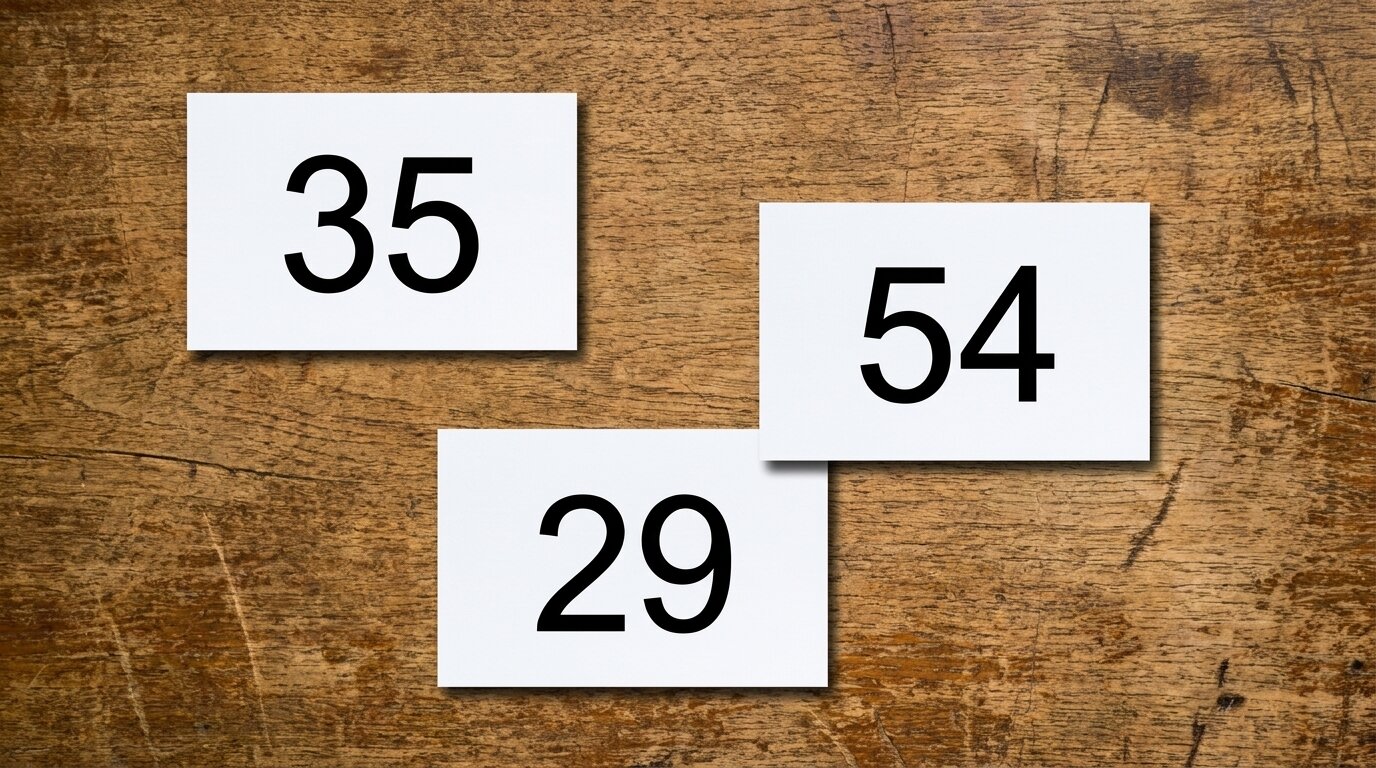}\\[2pt]
      {\tiny A human hand reaches into the frame and picks up the index cards one at a time, removing them from the desk. The hand must pick them up in numeric order, starting from the card with the smallest number.}
    \end{minipage}
    &
    \begin{minipage}[c]{0.17\linewidth}\centering
      \includegraphics[width=\linewidth]{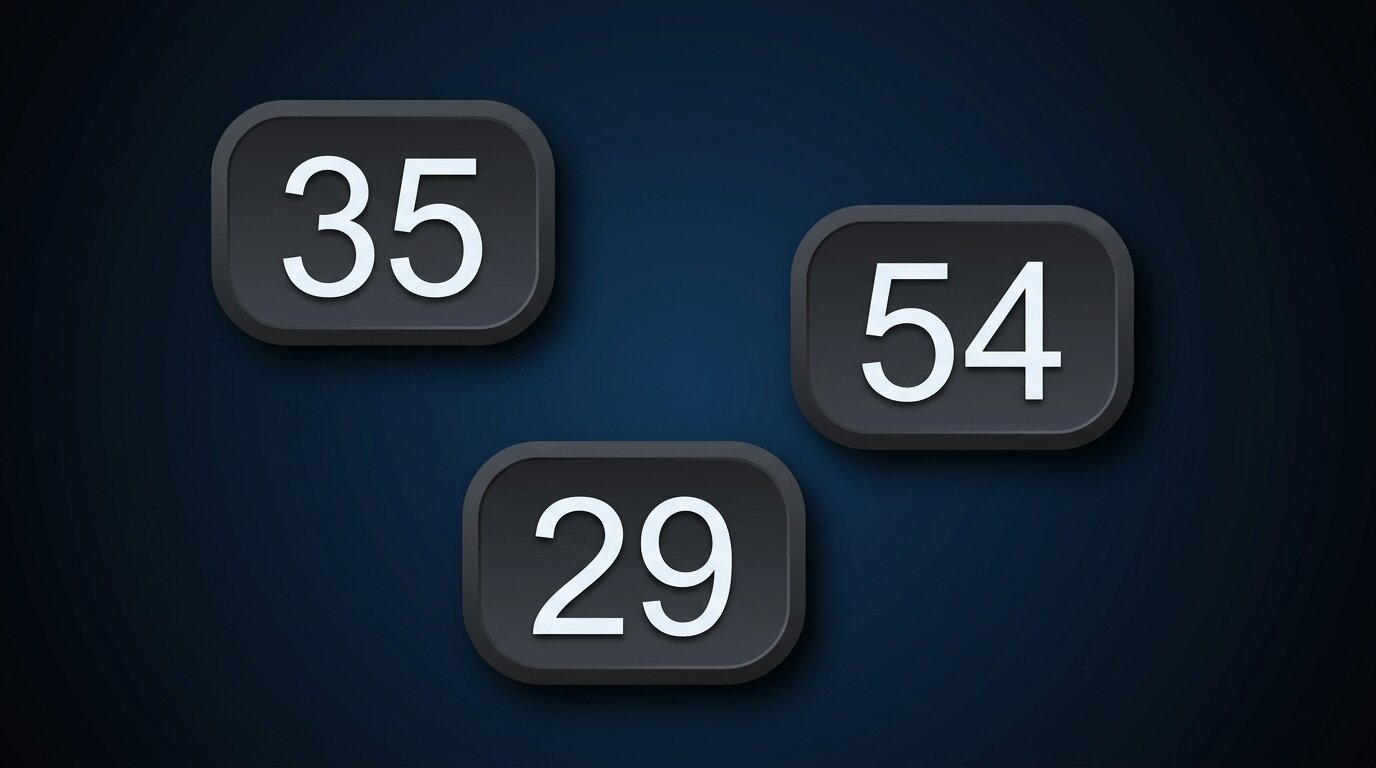}\\[2pt]
      {\tiny A digital mouse cursor moves across the screen and clicks on the buttons one at a time, causing each button to vanish immediately after being clicked. The cursor must click the buttons in numeric order, starting from the smallest number.}
    \end{minipage}
    &
    \begin{minipage}[c]{0.17\linewidth}\centering
      \includegraphics[width=\linewidth]{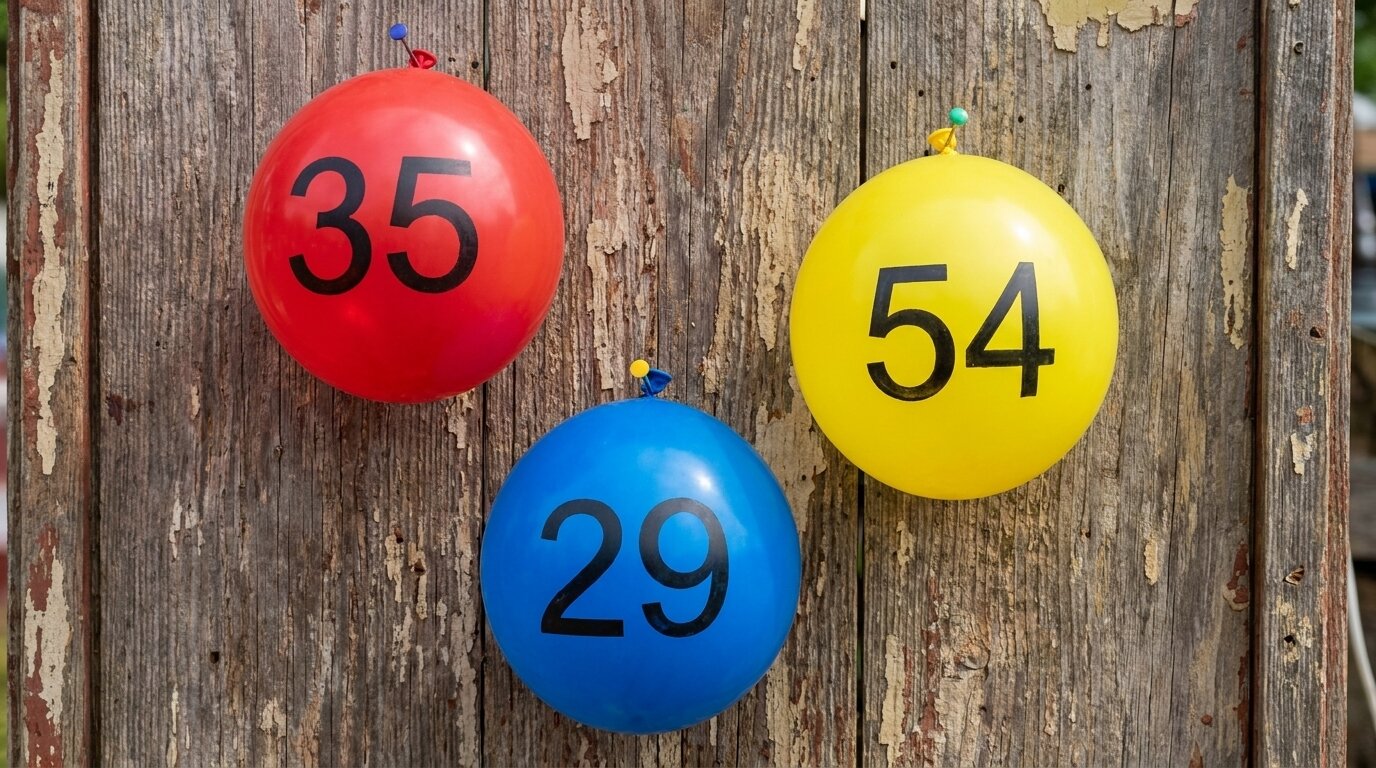}\\[2pt]
      {\tiny Metal darts fly into the frame from off-screen and strike the balloons, causing them to burst and disappear one at a time. The balloons must be popped in ascending numeric order, starting from the balloon with the smallest number.}
    \end{minipage}
    \\
    \midrule
    \rotatebox[origin=c]{90}{\textbf{\taskname{Connect the Dots}}}
    &
    \begin{minipage}[c]{0.17\linewidth}\centering
      \includegraphics[width=\linewidth]{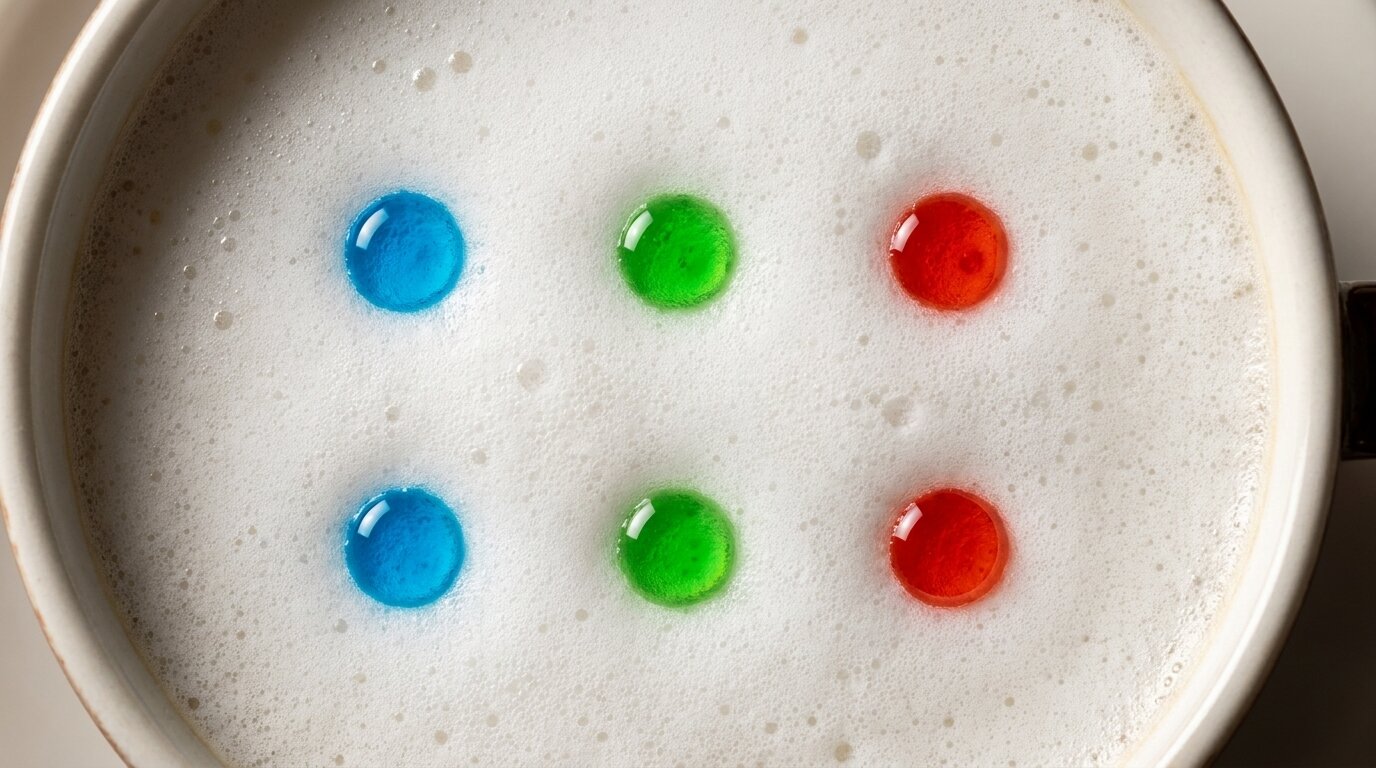}\\[2pt]
      {\tiny A top-down macro view of smooth, white milk foam with several brightly colored circular droplets of syrup resting on the surface. A thin metal barista etching tool dips into the foam and actively drags a continuous path from one colored droplet to its matching pair, leaving a distinct, visible indented line in the foam to connect them.}
    \end{minipage}
    &
    \begin{minipage}[c]{0.17\linewidth}\centering
      \includegraphics[width=\linewidth]{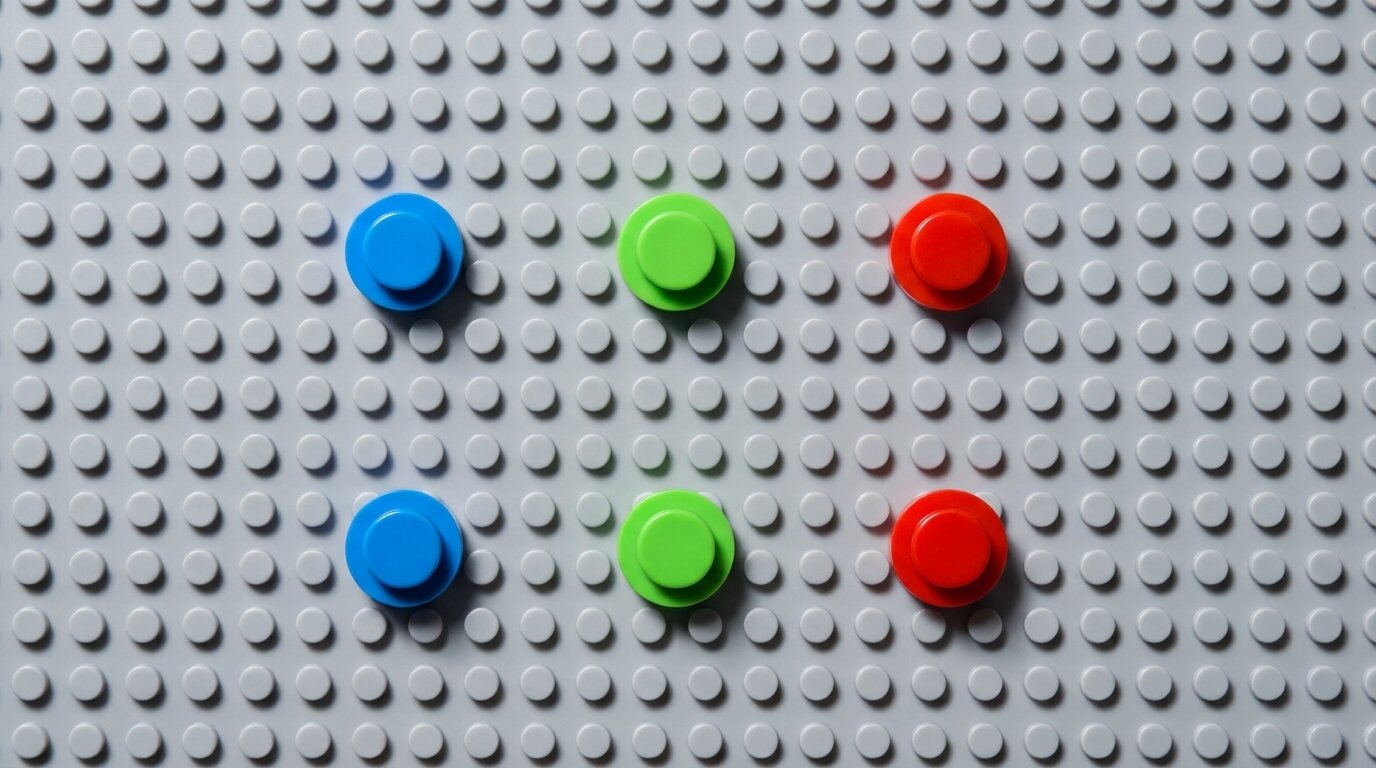}\\[2pt]
      {\tiny Connect each pair of same-colored round pegs. The video must be in a choppy, stop-motion animation style. For each matching pair, rectangular plastic building bricks of the same color should snap onto the baseplate piece-by-piece, building a continuous, jagged path between the two pegs.}
    \end{minipage}
    &
    \begin{minipage}[c]{0.17\linewidth}\centering
      \includegraphics[width=\linewidth]{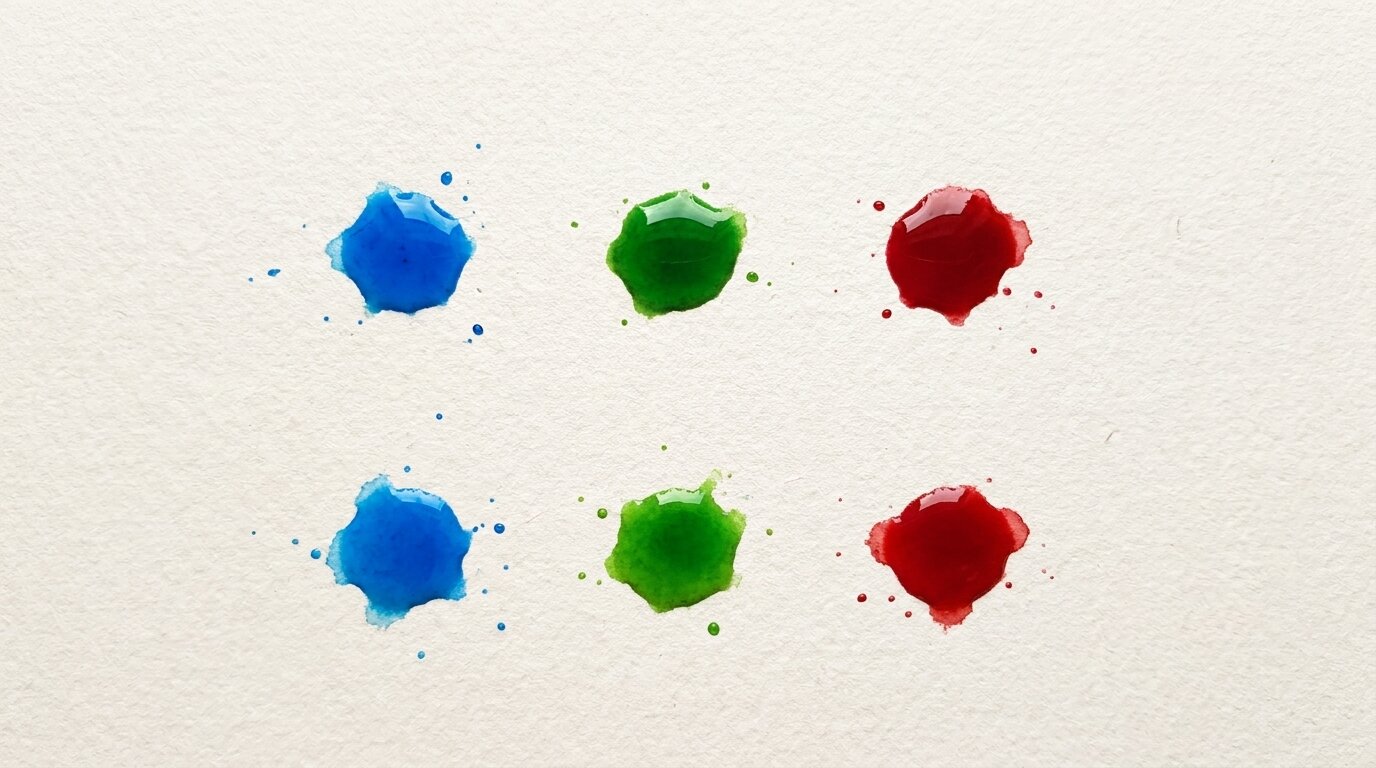}\\[2pt]
      {\tiny Draw a flowing ink stroke to connect each pair of same-colored ink pools.}
    \end{minipage}
    &
    \begin{minipage}[c]{0.17\linewidth}\centering
      \includegraphics[width=\linewidth]{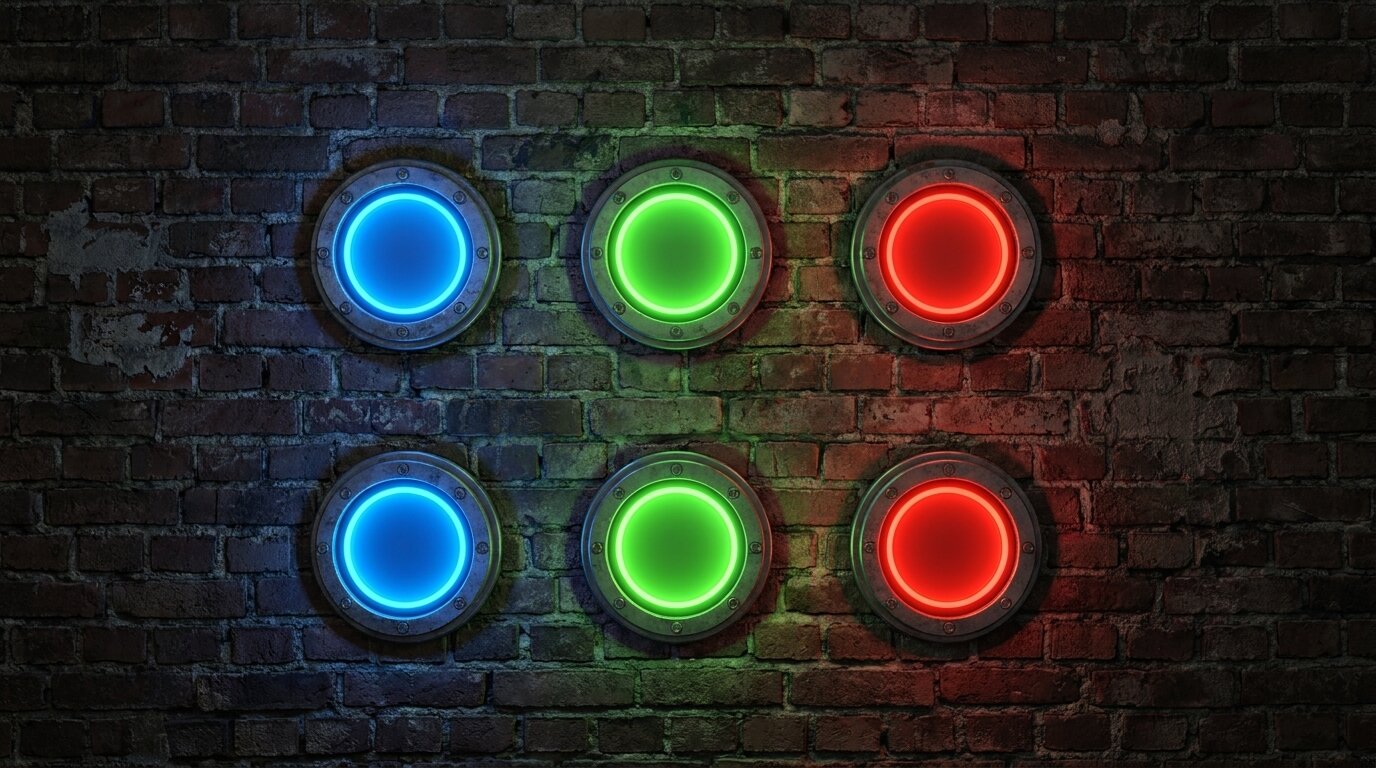}\\[2pt]
      {\tiny Route a continuous glass neon tube between each pair of same-colored mounting brackets, and then illuminate the tubes.}
    \end{minipage}
    &
    \begin{minipage}[c]{0.17\linewidth}\centering
      \includegraphics[width=\linewidth]{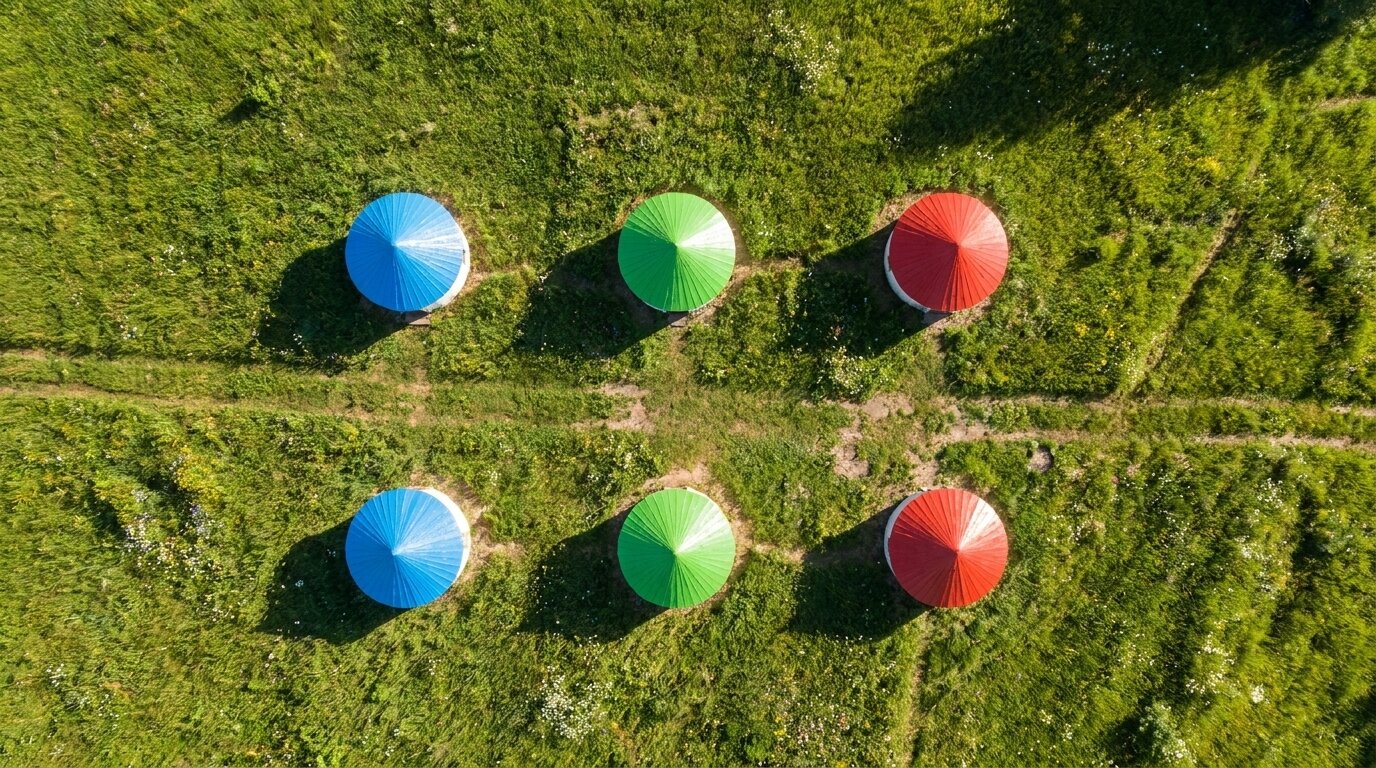}\\[2pt]
      {\tiny Animate asphalt roads dynamically paving themselves across the grassy terrain to connect each pair of identically colored buildings.}
    \end{minipage}
    \\
  \end{tabular}
  \caption{Joint text and image ideation variants. Each row shows a task with five
  example variants that modify both the input image and prompt simultaneously.}
  \label{fig:textimage-variants}
\end{figure*}

We further adapt the image-only and text-only freeform prompt ideators discussed in~\cref{subsec:freeform-ideation} and~\cref{sec:images-vs-words} to modify both the image and the text prompt simultaneously and produce alternative representations of a task. For example, the task variants transformed \taskname{Conjunctive Search} as blooming flower buds or \taskname{Sort 3 Numbers} as powering down neon signs. \cref{fig:textimage-variants} shows example joint variants across three tasks, and~\cref{tab:textimage-pass1-summary} reports the best and average pass@1 rates across $n=20$ joint proposals per task.

\begin{promptbox}{Joint text+image edit ideation prompt}
\textsc{system instruction}

You are an expert AI researcher designing task variants for a visual reasoning benchmark. Each task is presented through two levers: the visual scene (how images look) and the video event (what happens to express the answer). Your goal is to explore diverse alternative presentations where a video generation model might perform better, while preserving the core reasoning challenge.
Each proposal should be distinct from previous ones. Aim for diversity in visual domains, interaction styles, and prompt strategies.
\tcbline
\textsc{user content}

Based on the base task and the sample images provided, propose a new task variant.
The variant should:
\begin{itemize}[nosep]
\item Preserve the core reasoning challenge
\item Consider alternative visual scenes and/or video events the model may handle better
\end{itemize}
Output your proposal as a JSON block:
\begin{verbatim}
{"variant_id": "short_snake_case_id",
 "variant_description": "Brief description of the variant
  and the visual domain change",
 "instruction_variant": "Text prompt for the video generation
  model. Describe what event should unfold over time without
  leaking the solution to the task. Use the Base task
  description as a reference.",
 "image_edit_prompt": "Detailed prompt for an image editing
  model to transform the original task images into this
  variant's visual style. Be specific about what each visual
  element should become.",
 "autorater_rubric": "Criteria for evaluating the generated
  video. Describe desired behavior (what correct/partial/
  incorrect outcomes look like) AND undesired behavior to
  penalize (e.g., visual artifacts, glitches, unrealistic
  motion, object distortion)."}
\end{verbatim}
\begin{itemize}[nosep, leftmargin=0pt]
\item[] Base task instruction: \texttt{[base task description]}
\item[] Sample images: \texttt{[sample images]}
\item[] Past proposals: \texttt{[descriptions of all prior proposals]}
\end{itemize}
\end{promptbox}
\FloatBarrier

\begin{table}[t]
  \centering
  \small
  \caption{Pass@1 for $n=20$ freeform joint text and image variants. Best = highest single variant; Avg = mean $\pm$ std.}
  \label{tab:textimage-pass1-summary}
  \begin{tabular}{l cc}
    \toprule
    \textbf{Task} & \textbf{Best} & \textbf{Avg $\pm$ Std} \\
    \midrule
    \taskname{Conjunctive Search} & 12.0\% & 2.5 $\pm$ 3.7\% \\
    \taskname{Sort 3 Numbers} & 82.0\% & 24.5 $\pm$ 22.4\% \\
    \taskname{Connect the Dots} & 6.7\% & 1.4 $\pm$ 2.2\% \\
    \bottomrule
  \end{tabular}
\end{table}

\section{Test-time scaling details}
\label{app:test-time-scaling}
\begin{figure}[h]
    \centering
    \includegraphics[width=\linewidth]{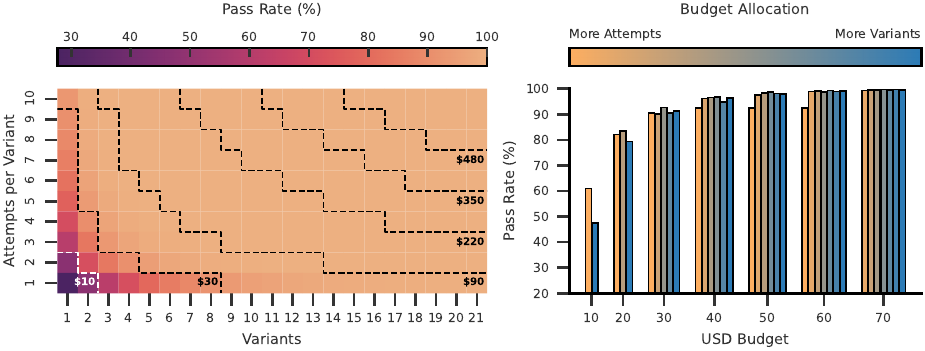}
    \caption{Test-time scaling budget analysis for \taskname{Maze 3$\times$3}.}
    \label{fig:test-time-scaling-maze}
\end{figure}

\begin{figure}[h]
    \centering
    \includegraphics[width=\linewidth]{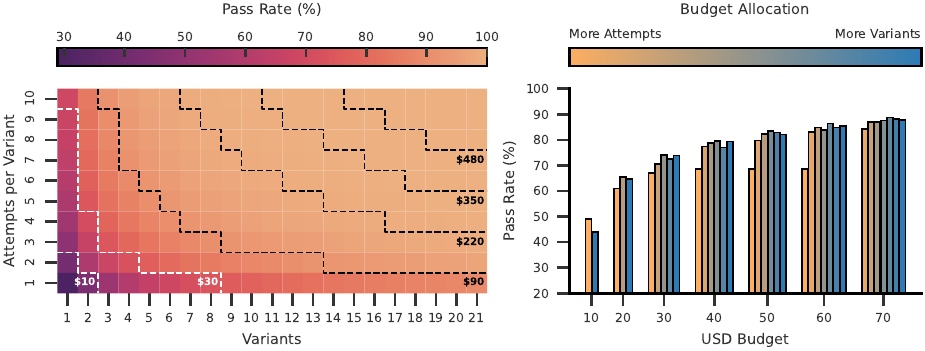}
    \caption{Test-time scaling budget analysis for \taskname{RushHour} Level 2.}
    \label{fig:test-time-scaling-rushhour}
\end{figure}

We add test-time scaling budget allocation results for \taskname{Maze} and \taskname{RushHour} similar to~\cref{fig:budget-tradeoffs} in~\cref{fig:test-time-scaling-maze,fig:test-time-scaling-rushhour}. Since self-consistency (majority voting) is not straight-forward for these tasks (outcomes are not simple choices but trajectories/sequences), we report pass rate as a proxy instead. Assuming a heuristic or algorithm for aggregating multiple \taskname{Maze} trajectory or \taskname{RushHour} sequences, VIPE also represents a viable and cost-effective test-time scaling strategy for these tasks.

\FloatBarrier

\section{Image model comparison}
\label{app:nano_banana}
This section contains a more detailed discussion of the results from~\cref{sec:image-gen-models}.

\paragraph{VIPE provides diminished or negative benefit for image generation
models}
For NB~Pro in image mode, VIPE yields only a modest +5\,pp improvement in
Pass@1 over raw sketches (50\% vs.\ 45\%). This is far smaller than the
+18\,pp improvement that VIPE provides for Veo~3.1 on the same task
(\cref{fig:vpct}). More strikingly, for NB~Flash in image mode,
VIPE hurts; Pass@1 drops from 37\% (sketch) to 21\% (VIPE), a
$-16$\,pp degradation. Majority vote accuracy tells a similar story, falling
from 48\% to 32\%. Why does this happen? 

\paragraph{The image generation pathway is a bottleneck}
The comparison between NB~Flash in image mode and text mode is especially
revealing, since these use the \emph{exact same underlying model}. In text mode, the model simply outputs a label; in
image mode, it must generate an image depicting the ball's final position.
On VIPE inputs, switching from image to text mode improves Pass@1 from 21\%
to 50\% (+29\,pp). On sketch inputs, the improvement is +21\,pp (37\%
$\rightarrow$ 58\%). The text-mode model matches NB~Pro in image mode (50\% Pass@1 on VIPE), demonstrating that the physics
reasoning capability is present within NB~Flash but is obscured by
the image generation pathway. Generating a physically accurate image
requires not only knowing the answer but also rendering it
correctly by composing the scene, placing the ball in the right bucket, and
maintaining visual consistency which is an additional burden that degrades effective
accuracy.

Interestingly, we also observe that Gemini~3.1~Pro (text-only VLM) achieves 96\%
accuracy on \taskname{VPCT} sketches (\cref{tab:vpct_control_conditions}), while NB~Pro
(image mode) achieves only 50\%, a 46\,pp gap. This gap highlights a fundamental asymmetry: expressing an answer through text is far easier than expressing it through image generation. The
image generation pathway introduces an additional ``rendering bottleneck'' that
degrades effective accuracy, even when the underlying reasoning is correct. This is
analogous to the performance vs.\ competence distinction~\citep{firestone2020performance}: the model has the competence but sometimes cannot express
it reliably through the image modality.

\paragraph{Sketch + text is the optimal configuration}
The best overall configuration was NB~Flash in text mode on sketch inputs which
achieves 58\% Pass@1 and 80\% majority vote accuracy. This outperforms
NB~Pro in image mode on VIPE inputs (50\% Pass@1, 54\% majority vote) without any visual prompt
engineering. The 80\% majority vote accuracy suggests that the model's
errors are well-distributed and can be effectively corrected through
ensembling.
Taken together, these results suggest that for physics reasoning on \taskname{VPCT},
the optimal strategy is to bypass both the VIPE translation and the
image generation pathway, instead using the image model's native text
reasoning capabilities on the original sketches.

\section{Autoraters}
\label{sec:autorater-details}
Each video is scored based on a task-specific definition of success, the engineered image prompt, and sample-specific information such as initial object positions. For VLM-based autoraters, the rubric has to be adjusted to fit the final image prompt variant, e.g., to mention specific appearances of various objects of interest. In this case, we instruct an LLM to adjust a base rubric based on the ideator output. \taskname{Maze}, \taskname{Connect the Dots}, \taskname{Sort 3 Numbers}, \taskname{Conjunctive Search}, and \taskname{VPCT} use Gemini~3.1~Pro as an autorater. \taskname{RushHour} uses an algorithmic autorater.

\subsection{Rubric-based autorater}
\label{sec:rubric-autorater}

For \taskname{Conjunctive Search}, \taskname{Sort 3 Numbers}, and \taskname{Connect the Dots}, we use a VLM-based autorater to evaluate whether each generated video correctly solves the task. The autorater operates in two steps: (1)~\emph{frame captioning}: each video is sampled at 1 FPS, and a VLM generates a caption for each frame describing the visual changes relative to the starting frame; (2)~\emph{scoring}: the autorater receives the input image, the frame captions, the full video, and a task-specific evaluation rubric, then outputs a score on a three-point scale (0, 0.5, 1) with a brief reasoning justification. A video is counted as a pass if and only if its score exceeds 0.5. We used Gemini~3.1~Pro for all autorater calls. For these tasks, the autorater rubric was generated alongside the freeform ideator's task variant proposals to cater the rubric for the particular text instruction or visual context.

\begin{promptbox}{Example generated autorater rubric (conjunctive search)}
Correct: Both blue spherical bulbs emit a clear, radiant glow. Partial: Only one blue bulb glows, or the glow is extremely weak. Incorrect: Red bulbs glow, square gels glow, no illumination occurs, or severe morphological artifacts distort the objects.
\end{promptbox}

\begin{promptbox}{Example generated autorater rubric (sort 3 numbers)}
Correct: Number sprites explode into particles (pop) and vanish one by one in strictly ascending numerical order. Partial: Sprites pop in the wrong order or simultaneously. Incorrect: Sprites fade out smoothly instead of popping, remain static, or exhibit non-retro visual artifacts.
\end{promptbox}

\begin{promptbox}{Example generated autorater rubric (connect dots)}
Correct: A continuous line (e.g., a physical patch cable or drawn wire) connects only the identically colored jacks. Partial: Incomplete connections or lines connecting mismatched colors. Incorrect: No lines formed, visual artifacts, jacks moving, or the background panel warping.
\end{promptbox}

\begin{promptbox}{Frame captioning prompt}
\textsc{system instruction}

Compare this frame from a video to the provided starting frame (Input Image). Describe the visual change in this frame relative to the starting frame. Be brief and concise. Focus only on significant visual differences in objects and their motion or state.

Output your description as a JSON block:
\begin{verbatim}
{"caption": "Brief description of the visual change"}
\end{verbatim}
\tcbline
\textsc{user content}
\begin{itemize}[nosep,leftmargin=*]
\item[] Starting Frame (Input Image): \texttt{[input image]}
\item[] Current Frame (Frame $i$ at $t$s): \texttt{[video frame $i$]}
\end{itemize}
\end{promptbox}

\begin{promptbox}{Scoring prompt}
\textsc{system instruction}

You are evaluating whether a video correctly solves a visual reasoning task. You will see the input image (the starting state) and the generated video (the model's output).

Evaluation rubric:
\texttt{\{autorater\_rubric\}}

IMPORTANT: Be lenient about visual or motion artifacts. Minor glitches, flickering, warping, or unnatural motion should NOT reduce the score unless they directly indicate a failure to understand the core reasoning requirement described in the rubric.

Rate the video's correctness on a 0--1 scale:
\begin{itemize}[nosep,leftmargin=*]
\item 1.0: Correctly demonstrates the task behavior
\item 0.5: Partially correct (some task elements right, some wrong)
\item 0.0: Completely incorrect or shows clear misunderstanding of the task
\end{itemize}

Output your evaluation as JSON:
\begin{verbatim}
{"score": 0.0,
 "reasoning": "Brief explanation of what was correct/incorrect"}
\end{verbatim}
\tcbline
\textsc{user content}
\begin{itemize}[nosep,leftmargin=*]
\item[] Input image for the task: \texttt{[input image]}
\item[] Detailed frame captions describing visual changes: \texttt{[frame captions text]}
\item[] Generated video: \texttt{[video, sampled at 1 FPS]}
\end{itemize}
\end{promptbox}

\paragraph{Human agreement}
We validated the autorater against human ratings on a sample of 216 videos drawn from the three tasks (see~\cref{tab:rubric-autorater-agreement}). Across all videos, the autorater achieves 85.6\% agreement with the human rater (Cohen's $\kappa = 0.711$). When considering only clear-cut cases where we exclude videos that the human rater marked as debatable (177 cases), agreement rises to 93.8\% ($\kappa = 0.874$). Disagreements show that the autorater tends to be more conservative: the autorater does not produce false positives (i.e., it does not pass videos that the human rater fails). We do not see disagreement disproportionally affecting tasks or text vs. image variants.

\paragraph{VIPE for autorating} In some cases, we found a simple deterministic VIPE procedure to be beneficial for the VLM autorater: in the \taskname{Maze} task splits, overlaying a static $16 \times 9$ grid (independent of the maze location or grid size) on the video before rating increased autorater alignment with human ratings from 0.25 to 0.58 Cohen's Kappa. This simple trick is not universally effective, though. On the \taskname{RushHour} task, the same grid led to a degradation in rater alignment by -0.12 Cohen's Kappa, and both variants of Gemini raters had lower agreement than our final deterministic autorater.

\begin{table}[h]
\centering
\small
\caption{Rubric-based autorater--human agreement on 216 videos.}
\label{tab:rubric-autorater-agreement}
\begin{tabular}{lr}
\toprule
\textbf{Metric} & \textbf{Value} \\
\midrule
Agreement & 188/216 (87.0\%) \\
Cohen's $\kappa$ & 0.732 \\
Agreement (clean-cut cases) & 168/177 (94.9) \\
Cohen's $\kappa$ (clean-cut cases) & 0.896 \\
Agreement on image variants & 58/68 (85.3\%) \\
Agreement on text variants & 124/140 (88.6\%) \\
\bottomrule
\end{tabular}
\end{table}

\subsection{VPCT container-choice autoraters}
\label{sec:vpct-autorater}

\paragraph{VLM-based autorater}
To score video generations across a variety of \taskname{VPCT} variants in~\cref{subsec:freeform-ideation} and~\cref{sec:images-vs-words}, we use a VLM-based autorater that identifies the \emph{first} object-container contact event, and compare this container choice with the ground-truth labels. Similar to the rubric-based autorater in~\cref{sec:rubric-autorater}, we first sample frames at 6 FPS and generate frame captions tracking the ball position and ball-container interactions. Then, we feed the input image, the frame captions, and the generated video for the autorater to extract the selected container from the video prediction. In all experiments, we use Gemini~3.1~Pro as the autorater.

\begin{promptbox}{VPCT frame captioning prompt}
\textsc{system instruction}

You are analyzing a single frame from a ball-drop physics simulation video. The video shows a ball or the main round target object falling through obstacles toward three containers arranged at the bottom of the scene: left, center, and right.

You are provided with:
\begin{enumerate}[nosep,leftmargin=*]
    \item \textbf{Starting Frame (Input Image):} Shows the initial scene with the ball, obstacles, and three containers.
    \item \textbf{Current Frame:} The frame you must analyze.
\end{enumerate}

Describe what you observe in the current frame, focusing on:
\begin{itemize}[nosep,leftmargin=*]
    \item Where is the ball relative to the obstacles and containers?
    \item Is the ball touching or entering any container? If so, which one (left, center, or right)?
    \item Are all three original containers still visible in the frame?
    \item Has the camera moved significantly or has the scene changed from the starting frame?
\end{itemize}

Output your description as a JSON block:
\begin{verbatim}
{"caption": "Description of ball position, container visibility,
 any ball-container contact, or any significant scene changes"}
\end{verbatim}
\tcbline
\textsc{user content}
\begin{itemize}[nosep,leftmargin=*]
\item[] Starting Frame (Input Image): \texttt{[input image]}
\item[] Current Frame (Frame $i$ at $t$s): \texttt{[video frame $i$]}
\end{itemize}
\end{promptbox}

\begin{promptbox}{VPCT judgment prompt}
\textsc{system instruction}

You are evaluating a ball-drop physics simulation video. The video shows a ball or the main round target object falling through obstacles toward three containers arranged at the bottom of the scene: left, center, and right.

Your task is to determine which container the ball falls into FIRST. Specifically, identify the first moment the ball touches or enters any container. The ball must be actually touching the container edge or going into the container --- floating nearby without contact does not count.

Important guidelines:
\begin{itemize}[nosep,leftmargin=*]
    \item Focus on the FIRST ball-container contact event.
    \item If the ball splits into copies, track the first contact by any copy.
    \item If obstacle boards move or morph, that is okay --- still look for the first ball-container contact.
    \item If the ball never reaches any container by the end of the video, report ``unknown.''
    \item If the camera moves significantly or the scene transforms to a different setting so that all three original containers leave the frame, report ``unknown.''
\end{itemize}

Output your evaluation as JSON:
\begin{verbatim}
{"predicted_container": "left or center or right or unknown",
 "reasoning": "Explanation of what you observed"}
\end{verbatim}
\tcbline
\textsc{user content}
\begin{itemize}[nosep,leftmargin=*]
\item[] Input image showing the initial scene with ball, obstacles, and three containers: \texttt{[input image]}
\item[] Detailed frame-by-frame captions describing ball movement: \texttt{[captions text]}
\item[] Generated video: \texttt{[video, sampled at 6 FPS]}
\end{itemize}
\end{promptbox}

\paragraph{Human agreement}
We validated the \taskname{VPCT} autorater against human ratings on 100 videos drawn from 20~image-variants and 20~text-variants. \Cref{tab:vpct-autorater-agreement} summarizes the results. The autorater-human consistency is similar across text and image variants. We observe that the autorater tends to be more conservative, marking "unknown" predictions more often than the human rater due to scene distortion.

\begin{table}[h]
\centering
\small
\caption{\taskname{VPCT} autorater--human agreement on 100 videos.}
\label{tab:vpct-autorater-agreement}
\begin{tabular}{lr}
\toprule
\textbf{Metric} & \textbf{Value} \\
\midrule
Agreement (4-class) & 86/100 (86.0\%) \\
Cohen's $\kappa$ (4-class) & 0.793 (substantial) \\
Agreement on image variants & 43/50 (86.0\%) \\
Agreement on text variants & 43/50 (86.0\%) \\
\bottomrule
\end{tabular}
\end{table}

\subsection{Maze autoraters}
\label{app:maze-autorater}
\paragraph{VLM-based autorater}
We first explore the use of an VLM-based autorater for this task, similar to~\cref{sec:vpct-autorater}. Due to the large number of generated videos across four splits, we forgo frame captions and directly ask the VLM to rate the correctness of the entire video using the prompt below. The placeholders \texttt{\{runner\}}, \texttt{\{goal\}}, \texttt{\{path\}}, \texttt{\{walls\}} are dynamically filled based on the maze variant, e.g., \texttt{red circle}, \texttt{green circle}, \texttt{white path}, \texttt{black walls} for the original variant but \texttt{glowing red orb/circle}, \texttt{glowing green orb/circle}, \texttt{glowing white path/strip}, \texttt{black/dark background} for the \texttt{neon\_motion\_graphics} variant. The specific fill values are determined by Gemini~3.1~Pro based on the description of each variant. We use Gemini~3.1~Pro for rating.

\begin{promptbox}{Maze scoring prompt}
\texttt{[video]}

You are an expert video analyzer. Analyze the video which shows a maze puzzle being solved. In this puzzle:
\begin{itemize}[nosep,leftmargin=*]
    \item The runner is a \texttt{\{runner\}}.
    \item The goal is a \texttt{\{goal\}}.
    \item The path is \texttt{\{path\}}.
    \item The walls are \texttt{\{walls\}}.
    \item The runner needs to reach the goal.
\end{itemize}

Your task is to evaluate if the video remains valid and if the runner successfully solves the maze. Only consider the part of the video from the beginning until the scene becomes invalid. A scene becomes invalid if:
\begin{itemize}[nosep,leftmargin=*]
    \item The layout of the maze (walls and paths) changes.
    \item The runner or goal appears/disappears, morphs, deforms, or changes color.
    \item The runner crosses or slides into the walls.
    \item There are significant glitches, noise, or artifacts.
    \item The camera pans, zooms, or moves (it should be a static top-down view).
\end{itemize}

Output your response as a JSON object with the following fields:
\begin{enumerate}[nosep]
    \item ``success'': A boolean indicating if the runner successfully reached the goal without any rule violations.
    \item ``invalid\_after\_seconds'': A float representing the timestamp (in seconds) when the scene first became invalid. If the video remains fully valid and does not violate any rules until the end, set this to null.
    \item ``justification'': A text explanation of your analysis. Explain why the video was marked valid/invalid or success/failure.
\end{enumerate}

Only output the JSON object, nothing else. Do not wrap it in markdown block.
\end{promptbox}

\begin{figure}
    \centering
    \includegraphics[width=0.5\linewidth]{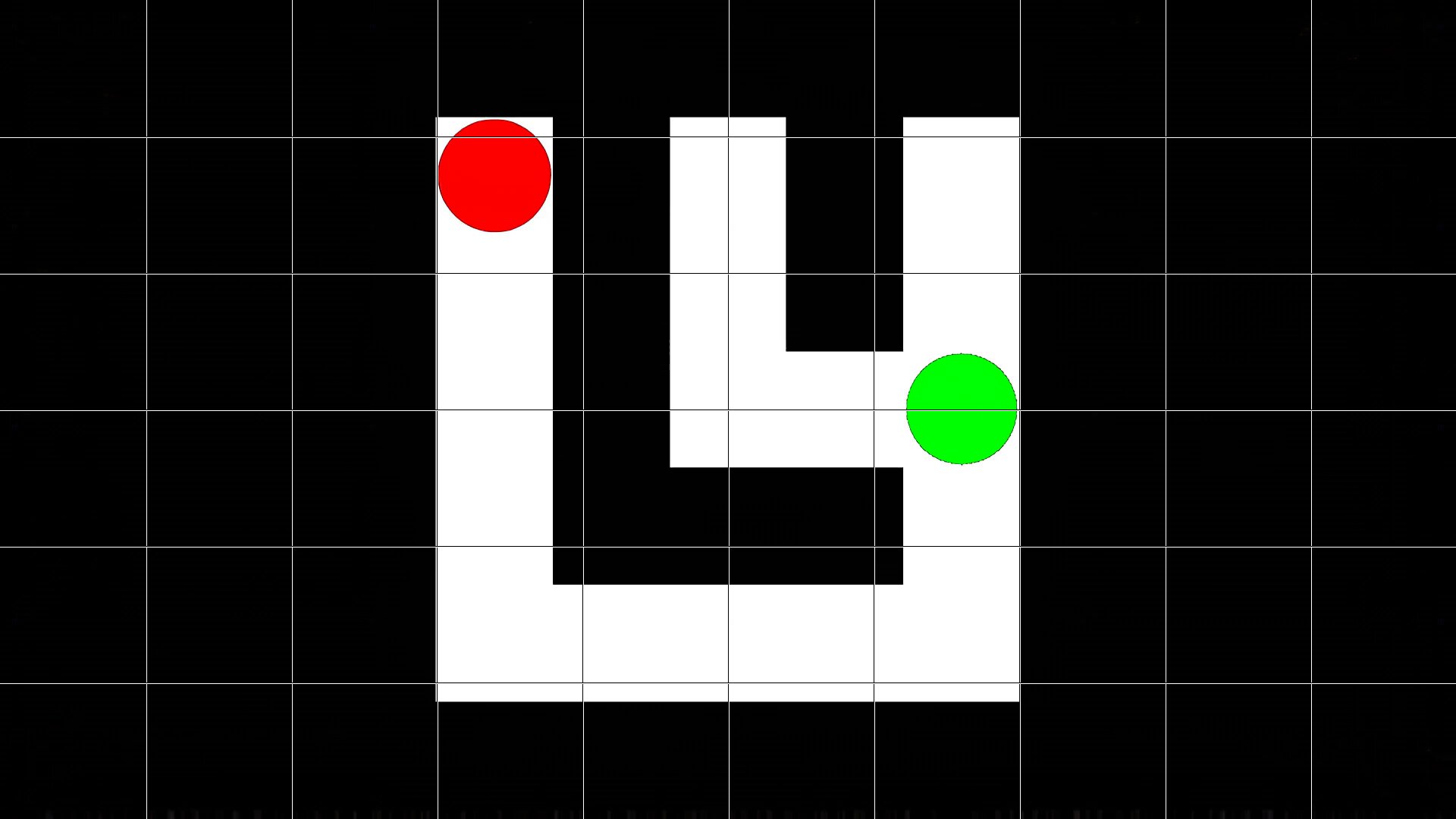}
    \caption{The first frame of a maze video, overlaid with a static grid. The same grid is used for all maze splits.}
    \label{fig:grid-overlay}
\end{figure}

\paragraph{VLM-based autorater with grid overlay}
We experiment with using overlaying the entire video with a static $16\times9$ grid with black-and-white \si{2}{px} grid lines, see~\cref{fig:grid-overlay}. We otherwise use the same autorater prompt and model (Gemini~3.1~Pro).

\paragraph{Human agreement}
We validate the VLM-based autoraters against human ratings on a sample of 100 videos drawn across all four splits (see~\cref{tab:maze-autorater-agreement}). Interestingly, while Gemini on the original videos only achieved `fair' agreement with a human rater (Cohen's $\kappa$ of 0.252), the grid overlay improved the rater to `moderate' agreement (Cohen's $\kappa$ of 0.583). Results in~\cref{sec:automated-vipe} consequently employ the latter autorater. This simple grid overlay can be viewed as a form of VIPE that is evidently beneficial for video understanding.

\begin{table}[]
    \centering
    \small
    \caption{VLM-based autorater-human agreement on 100 maze videos.}
    \label{tab:maze-autorater-agreement}
    \begin{tabular}{lrr}
        \toprule
        & \multicolumn{2}{c}{\textbf{VLM-based autorater}}\\
        \cmidrule(lr){2-3}
        \textbf{Metric} & \textbf{Original} & \textbf{With grid} \\
        \midrule
        Agreement & 76.00\% & 85.00\% \\
        Cohen's $\kappa$ & 0.252 & 0.583 \\
        \bottomrule
    \end{tabular}
\end{table}

\subsection{RushHour autoraters}
Solutions in \taskname{RushHour} are not necessarily unique (in some samples, obstacle cars can be moved independently in various orders). The task as formulated by~\citet{zeller2026mentisoculi} also does not penalize non-optimal solutions that, e.g., move cars back-and-forth or (unnecessarily) move distractor cars. Fortunately, the dataset comes with a validator that determines whether a given move sequence (e.g., \texttt{Car A forward, Car B backward, Car R forward}) solves a puzzle. The autoraters for this task are, thus, better understood as parsers that extract the move sequence from a video. While this task sounds trivial at first, many generated videos suffer from arbitrary perspective changes, non-continuous movements, duplicating or vanishing objects, etc.

\paragraph{VLM-based parser}
As in~\cref{app:maze-autorater}, we first test a VLM-based autorater, asking Gemini~3.1~Pro to parse the move sequence from the video (prompt detailed below). Gemini only has to parse each car's initial movement (following~\citet{zeller2026mentisoculi}), and provides absolute movement directions on an eight-point compass rose, which we can map to forward/backward based on each car's predefined axis of movement. This prompt, too, contains placeholders populated with the appropriate terms for the puzzle variant, e.g., using \texttt{anodized metal slides/deadbolts}, \texttt{metal slider}, \texttt{puzzle box backplate}, and \texttt{brass exit slot} for \texttt{objects}, \texttt{object\_singular}, \texttt{container}, and \texttt{exit} for the \texttt{anodized\_metal\_puzzle\_box} variant. We also experiment with a version of this parser for which each input video is overlaid with a static grid as in~\cref{app:maze-autorater}, \cref{fig:grid-overlay}.

\begin{promptbox}{RushHour sequence parsing prompt}
\texttt{[video]}

You are an expert video analyzer. Analyze the video which shows a puzzle being solved.
The puzzle is a variation of \taskname{RushHour}. In this variation:
\begin{itemize}[nosep,leftmargin=*]
    \item The objects are \texttt{\{objects\}}.
    \item The board/container is \texttt{\{container\}}.
    \item The exit is \texttt{\{exit\}}.
    \item One of the objects is the target (usually red/pinkish, labeled `R' or `red\_car'), which needs to reach the \texttt{\{exit\}}.
\end{itemize}

Your task is to extract the sequence of moves of the \texttt{\{objects\}} from the video and evaluate if the video remains valid.
Only consider the part of the video from the beginning until the scene becomes invalid.
A scene becomes invalid if:
\begin{itemize}[nosep,leftmargin=*]
    \item The layout of the \texttt{\{container\}} changes.
    \item \texttt{\{objects\}} appear or disappear.
    \item \texttt{\{objects\}} morph, deform, or change color.
    \item \texttt{\{objects\}} leave their defined tracks/axes.
\end{itemize}

For the valid part of the video, extract each \texttt{\{object\_singular\}}'s first move.
Each move should specify the object label and the absolute direction of movement on an 8-point compass rose (N, NE, E, SE, S, SW, W, NW), where N always points to the top of the frame.
Format each move as a string, e.g., ``A N'', ``B NW''.

Output your response as a JSON object with the following fields:
\begin{enumerate}[nosep]
    \item ``moves'': A list of strings representing the extracted moves in order, e.g., [``A N'', ``B NW'']. If no moves occur before the scene becomes invalid, this should be [].
    \item ``invalid\_after\_seconds'': A float representing the timestamp (in seconds) when the scene first became invalid. If the video remains fully valid and does not violate any rules until the end, set this to null.
    \item ``justification'': A text explanation of your analysis. Explain why the video was marked valid/invalid (e.g., if an object morphed, specify which object and at what time), and describe the moves you observed.
\end{enumerate}

Only output the JSON object, nothing else. Do not wrap it in markdown block.
\end{promptbox}

\paragraph{Algorithmic parser}
The algorithmic parser uses computer vision techniques to extract the move sequence and validate video consistency. The parser first aligns each frame to the starting frame to compensate for camera movement. Initial car positions are masked out during alignment to avoid errors from moving cars. Based on the first frame, the parser initializes a tracker for each movable car by cropping the car's template from the first frame. In subsequent frames, the tracker attempts to match this template along the cars permissible motion axis, restricting the problem to a 1D search space. A move (\texttt{forward} or \texttt{backward}) is registered when the displacement along this axis exceeds $0.5$ world units. The parser detects video invalidity using two main checks:
\begin{itemize}[nosep,leftmargin=*]
    \item \textbf{Object morphing}: If the template matching normalized cross-correlation (NCC) score falls below $0.6$, the object is considered morphed or vanished, and the video is marked invalid. In \emph{lenient} mode, tracking failures are ignored for cars that have already completed an initial move.
    \item \textbf{Background consistency}: The parser computes the absolute difference between the current frame and the starting frame in the (aligned) background region (excluding cars and rails). It applies an HSV-based shadow filter to ignore illumination changes caused by moving shadows. A non-shadow connected region of $>1000$ pixels is interpreted as a structural background change, marking the video as invalid.
\end{itemize}
The extracted move sequence is verified logically against the puzzle's initial state, and the parser terminates early as soon as a valid solution sequence is found.

\paragraph{Human agreement}
We validate the candidate autoraters against human ratings on a sample of 120 videos from both \taskname{RushHour} levels (see~\cref{tab:rushhour-autorater-agreement}). In contrast to~\cref{app:maze-autorater}, we don't find the grid overlay to be beneficial in this task for Gemini's video understanding, possible because most cars' motion axes are not axis-aligned. Both Gemini-based raters achieve only `fair' agreement (Cohen's $\kappa < 0.4$). The strict algorithmic parser fails entirely, since in the vast majority of generated videos, the first-moved car strays off its valid motion axis before the second car is moved. In the lenient case, where we only ensure that each car's initial move is in a valid direction and ignore it entirely afterwards (aligning with~\citep{zeller2026mentisoculi}), the parser achieves `moderate' agreement (Cohen's $\kappa$ 0.570). We use this last parser for all experiments in~\cref{sec:automated-vipe}.

\begin{table}[]
    \centering
    \caption{Autorater-human agreement on 120 \taskname{RushHour} videos.}
    \label{tab:rushhour-autorater-agreement}
    \begin{tabular}{lrrrr}
    \toprule
    & \multicolumn{2}{c}{\textbf{VLM-based parser}} & \multicolumn{2}{c}{\textbf{Algorithmic parser}} \\
    \cmidrule(lr){2-3} \cmidrule(lr){4-5}
    \textbf{Metric} & \textbf{Original} & \textbf{With grid} & \textbf{Strict} & \textbf{Lenient} \\
    \midrule
    Agreement & 82.50\% & 80.83\% & 76.67\% & 87.50\% \\
    Cohen's $\kappa$ & 0.380 & 0.250 & 0.000 & 0.570 \\
    \bottomrule
    \end{tabular}
\end{table}

\end{document}